\documentclass[lettersize,journal]{IEEEtran}
\usepackage{times}
\usepackage{epsfig}
\usepackage{graphicx}
\usepackage{amsmath}
\usepackage{amssymb}
\usepackage{graphicx}
\usepackage{amsmath}
\usepackage{amssymb}
\usepackage{booktabs}
\usepackage{multirow}
\usepackage{caption,subcaption}
\def\BibTeX{{\rm B\kern-.05em{\sc i\kern-.025em b}\kern-.08em
		T\kern-.1667em\lower.7ex\hbox{E}\kern-.125emX}}
\usepackage{balance}
\usepackage[normalem]{ulem}
\useunder{\uline}{\ul}{}
\usepackage{url}

\begin{document}
	\title{SR-R$^2$KAC: Improving Single Image Defocus Deblurring}
	\author{Peng TANG, Zhiqiang Xu, Pengfei Wei, Xiaobin Hu, Peilin Zhao, Xin Cao, Chunlai Zhou, Tobias Lasser
		\thanks{Peng TANG, Xiaobin Hu, Tobias Lasser are with Department of Informatics,  School of Computation, Information and Technology, Technical University of Munich, Garching, Germany.
			Zhiqiang Xu is the corresponding author (zhiqiangxu2001@gmail.com)}}
	
	\markboth{Journal of \LaTeX\ Class Files,~Vol.~18, No.~9, September~2020}%
	{How to Use the IEEEtran \LaTeX \ Templates}
	
	\maketitle
	
	\begin{abstract}
		We propose an efficient deep learning method for single image defocus deblurring (SIDD) by further exploring inverse kernel properties. 
		Although the current inverse kernel method, i.e., kernel-sharing parallel atrous convolution (KPAC), can address spatially varying defocus blurs, it has difficulty in handling large blurs of this kind. 
		To tackle this issue, we propose a Residual and Recursive Kernel-sharing Atrous Convolution (R$^2$KAC).
		R$^2$KAC builds on a significant observation of inverse kernels, that is, successive use of inverse-kernel-based deconvolutions with fixed size helps remove unexpected large blurs but produces ringing artifacts. 
		Specifically, on top of kernel-sharing atrous convolutions, R$^2$KAC stacks atrous convolutions recursively to simulate a large inverse kernel.
		To further alleviate the contingent effect of recursive stacking, i.e., ringing artifacts, we add identity shortcuts between atrous convolutions to simulate residual deconvolutions. 
		Lastly, a scale recurrent module is embedded in the R$^2$KAC network, leading to SR-R$^2$KAC, so that multi-scale information from coarse to fine is exploited to progressively remove the spatially varying defocus blurs.
		Extensive experimental results show that our method achieves the state-of-the-art performance. 
	\end{abstract}
	

	\section{Introduction}
	\IEEEPARstart{D}{efocus} blur is one of the common blur effects in photographs.
	It usually occurs when a large camera aperture is used.
	The large aperture may make a shallow depth of field (DoF), and defocus blur will be generated outside DoF.
	The farther the object is away from the DoF, the stronger the defocus blur is.
	Removing undesired defocus blurs is not only highly demanded by photographers but also beneficial for high-level vision applications \cite{Abuolaim2020,Lee2021} such as object detection \cite{Cao2020, Dai2016} and semantic segmentation \cite{Noh2015,Wang2020}. 
	However, spatially-varying and large blurs make single image defocus deblurring (SIDD) challenging.

	Before deep learning based SIDD, two-stage approaches \cite{Bando2007, Shi2015, Park2017, Andres2016, Cho2017, Quan2021, Lee2019} dominate the field, where the first stage is defocus map estimation (DME) and the second stage is to incorporate the estimated defocus map into the non-blind deconvolution \cite{Fish1995, Krishnan2009} to restore a blurry image.
	However, accurate DME is difficult to achieve since the practical defocus blurs are spatially varying and thus cannot be approximated by a simple predefined kernel.
	Therefore, two-stage approaches are highly limited by DME errors, and moreover the high computational cost occurred in inference.
	Some recent approaches apply DME into deep learning to improve deblurring accuracy, such as \cite{Lee2019,Quan2021}, however they still suffer from the same difficulty in obtaining accurate estimated defocus maps.
	
	DPDNet \cite{Abuolaim2020} is the first end-to-end SSID method that directly learns a sharp image from a blurry image, and it achieves much better results than previous two-stage approaches.
	However, as indicated in \cite{Lee2021,Zhou2019,Son2021}, the common UNet structure of DPD makes it not adaptable enough to handle spatially varying and large blurs. 
	To improve, more advanced modules including Kernel-Sharing Parallel Atrous Convolution (KPAC) \cite{Son2021} and Iterative Filter Adaptive Nnetwork (IFAN) \cite{Lee2021} are proposed. 
	These two methods outperform DPDNet by successfully handling spatially-varying blurs, but leave the issue of large defocus blurs not addressed.
	In this paper, we focus on the solutions to eliminate large defocus blurs.
	We investigate and improve based on KPAC as it achieves comparable performance with less more compact structure compared with IFAN.
	
	In this paper, we propose an end-to-end deep learning framework, Residual and Recursive Kernel-Sharing Recursive Atrous Convolution with scale recurrent module (SRM) simplified as SR-R$^2$KAC, for SIDD.
	SR-R$^2$KAC is build on KPAC so that it inherits the capability of KPAC in handling spatially-varying defocus blurs.
	We further propose improvements to make R-R$^2$KAC also qualified in addressing large defocus blurs.
	The improvements in SR-R$^2$KAC is motivated by an observation of inverse kernel decomposition, that is, we find that a large-sized inverse-kernel-based deconvolution can be approximated by successively stacking small-sized inverse-kernel-based deconvolutions, accompany with an undesired effect of ringing artifacts.
	Specifically, we propose to apply the kernel-sharing atrous convolutions recursively to simulate a large inverse kernel to tackle large defocus blurs, and simultaneously add identity shortcuts between atrous convolutions for residual connection simulation to reduce the contingent effect of ringing artifacts.
	KPAC with such an improvement gives us R$^2$KAC.
	To further boost the performance, we propose to take advantage of a scale recurrent mechanism \cite{Tao2018,Quan2021} that fully exploits multi-scale information to progressively remove defocus blurs in a coarse-to-fine manner.
	
	We conduct experimental studies to demonstrate the superiority of our SR-R$^2$KAC over existing methods using multiple public defocus blur datasets, including DPDD \cite{Abuolaim2020}, ReadDOF \cite{Lee2021}, PixelDP \cite{Abuolaim2020}, and CUHK \cite{Shi2014}. 
	Experimental results show the effectiveness of our SR-R$^2$KAC in handling both spatially-varying and large blurs.
	
	\section{Related Works}
	
	We focus on the literature of defocus deblurring due to the space limit. 
	Previous SIDD methods can be roughly divided into two categories: two-stage based and end-to-end learning-based approaches. 
	Most of two-stage based approaches \cite{Bando2007, Shi2015, Park2017, Andres2016, Cho2017, Krishnan2009, Lee2019} put emphasis on the first stage, i.e., DME, as it is the basis for the second stage of non-blind deconvolution \cite{Fish1995, Krishnan2009}. 
	One typical scheme of improving the DME accuracy is to capture more hand-crafted features \cite{Bando2007, Shi2015, Park2017, Andres2016, Cho2017, Krishnan2009}.
	Other methods \cite{Lee2019, Quan2021} also incorporate predefined kernel approximation into CNN to achieve better performance.
	However, all these approaches use a predefined blur model without considering the complex property of real-world defocus images, such as irregular shapes and varying sizes, and thus achieve unsatisfactory performance in practice. 
	
	Abuolaim and Brown \cite{Abuolaim2020} proposed the first end-to-end deep learning based method, called DPDNet, for SIDD. 
	It significantly outperforms typical two-stage based approaches, but still fails to deal with spatially varying and large blurs.
	Recently, some more advanced methods, e.g., KPAC \cite{Son2021} and IFAN \cite{Lee2021}, were proposed.
	To well address spatially-varying blur, KPAC parallely combines the outputs of multiple kernel-sharing atrous convolutions and learns attention maps to simulate a linear combination of multiple inverse-kernel deconvolutions.  
	In contrast, IFAN \cite{Lee2021} adopts a filter prediction network (FAN) that predicts pixel-wise deblurring filters for adaptively handling the defocused features and then feeds these features into a restoration network for deblurring defocus blurs.
	Both KPAC and IFAN are specifically proposed to handle spatially-varying blurs, but leaving the issue of large defocus blur not well addressed.
	Moreover, compared with KPAC, IFAN needs much more model parameters and extra dual-pixel image for training, which limits its application in practical scenarios, especially those without dual-pixel images.
	Thus, we select KPAC as our backbone model.
	
	\subsection{Analysis of KPAC}
	
	The core of KPAC is inverse kernel deconvolutions, and thus we analyze the possible limitation of KPAC in handling large blur from the perspective of inverse kernel deconvolutions.
	Let us first review two main properties of KPAC, i.e., Eqs.~(\ref{size_change}) and (\ref{eq15}).
	\begin{equation} \label{size_change}
		\left(\frac{1}{s^{2}} k_{\uparrow s}\right)^{\dagger}=\frac{1}{s^{2}}\left(k_{\ \uparrow s}^{\dagger}\right),
	\end{equation}
	\begin{equation} \label{eq15}
		I_S \approx \sum_{i}\Big\{w_{i} \cdot\big(\frac{1}{s_{i}^{2}} k^{\dagger}_{\ \uparrow s_{i}} * I_B\big)\Big\},
	\end{equation}
	where $k$ is the blur kernel and $k^{\dagger}$ is the corresponding inversed kernel approximated by:
	\begin{equation} \label{inverse_kernel}
		k^{\dagger} = f^{-1}(\frac{1}{f(k)})
	\end{equation}
	with $f(\cdot)$ and $f^{-1}(\cdot)$ indicating discrete Fourier transformation (DFT) and inverse Fourier transformation (IFT), respectively,
	$s \in \{1,2,3,...,N\}$ is the scaling factor, $\uparrow s$ is the upsampling operation, the symbol $*$ represents the convolution operation, $w_{i}$ is the weight factor, and $I_S$ and $I_B$ indicate the sharp and blurry images. 
	Eq.~(\ref{size_change}) presents an observed property, that is, the shape of the corresponding inverse kernel $k^{\dagger}$ remains the same when the spatial support size of a blur kernel $k$ changes.
	Eq.~\ref{eq15} presents	a linear combination of inverse kernels $k^{\dagger}_{\ \uparrow s_{i}}$ with different discrete scaling factors to approximate the deconvolution.
	Note that KPAC uses mulitple scaling factors with shared-shape inverse kernels to model the defocus blur with varying size but the same shape.
	As it is unrealistic to include infinite scaling factors for all possible sizes, a limited scaling factor set is used.
	This makes KPAC fail to handle those large defocus blurs whose scaling factor is larger than the maximum of the scaling factor set.
	
	We further validate the above analyses by experiments. 
	We follow \cite{Son2021} to use Lanczos upsampling for scaling inverse kernels and Wiener deconvolutions for computing spatial inverse kernels. 
	We set the maximum of the scaling factor to 5, i.e., the maximum $s_i$ is 5, in the learning phase, but test on two cases where the target scaling factors $s_t$ are 7 and 3.5, respectively.
	The former test case represents the blur with an unexpected large blur, while the latter represents the blur with normal spatially-varying size.
	We use the approximated accuracy as the evaluation metric following \cite{Son2021}.
	Fig.~\ref{fig1}(1) shows the deblur results of KPAC where the first and second rows correspond to the results for the target scaling factor 7 and 3.5, respectively.
	As can be seen, KPAC obtains a visually pleasing result for the second row (approximated accuracy is 98.0$\%$) but suffers a significant performance degradation for the first row (approximated accuracy is 88.7$\%$).
	
	Theoretically, an intuitive way to tackle this issue is to increase the scaling factor $s_i$. 
	Increasing $s_i$ in the inverse kernel deconvolution is equivalent to increasing the dilation rate of the corresponding convolution in the CNN structure.
	We empirically find that doing so does not help KPAC to handle large defocus blurs.
	The experimental results indicate that using a larger dilation rate does not improve the deblurring performance but increases the computational complexity. 
	More analyses can be found in the Sec.III.J.
	Therefore, we aim to design a new deep structure that can effectively handle large blurs without increasing scaling factors.

	\subsection{Derivation of R$^2$KAC}
	
	\begin{figure*}[h]
		\centering
		
		\begin{subfigure}[t]{0.16\linewidth}	
			\includegraphics[width=2.6cm,height=3cm]{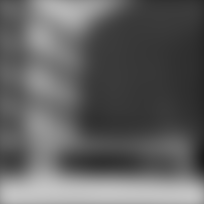}\vspace{4pt}
		\end{subfigure}
		\hfill
		\begin{subfigure}[t]{0.16\linewidth}	
			\includegraphics[width=2.6cm,height=3cm]{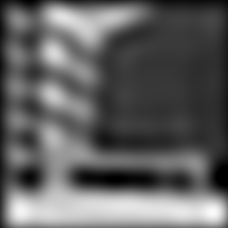}\vspace{4pt}
		\end{subfigure}
		\hfill
		\begin{subfigure}[t]{0.16\linewidth}	
			\includegraphics[width=2.6cm,height=3cm]{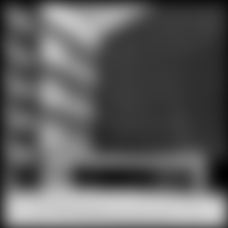}\vspace{4pt}
		\end{subfigure}
		\hfill
		\begin{subfigure}[t]{0.16\linewidth}	
			\includegraphics[width=2.6cm,height=3cm]{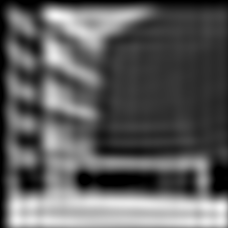}\vspace{4pt}
		\end{subfigure}
		\hfill
		\begin{subfigure}[t]{0.16\linewidth}	
			\includegraphics[width=2.6cm,height=3cm]{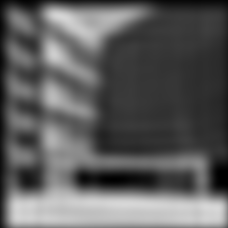}\vspace{4pt}
		\end{subfigure}
		
		\vfill
		\begin{subfigure}[t]{0.16\linewidth}	
			\includegraphics[width=2.6cm,height=2.6cm]{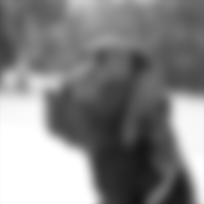}\vspace{4pt}
			\caption{By  $\frac{1}{s_t^{2}} k_{\uparrow s_t}$}
		\end{subfigure}
		\hfill
		\begin{subfigure}[t]{0.16\linewidth}	
			\includegraphics[width=2.6cm,height=3cm]{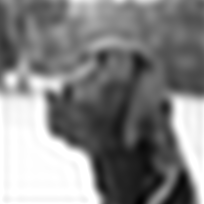}\vspace{4pt}
			\caption{By $\frac{1}{s_t^{2}} k^{\dagger}_{\ \uparrow s_t}$}
		\end{subfigure}
		\hfill
		\begin{subfigure}[t]{0.16\linewidth}	
			\includegraphics[width=2.6cm,height=3cm]{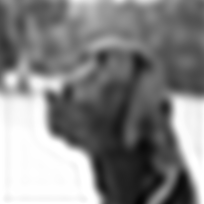}\vspace{4pt}
			\caption{By KPAC}
		\end{subfigure}
		\hfill
		\begin{subfigure}[t]{0.16\linewidth}	
			\includegraphics[width=2.6cm,height=3cm]{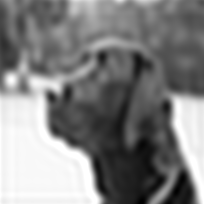}\vspace{4pt}
			\caption{By RKAC}
		\end{subfigure}
		\hfill
		\begin{subfigure}[t]{0.16\linewidth}	
			\includegraphics[width=2.6cm,height=3cm]{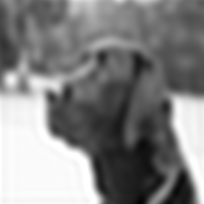}\vspace{4pt}
			\caption{By R$^2$KAC}
		\end{subfigure}
		\caption{Preliminary emprical comparsions.}
		\vspace{-0.2in}
		\label{fig1}
	\end{figure*}
	
	Our solution to handle large blurs is based on our new observations on inverse kernel properties. 
	Specifically, we find that a large inverse kernel can be approximated by two small-sized inverse kernel deconvolutions successively.
	We mathematically formulate this observation as follows.
	We first model the blurry image by applying a blur kernel on the sharp image \cite{Son2021,Ren2018}.
	\begin{equation} \label{eq16_1}
		I_B = k * I_S
	\end{equation}	
	We then use a Gaussian kernel to approximate the defocus blur shape, as Gaussian kernel approximation has proven to be effective in SIDD \cite{Shi2015,Park2017,Krishnan2009,Lee2019, Son2021}:
	\begin{equation} \label{eq16}
		I_B = k * I_S \approx g_{s_3} * I_S.
	\end{equation}	
	Since applying two Gaussian blurs with small standard deviations can achieve the same effect as applying the Gaussian blur with a larger standard deviation \footnote{\url{https://math.stackexchange.com/q/3159846}},  
	we obtain that: 
	\begin{equation} \label{eq17}
		I_B \approx g_{s_3} * I_S = g_{s_2} * g_{s_1} * I_S,
	\end{equation} 
	where the standard deviations of $g_{s_1}$, $g_{s_2}$, and $g_{s_3}$ are $s_1$, $s_2$ and $s_3$, respectively, and $s_3 = \sqrt{s_2^2 +s_1^2}$.
	After performing DFT on Eq.~(\ref{eq17}), we obtain Eq.~(\ref{eq18}):
	\begin{equation} \label{eq18}
		f(I_B) \approx f(g_{s_2}) \odot f(g_{s_1}) \odot f(I_S), f(I_B) \approx f(g_{s_3}) \odot f(I_S) 
	\end{equation}	
	where $\odot$ is element-wise multiplication. Then, we have:
	\begin{equation} \label{eq19}
		f(I_S) \approx \frac{1}{f(g_{s_2})} \odot \frac{1}{f(g_{s_1})} \odot f(I_B), f(I_S) \approx \frac{1}{f(g_{s_3})} \odot f(I_B) 
	\end{equation}	
	Next, with IFT on Eq.~(\ref{eq19}) and the pseudo inverse kernel in Eq.~(\ref{inverse_kernel}), we derive:
	\begin{equation} \label{eq20}
		I_S  \approx (g_{s_3})^{\dagger} * I_B \approx (g_{s_2})^{\dagger} * (g_{s_1})^{\dagger} *I_B,
	\end{equation}
	where $(g_{s_j})^{\dagger}$ is the spatial inverse kernel of $g_{s_j}$ where $j=1,2,3$. 
	We then consider a Gaussain blur $g$ with the unit standard deviation, i.e., $\sigma(g) = 1$, and increase $\sigma(g)$ by $s_j$ times to get $g_{s_j}$ for $j=1,2,3$. 
	As shown in \cite{Son2021}, changing the standard deviation of a Gaussian kernel by $s$ times is equivalent to upsampling the Gaussian kernel by a sinc filter with the scaling factor $s$. Thus, we have:
	\begin{equation} \label{eq21}
		I_S  \approx (\frac{1}{s_3^{2}}g_{\uparrow s_3})^{\dagger} * I_B \approx  (\frac{1}{s_2^{2}}g_{\uparrow s_2})^{\dagger} * (\frac{1}{s_1^{2}}g_{\uparrow s_1})^{\dagger} *I_B.
	\end{equation}
	With Eq. (\ref{size_change}), we finally obtain:
	\begin{equation} \label{eq22}
		I_S  \approx \frac{1}{s_3^{2}}g_{\uparrow s_3}^{\dagger} * I_B \approx  \frac{1}{s_2^{2}}g_{\uparrow s_2}^{\dagger} * \frac{1}{s_1^{2}}g_{\uparrow s_1}^{\dagger} *I_B.
	\end{equation}
	As shown in Eq.~(\ref{eq22}), we may use multiple small shape-sharing inverse kernels\footnote{We showcase $2$ here, but it can be easily extended to more kernels.} to model a larger inverse kernel for effectively removing larger Gaussian blurs.
	
	To manage spatially-varying blurs, we maintain the linear combination of multiple inverse kernel-based deconvolutions following Eq.~(\ref{eq15}). 
	We further adopt the proposed multiple small inverse kernel deconvolutions with fixed size successively stated above, and we obtain:
	\begin{equation} \label{definition_1}
		f_{i}(x) =  \frac{1}{s_{i}^{2}} g^{\dagger}_{\ \uparrow s_{i}} * x
	\end{equation}
	\begin{equation} \label{RKAC}
		I_S \approx \sum_{i}\Big\{w_{i} \cdot \big(f_i(f_{i-1}(...f_1(I_B)...)) \big)\Big\}
	\end{equation}
	
	To reduce the contingent ringing artifacts, we follow \cite{Yuan2007} to add residual deconvolutions by reducing the absolute amplitude of the Gibbs oscillations. 
	Therefore, the final form of R$^2$KAC is formulated as follows,
	\begin{equation} \label{definition_2}
		fr_{i}(x) =  \frac{1}{s_{i}^{2}} g^{\dagger}_{\ \uparrow s_{i}} * x + x
	\end{equation}	
	\begin{equation} \label{R2KAC}
		I_S \approx \sum_{i}\Big\{w_{i} \cdot \big(fr_i(fr_{i-1}(...fr_1(I_B)...)) \big)\Big\}
	\end{equation}
	
	We also empirically validate the R$^2$KAC block following the experimental setting stated in Section 3.1.
	We compare with KPAC and a variant of R$^2$KAC, RKAC (i.e., R$^2$KAC without the residual deconvolution).
	Fig.~\ref{fig1}(c-e) show the comparison results. 
	As can be seen from the first row, none of methods can perfectly fit the size of the large blur.
	However, R$^2$KAC achieves the best approximation accuracy $93.3\%$ compared with KPAC ($88.7\%$) and RKAC ($85.4\%$), which shows the superiority of R$^2$KAC in handling large blurs. 
	Moreover, from the second row, we find that the approximation accuracy of KPAC, RKAC, and R$^2$KAC are comparable, namely 98.0$\%$, 96.1$\%$, and 99.1$\%$
	This shows that R$^2$KAC can also handle the blur with normal spatially-varying size. 
	\begin{figure}[t]
		\centering
		\includegraphics[width=8cm, height=5cm]{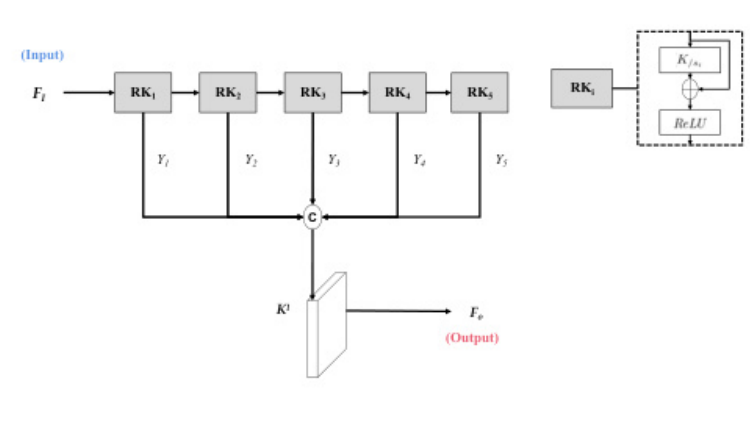}	
		\caption{Structure of R$^2$KAC. } 
		\vspace{-0.2in}
		\label{fig2}
	\end{figure}	
	\begin{figure*}[t]
		\centering
		\includegraphics[width=18cm, height=9cm]{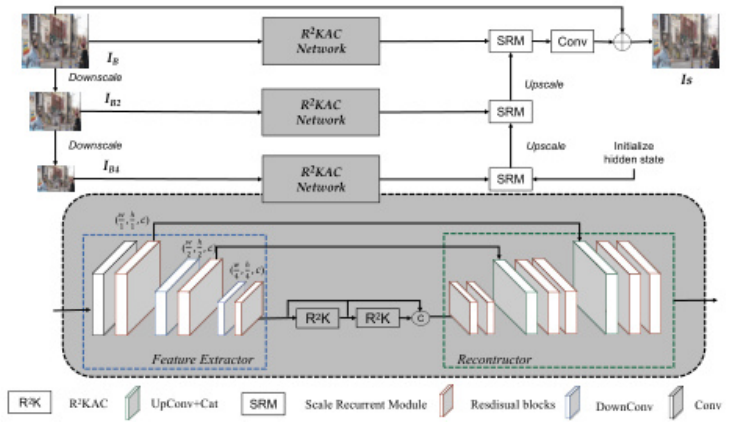}
		\caption{Defocus deblurring network with SR-R$^2$KAC module. The R$^2$KAC network is weight-shared.
			The size of $I_{B2}$ and $I_{B4}$ is 1/2, 1/4 of that of $I_B$ respectively.} 
		\label{fig3}
	\end{figure*}

	\subsection{Implementing R$^2$KAC with CNN}
	We design a CNN based network to implement R$^2$KAC.
	As shown in Fig. \ref{fig2}, R$^2$KAC takes a defocused feature map $F_1 \in \mathbb{R}^{W \times H \times C}$ as input, where $W$, $H$, and $C$ are width, height, and the number of channels of the feature map, respectively. 
	It then recursively stacks five kernel-sharing atrous convolutions with identity shortcut, denoted as $RK_i$ ($i={1,2,..., N}$), to simulate the combination of inverse-kernel-based successive deconvolutions as well as residual deconvolutions for handling large blurs.
	Each $RK_i$ contains a kernel-sharing atrous convolution $K_{/ s_i} \in \mathbb{R}^{k \times k \times C \times C}$, a nonlinear activation $ReLU$ and a residual connection, where $s_i$ is the dilation rate of the kernel-sharing atrous convolution.
	Finally, R$^2$KAC uses the feature concatenation operation $CAT(\cdot)$ and a common convolution $K^l \in \mathbb{R}^{3 \times 3 \times NC \times C}$ to achieve a nonlinear aggregation of the outputs $Y_i$ for dealing with spatially varying blurs. 
	As shown in Fig. \ref{fig2}, the generated output feature map $F_2  \in \mathbb{R}^{W \times H \times C}$ can be obtained as follows:
	\begin{equation} \label{eq2}
		F_{2}=K^{l} * CAT\left(Y_1,...,Y_N \right).
	\end{equation}
	Mathematically, each $Y_i$ is represented as:
	\begin{equation} \label{eq3}
		Y_i =
		\begin{cases}
			ReLU(RK_{1} * F_1),  &\hskip -.3em \text{if} \  i=1 \\
			ReLU(K_{/s_i}*RK_{i-1}(F_1) + RK_{i-1}(F_1)), & \hskip -.3em \text{if } \ i>1
		\end{cases}
	\end{equation}
	
	Note that R$^2$KAC does not learn the scale and shape attention maps to combine the outputs as we empirically find that removing attention modules increases the performance and reduces computational cost.
	We believe this is due to the following reasons: (1) excessive scale and shape attention modules make the network overfit to the training set and thus limit the generalization ability, and (2) the deep learning process can make the model implicitly learn the combination of different kernel-sharing layers.
	
	\subsection{Defocus Deblurring Network with SR-R$^2$KAC}
	To further improve the deblurring performance,  we adopt the scale recurrent mechanism (SRM) \cite{Tao2018,Quan2021} to enable R$^2$KAC progressively conduct defocus deblurring from coarse to fine (from low resolution $I_{B4}$ to high resolution $I_B$).
	Combining R$^2$KAC and SRM, Fig. \ref{fig3} shows our complete defocus deblurring network, SR-R$^2$KAC.
	Precisely, R$^2$KAC network contains three parts: Feature Extractor, Reconstructor, and stacked $R^2K$ blocks. 
	To facilitate the processing in multi-scale feature spaces, we use a UNet-like structure \cite{Ronneberger2015} as the backbone and embed stacked $R^2K$ blocks between Feature Extractor and Reconstructor parts. 
	All convolution layers in R$^2$KAC use Rectified Linear Unit (ReLU) \cite{Nair2010} for nonlinear activation, except for the final layer that uses Softmax function.
	Following \cite{Zamir2022,Lee2021}, we also add some residual blocks in the Reconstructor and Feature Extractor modules, as indicated by the red square block in Fig.~\ref{fig3}.
	For SRM, as shown in Fig.~\ref{fig3}, SR-R$^2$KAC extracts the information from low-resolution, such as $I_{B4}$ and $I_{B2}$, via R$^2$KAC network and then progressively pass the extracted information to the next stage of higher-resolution by SRM (here we uses the attention processing unit \cite{Quan2021}) and Upscaling operation.
	
	\section{Experiments}
	We implement our SR-R$^2$KAC using Pytorch \cite{Paszke2017}.
	The final network consists of $2$ stacked R$^2$KAC blocks, and the kernel size for the stacked blocks is $5$. 
	The channel $C$ is set to 48.
	We used Adam optimizer \cite{Kingma2014} with a batch size of 4 for training.
	The network is trained for 2000 epochs with an initial learning rate of $1.0 \times 10^{-4}$.
	The learning rate is step-decayed by half at the 1000th and 1500th epochs. 
	Stochastic weights averaging \cite{Izmailov2018} is used at the last 100 epochs to obtain final weights for evaluation.
	The images are cropped into $256 \times 256$ patches for training.
	We use mean-squared error (MSE) between the output and ground-truth sharp images as a content loss function. 
	Recent studies show that the frequency reconstruction loss \cite{Cho2021, Ichimura2018, Jiang2020} and the perceptual loss \cite{Johnson2016, Son2021} can bring performance improvement in many image enhancement and restoration tasks. 
	Thus, we also incorporate the frequency reconstruction loss and perceptual loss to co-supervise our defocus deblurring network. 
	The balancing factors of content loss, frequency reconstruction loss, and perceptual loss are empirically set to 1, 0.2, and 0.2, respectively. 
	
	Four public datasets including DPDD \cite{Abuolaim2020}, ReadDOF \cite{Lee2021}, PixelDP \cite{Abuolaim2020}, and CUHK defocus blur detection dataset \cite{Shi2014} are used in the experiments. 
	We use the training set of DPDD for training.
	The test set of DPDD and the other three datasets are used for evaluation.
	
	\subsection{Experiment Analysis}
	
	We first conduct ablation studies to analyze the effectiveness of each component in SR-R$^2$KAC.
	
	\begin{table}[t] \tiny
		\centering
		\resizebox{\linewidth}{!}{
			\begin{tabular}{c|cc|cc|c}
				\toprule
				\multirow{3}{*}{Methods} & \multicolumn{2}{c|}{\multirow{2}{*}{DPDD}} & \multicolumn{2}{c|}{\multirow{2}{*}{RealDoF}} & \multirow{3}{*}{Param(M)} \\
				& \multicolumn{2}{c|}{}        & \multicolumn{2}{c|}{}         &                                     \\
				& PSNR                         & SSIM        & PSNR                          & SSIM          &                                     \\ \midrule
				R2KAC                   & 26.04                        & 0.801       & 25.06                         & 0.753         & 1.72                                \\
				KPAC                    & 25.90                        & 0.796       & 24.77                         & 0.740         & 1.90                                \\
				RB                      & 25.58                        & 0.784       & 23.99                         & 0.702         & 1.74                                \\ \bottomrule
		\end{tabular}}
		\caption{Comparisons of R$^2$KAC, KPAC and RB.
		}	
		\label{tab2}
	\end{table}
	\begin{table}[t]
		\centering
		\resizebox{\linewidth}{!}{
			\begin{tabular}{cccc|cc|cc|c}
				\toprule
				\multicolumn{3}{c}{R$^2$KAC}                                         & \multirow{3}{*}{SRM} & \multicolumn{2}{c|}{\multirow{2}{*}{DPDD}} & \multicolumn{2}{c|}{\multirow{2}{*}{RealDoF}} & \multirow{3}{*}{Param(M)} \\ \cmidrule{1-3}
				\multirow{2}{*}{RKAC} & \multirow{2}{*}{RC} & \multirow{2}{*}{AM} &                      & \multicolumn{2}{c|}{}        & \multicolumn{2}{c|}{}         &                                     \\
				&                     &                     &                      & PSNR                         & SSIM        & PSNR                          & SSIM          &                                     \\ \midrule
				$\checkmark$          &                     &                     &                      & 25.92                        & 0.797       & 24.73                         & 0.739         & 1.72                                \\
				$\checkmark$          & $\checkmark$        &                     &                      & 26.04                        & 0.801       & 25.06                         & 0.753         & 1.72                                \\
				$\checkmark$          & $\checkmark$        & $\checkmark$        &                      & 25.99                        & 0.799       & 24.98                         & 0.751         & 1.90                                \\ \midrule
				$\checkmark$          & $\checkmark$        &                     & $\checkmark$         & 26.19                        & 0.804       & 25.32                         & 0.761         & 2.21                                \\ \bottomrule
		\end{tabular}}
		
		\caption{Ablation study of SR-R$^2$KAC.
		}	
		\label{tab2_1}
	\end{table}

	\begin{table}[t]
		\centering
		
		\scalebox{1}{
			\renewcommand\arraystretch{1.5}
			\setlength{\tabcolsep}{2mm}{	
				\begin{tabular}{cccccc}
					\toprule
					\multirow{2}{*}{Number} & \multicolumn{2}{c}{DPDD} & \multicolumn{2}{c}{RealDof} & \multirow{2}{*}{Param(M)} \\ \cmidrule{2-5}
					& PSNR (dB)     & SSIM     & PSNR (dB)      & SSIM       &                           \\ \midrule
					1                       & 26.01         & 0.799    & 25.01          & 0.752      & 1.56                      \\
					2                       & 26.04         & 0.801    & 25.06          & 0.753      & 1.72                      \\
					3                       & 26.04        & 0.800    & 25.06          & 0.753      & 1.89                      \\ \bottomrule
		\end{tabular}}}
		\caption{The effect of the number of R$^2$KAC blocks.}	
		\label{tab_44}
		\vspace{-0.2in}
	\end{table}

	\begin{table*}[t]
		\centering
		
		\resizebox{\linewidth}{!}{
			\begin{tabular}{ccccc|cccccc}
				\toprule
				\multirow{2}{*}{Model} & \multicolumn{4}{c|}{DPDD}                                         & \multicolumn{4}{c}{RealDoF}                                       & \multirow{2}{*}{Parameter (M)} & \multirow{2}{*}{Time (Sec)} \\
				& PSNR (dB)      & SSIM           & LPIPs          & MAE            & PSNR (dB)      & SSIM           & LPIPs          & MAE            &                                &                            \\ \midrule
				Blurry Input           & 23.89          & 0.725          & 0.349          & 0.471          & 22.33          & 0.633          & 0.524          & 0.513          & -                              & -                          \\
				JNB                    & 23.69          & 0.707          & 0.442          & 0.480          & 22.36          & 0.635          & 0.601          & 0.511          & -                              & -                          \\
				EBDB                   & 23.94          & 0.723          & 0.402          & 0.468          & 22.38          & 0.638          & 0.594          & 0.509          & -                              & -                          \\
				DMENet                 & 23.90          & 0.720          & 0.410          & 0.470          & 22.41          & 0.639          & 0.597          & 0.508          & 26.94                          & -                          \\
				DPDNet                 & 24.03          & 0.735          & 0.279          & 0.461          & 22.67          & 0.666          & 0.420          & 0.506          & 35.25                          & -                          \\
				KPAC                   & 25.22          & 0.774          & 0.227          & -              & 23.89          & 0.718          & 0.336          & -              & 1.58                           & 0.090                      \\
				IFAN                   & 25.37          & 0.789          & 0.217          & 0.217          & 24.71          & 0.748          & 0.306          & 0.407          & 10.48                          & 0.016                      \\
				GKMNet                 & 25.36          & 0.774          & 0.276          & 0.401          & 24.30          & 0.716          & 0.399          & 0.413          & 1.41                           & 0.036                      \\
				DRBNet                 & 25.47          & 0.787          & 0.246          & -              & 24.71          & 0.744          & 0.337          & -              & 11.69                          & 0.011                      \\
				Restormer              & 25.98          & 0.811          & \textbf{0.178} & 0.378          & -              & -              & -              & -              & 26.10                          & 0.45                      \\ \midrule
				R$^2$KAC               & 26.04          & 0.801          & 0.191          & 0.376          & 25.06          & 0.753          & 0.303          & 0.391          & 1.72                           & 0.009                      \\
				SR-R$^2$KAC            & \textit{26.19} & 0.803          & 0.194          & \textit{0.371} & \textit{25.41} & \textit{0.765} & \textit{0.297} & \textit{0.377} & 2.21                           & 0.039                      \\
				SR-R$^2$KAC-B          & \textbf{26.41} & \textbf{0.811} & \textit{0.179} & \textbf{0.363} & \textbf{25.59} & \textbf{0.772} & \textbf{0.279} & \textbf{0.372} & 8.81                           & 0.043                      \\ \bottomrule
		\end{tabular}}
		\caption{Comparison between our method and previous defocus deblurring methods on the DPDD dataset. \textbf{Bold numbers} represent the best values, and {\it italics numbers} represent the second best values.
		}		  
		\label{tab6}
	\end{table*}
	
	\begin{figure*}[t]
		\centering
		
		\begin{subfigure}[t]{0.16\linewidth}	
			\includegraphics[width=2.6cm,height=3cm]{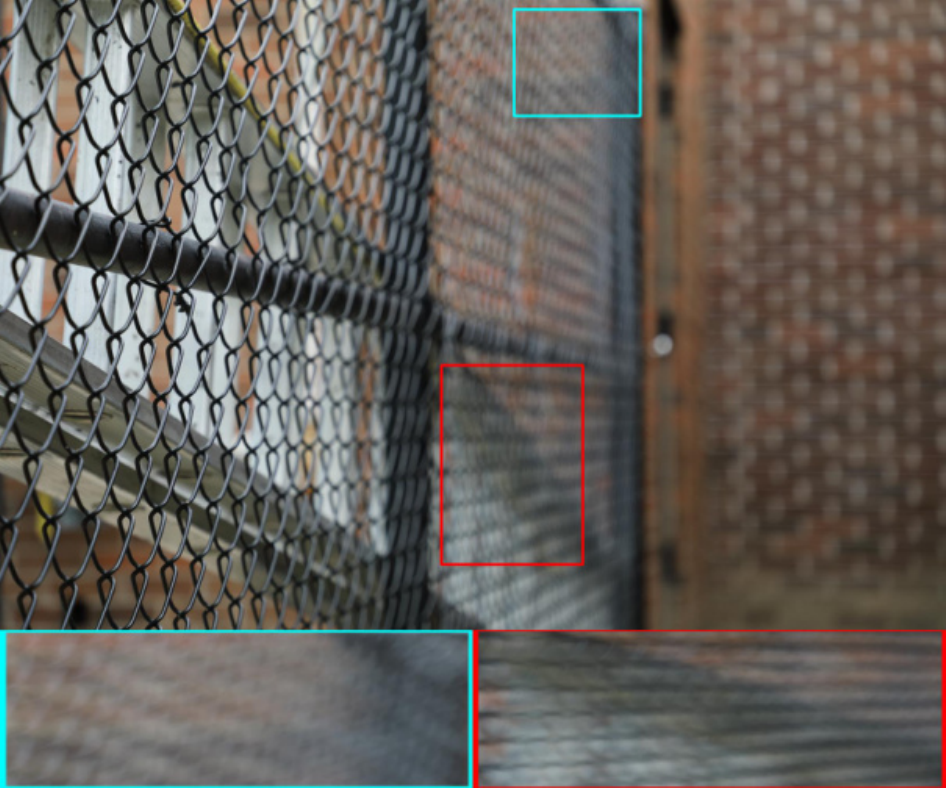}\vspace{4pt}
		\end{subfigure}
		\hfill
		\begin{subfigure}[t]{0.16\linewidth}	
			\includegraphics[width=2.6cm,height=3cm]{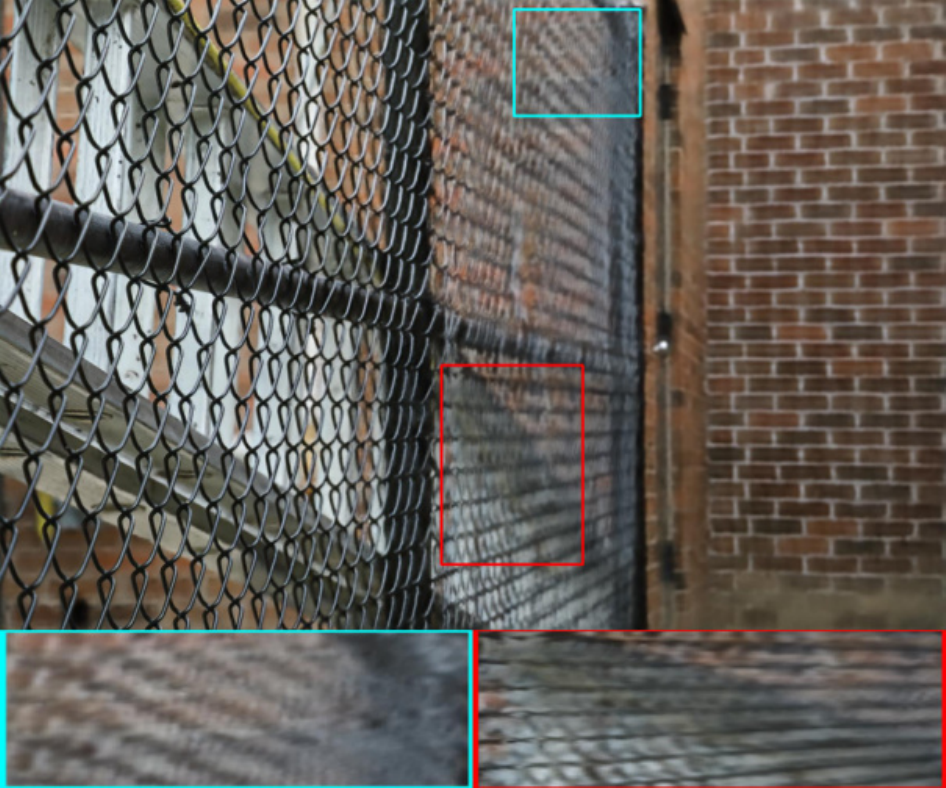}\vspace{4pt}
		\end{subfigure}
		\hfill
		\begin{subfigure}[t]{0.16\linewidth}	
			\includegraphics[width=2.6cm,height=3cm]{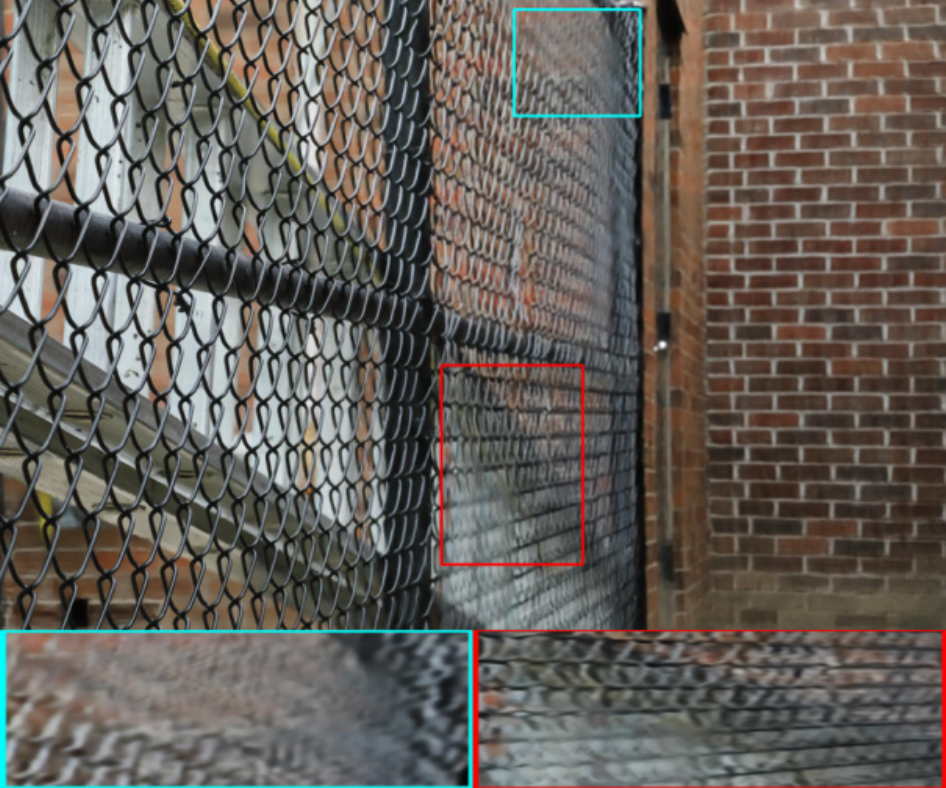}\vspace{4pt}
		\end{subfigure}
		\hfill
		\begin{subfigure}[t]{0.16\linewidth}	
			\includegraphics[width=2.6cm,height=3cm]{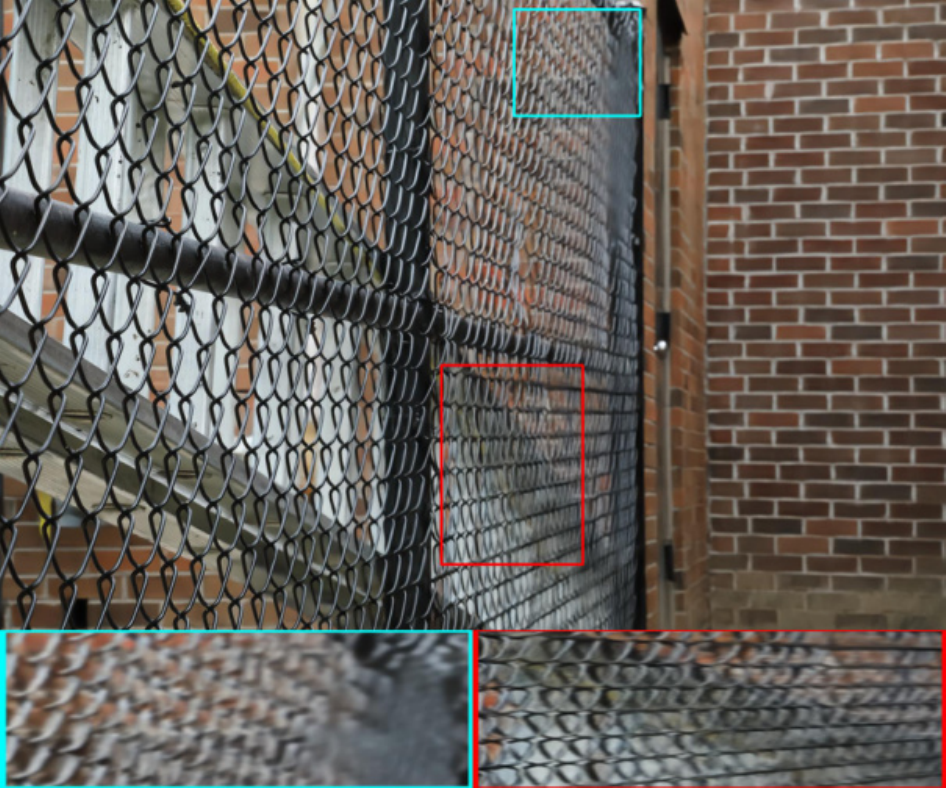}\vspace{4pt}
		\end{subfigure}
		\hfill
		\begin{subfigure}[t]{0.16\linewidth}	
			\includegraphics[width=2.6cm,height=3cm]{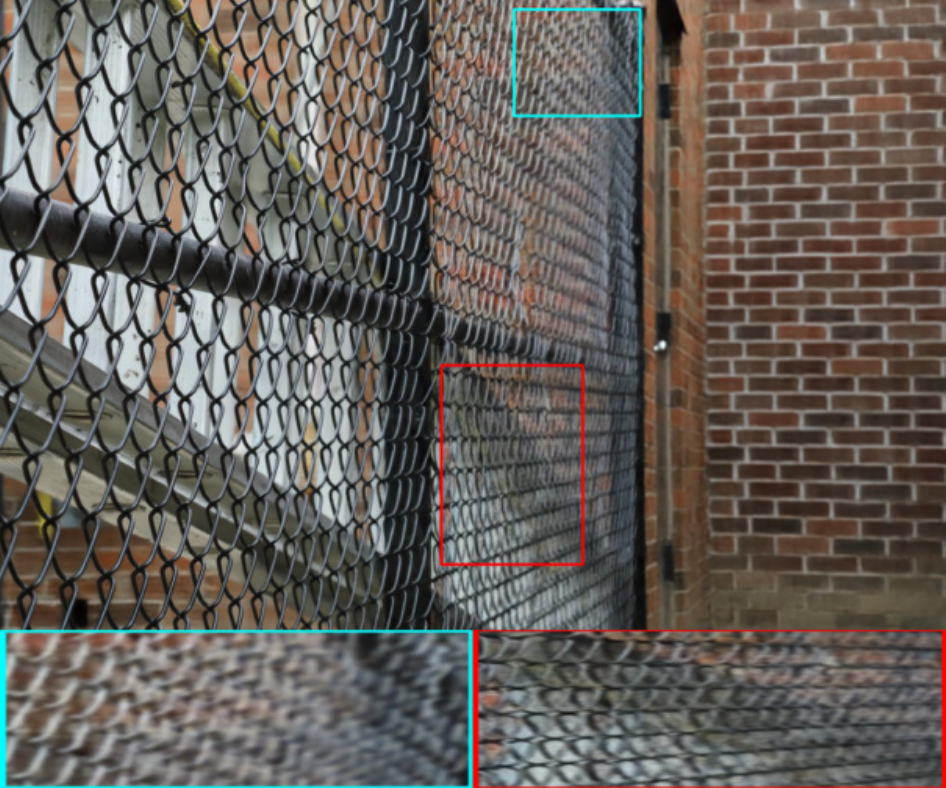}\vspace{4pt}
		\end{subfigure}
		\hfill
		\begin{subfigure}[t]{0.16\linewidth}	
			\includegraphics[width=2.6cm,height=3cm]{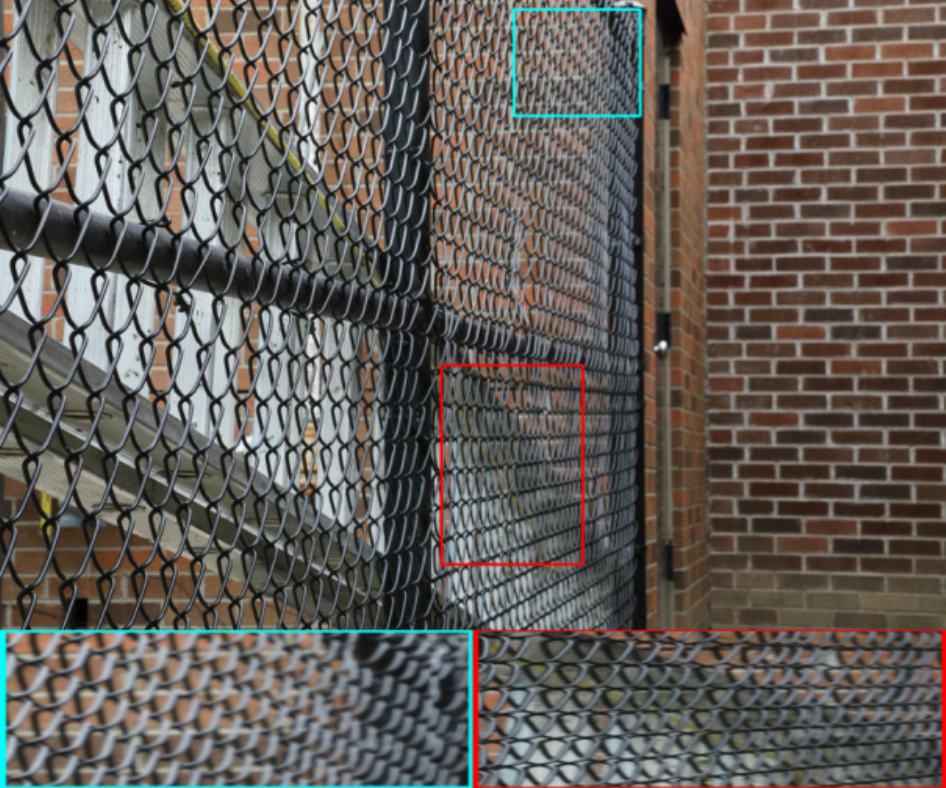}\vspace{4pt}
		\end{subfigure}
		\vfill
		
		\begin{subfigure}[t]{0.16\linewidth}	
			\includegraphics[width=2.6cm,height=3cm]{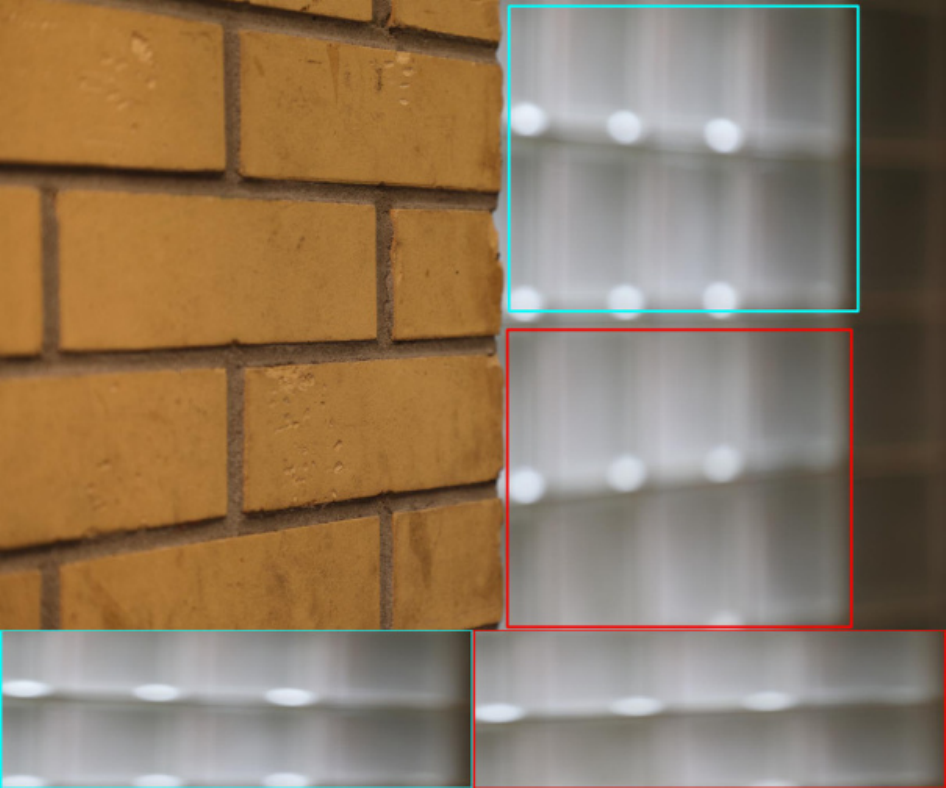}\vspace{4pt}
		\end{subfigure}
		\hfill
		\begin{subfigure}[t]{0.16\linewidth}	
			\includegraphics[width=2.6cm,height=3cm]{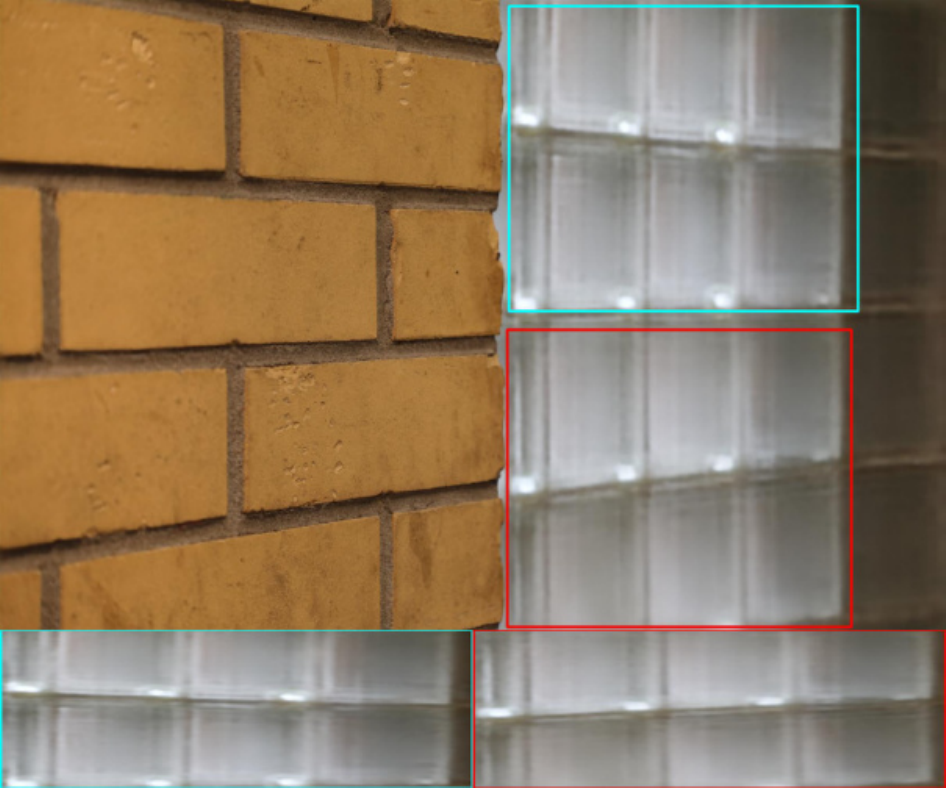}\vspace{4pt}
		\end{subfigure}
		\hfill
		\begin{subfigure}[t]{0.16\linewidth}	
			\includegraphics[width=2.6cm,height=3cm]{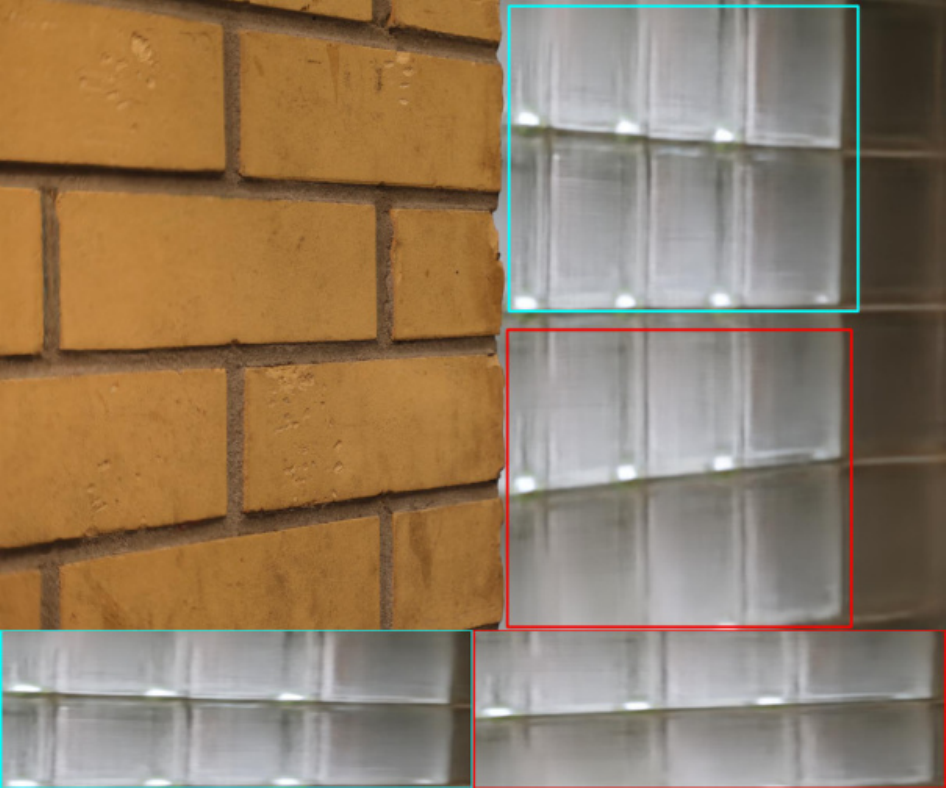}\vspace{4pt}
		\end{subfigure}
		\hfill
		\begin{subfigure}[t]{0.16\linewidth}	
			\includegraphics[width=2.6cm,height=3cm]{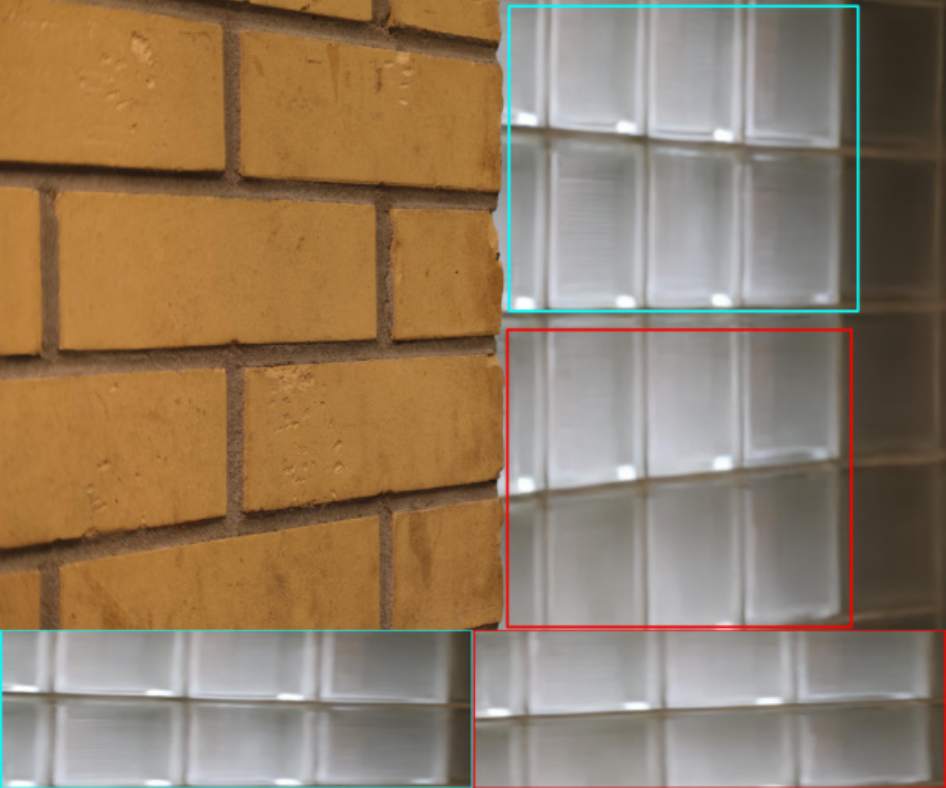}\vspace{4pt}
		\end{subfigure}
		\hfill
		\begin{subfigure}[t]{0.16\linewidth}	
			\includegraphics[width=2.6cm,height=3cm]{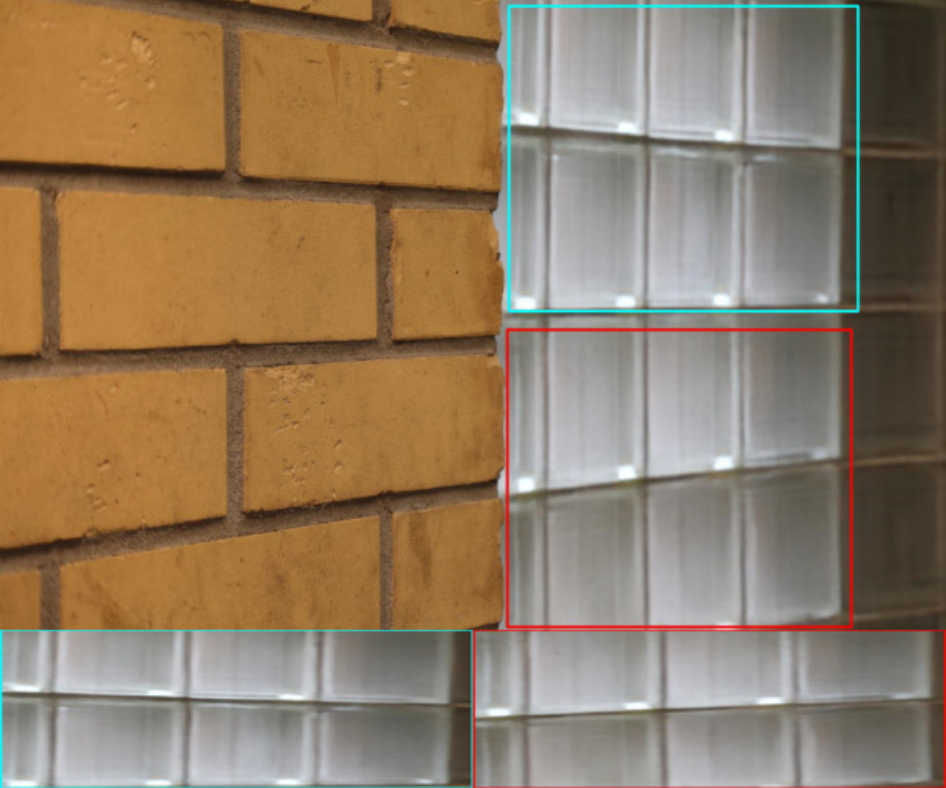}\vspace{4pt}
		\end{subfigure}
		\hfill
		\begin{subfigure}[t]{0.16\linewidth}	
			\includegraphics[width=2.6cm,height=3cm]{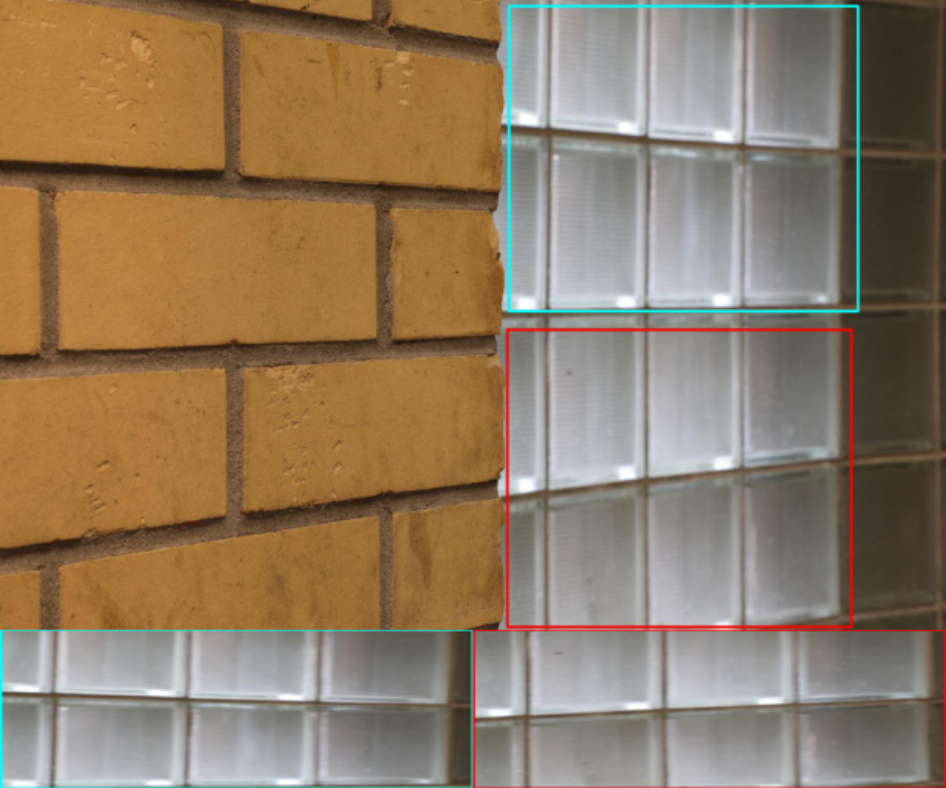}\vspace{4pt}
		\end{subfigure}
		\vfill
		
		\begin{subfigure}[t]{0.16\linewidth}	
			\includegraphics[width=2.6cm,height=3cm]{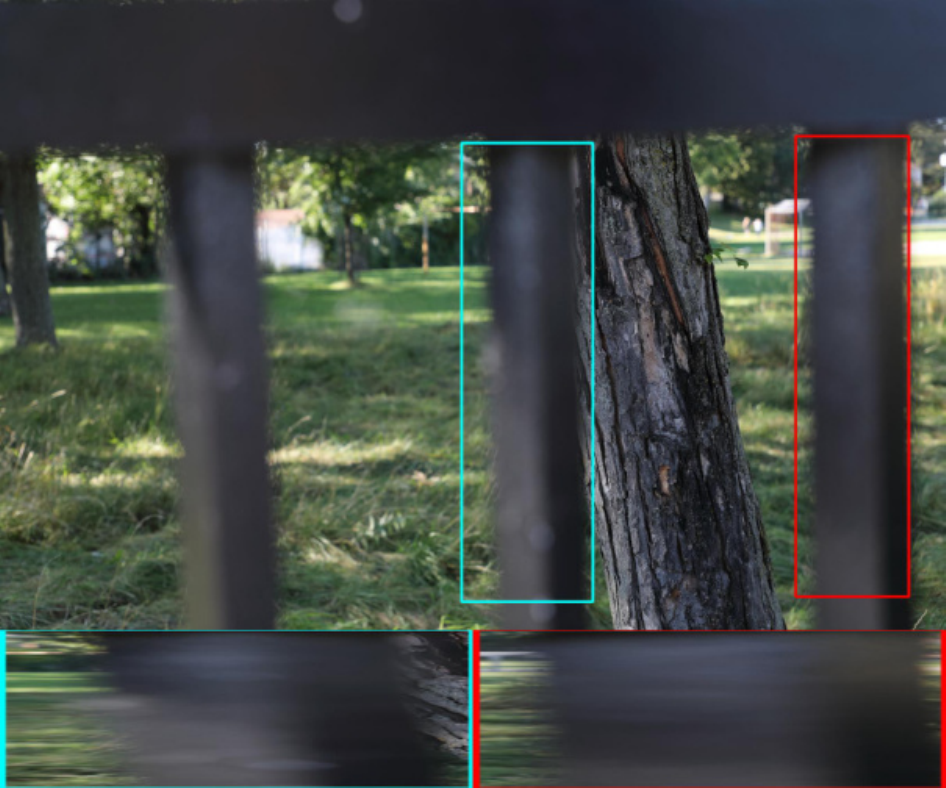}\vspace{4pt}
			\caption{Input}
		\end{subfigure}
		\hfill
		\begin{subfigure}[t]{0.16\linewidth}	
			\includegraphics[width=2.6cm,height=3cm]{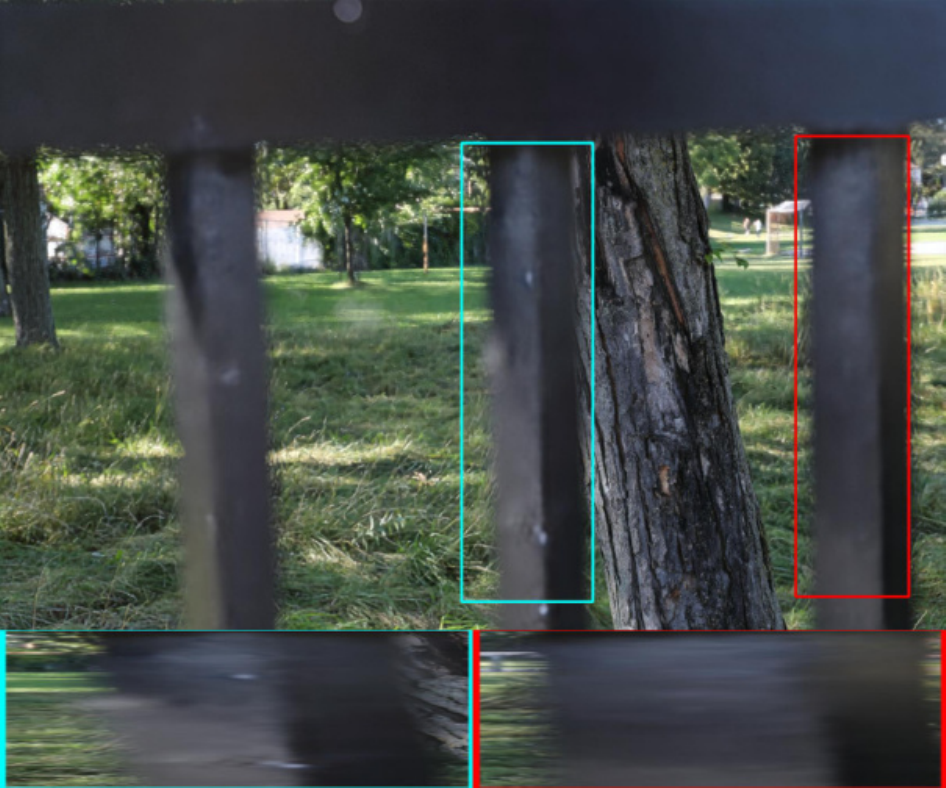}\vspace{4pt}
			\caption{GKMNet}
		\end{subfigure}
		\hfill
		\begin{subfigure}[t]{0.16\linewidth}	
			\includegraphics[width=2.6cm,height=3cm]{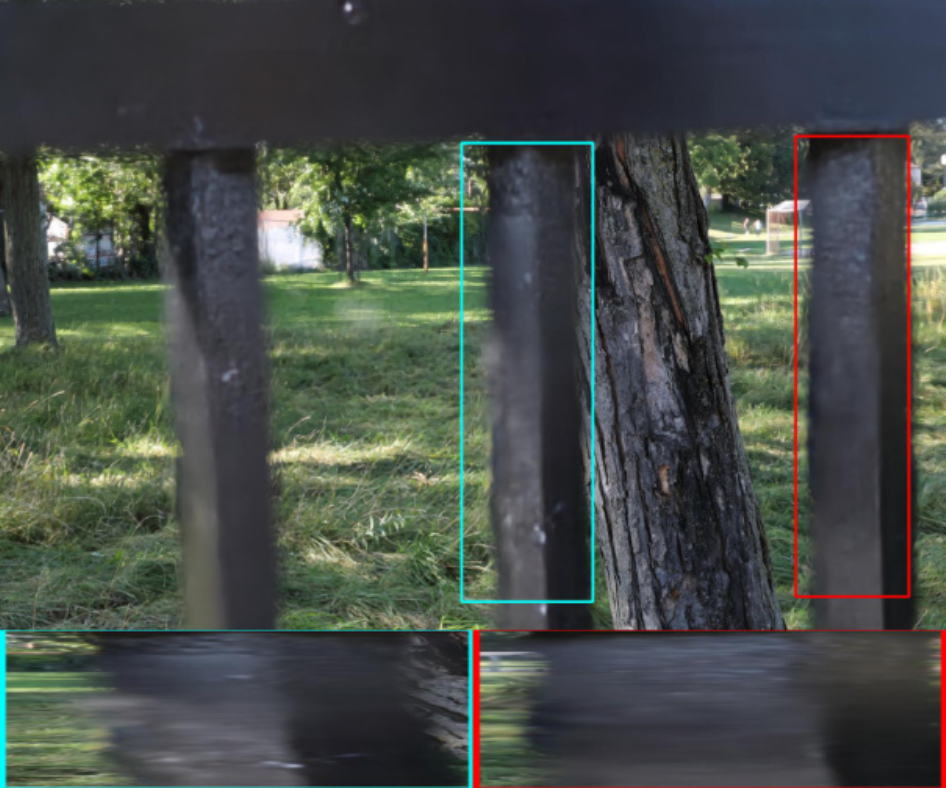}\vspace{4pt}
			\caption{IFAN}
		\end{subfigure}
		\hfill
		\begin{subfigure}[t]{0.16\linewidth}	
			\includegraphics[width=2.6cm,height=3cm]{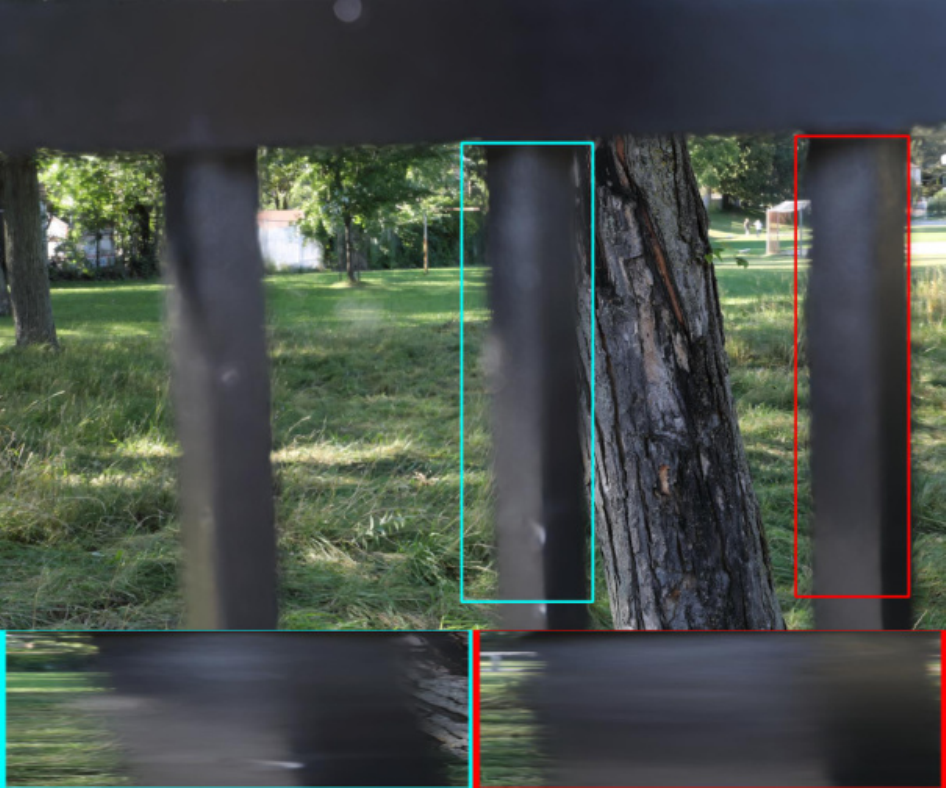}\vspace{4pt}
			\caption{Restormer}
		\end{subfigure}
		\hfill
		\begin{subfigure}[t]{0.16\linewidth}	
			\includegraphics[width=2.6cm,height=3cm]{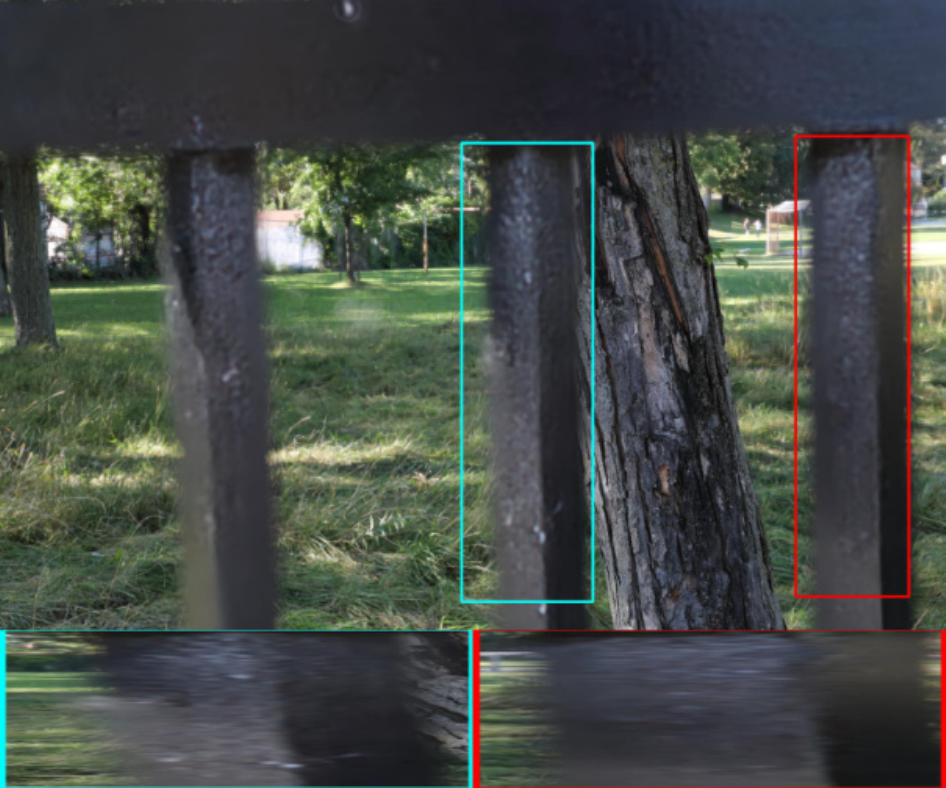}\vspace{4pt}
			\caption{SR-R$^2$KAC-B}
		\end{subfigure}
		\hfill
		\begin{subfigure}[t]{0.16\linewidth}	
			\includegraphics[width=2.6cm,height=3cm]{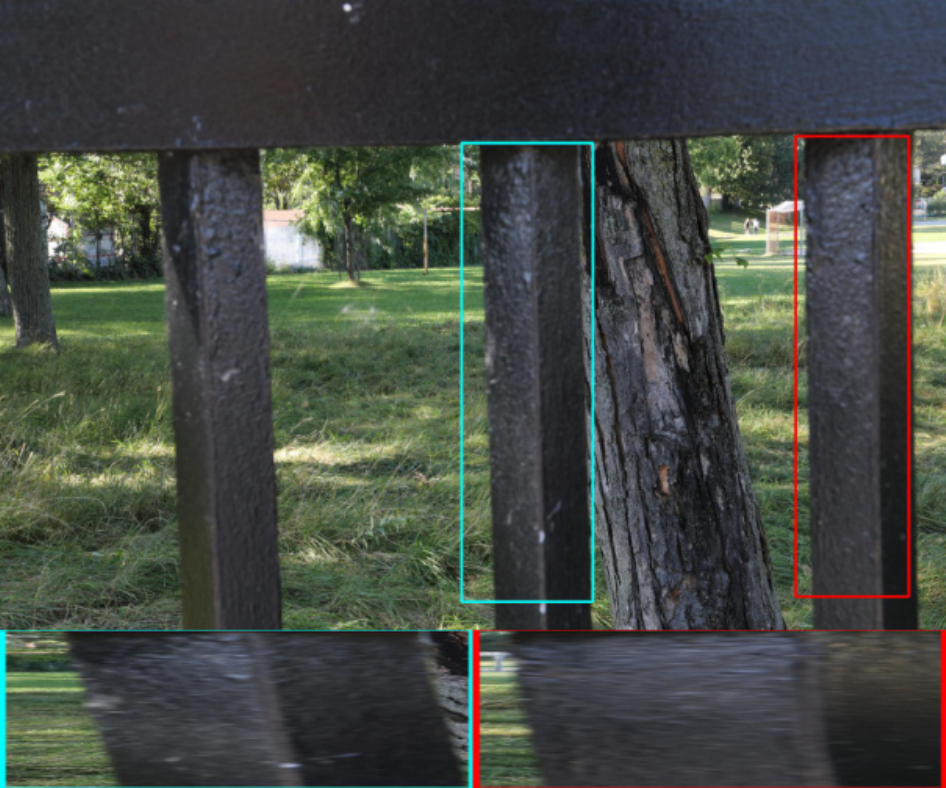}\vspace{4pt}
			\caption{Sharp}
		\end{subfigure}
		\vfill
		
		\caption{Visualization results of different defocus deblurring methods on DPDD dataset.}
		\label{fig_4}
	\end{figure*}
	
	\begin{figure*}[t]
		\centering
		
		\begin{subfigure}[t]{0.16\linewidth}	
			\includegraphics[width=2.6cm,height=3cm]{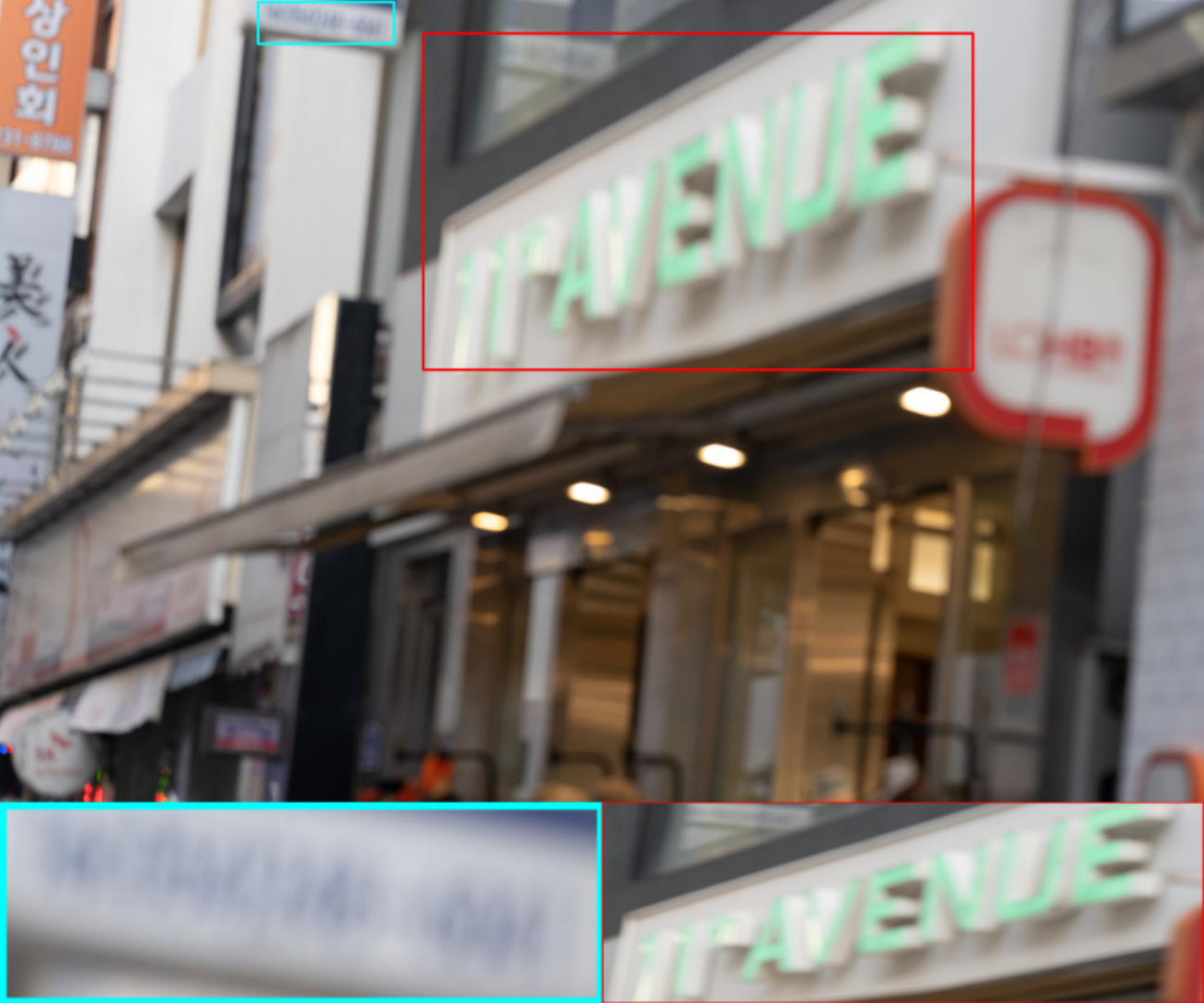}\vspace{4pt}
		\end{subfigure}
		\hfill
		\begin{subfigure}[t]{0.16\linewidth}	
			\includegraphics[width=2.6cm,height=3cm]{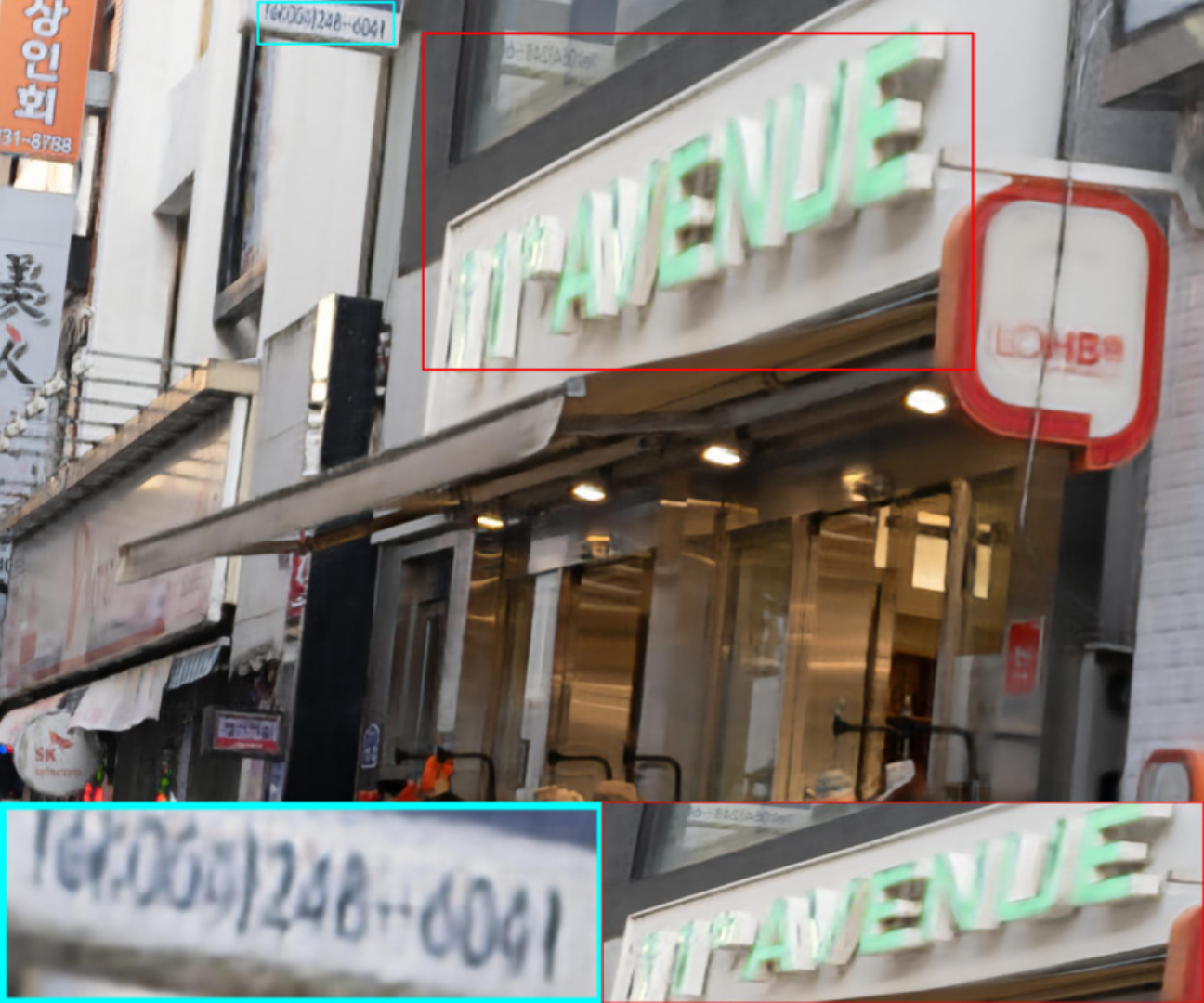}\vspace{4pt}
		\end{subfigure}
		\hfill
		\begin{subfigure}[t]{0.16\linewidth}	
			\includegraphics[width=2.6cm,height=3cm]{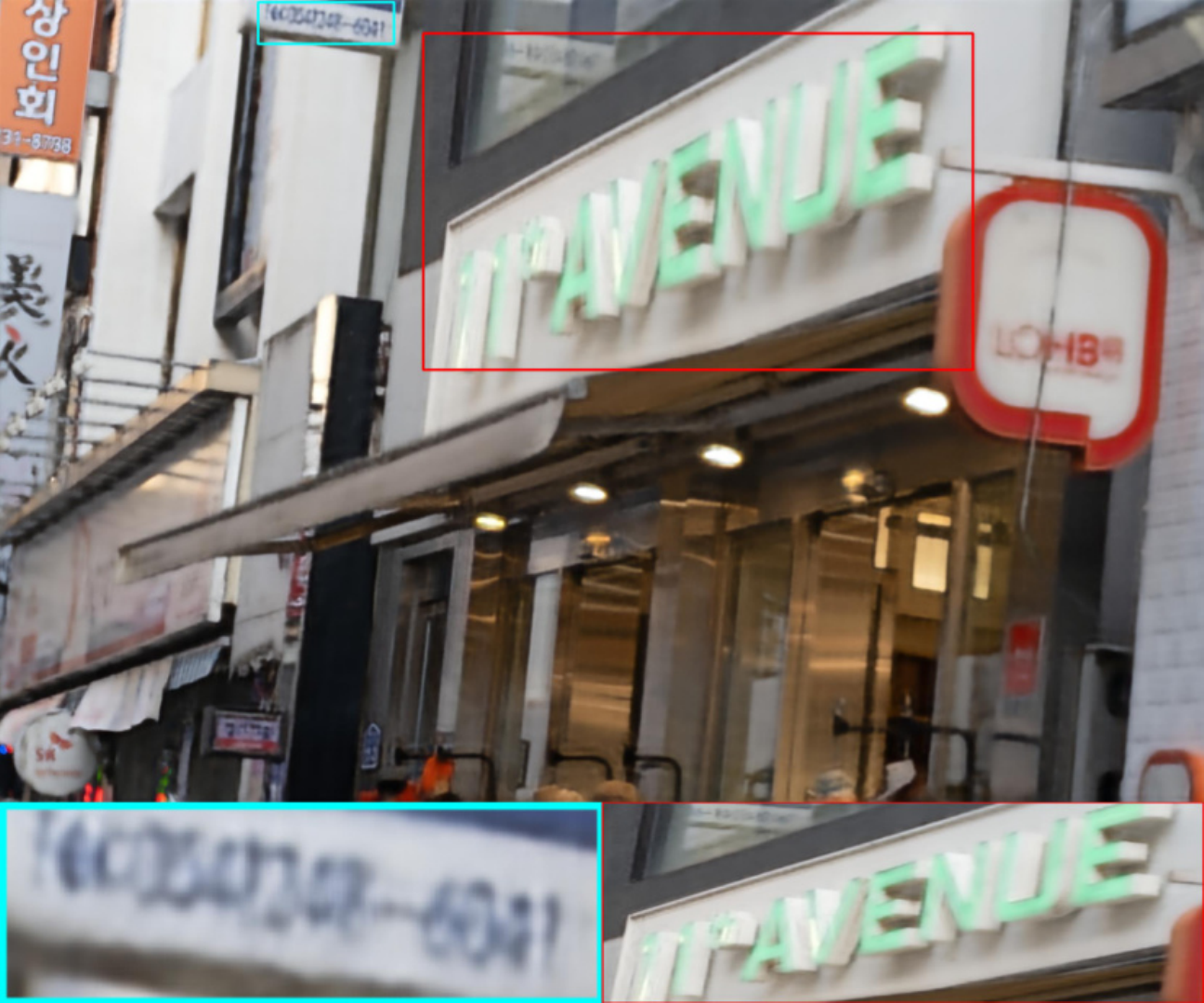}\vspace{4pt}
		\end{subfigure}
		\hfill
		\begin{subfigure}[t]{0.16\linewidth}	
			\includegraphics[width=2.6cm,height=3cm]{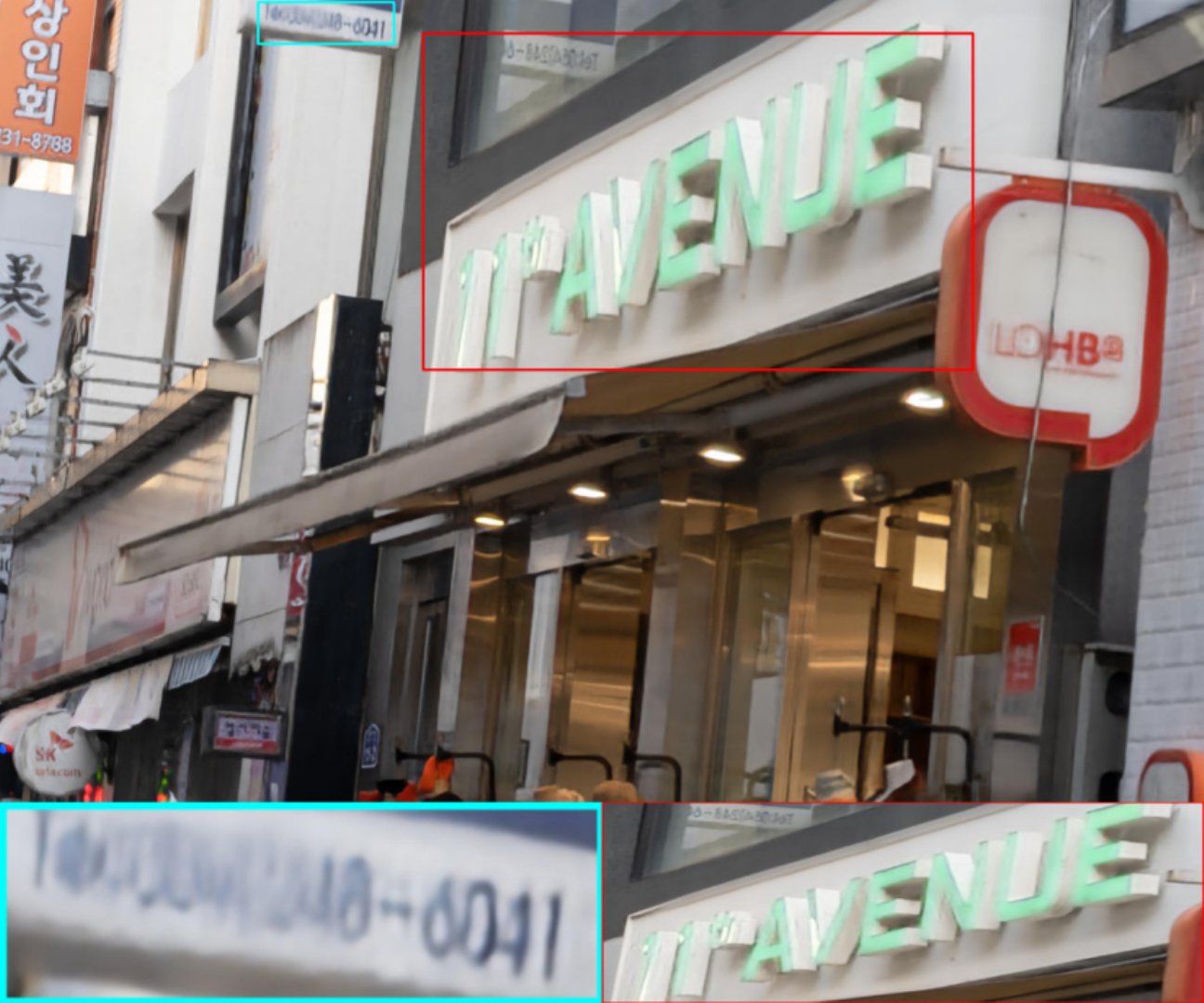}\vspace{4pt}
		\end{subfigure}
		\hfill
		\begin{subfigure}[t]{0.16\linewidth}	
			\includegraphics[width=2.6cm,height=3cm]{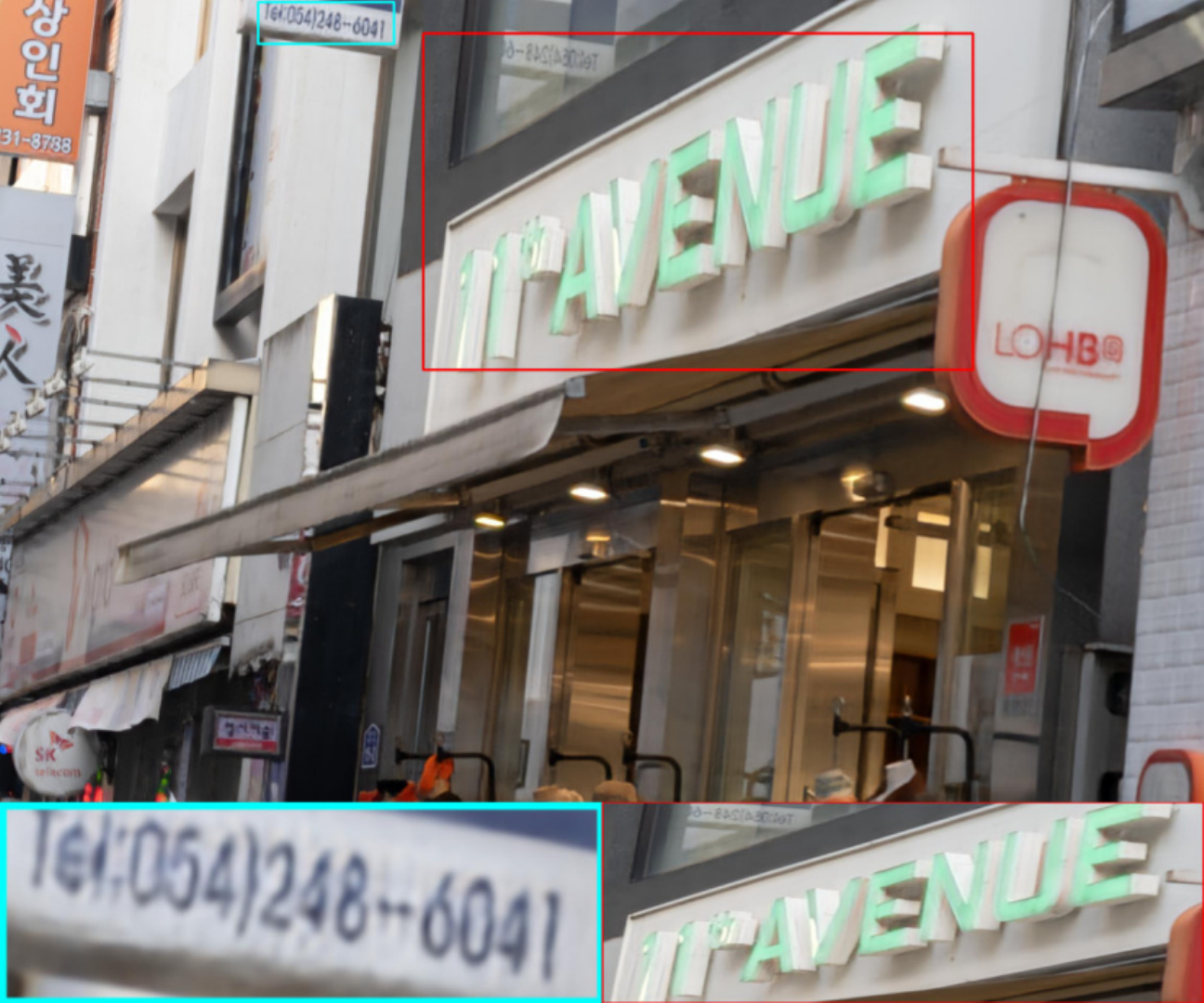}\vspace{4pt}
		\end{subfigure}
		\hfill
		\begin{subfigure}[t]{0.16\linewidth}	
			\includegraphics[width=2.6cm,height=3cm]{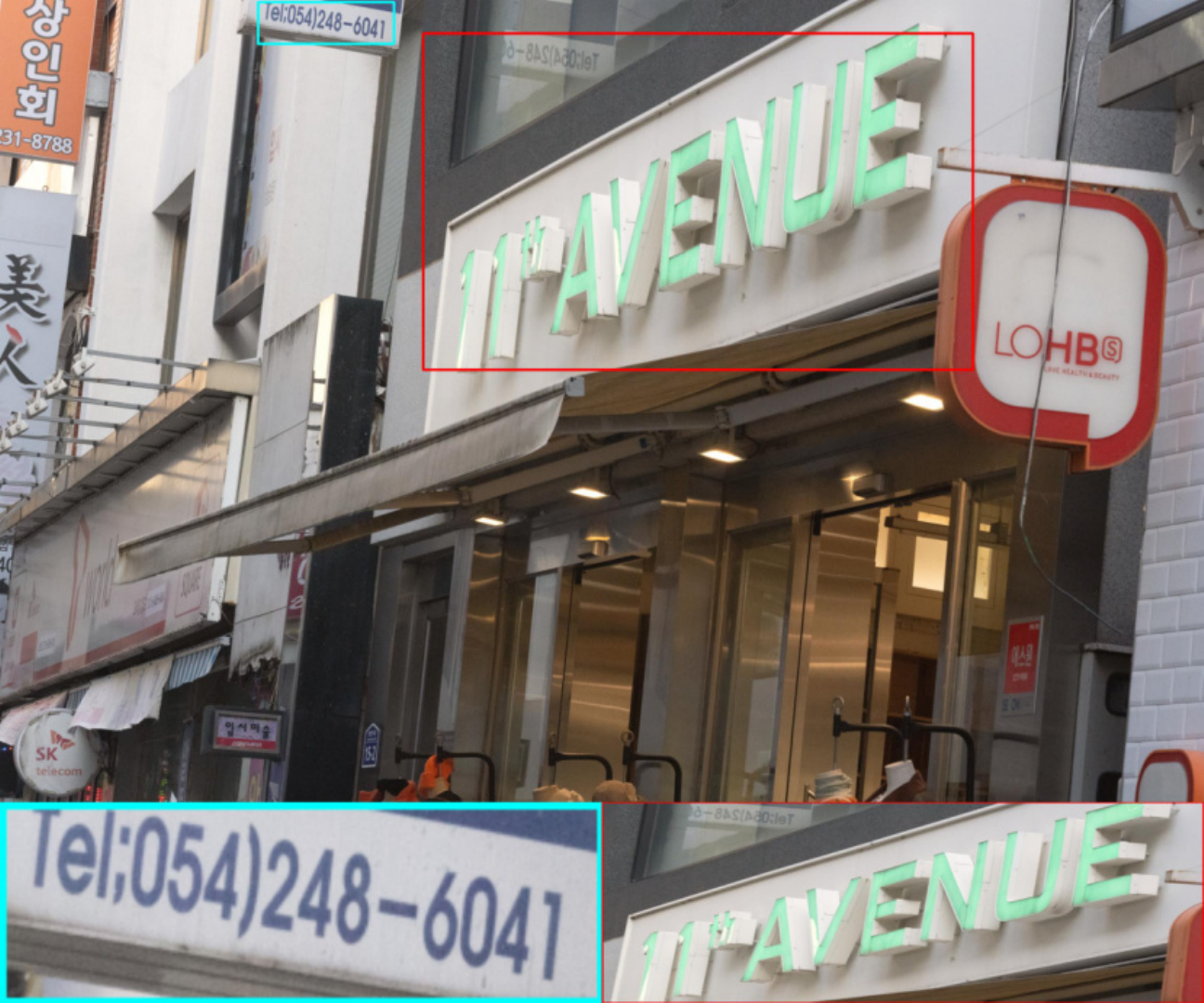}\vspace{4pt}
		\end{subfigure}
		\vfill
		
		\begin{subfigure}[t]{0.16\linewidth}	
			\includegraphics[width=2.6cm,height=3cm]{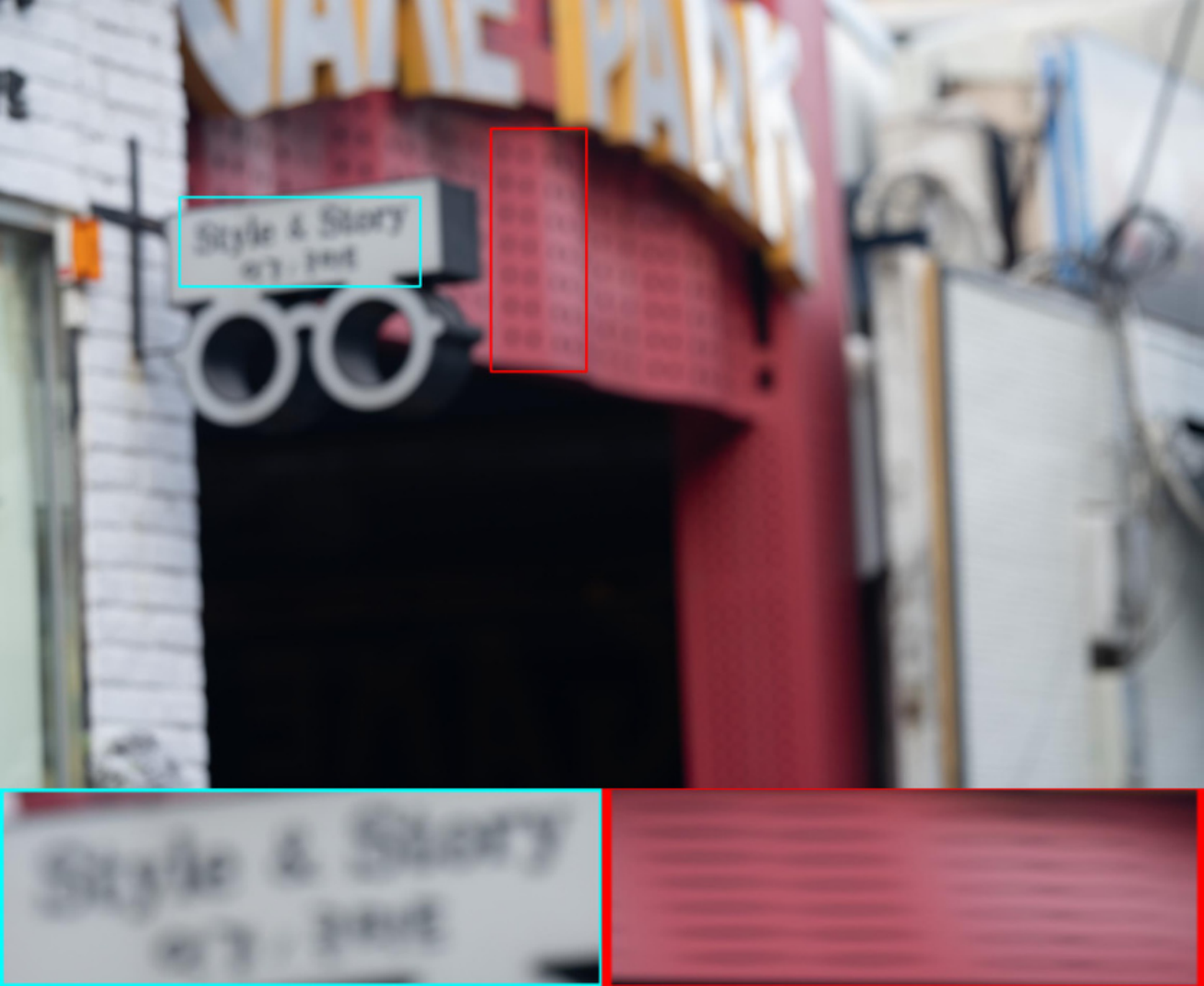}\vspace{4pt}
		\end{subfigure}
		\hfill
		\begin{subfigure}[t]{0.16\linewidth}	
			\includegraphics[width=2.6cm,height=3cm]{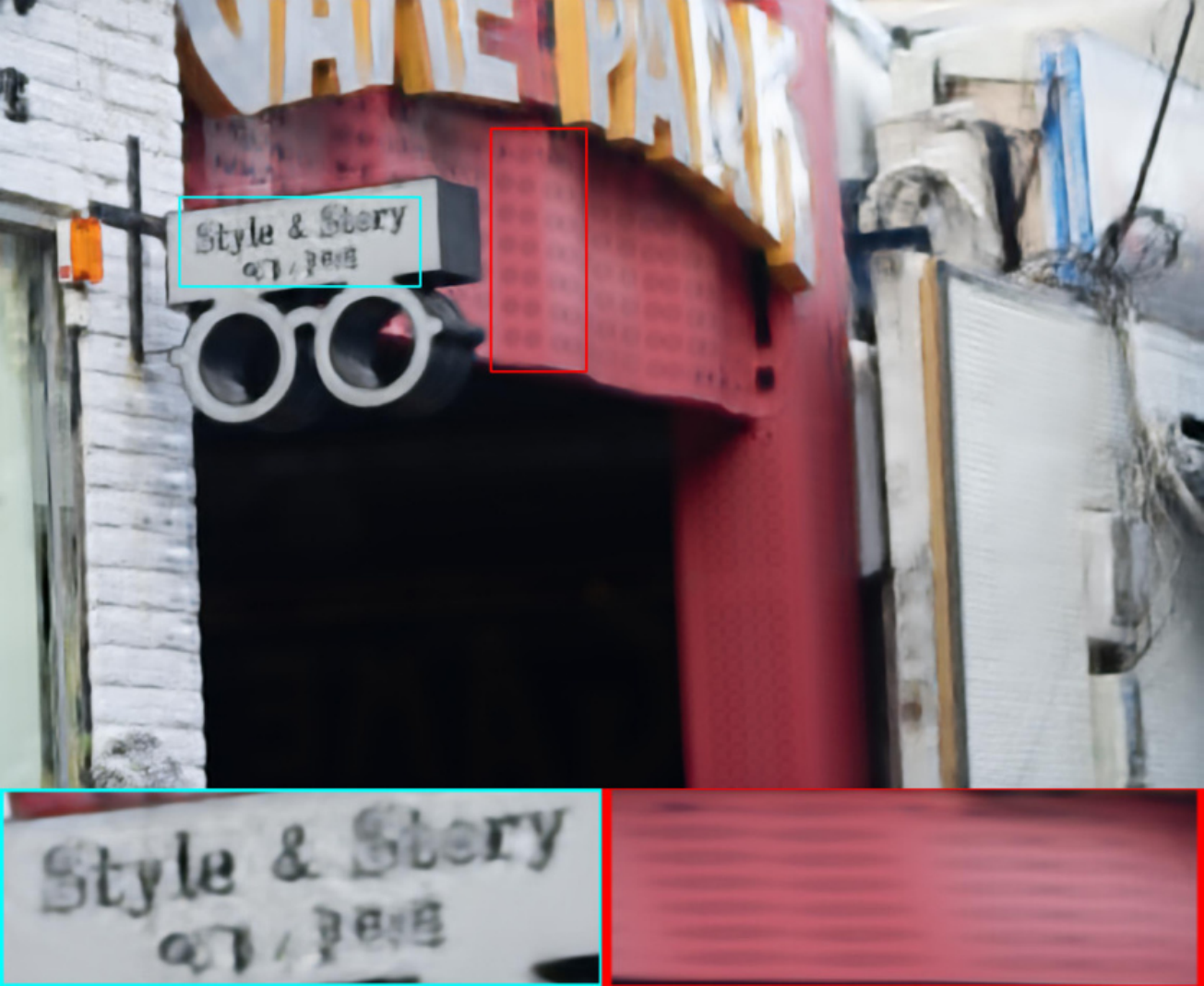}\vspace{4pt}
		\end{subfigure}
		\hfill
		\begin{subfigure}[t]{0.16\linewidth}	
			\includegraphics[width=2.6cm,height=3cm]{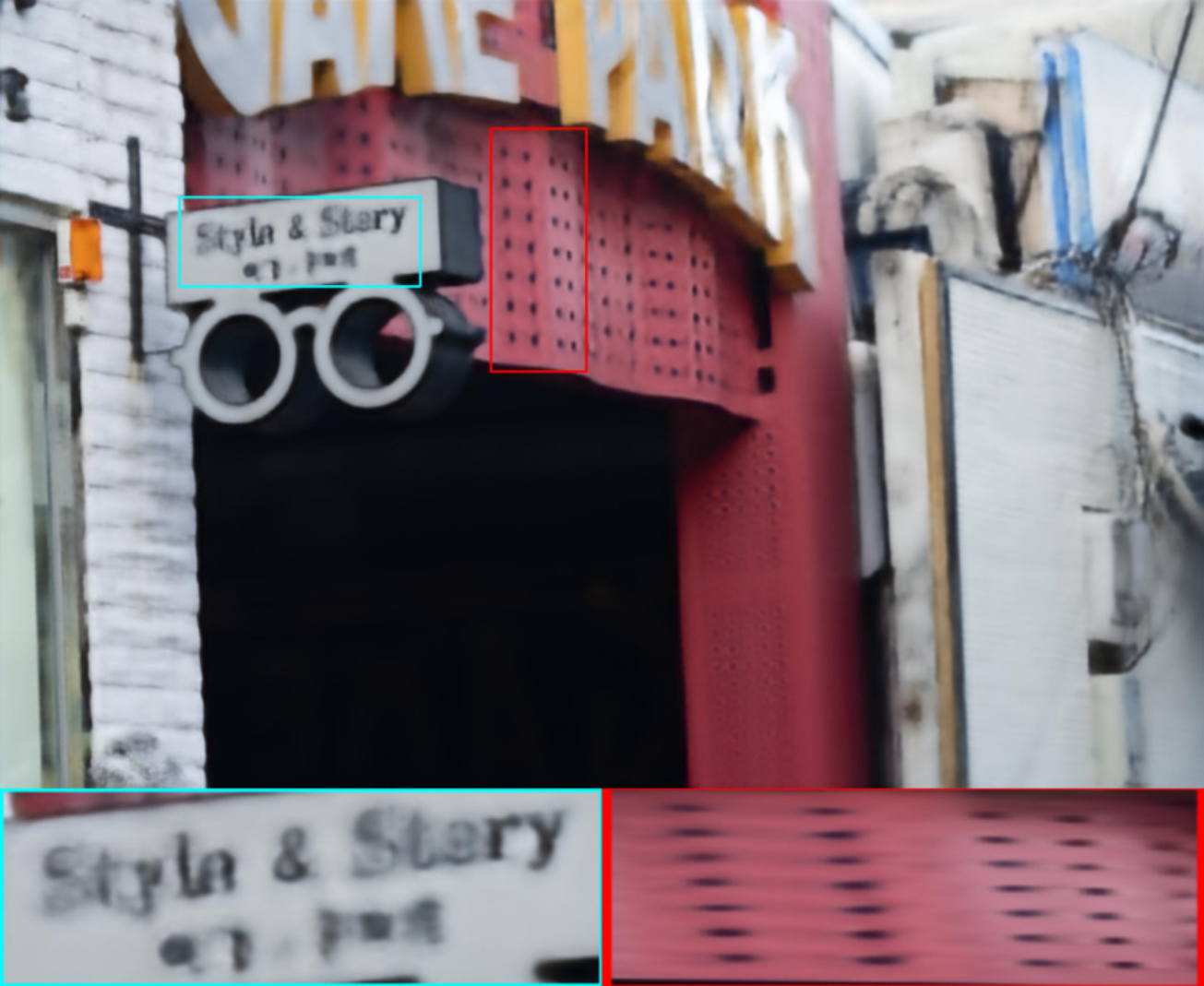}\vspace{4pt}
		\end{subfigure}
		\hfill
		\begin{subfigure}[t]{0.16\linewidth}	
			\includegraphics[width=2.6cm,height=3cm]{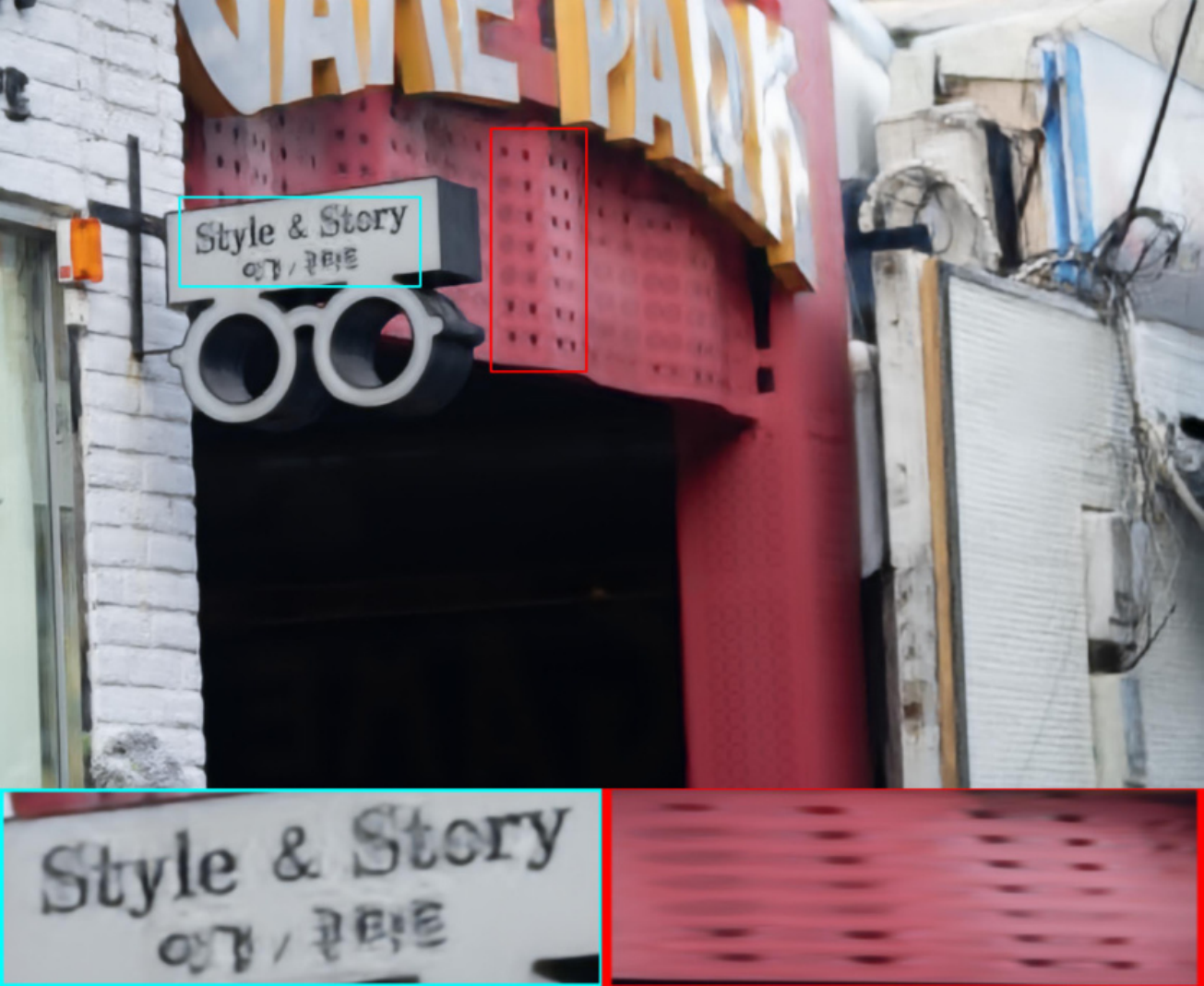}\vspace{4pt}
		\end{subfigure}
		\hfill
		\begin{subfigure}[t]{0.16\linewidth}	
			\includegraphics[width=2.6cm,height=3cm]{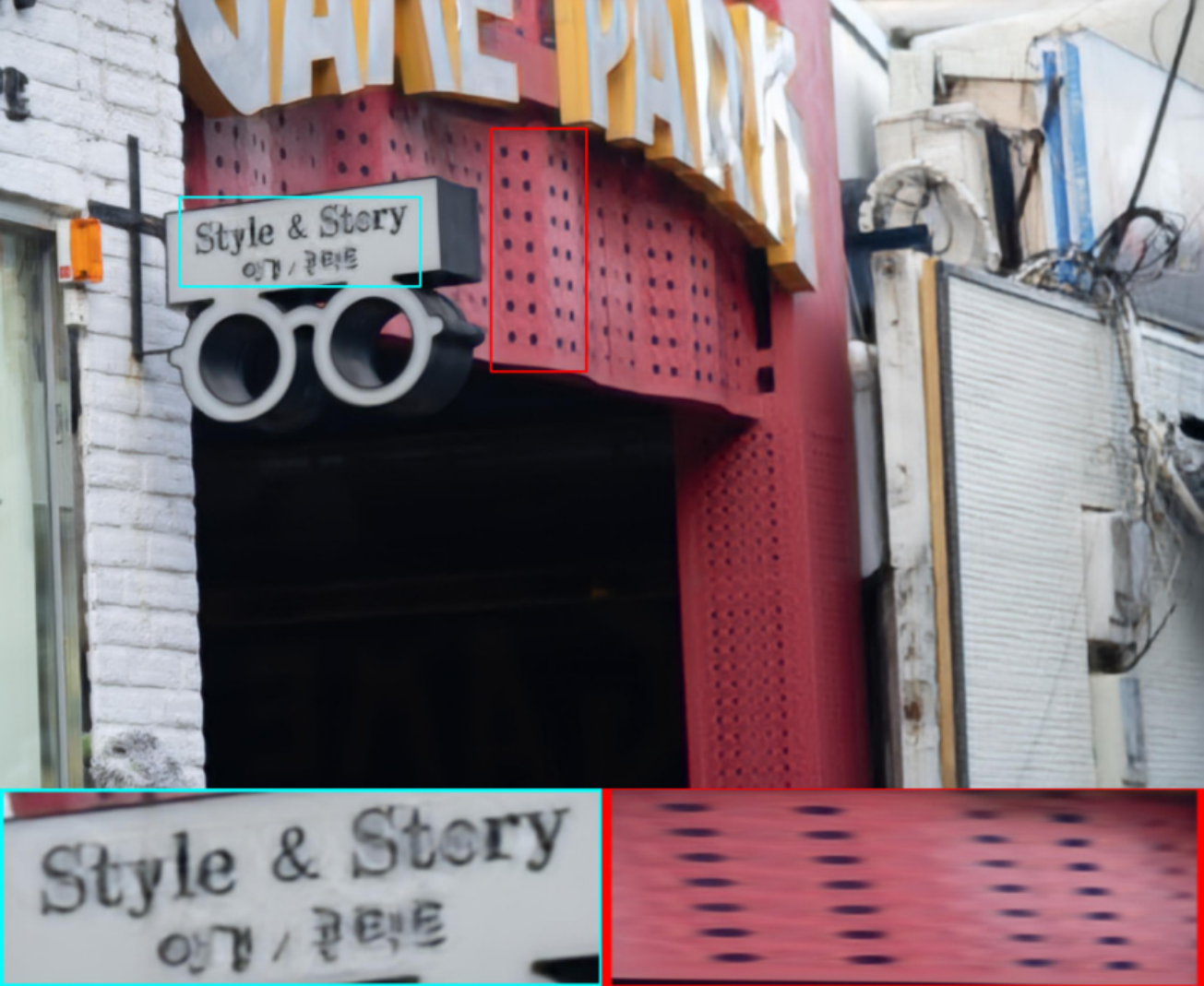}\vspace{4pt}
		\end{subfigure}
		\hfill
		\begin{subfigure}[t]{0.16\linewidth}	
			\includegraphics[width=2.6cm,height=3cm]{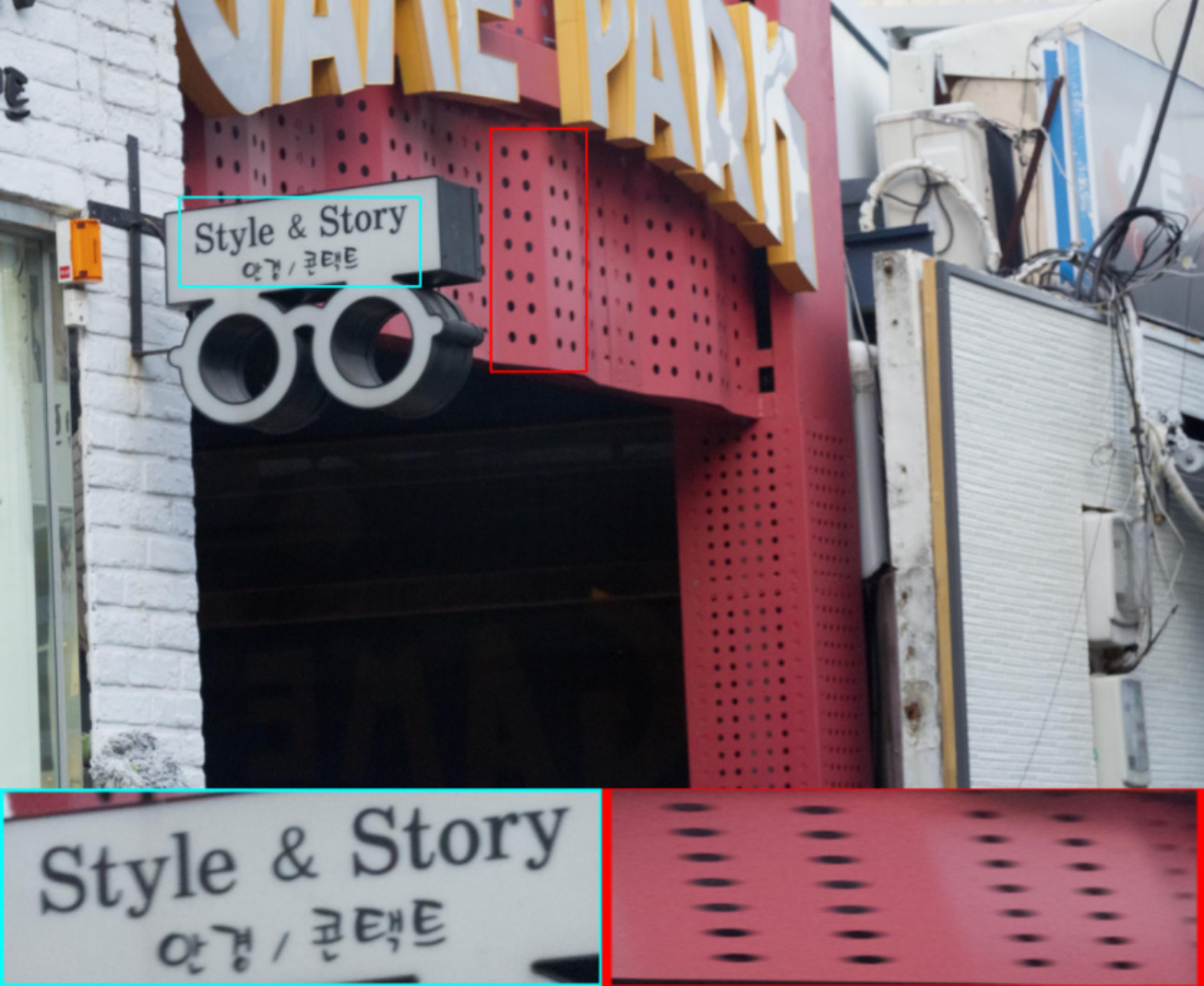}\vspace{4pt}
		\end{subfigure}
		\vfill
		
		\begin{subfigure}[t]{0.16\linewidth}	
			\includegraphics[width=2.6cm,height=3cm]{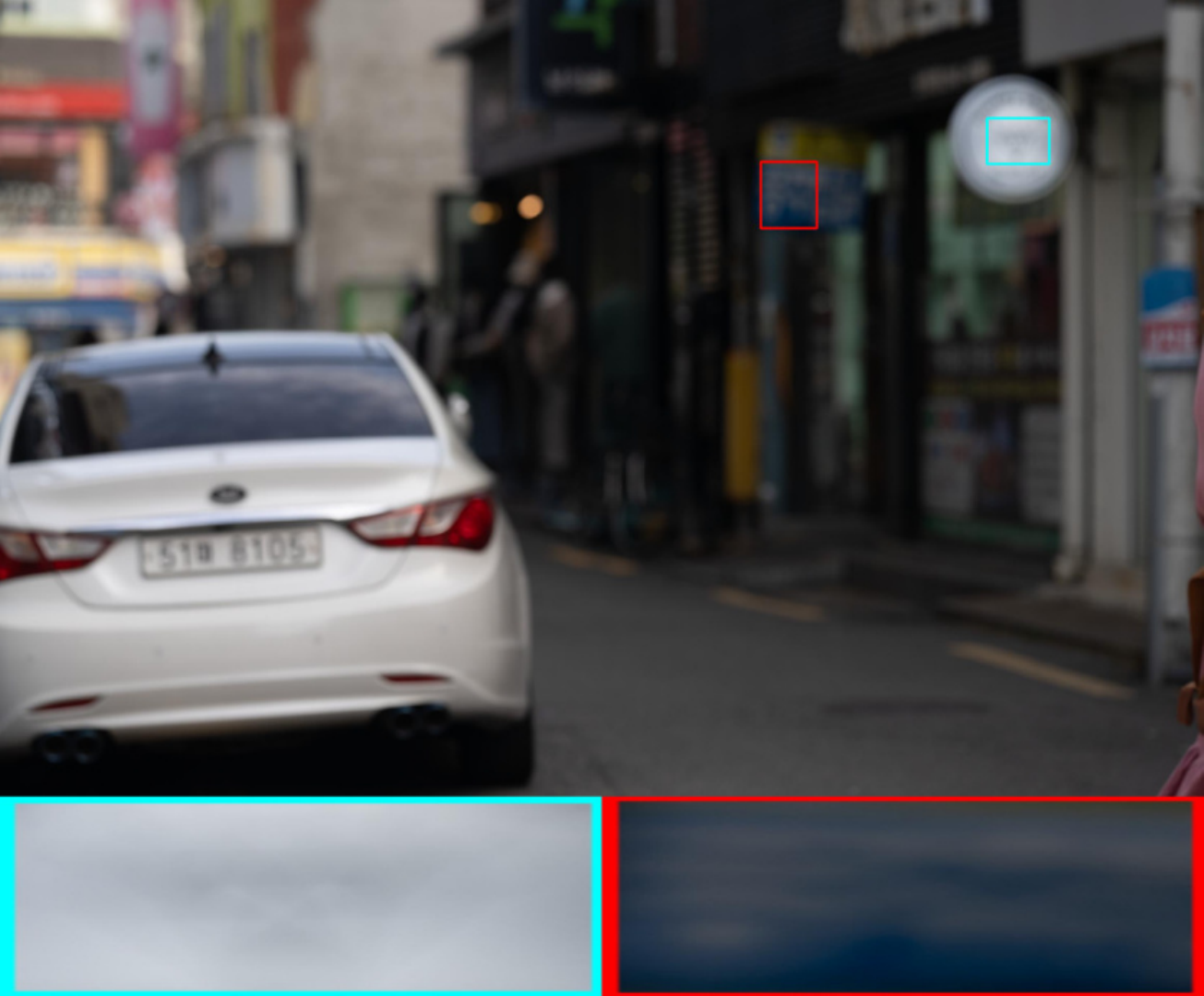}\vspace{4pt}
			\caption{Input}
		\end{subfigure}
		\hfill
		\begin{subfigure}[t]{0.16\linewidth}	
			\includegraphics[width=2.6cm,height=3cm]{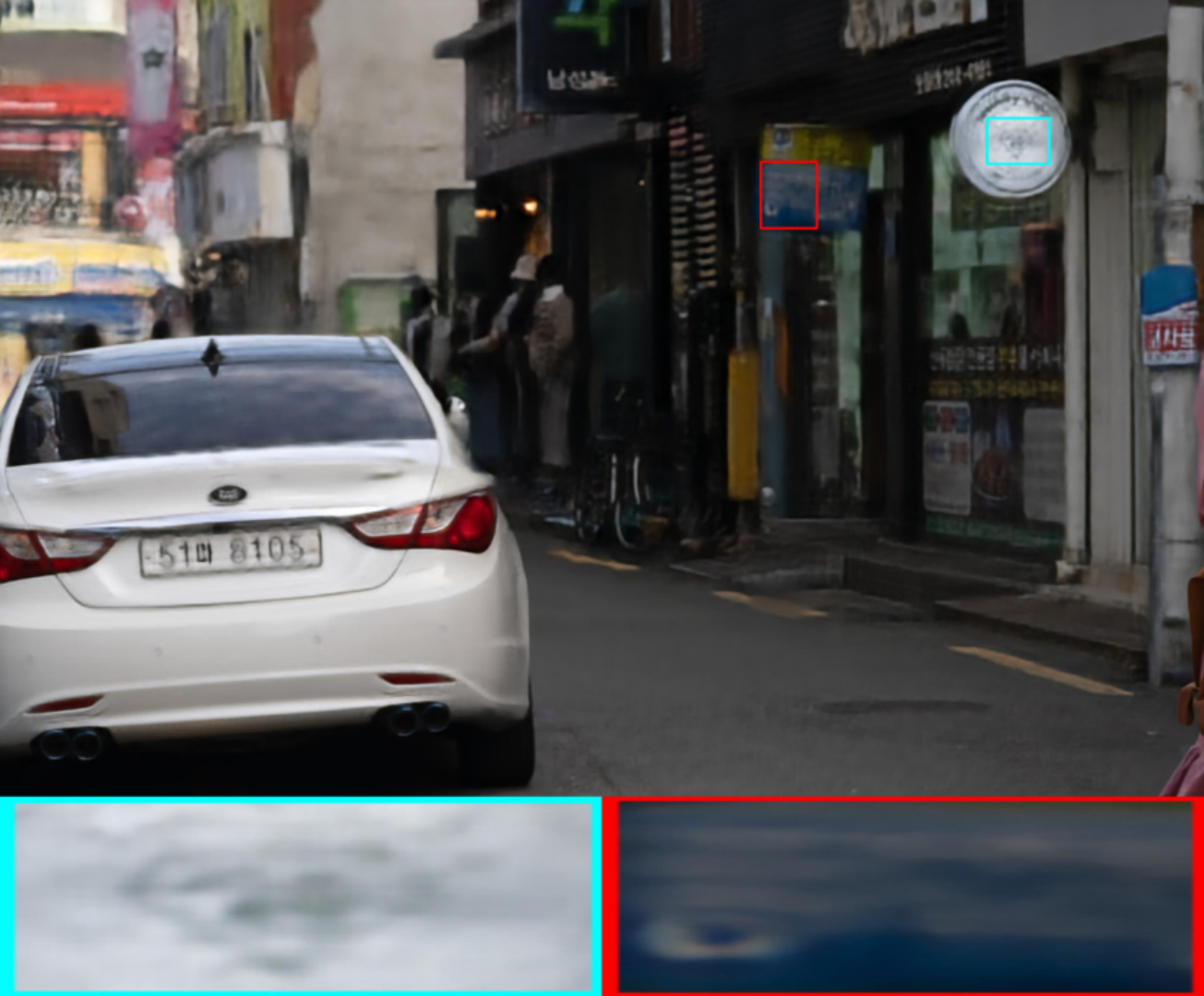}\vspace{4pt}
			\caption{KPAC}
		\end{subfigure}
		\hfill
		\begin{subfigure}[t]{0.16\linewidth}	
			\includegraphics[width=2.6cm,height=3cm]{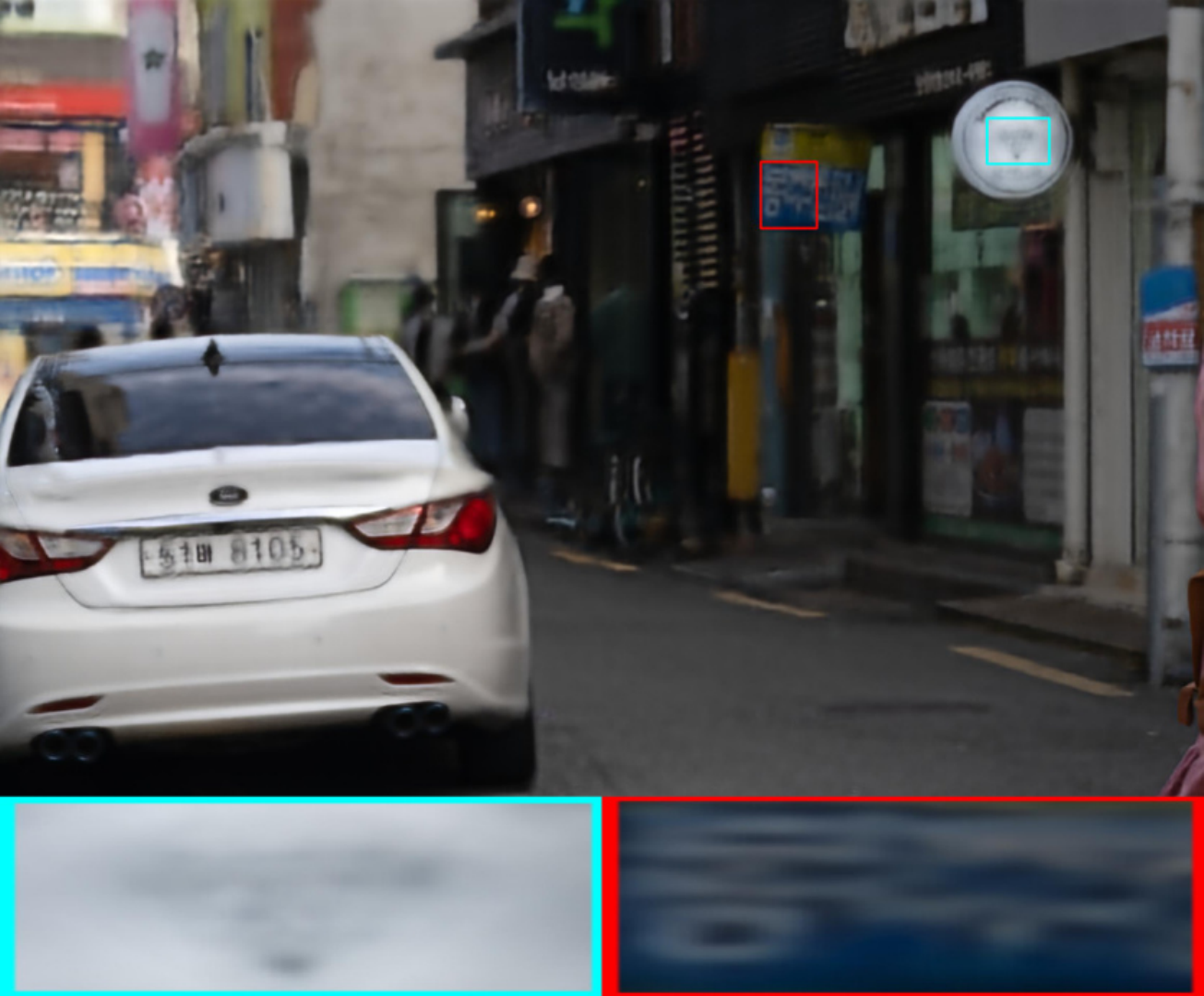}\vspace{4pt}
			\caption{GKMNet}
		\end{subfigure}
		\hfill
		\begin{subfigure}[t]{0.16\linewidth}	
			\includegraphics[width=2.6cm,height=3cm]{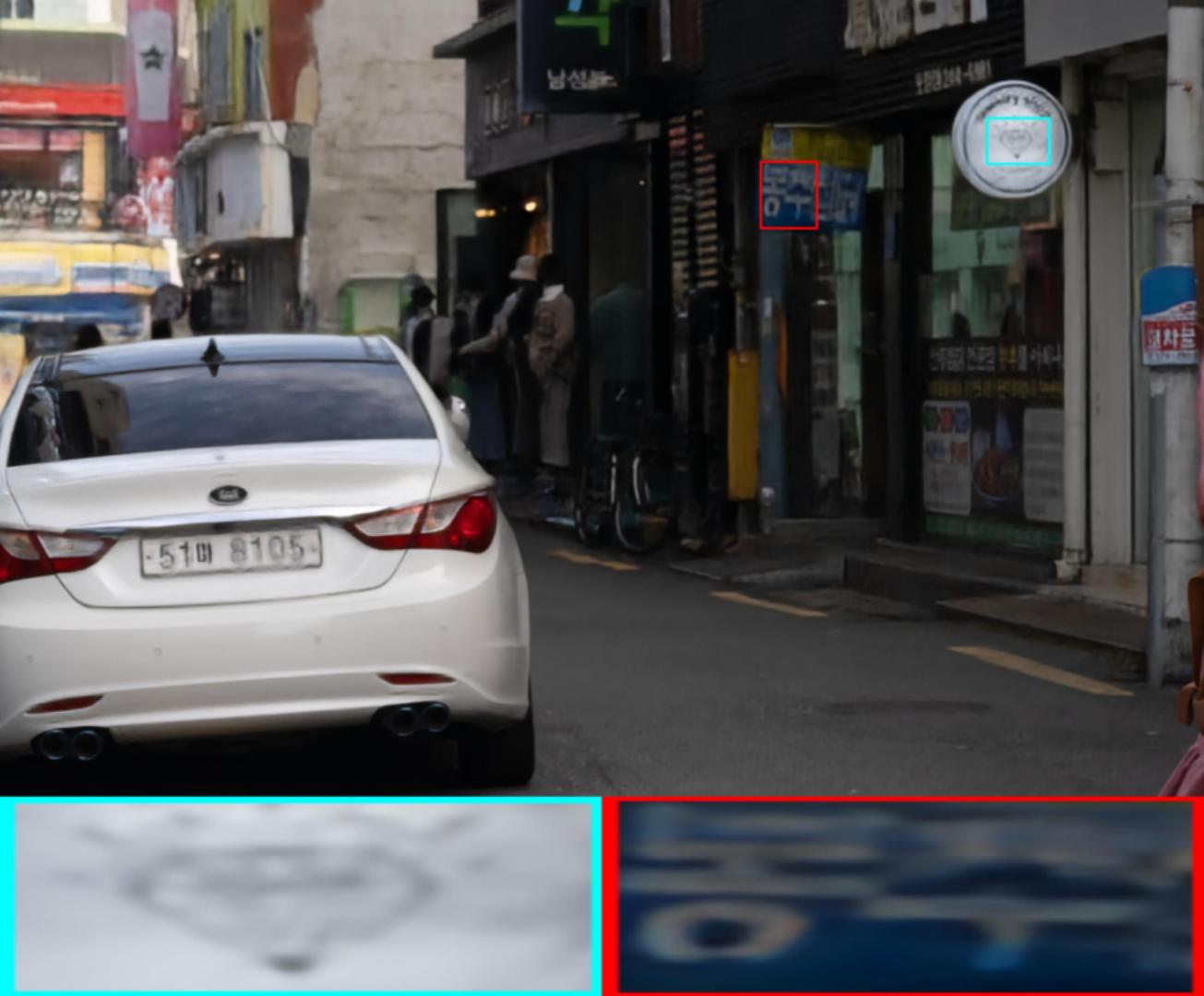}\vspace{4pt}
			\caption{IFAN}
		\end{subfigure}
		\hfill
		\begin{subfigure}[t]{0.16\linewidth}	
			\includegraphics[width=2.6cm,height=3cm]{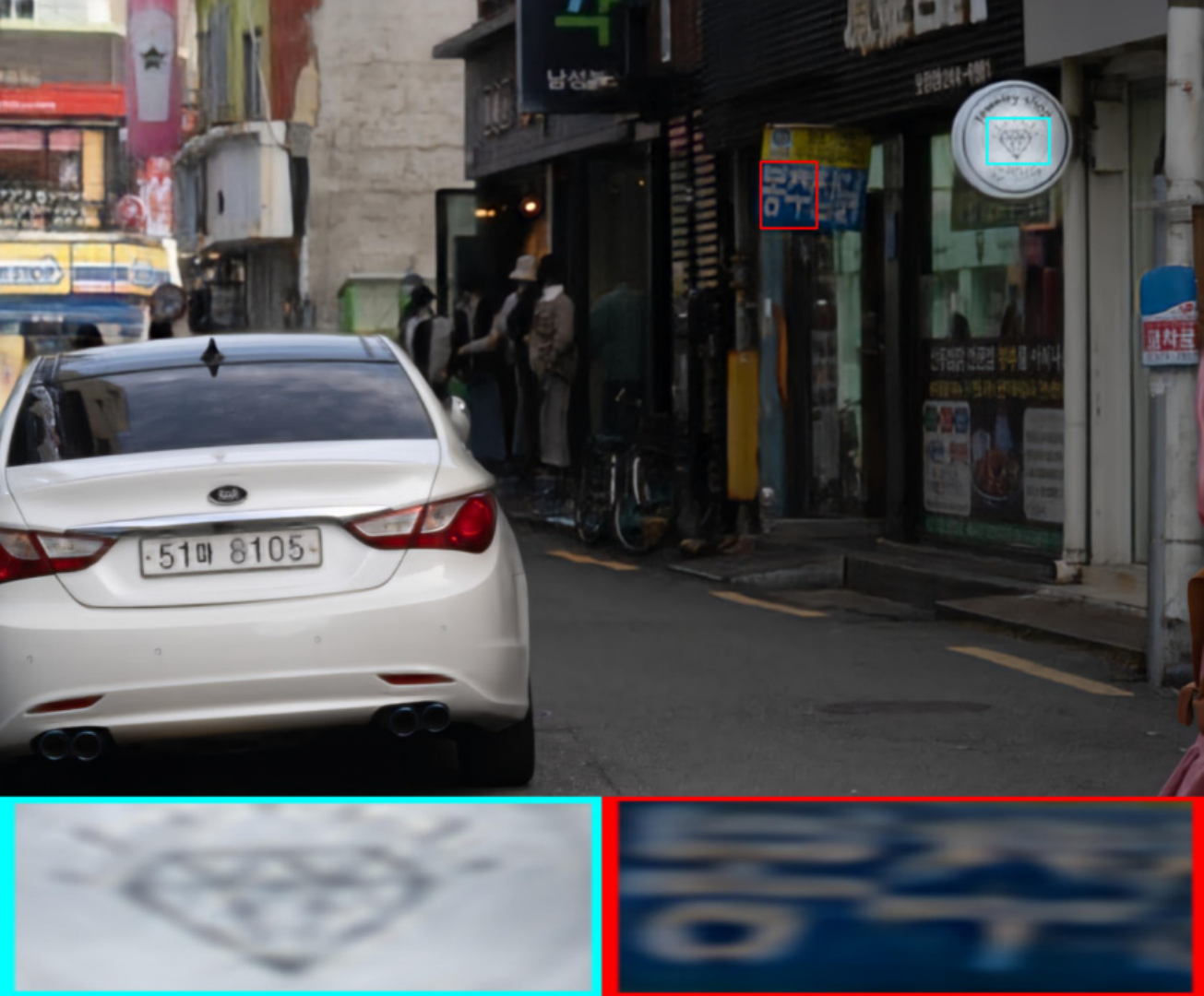}\vspace{4pt}
			\caption{SR-R$^2$KAC-B}
		\end{subfigure}
		\hfill
		\begin{subfigure}[t]{0.16\linewidth}	
			\includegraphics[width=2.6cm,height=3cm]{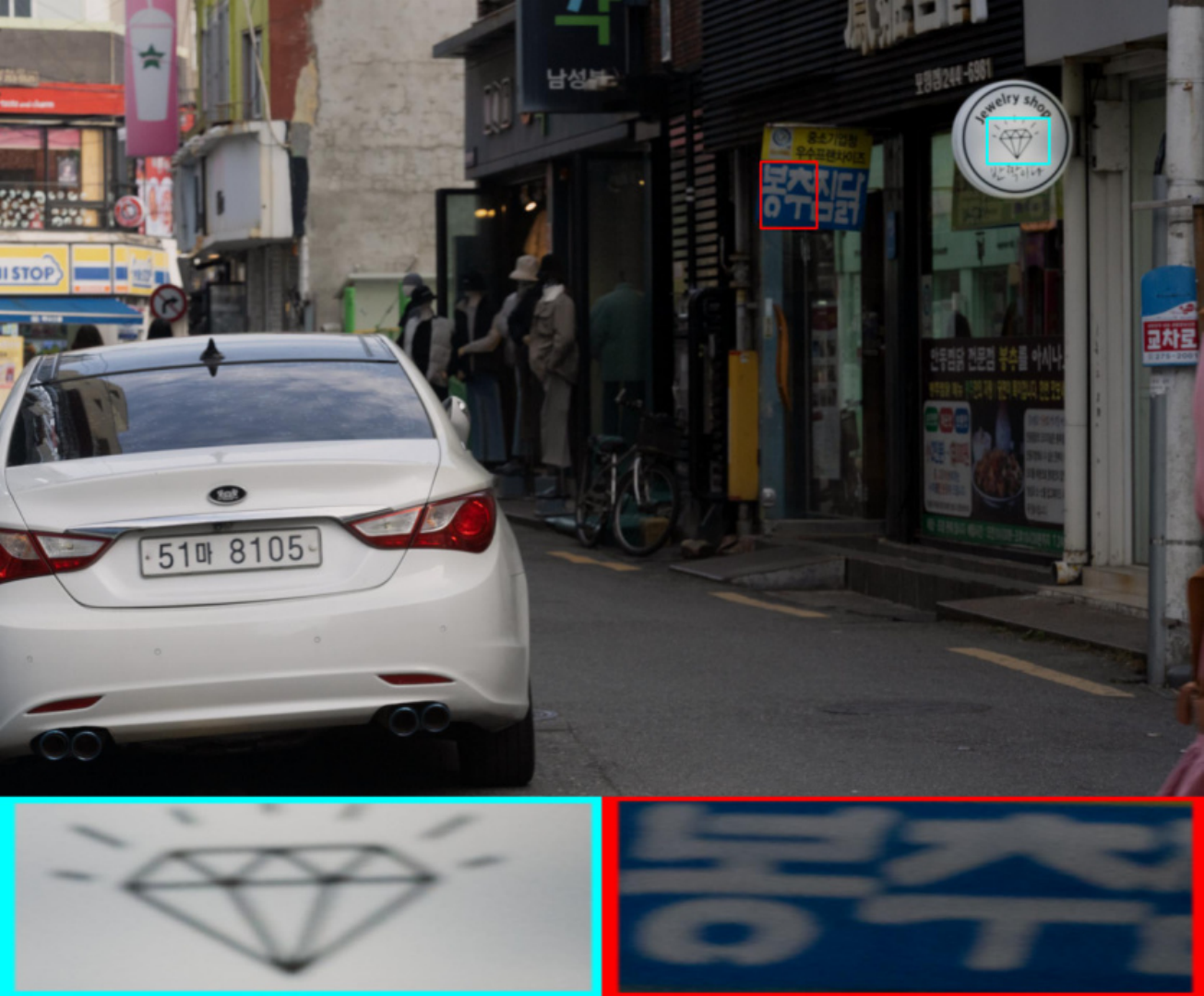}\vspace{4pt}
			\caption{Sharp}		
		\end{subfigure}
		
		\vfill
		
		\caption{Visualization results of different defocus deblurring methods on the RealDoF dataset.}
		\label{fig_5}
	\end{figure*}
	
	\begin{figure*}[h]
		\centering
		
		\begin{subfigure}[t]{0.16\linewidth}	
			\includegraphics[width=2.6cm,height=3cm]{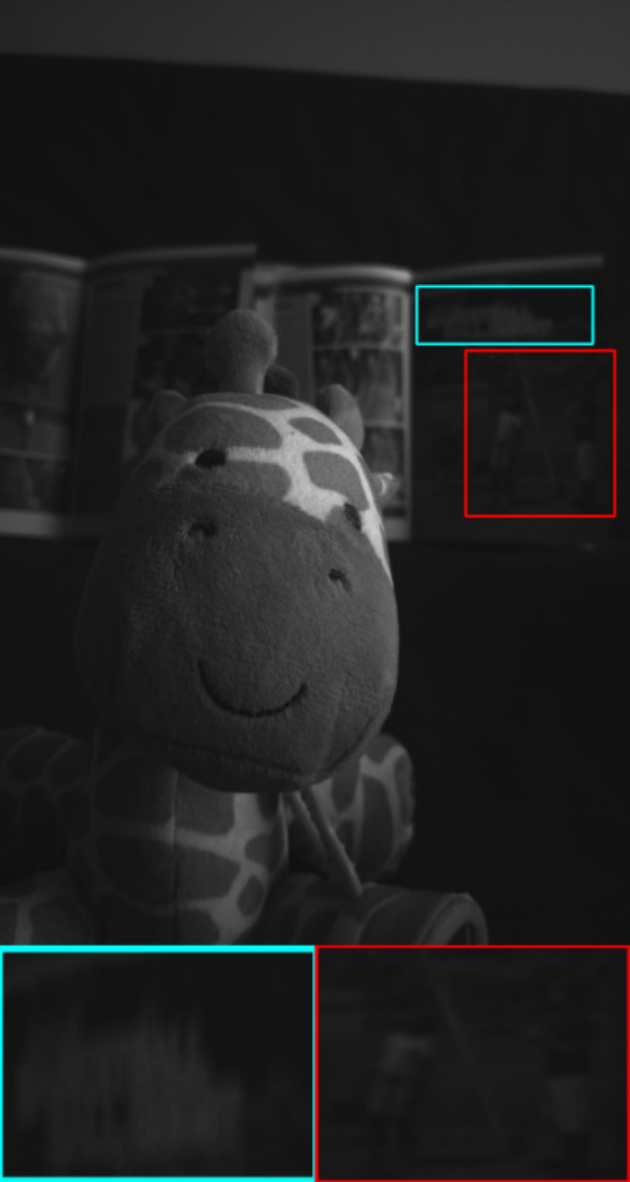}\vspace{4pt}
		\end{subfigure}
		\hfill
		\begin{subfigure}[t]{0.16\linewidth}	
			\includegraphics[width=2.6cm,height=3cm]{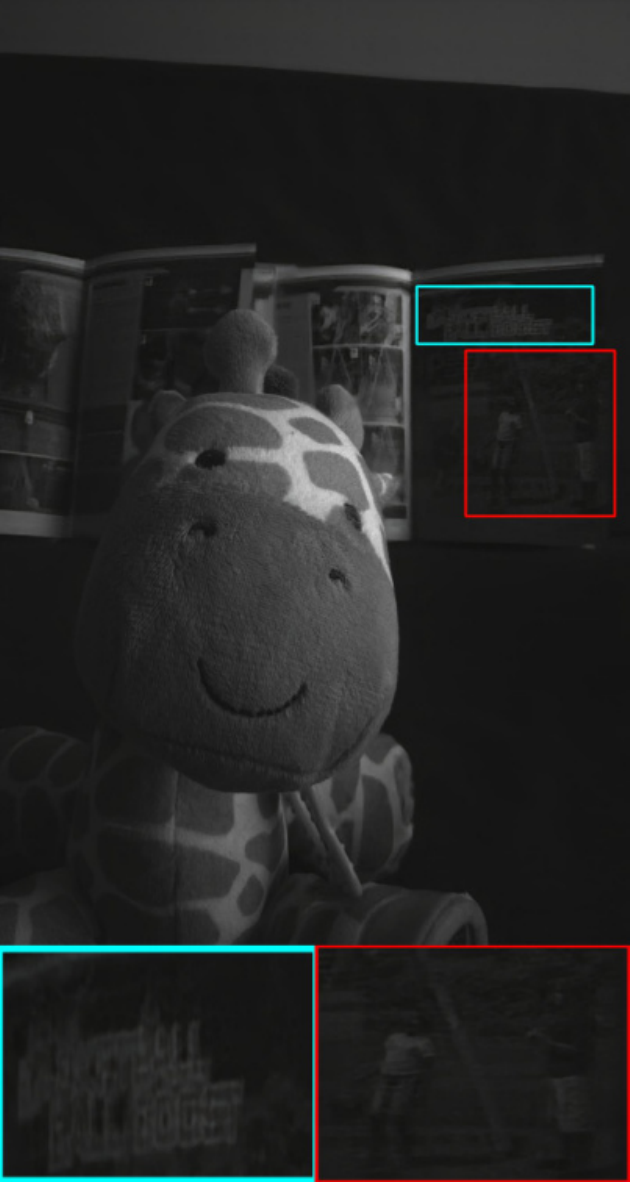}\vspace{4pt}
		\end{subfigure}
		\hfill
		\begin{subfigure}[t]{0.16\linewidth}	
			\includegraphics[width=2.6cm,height=3cm]{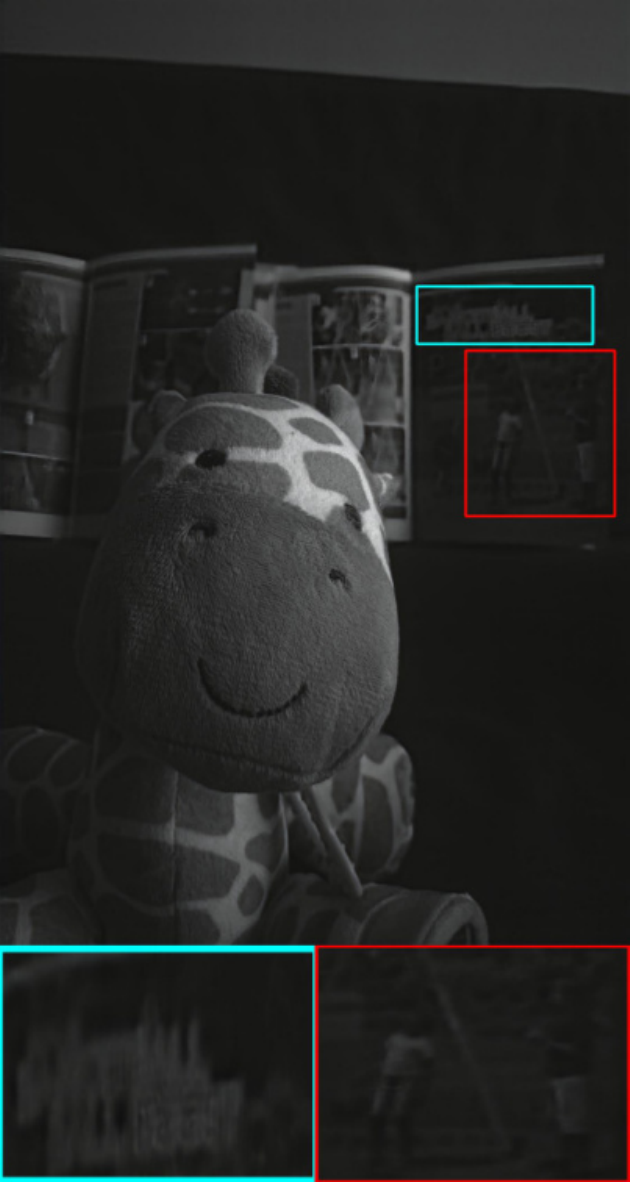}\vspace{4pt}
		\end{subfigure}
		\hfill
		\begin{subfigure}[t]{0.16\linewidth}	
			\includegraphics[width=2.6cm,height=3cm]{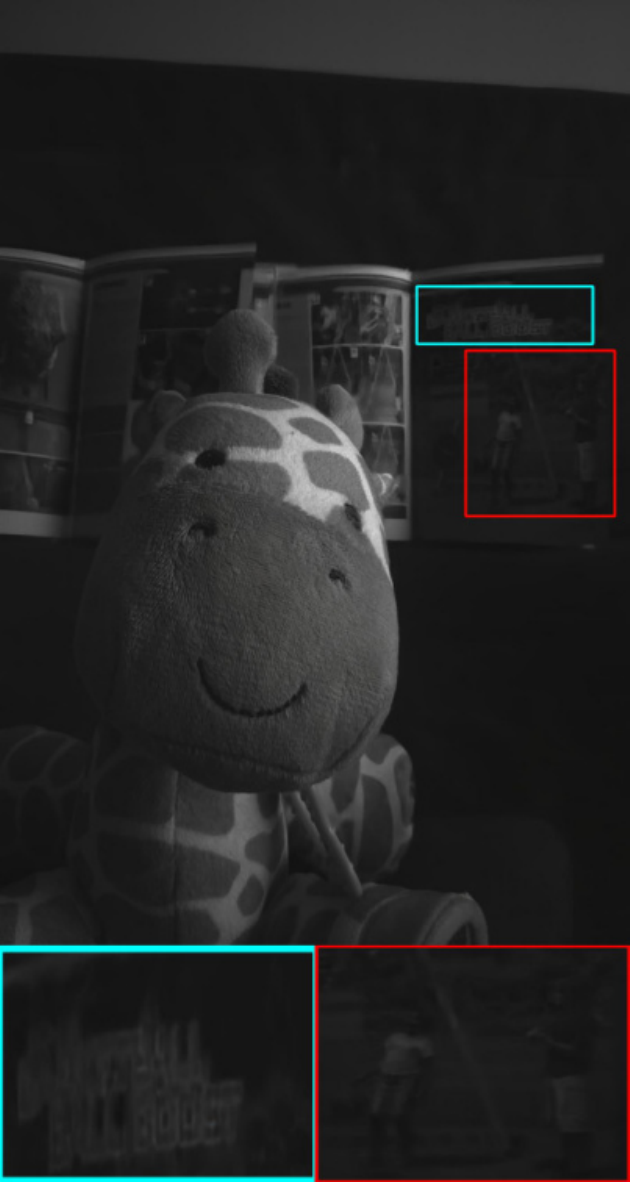}\vspace{4pt}
		\end{subfigure}
		\hfill
		\begin{subfigure}[t]{0.16\linewidth}	
			\includegraphics[width=2.6cm,height=3cm]{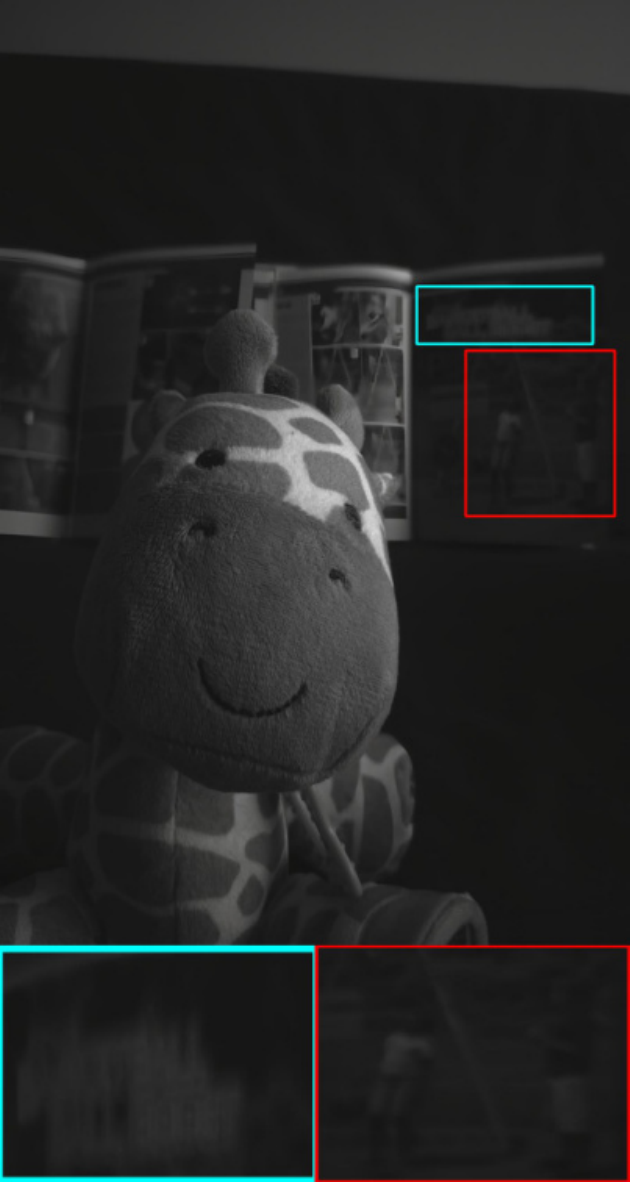}\vspace{4pt}
		\end{subfigure}
		\hfill
		\begin{subfigure}[t]{0.16\linewidth}	
			\includegraphics[width=2.6cm,height=3cm]{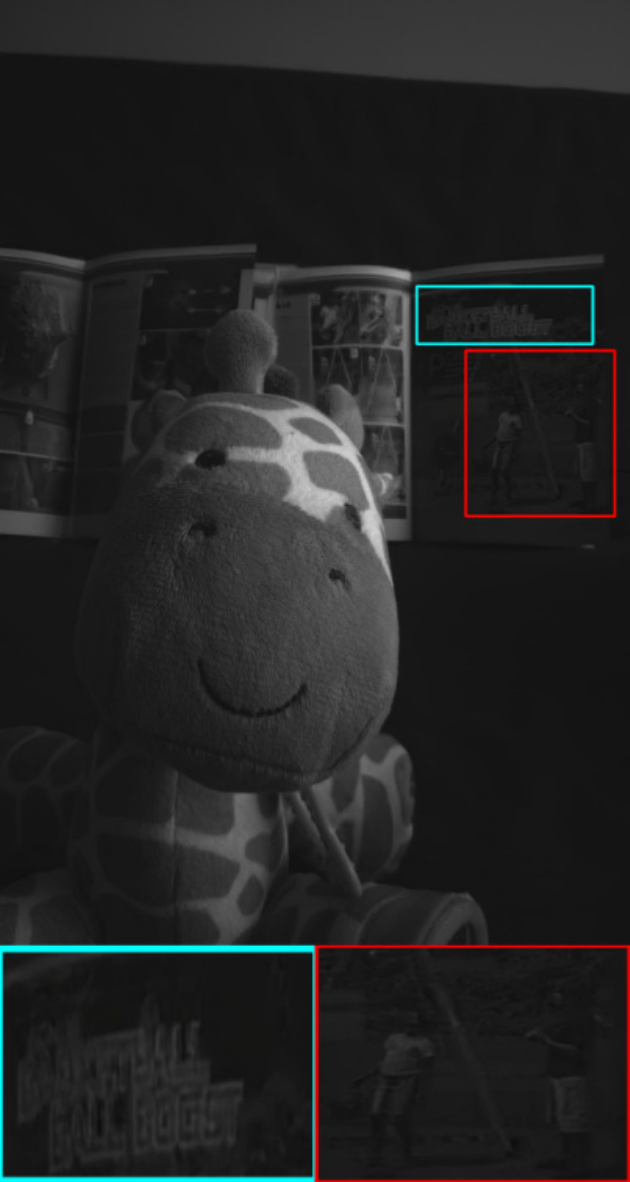}\vspace{4pt}
		\end{subfigure}	
		\vfill
		
		\begin{subfigure}[t]{0.16\linewidth}	
			\includegraphics[width=2.6cm,height=3cm]{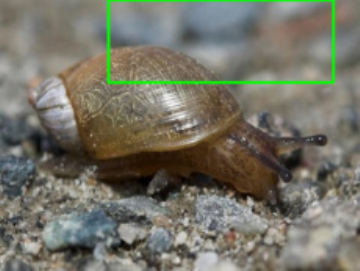}\vspace{4pt}
			\caption{Input}
		\end{subfigure}
		\hfill
		\begin{subfigure}[t]{0.16\linewidth}	
			\includegraphics[width=2.6cm,height=3cm]{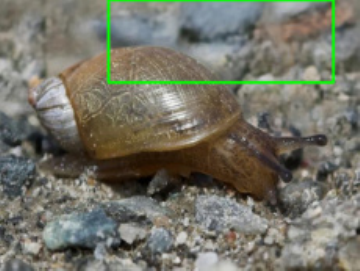}\vspace{4pt}
			\caption{KPAC}
		\end{subfigure}
		\hfill
		\begin{subfigure}[t]{0.16\linewidth}	
			\includegraphics[width=2.6cm,height=3cm]{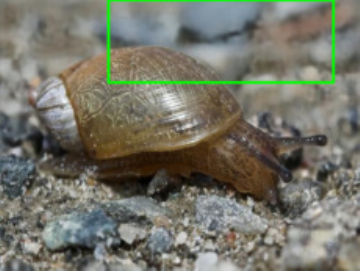}\vspace{4pt}
			\caption{GKMNet}
		\end{subfigure}
		\hfill
		\begin{subfigure}[t]{0.16\linewidth}	
			\includegraphics[width=2.6cm,height=3cm]{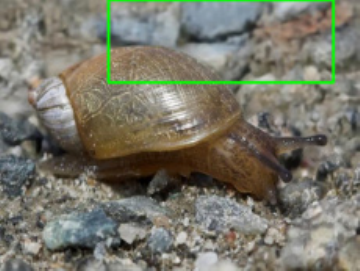}\vspace{4pt}
			\caption{IFAN}
		\end{subfigure}
		\hfill
		\begin{subfigure}[t]{0.16\linewidth}	
			\includegraphics[width=2.6cm,height=3cm]{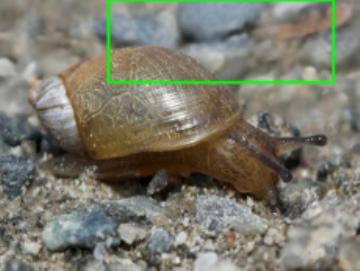}\vspace{4pt}
			\caption{Restormer}
		\end{subfigure}
		\hfill
		\begin{subfigure}[t]{0.16\linewidth}	
			\includegraphics[width=2.6cm,height=3cm]{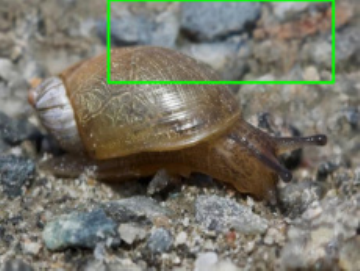}\vspace{4pt}
			\caption{SR-R$^2$KAC-B}
		\end{subfigure}
		
		\vfill		
		
		\caption{Qualitative results on PixelDP (1th row) \cite{Abuolaim2020} and CUHK (2nd row) \cite{Shi2015} datasets among KPAC \cite{Son2021}, IFAN \cite{Lee2021} , GKMNet \cite{Quan2021} and Restormer \cite{Zamir2022} and our SR-R$^2$KAC-B.}
		\label{fig_7}
	\end{figure*}
	
	\subsection{Ablation study}
	
	We start with the whole R$^2$KAC, i.e., directly using R$^2$KAC for SSID. 
	We compare R$^2$KAC with two more baseline modules, one using KPAC and another simply using several Residual blocks (RB) as \cite{Lee2021} does\footnote{Referring to Fig. 3, the first baseline replaces R$^2$KAC with KPAC, and the second baseline replaces R$^2$KAC with RB.}. 
	The comparsion results are shown in Table. \ref{tab2}.
	We can see that R$^2$KAC achieves better results with fewer model parameters than both baselines. On the one hand, this indicates the advantages of R$^2$KAC over KPAC. 
	On the other hand, it proves that simply using residual blocks fails to improve the debluring performance as it is even worse than KPAC.
	
	Next, we analyze the individual component, i.e., RKAC, residual connection (RC), and attention module (AM), with respect to R$^2$KAC.
	Except for our final model SR-R$^2$KAC (RKAC+RC+SRM), we have another three variants, namely RKAC, RKAC+RC, and RKAC+RC+AM.
	The comparison results are shown in Table \ref{tab2_1}.
	The performance improvement of RKAC+RC over RKAC indicates the necessity of residual connections. More analysis about the RC's importance can be found in Sec.III.F.
	From the comparison of RKAC+RC and RKAC+RC+AM, we find that it is better not using attention modules.
	In addition, the scale recurrent module (RKAC+RC+SRM v.s. RKAC+RC) further boosts the performances R$^2$KAC network from $26.04$dB to $26.19$dB on the DPDD dataset and from $25.06$dB to $25.32$dB on the RealDoF dataset.	
	
	\subsection{The number of R$^2$KAC blocks} 
	Multiple R$^2$KAC blocks can be stacked to utilize the
	iterative nature for reducing remaining blurs and thus achieve deblurring results of better quality. 
	Before using the SRM, we explore the effect of the number of R$^2$KAC blocks at the single-scale branch.
	Table \ref{tab_44} shows that by stacking $2$ R$^2$KAC blocks, the PSNR values increase from $26.01$dB to $26.04$dB on the DPDD dataset, and from $25.01$dB to $25.06$dB on the RealDoF dataset. 
	When stacking $3$ R$^2$KAC blocks, the PSNR values on both datasets do not increase, probably due to the increased training complexity \cite{Son2021}. 
	Based on these experiments, we build the SR-R$^2$KAC module by stacking two R$^2$KAC blocks. 
	
	\subsection{Comparison Studies}
	We compare R$^2$KAC, SR-R$^2$KAC and SR-R$^2$KAC-B\footnote{SR-R$^2$KAC-B is obtained by changing the channel number $C$ in SR-R$^2$KAC from 48 to 96.} with previous defocus deblurring methods, including deep learning based methods \cite{Abuolaim2020,Lee2019,Son2021, Quan2021, Zamir2022, Ruan2022} and two-stage based methods  \cite{Shi2015,Kingma2014,Lee2019}. 
	For IFAN \cite{Lee2021}, KPAC \cite{Son2021}, GKMNet \cite{Quan2021}, and Restormer \cite{Zamir2022}, we reproduce the results based on the released codes and pre-trained weights. 
	For JNB \cite{Shi2015}, EBDB \cite{Karaali2017}, DMENet \cite{Lee2019},  DPDNet \cite{Abuolaim2020} and DRBNet \cite{Ruan2022}, we directly report the results either from \cite{Lee2021} or from the corresponding paper for comparison. 
	All the methods are trained on the DPDD training set. 
	Specifically, IFAN needs to use dual views data for training, while other methods do not. 
	For a comprehensive comparison, we follow \cite{Lee2021, Son2021, Abuolaim2021} and use PSNR as the main evaluation metric, Structural Similarity (SSIM) \cite{Wang2004}, and Learned Perceptual Image Patch Similarity (LPIPS) \cite{Zhang2018} as complementary evaluation metrics. 
	With one NVIDIA A100 GPU, the average inference time is also reported for DPDD dataset.
	
	Table \ref{tab6} shows quantitative comparisons on the DPDD and RealDoF datasets. 
	As can be seen, two-stage methods (2nd-4th rows) generally do not perform well in terms of all the evaluation metrics on both DPDD and RealDoF datasets. 
	DPDNet (5th row) performs better than all the two-stage methods, and the following advanced deep learning-based methods (i.e., KPAC, IFAN, GKMNet and DRB, corresponding to the 6th-10th rows) further boost the performance. 
	Notably, Restormer and SR-R$^2$KAC-B significantly outperform the other methods on all metrics.  
	Compared with Restormer, SR-R$^2$KAC-B achieves much better PSNR (26.41dB vs 25.98dB) and MAE values (0.364 vs 0.378) and comparable SSIM and LPIPs values with much less computational cost (Parameter: 8.81M vs 26.10M, Time: 0.043s vs 0.45s), which indicate a clear winner of SR-R$^2$KAC-B.
	Moreover, on the RealDoF dataset, all the variants of our method significantly outperform the other baselines.
	Specifically, SR-R$^2$KAC-B outperforms the second-best method IFAN by 0.70dB in PSNR value (25.59dB vs 24.71dB).
	Since Restormer fails to work on the RealDoF dataset under our configuration, we compare with it on a downscaled RealDoF dataset in Sec.III.H.
	
	Fig. \ref{fig_4} shows the qualitative results of GKMNet, IFAN, Restormer and SR-R$^2$KAC-B on DPDD dataset.
	As can be seen (for the cases of spatially varying blurs), all the images obtained by other baselines still contain remaining blurs, while the images obtained from our model are much clearer. 
	Fig. \ref{fig_5} shows the qualitative results of KPAC, GKMNet, IFAN and SR-R$^2$KAC-B on the RealDoF dataset.
	As shown in Fig. \ref{fig_5} (for the cases of large blurs), SR-R$^2$KAC-B generates much clearer images than the other four baselines. 
	
	To investigate the generalizability, we further compare SR-R$^2$KAC-B with KPAC, IFAN, GKMNet and Restormer on PixelDP and CUHK datasets for more qualitative analysis.
	Fig. \ref{fig_7} shows that our model is better generalized than the other five baselines on both datasets. 
	
\begin{figure*}[t]
	\centering
	\includegraphics[width=18cm, height=11cm]{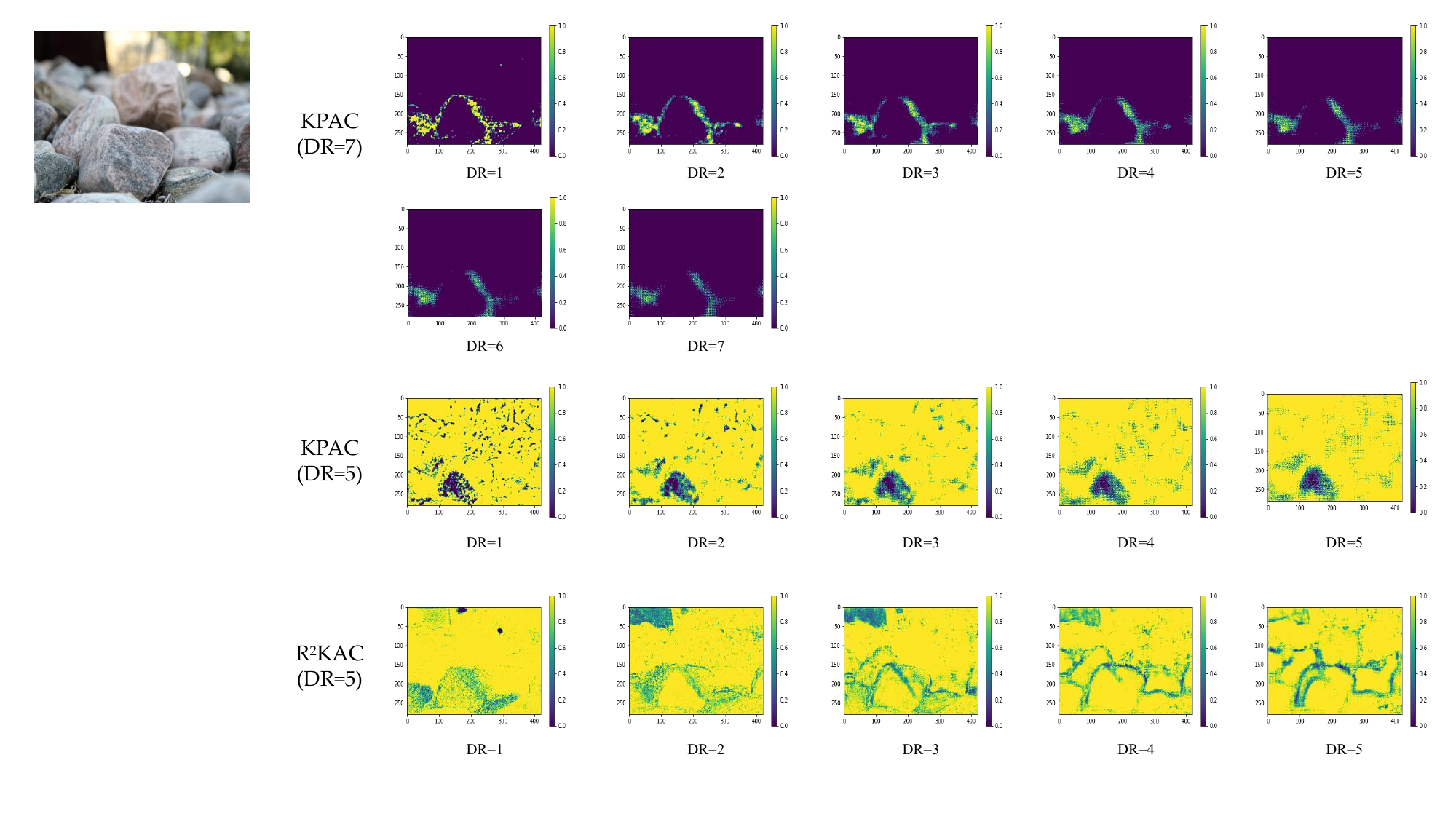}
	\caption{Feature maps extracted by KPAC (DR=5) and KPAC (DR=7). The light area indicates the area with strong response, and the dark area indicates the area with weak response for KPAC. } 
	\label{FM_KPAC}
\end{figure*}
	
	\subsection{Analysis of KPAC with larger dilation rate} 
	In this section, we provide more analyses of KPAC with different dilation rates (DR). Table~\ref{tab_s_4} compares KPAC with DR of 5, same as in \cite{Son2021}, and a larger DR of 7.
	The larger DR setting is used to evaluate the effect of increasing the dilation rate in KPAC (it is equivalent to enlarging the scaling factor in inverse kernel deconvolutions). 
	The results show that KPAC with DR of 7 does not improve KPAC with DR of 5, and requires more model parameters.
	In contrast, R$^2$KAC outperforms both of these two KPACs.
	
	We also visualize the feature maps (FMs) extracted by KPACs (DR = 5 and 7) and R$^2$KAC (DR = 5) in Fig.~\ref{FM_KPAC} for qualitative analysis. 
	FMs are obtained using the first KPAC block output, on which element-wise summation on channels followed by normalization are done.
	Fig.~\ref{FM_KPAC} shows that FMs extracted by KPAC (DR=7) have only a few responses focused on the boundary between the blurry and sharp areas, while FMs extracted by KPAC(DR=5) can roughly distinguish the blurry and sharp areas. 
	FMs extracted by R$^2$KAC (DR=5) classifies the blurry and sharp areas more accurately then KPAC (DR=5), and the FMs extracted by different rate focus on different areas. 
	These results illustrate that our proposed R$^2$KAC has a stronger ability to capture blurry feature maps compared with KPAC.
	
\begin{table}[t]
	\centering
	\caption{The effect of larger dilation rate (DR) in KPAC.}	
	\renewcommand\arraystretch{1}	
	\resizebox{\linewidth}{!}{
		\begin{tabular}{cccccc}
			\toprule
			\multirow{2}{*}{Number} & \multicolumn{2}{c}{DPDD} & \multicolumn{2}{c}{RealDof} & \multirow{2}{*}{Param(M)} \\ \cmidrule{2-5}
			& PSNR (dB)     & SSIM     & PSNR (dB)      & SSIM       &                           \\ \midrule
			KPAC origial (DR=5)             & 25.90       & 0.796      & 24.77        & 0.740 &  1.98     \\
			KPAC (DR=7)       & 25.88       & 0.796      & 24.74        & 0.738  &1.90\\
			R$^2$KAC (DR=5)     & 26.04                        & 0.801       & 25.06   & 0.753 &1.72 
			\\ \hline   
	\end{tabular}}
	\label{tab_s_4}
\end{table}
\begin{figure*}[t]
	\centering
	\includegraphics[width=18cm, height=11cm]{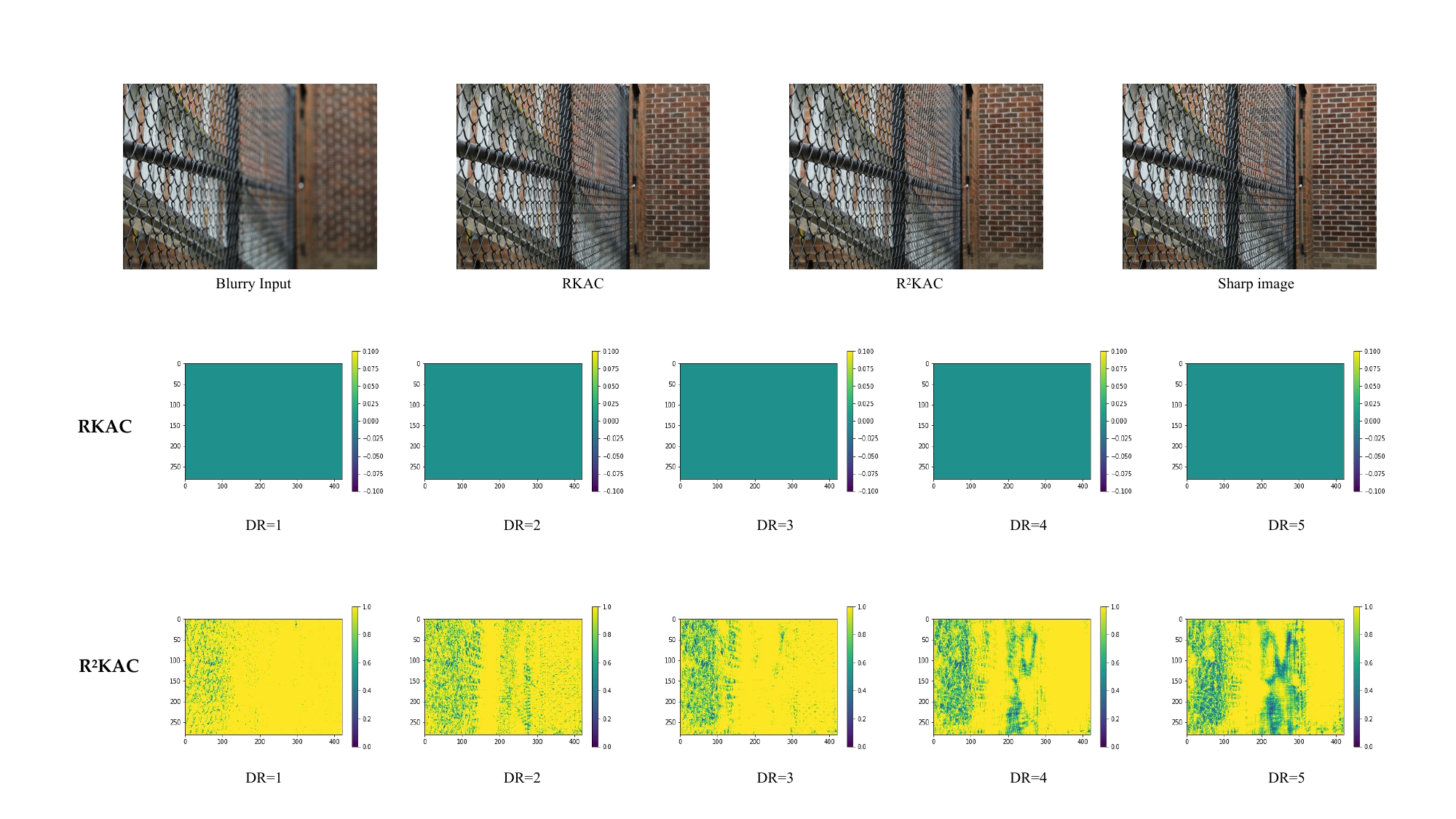}
	\caption{Feature maps extracted by R$^2$KAC and RKAC.} 
	\label{FM_R2KAC_RKAC}
\end{figure*}	
	\subsection{Analysis of Residual Connection in R$^2$KAC} 
	We have empirically shown the importance of RC in the single R$^2$KAC module in Section 1.
	Herein, we present the analysis of RC in the R$^2$KAC network.
	Quantitative results have already been presented in Table 1 of the main paper.
	We further show the feature maps extracted by R$^2$KAC and RKAC in Fig.~\ref{FM_R2KAC_RKAC} in this section.
	Fig.~\ref{FM_R2KAC_RKAC} displays that R$^2$KAC can extract the texture of the blurry image and then roughly classify blurry and sharp areas, but RKAC fails to do so.
	These qualitative results demonstrate that RC plays an important role in extracting the local details in the large kernel convolution, which is also shown in Guideline 2 of \cite{Ding2022} (R$^2$KAC can be viewed as with large kernel, as its kernel size is 5 with dilation with 2-5).
	There are other analyses showing that RC is important for large kernel convolution, e.g., \cite{Veit2016,Ding2022}.
	These works claim that RC makes the model an implicit ensemble composed of numerous models with different receptive fields (RFs), so that it can benefit from a much larger maximum RF while not losing the ability to capture small-scale patterns.
	

\begin{figure}[t]
	\begin{subfigure}[t]{0.45\linewidth}	
		\includegraphics[width=4cm,height=3cm]{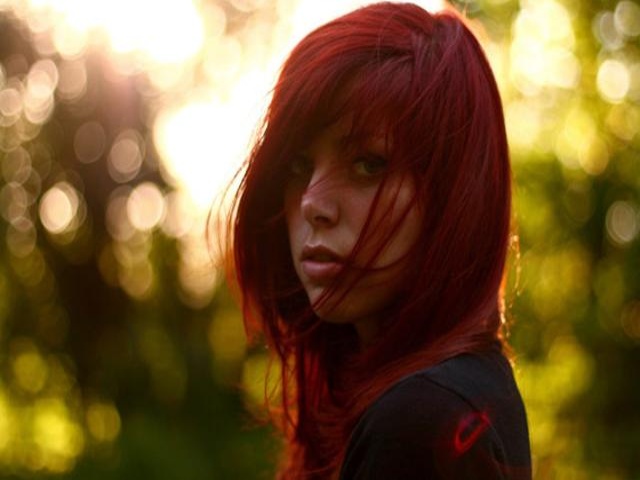}\vspace{4pt}
		\caption{Input}
	\end{subfigure}
	\hfill
	\begin{subfigure}[t]{0.45\linewidth}	
		\includegraphics[width=4cm,height=3cm]{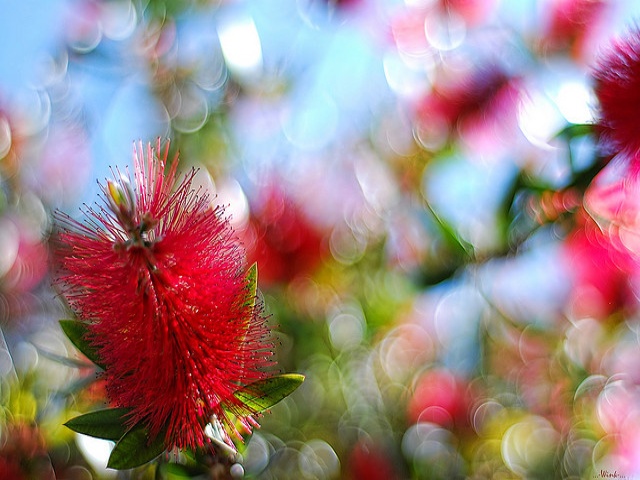}\vspace{4pt}
		\caption{Input}
	\end{subfigure}
	\vfill
	\begin{subfigure}[t]{0.45\linewidth}	
		\includegraphics[width=4cm,height=3cm]{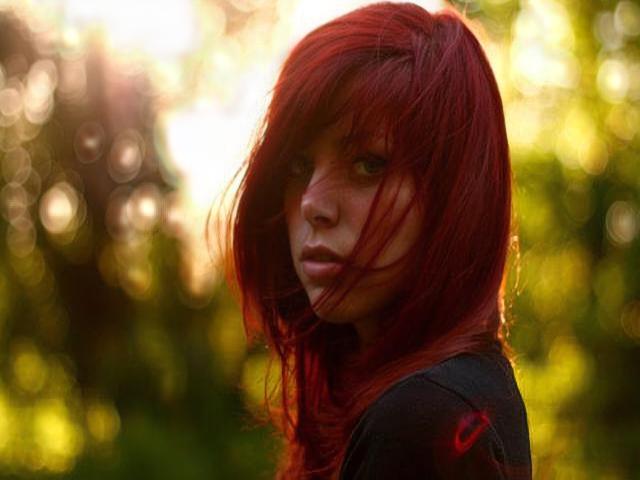}\vspace{4pt}
		\caption{SR-R$^2$KAC-B}
	\end{subfigure}
	\hfill
	\begin{subfigure}[t]{0.45\linewidth}	
		\includegraphics[width=4cm,height=3cm]{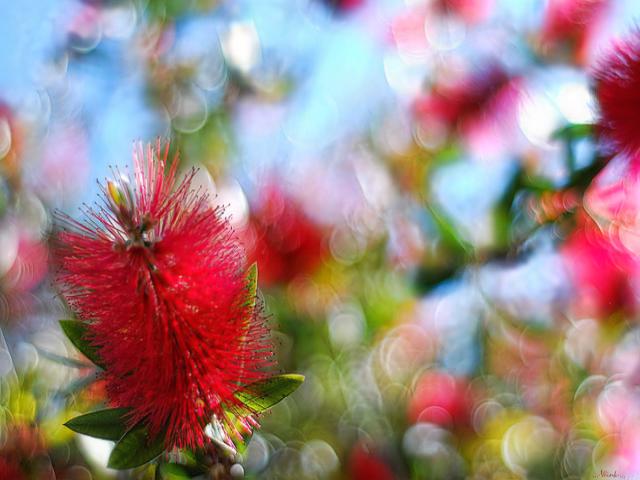}\vspace{4pt}
		\caption{SR-R$^2$KAC-B}
	\end{subfigure} 
	\caption{Failure cases.}
	\label{fig_s_6}
\end{figure}
	

	\subsection{Comparisons of SR-R$^2$KAC-B and Restormer on Downscaled RealDoF Dataset}
	As Restormer cannot infer 2300$\times$1500$\times$3 shaped images from the RealDoF dataset, we downscale the images to 1680$\times$1120$\times$3, which is the same as the size of the image from DPDD dataset.
	Table~\ref{table_s_4} shows that SR-R$^2$KAC-B also outperforms Restormer on the downscaled dataset with much less model parameters.
	\begin{table}[h]
		\centering
		\caption{Comparison of SR-R$^2$KAC-B and Restormer on downscaled RealDoF dataset}
		\scalebox{1}{
			\renewcommand\arraystretch{1.0}
			\setlength{\tabcolsep}{2mm}{
				\begin{tabular}{cccccc}
					\toprule
					\multirow{2}{*}{Model} & \multicolumn{4}{c}{RealDoF(downscaled)}                           & \multirow{2}{*}{Parameter (M)} \\ \cmidrule{2-5}
					& PSNR (dB)      & SSIM           & LPIPs          & MAE            &                                \\ \midrule
					Restormer              & 25.44          & \textbf{0.801} & \textbf{0.217} & 0.371          & 26.10                          \\
					SR-R$^2$KAC-B          & \textbf{25.88} & 0.795          & \textbf{0.217} & \textbf{0.362} & \textbf{8.81}                  \\ \bottomrule
		\end{tabular}}}
		\label{table_s_4}
	\end{table}

	\subsection{Comparisons on LF-DoF Dataset}
	To further evaluate the generalizability of SR-R$^2$KAC, we compare it with AirNet \cite{Ruan2021}, and DRBNet \cite{Ruan2022} on LF-DoF dataset \cite{Ruan2022}.
	The comparison on the LF-DoF dataset in Table \ref{table_s_2} shows that SR-R$^2$KAC outperforms the state-of-the-art methods on the LF-DoF dataset, i.e., DRBNet and AIFNet, which demonstrates the stronger generalization ability of our method.
	\begin{table}[h] 
		\centering
		\caption{Performance comparison on LF-DOF dataset}
		\scalebox{1}{
			\renewcommand\arraystretch{1.0}
			\setlength{\tabcolsep}{2mm}{
				\begin{tabular}{cccc}
					\toprule
					Method & PSNR  & SSIM  & LPIPs \\ \midrule
					AIFNet & 29.68 & 0.884 & 0.202 \\
					DRBNet & 30.40 & 0.891 & 0.145 \\
					SR-R$^2$KAC-B  & 31.11 & 0.899 & 0.112 \\ \bottomrule
		\end{tabular}}}
		\label{table_s_2}
	\end{table}

%

	\section{Failure Cases}
	SR-R$^2$KAC mainly aim to handle cases where blurs are with different sizes but the same shape. 
	However, in practice, the blur shape may also varies \cite{Son2021}.
	As shown in Fig.~\ref{fig_s_6}, SR-R$^2$KAC may not well handle the blurs with irregular shapes, which rarely appear in the DPDD training set \cite{Abuolaim2020}.

	\section{Conclusion}

	We proposed an end-to-end deep learning framework SR-R$^2$KAC, i.e., residual and recursive kernel-sharing atrous convolution (R$^2$KAC) with scale recurrent module (SRM), for SIDD.
	Inspired by an observation, that is, a large-sized inverse kernel can be approximated by the recursive stacking of small-sized inverse kernels, R$^2$KAC recursively uses kernel-sharing atrous convolutions with different dilation rates and adds identity shortcuts between atrous convolutions for efficiently handling both spatially varying and large defocus blurs. 
	To further improve the deblurring quality, a scale recurrent module is used to exploit the multi-scale information from low-resolution to high resolution thoroughly.
	In experiments, we show the superiority of SR-R$^2$KAC over existing methods, and the results indicate that SR-R$^2$KAC is a promising method for SSID.
	However, we also note that it is still challenging for SR-R$^2$KAC to handle blurs with irregular shapes. 

\clearpage
\section{Appendix}

\subsection{Overall Structure of KPAC}
\begin{figure}[h]
	\centering
	\includegraphics[width=9cm, height=5cm]{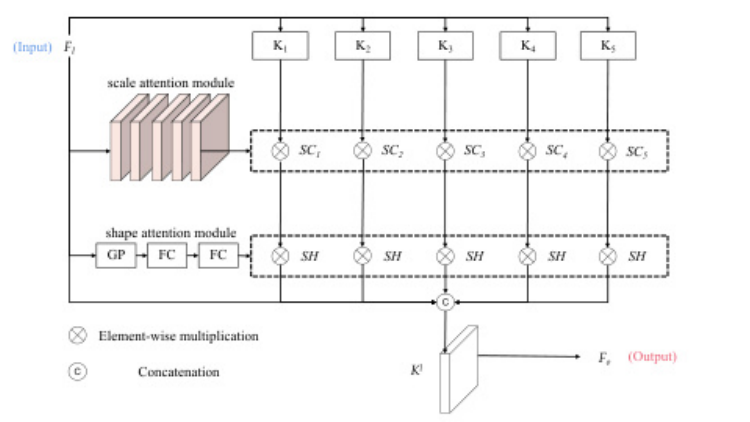}
	\caption{Structure of KPAC} 
	\label{fig10}
\end{figure}

Fig.~\ref{fig10} shows the overall structure of KPAC. 
As can be seen, KPAC combines the outputs from multiple atrous convolutions in parallel.
Moreover, KPAC learns spatial attention as the scale attention $SC_i \in \mathbb{R}^{1 \times 1 \times C}$ and a channel-wise attention as the shape attention $SH \in \mathbb{R}^{W \times H \times 1}$, where $i= (1,2,...,N)$ \cite{Son2021}.
Note that the number of channels for the input and output of each $RK_i$ are the same in R$^2$KAC.
To ensure that KPAC and R$^2$KAC only differ on the network structure, in our experiments, we also keep the number of channels unchanged for each $K_i$ in KPAC.
\subsection{More Comparisons bewteen KPAC and R$^2$KAC}
\begin{figure*}[h]
	\centering
	
	\begin{subfigure}[t]{0.22\linewidth}	
		\includegraphics[width=4cm,height=4cm]{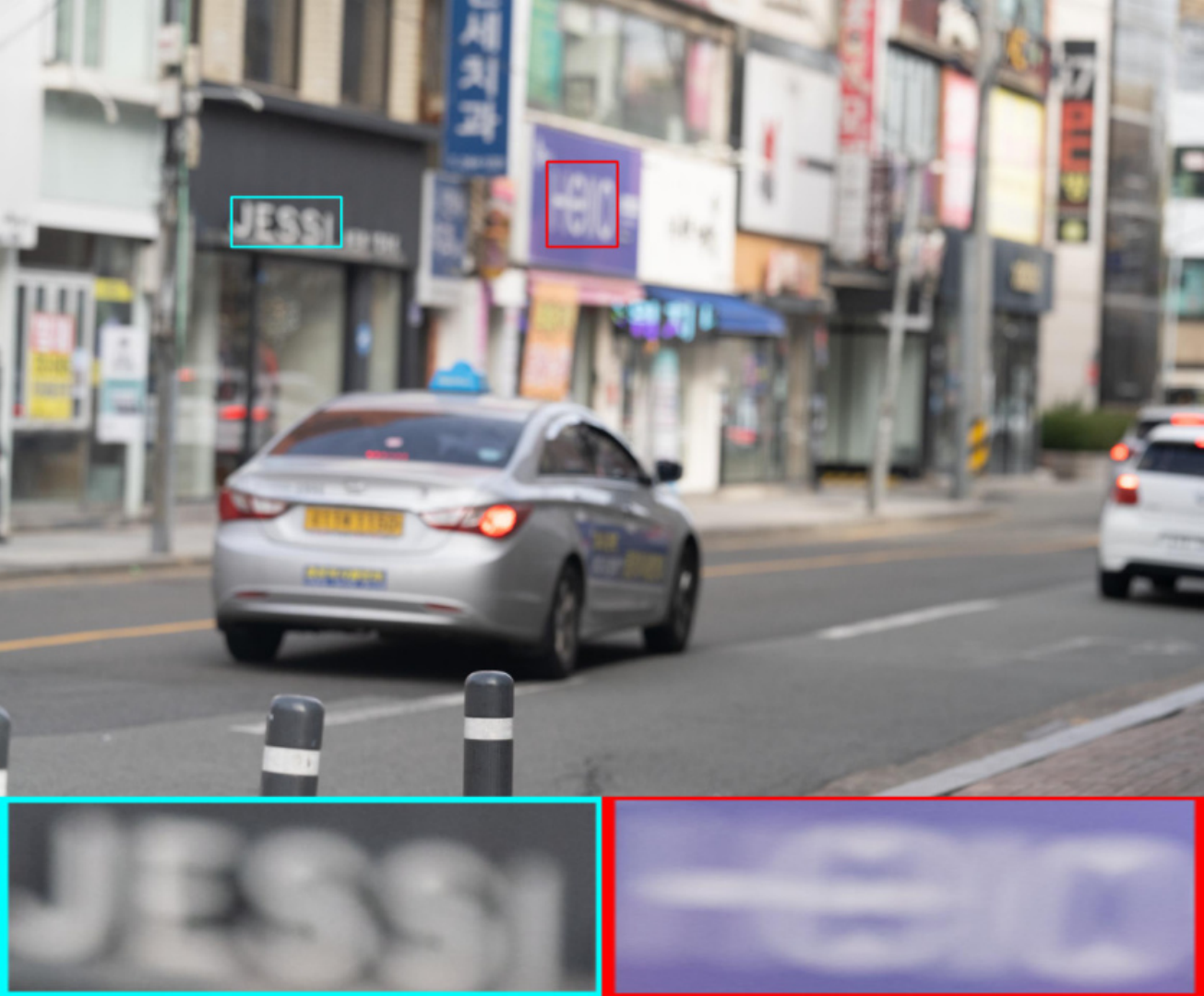}\vspace{4pt}
	\end{subfigure}
	\hfill
	\begin{subfigure}[t]{0.22\linewidth}	
		\includegraphics[width=4cm,height=4cm]{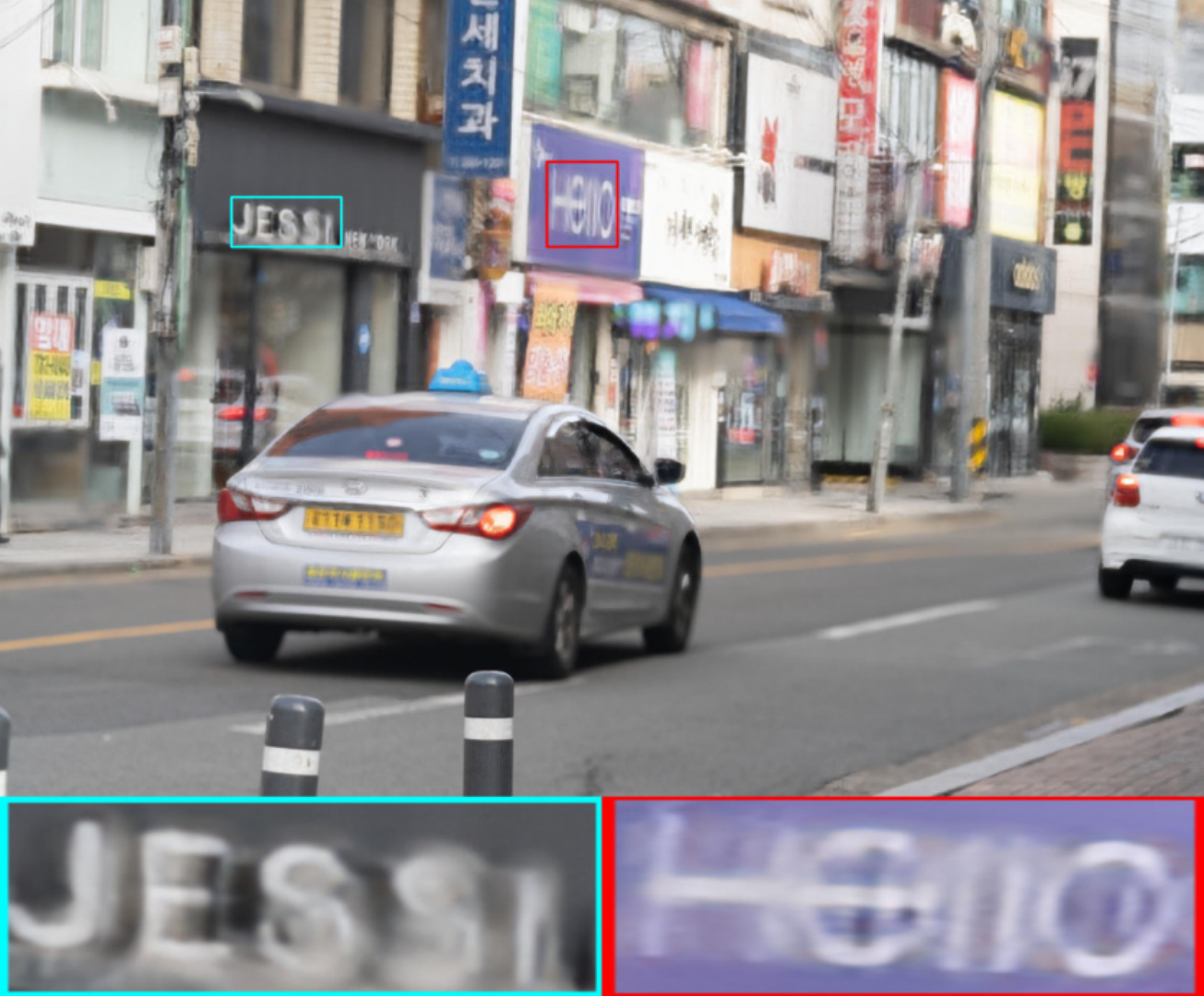}\vspace{4pt}
	\end{subfigure}
	\hfill
	\begin{subfigure}[t]{0.22\linewidth}	
		\includegraphics[width=4cm,height=4cm]{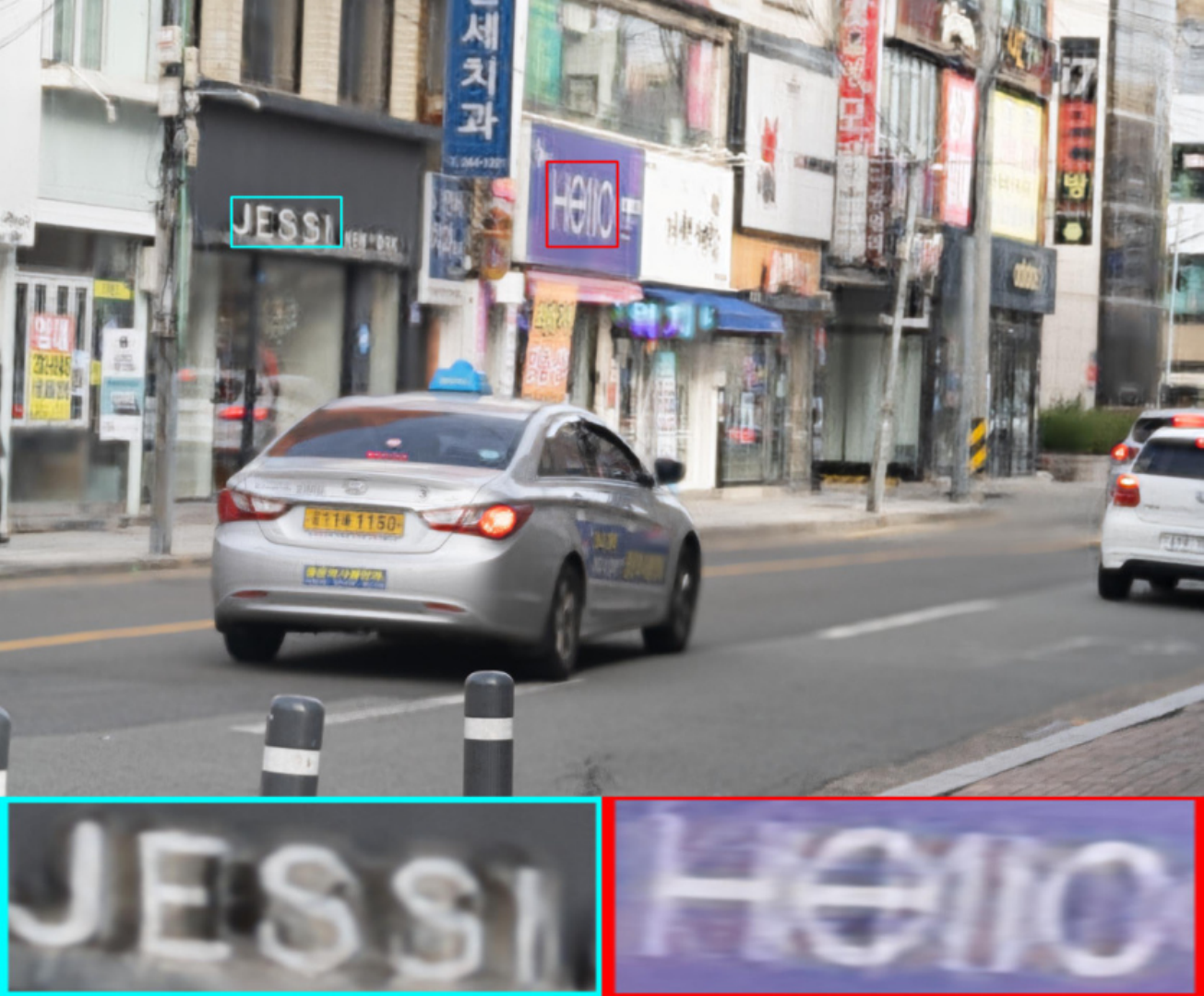}\vspace{4pt}
	\end{subfigure}
	\hfill
	\begin{subfigure}[t]{0.22\linewidth}	
		\includegraphics[width=4cm,height=4cm]{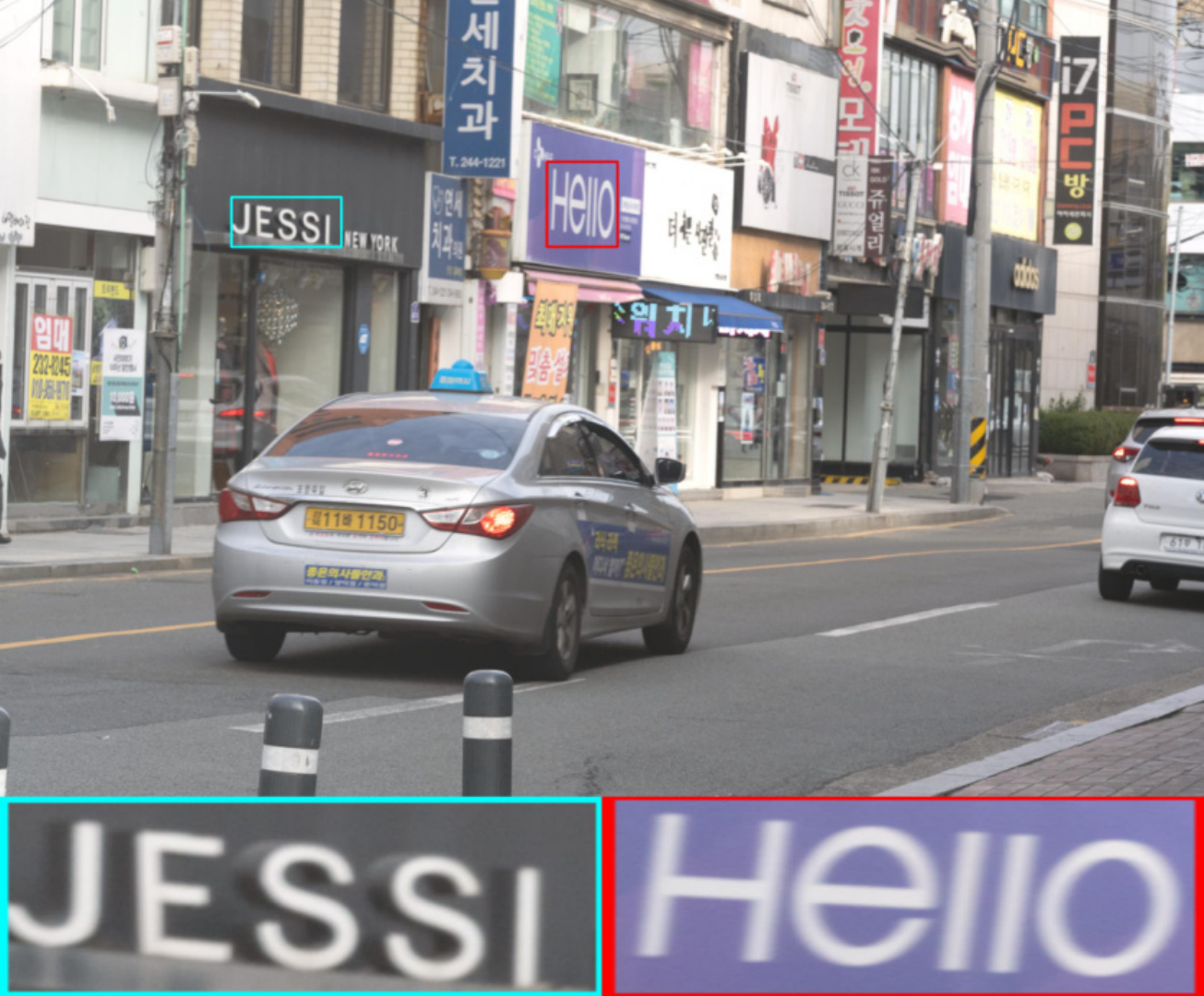}\vspace{4pt}
	\end{subfigure}
	\vfill
	
	\begin{subfigure}[t]{0.22\linewidth}	
		\includegraphics[width=4cm,height=4cm]{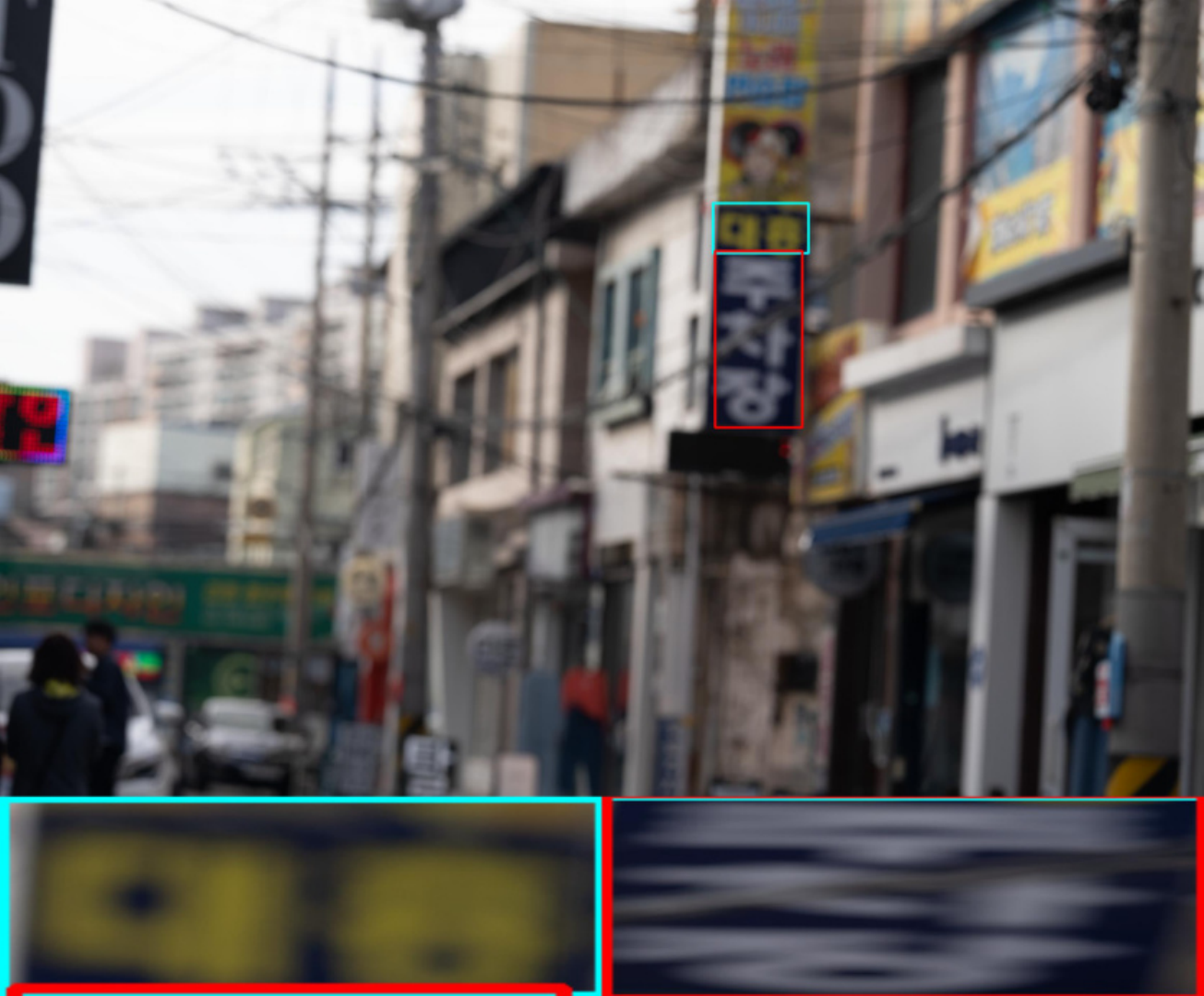}\vspace{4pt}
	\end{subfigure}
	\hfill
	\begin{subfigure}[t]{0.22\linewidth}	
		\includegraphics[width=4cm,height=4cm]{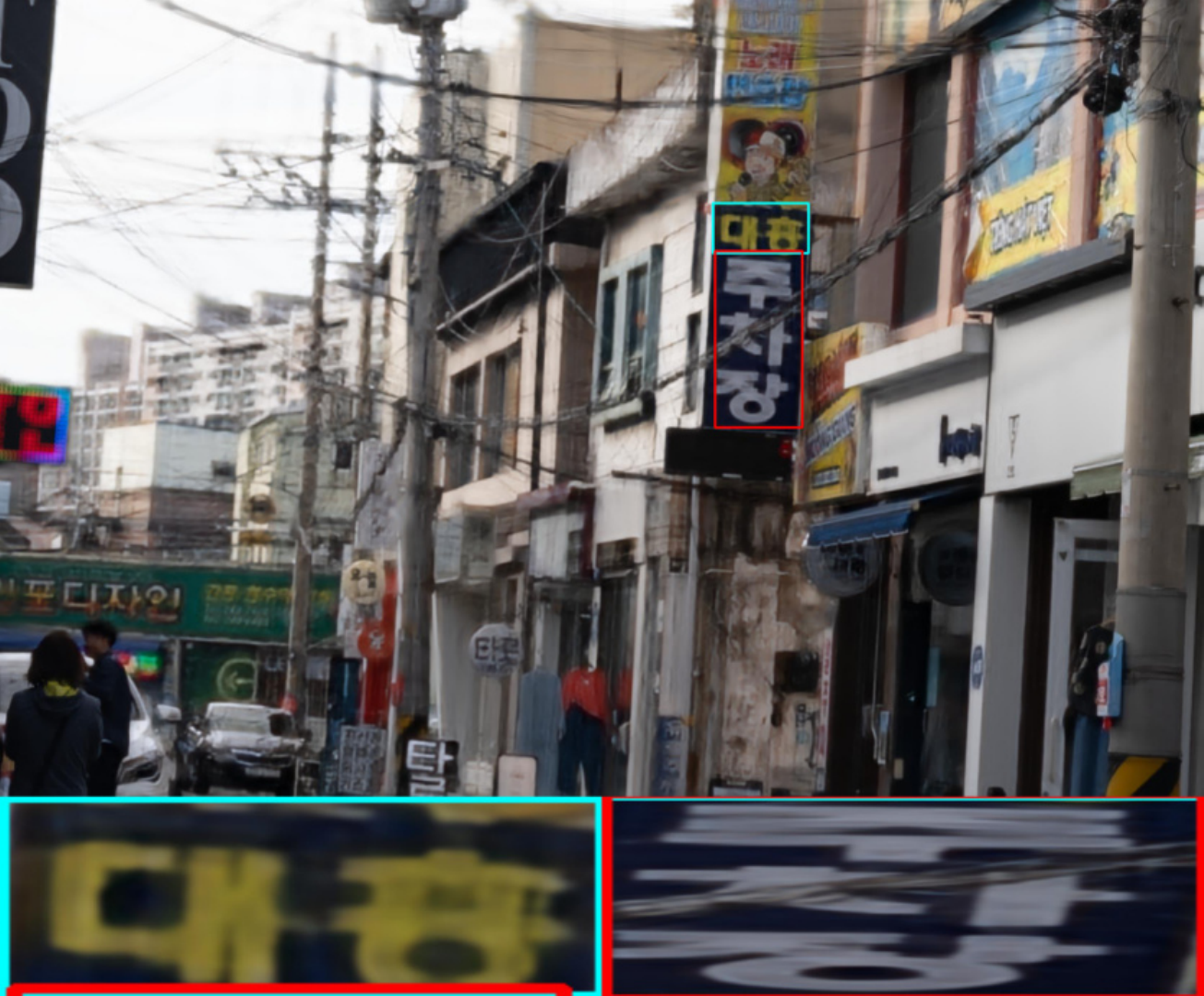}\vspace{4pt}
	\end{subfigure}
	\hfill
	\begin{subfigure}[t]{0.22\linewidth}	
		\includegraphics[width=4cm,height=4cm]{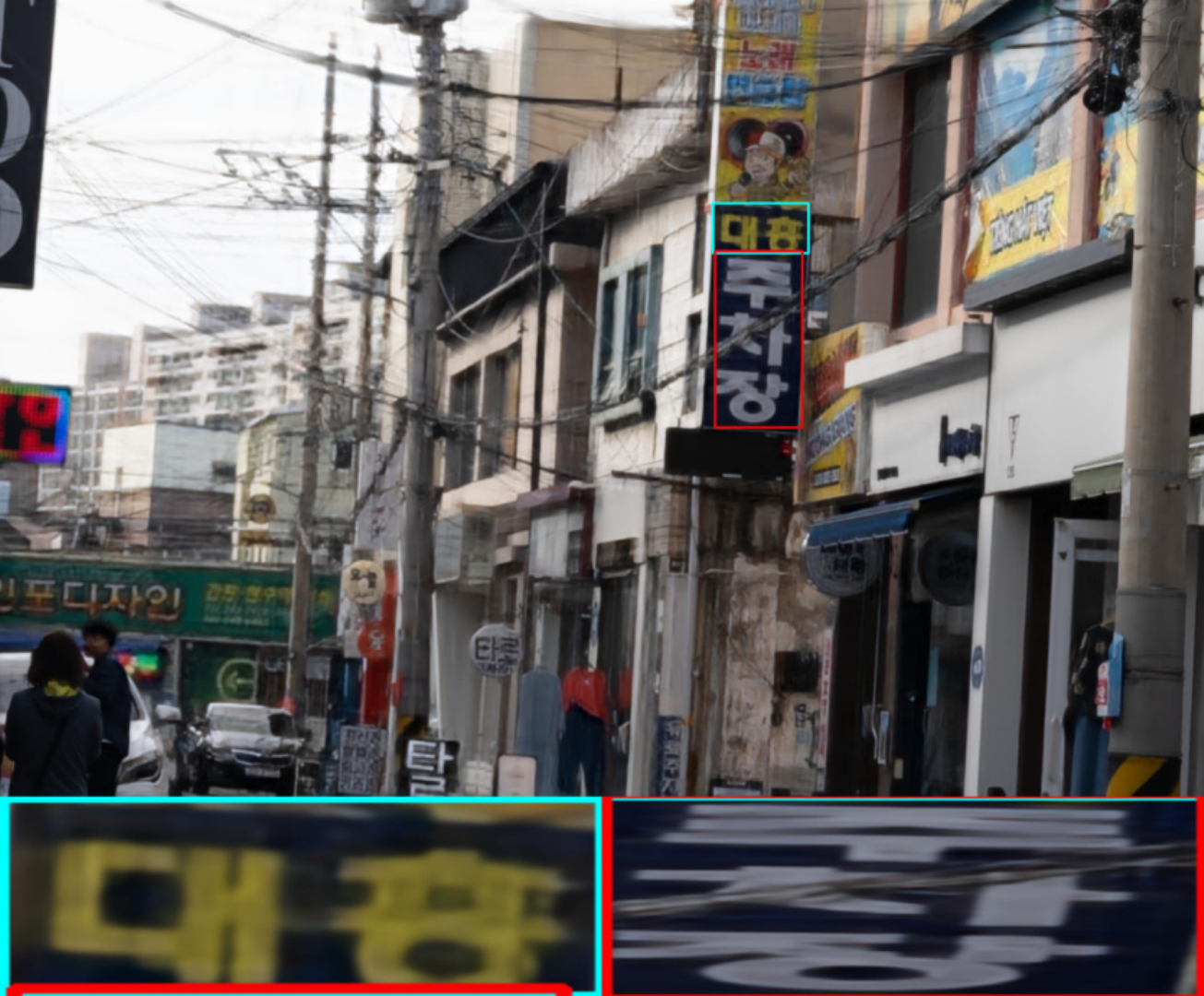}\vspace{4pt}
	\end{subfigure}
	\hfill
	\begin{subfigure}[t]{0.22\linewidth}	
		\includegraphics[width=4cm,height=4cm]{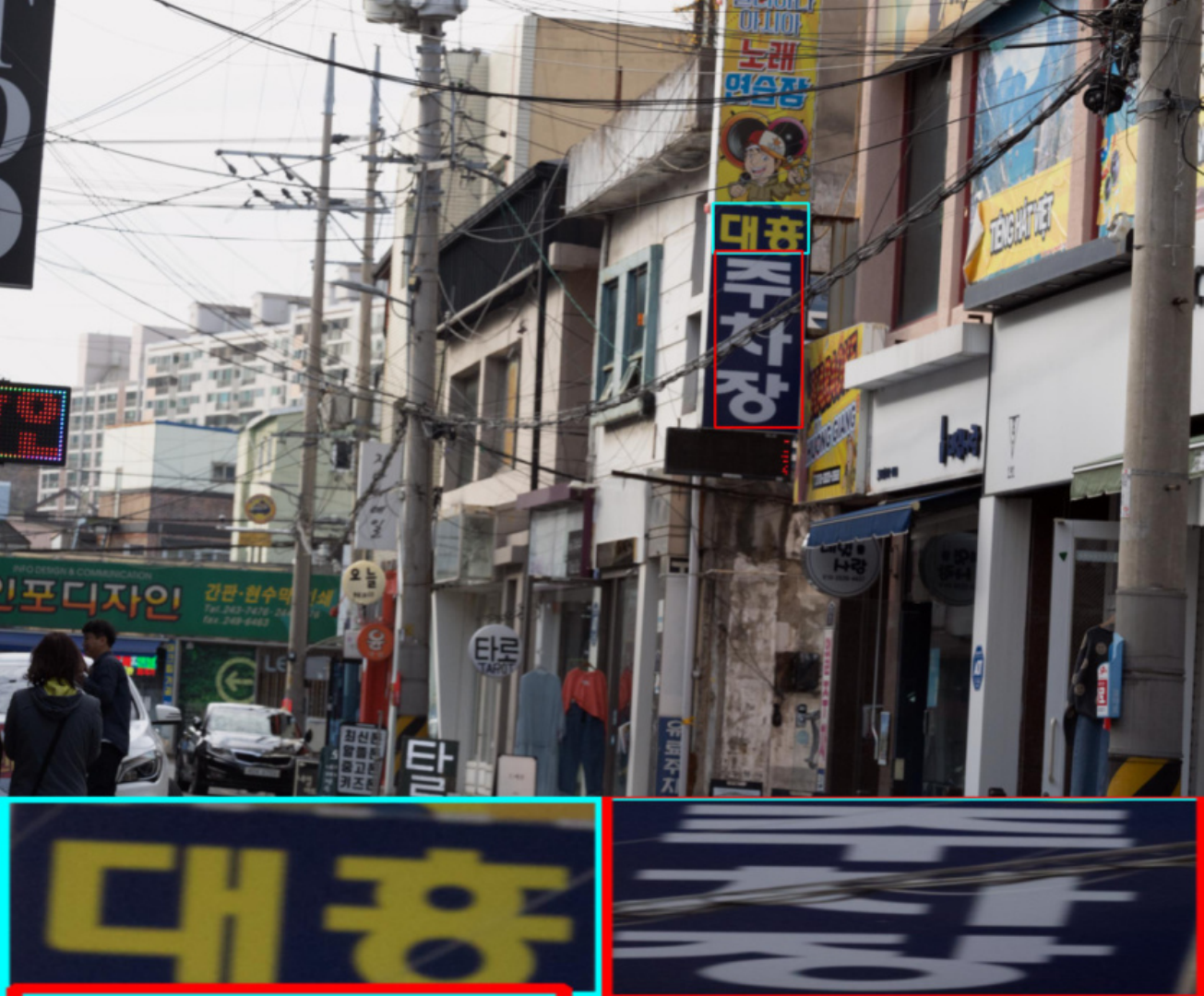}\vspace{4pt}
	\end{subfigure}
	\vfill
	
	\begin{subfigure}[t]{0.22\linewidth}	
		\includegraphics[width=4cm,height=4cm]{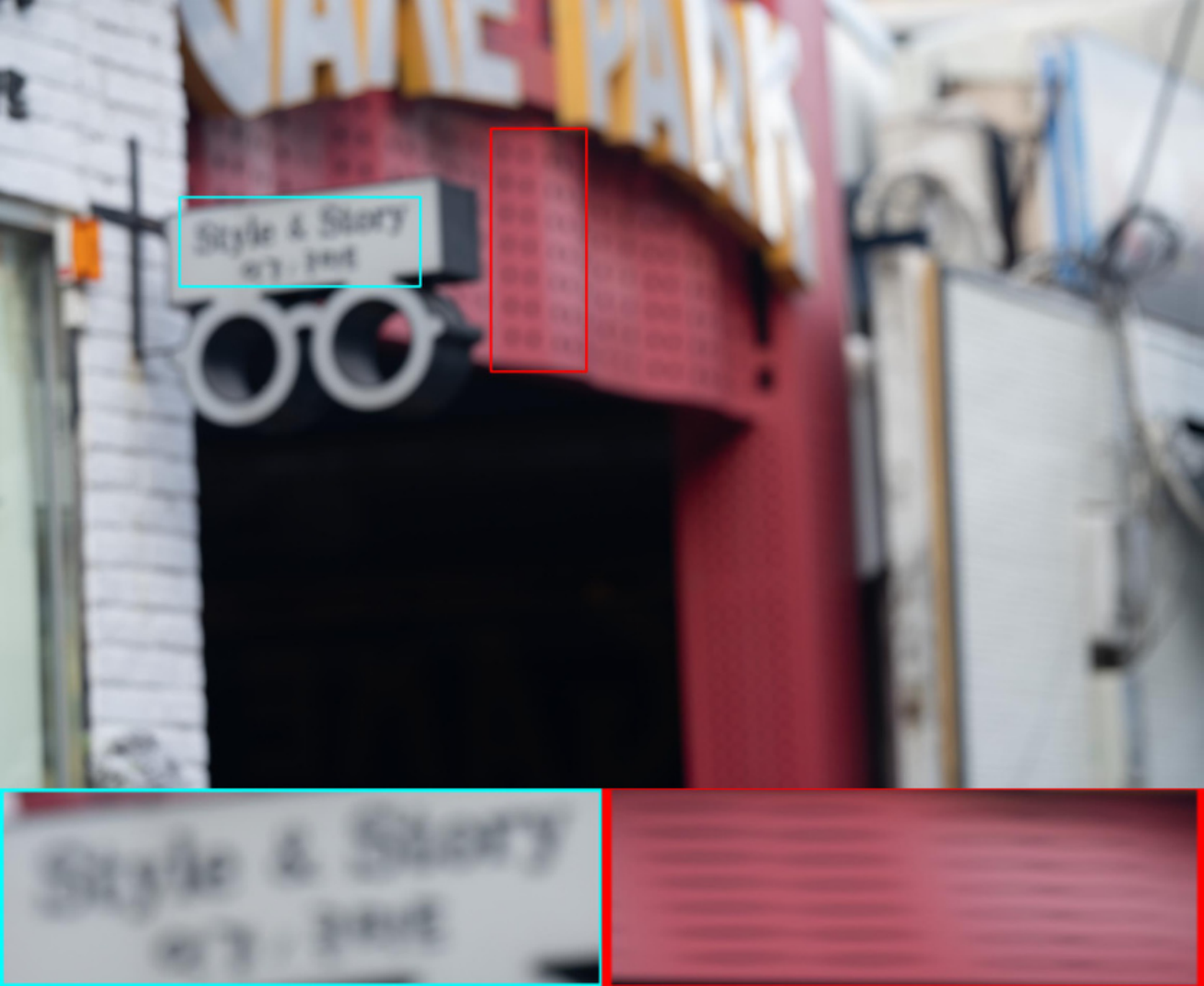}\vspace{4pt}
	\end{subfigure}
	\hfill
	\begin{subfigure}[t]{0.22\linewidth}	
		\includegraphics[width=4cm,height=4cm]{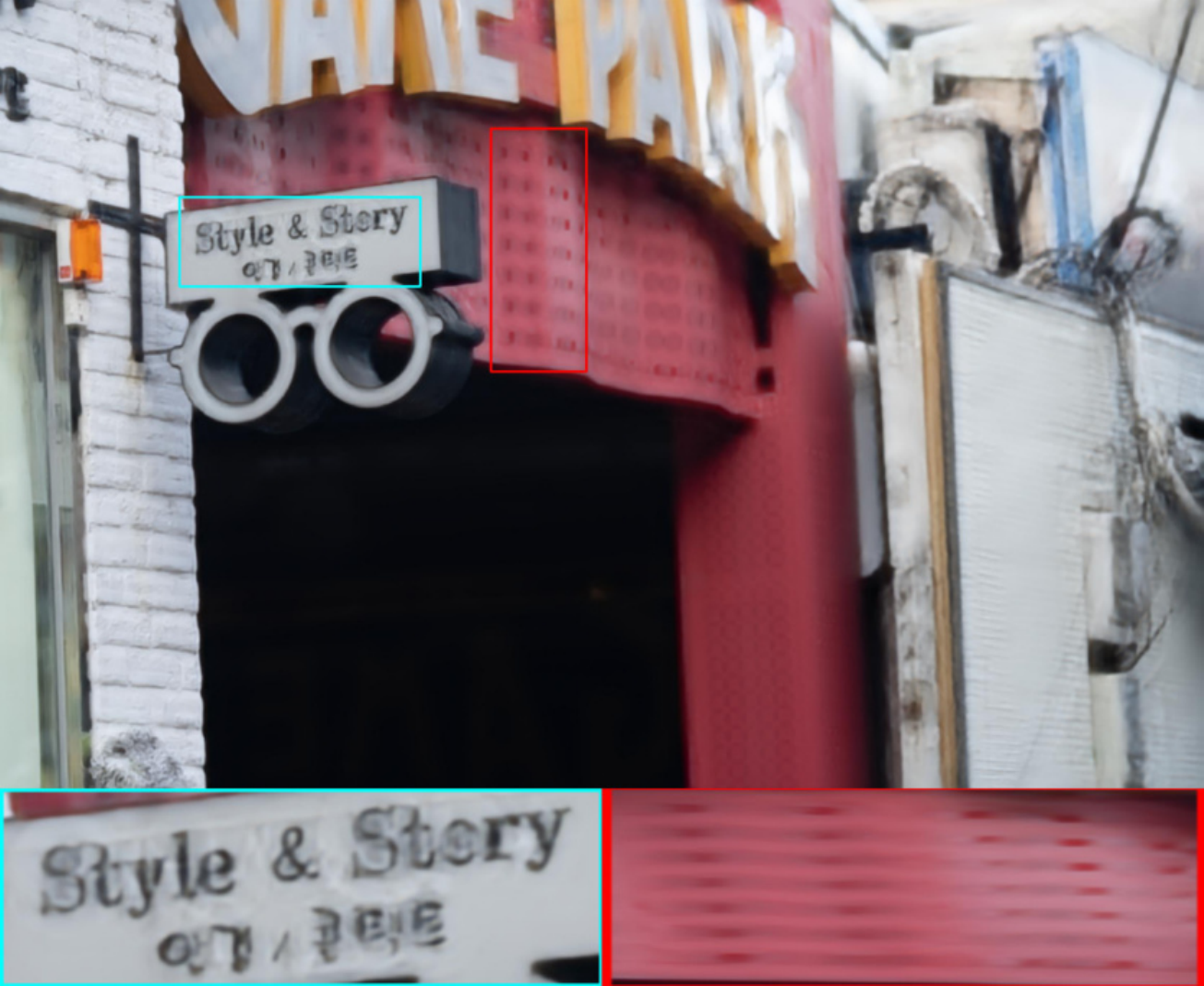}\vspace{4pt}
	\end{subfigure}
	\hfill
	\begin{subfigure}[t]{0.22\linewidth}	
		\includegraphics[width=4cm,height=4cm]{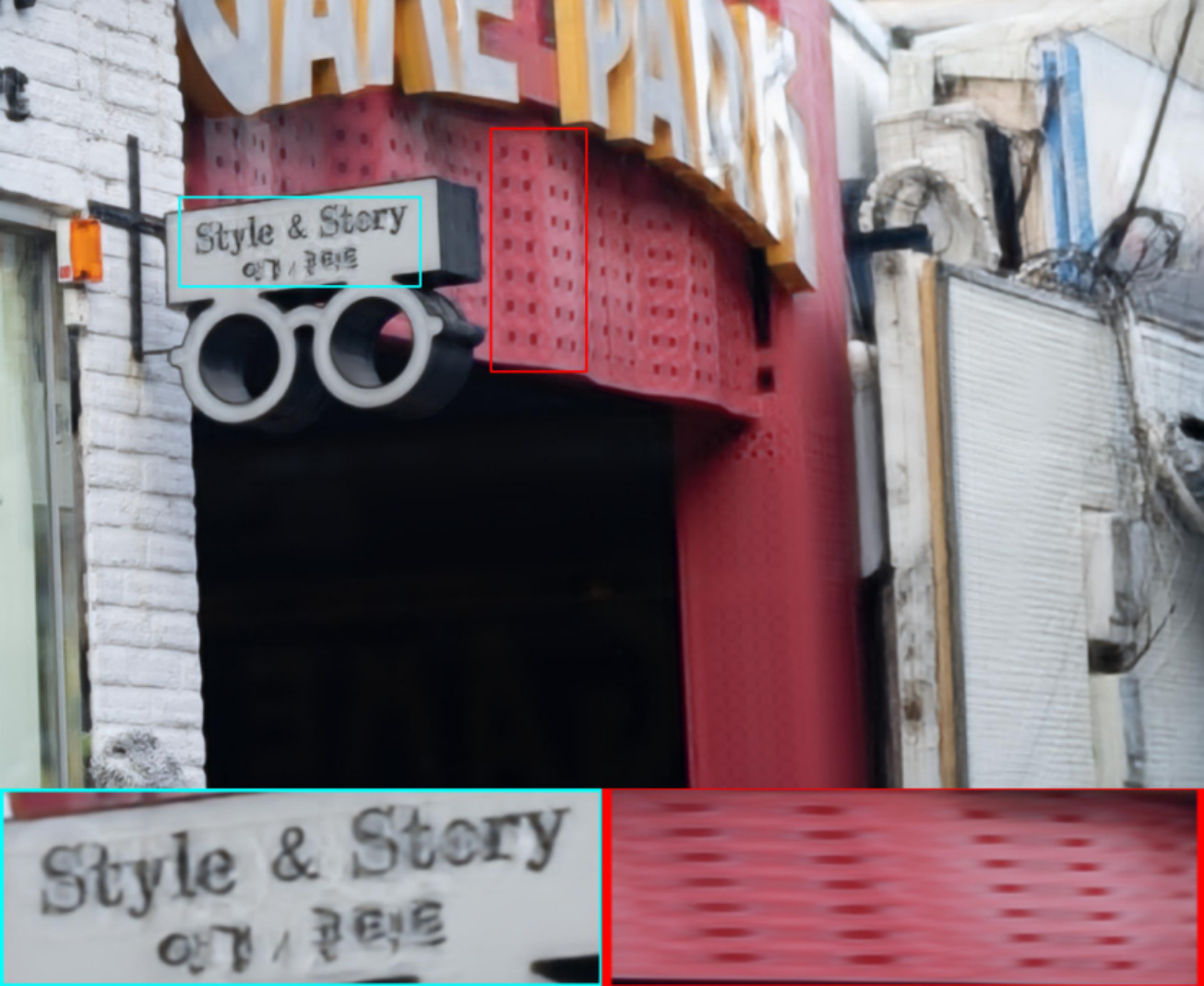}\vspace{4pt}
	\end{subfigure}
	\hfill
	\begin{subfigure}[t]{0.22\linewidth}	
		\includegraphics[width=4cm,height=4cm]{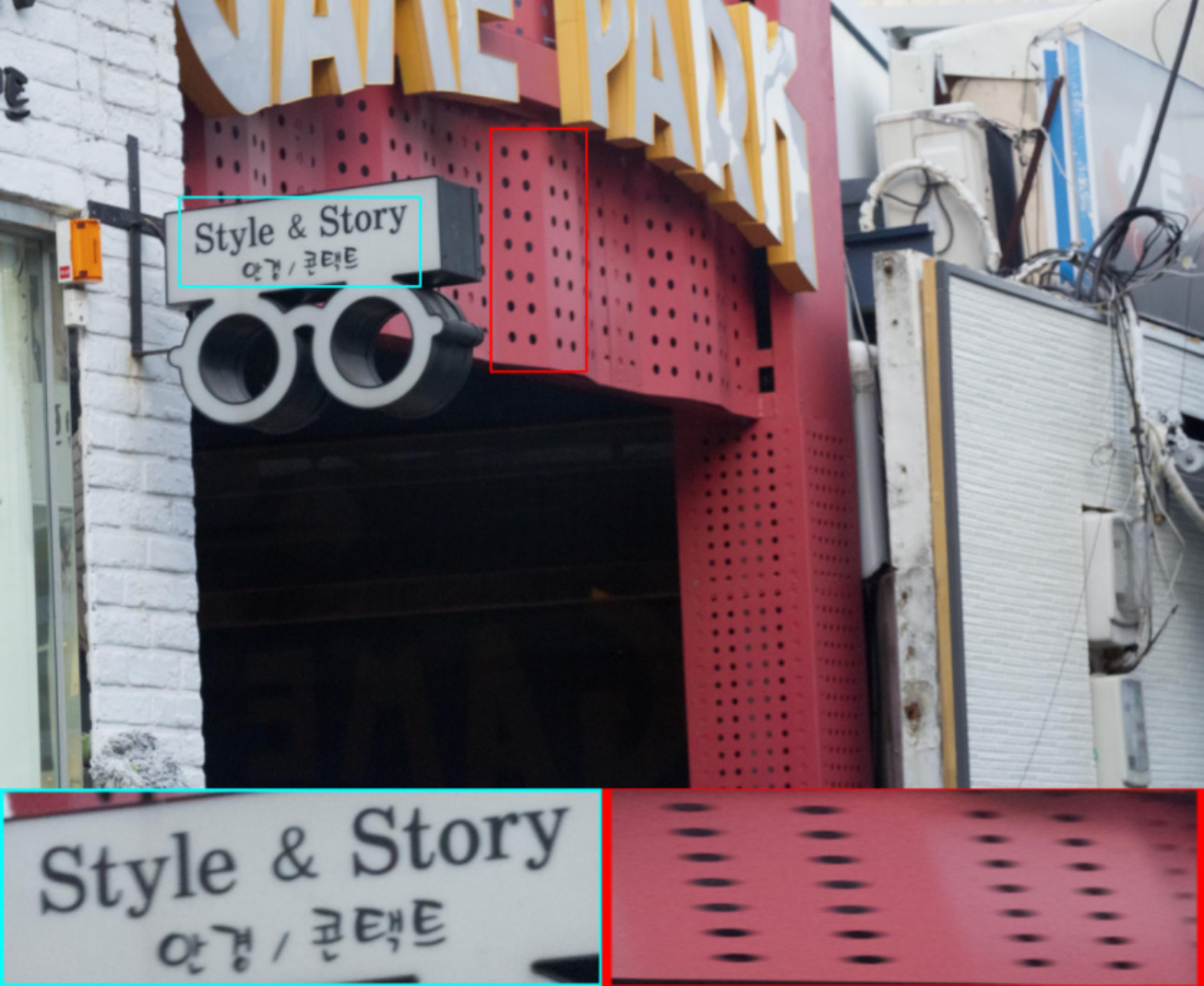}\vspace{4pt}
	\end{subfigure}
	\vfill

	\begin{subfigure}[t]{0.22\linewidth}	
		\includegraphics[width=4cm,height=4cm]{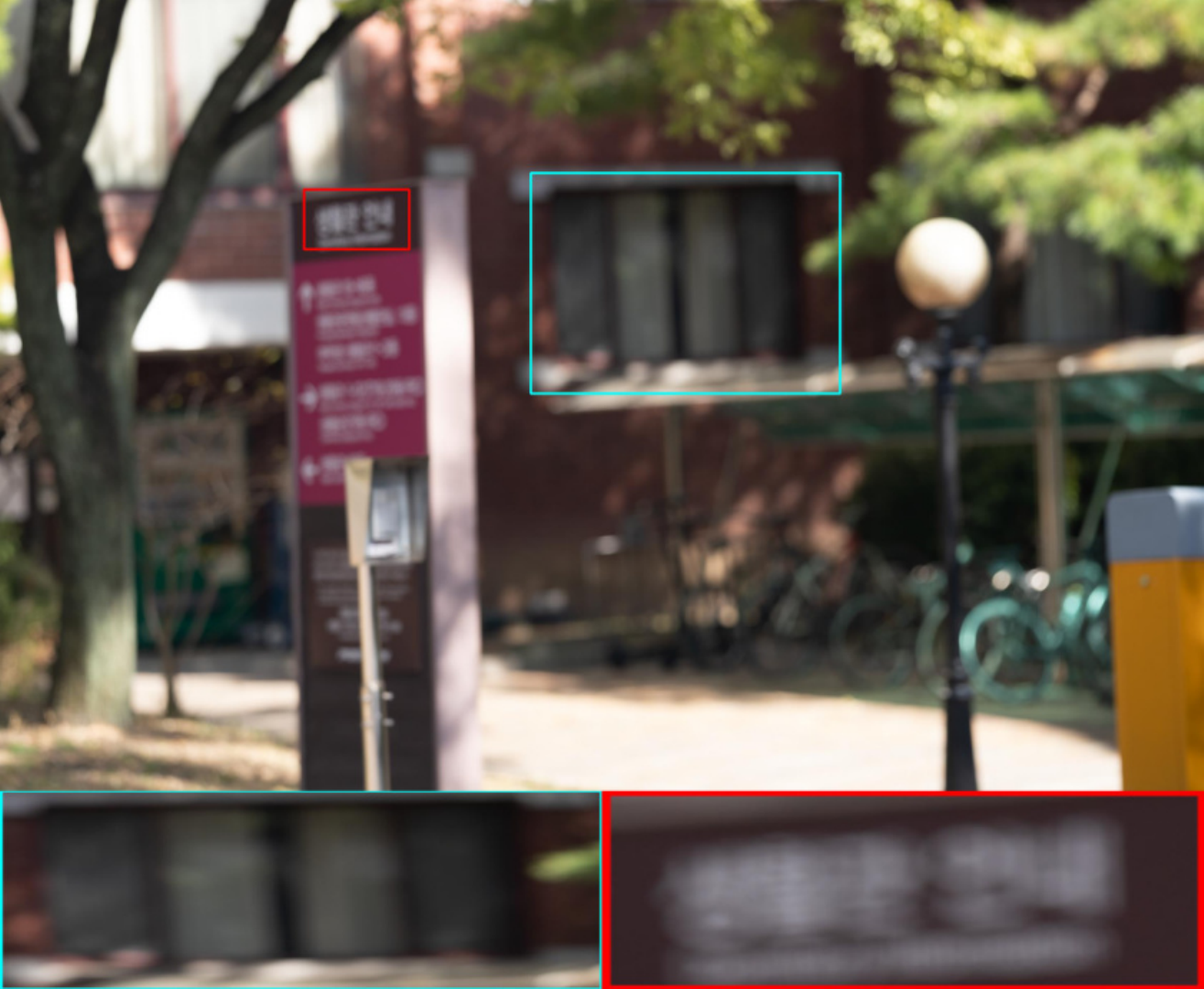}\vspace{4pt}
	\end{subfigure}
	\hfill
	\begin{subfigure}[t]{0.22\linewidth}	
		\includegraphics[width=4cm,height=4cm]{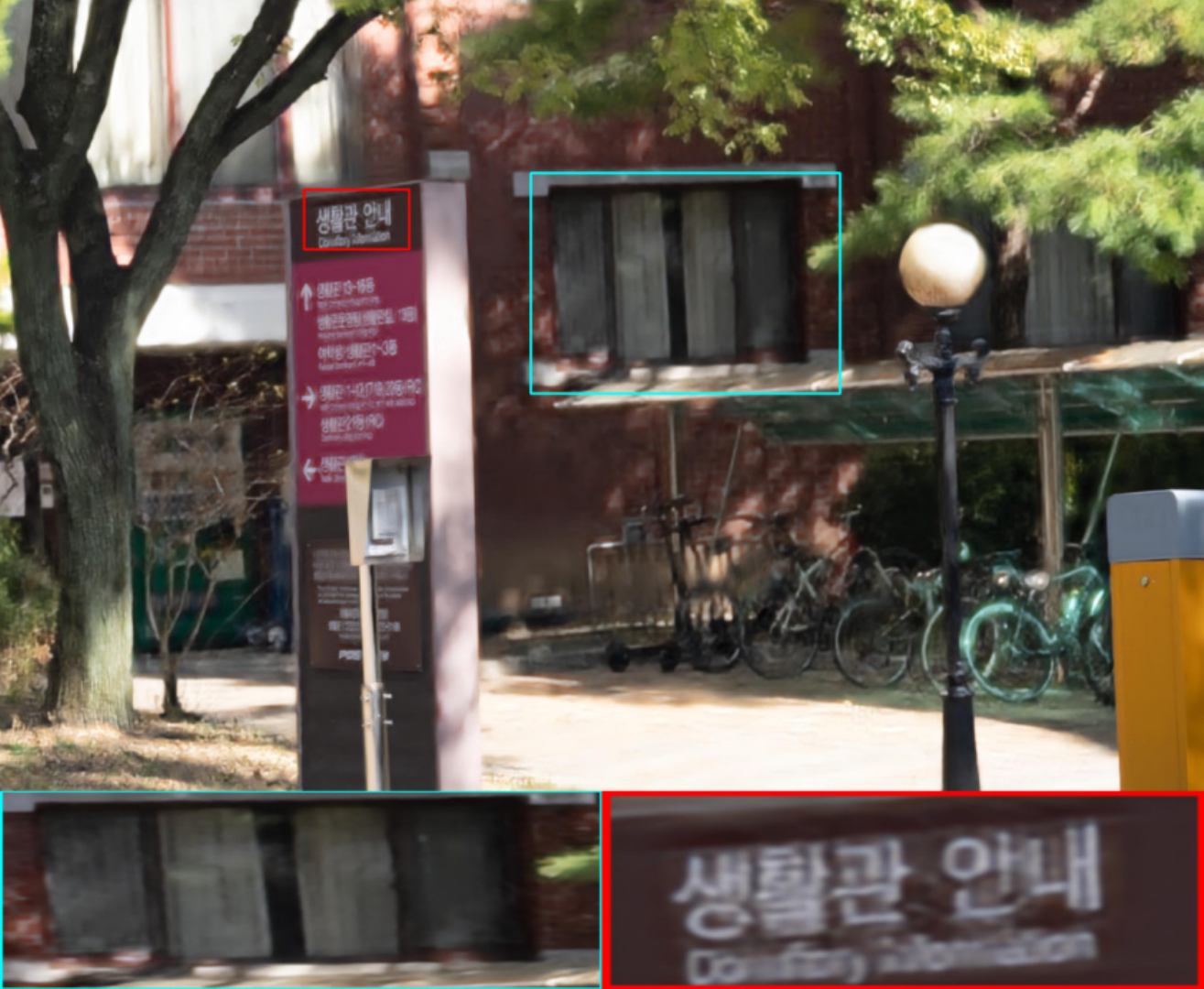}\vspace{4pt}
	\end{subfigure}
	\hfill
	\begin{subfigure}[t]{0.22\linewidth}	
		\includegraphics[width=4cm,height=4cm]{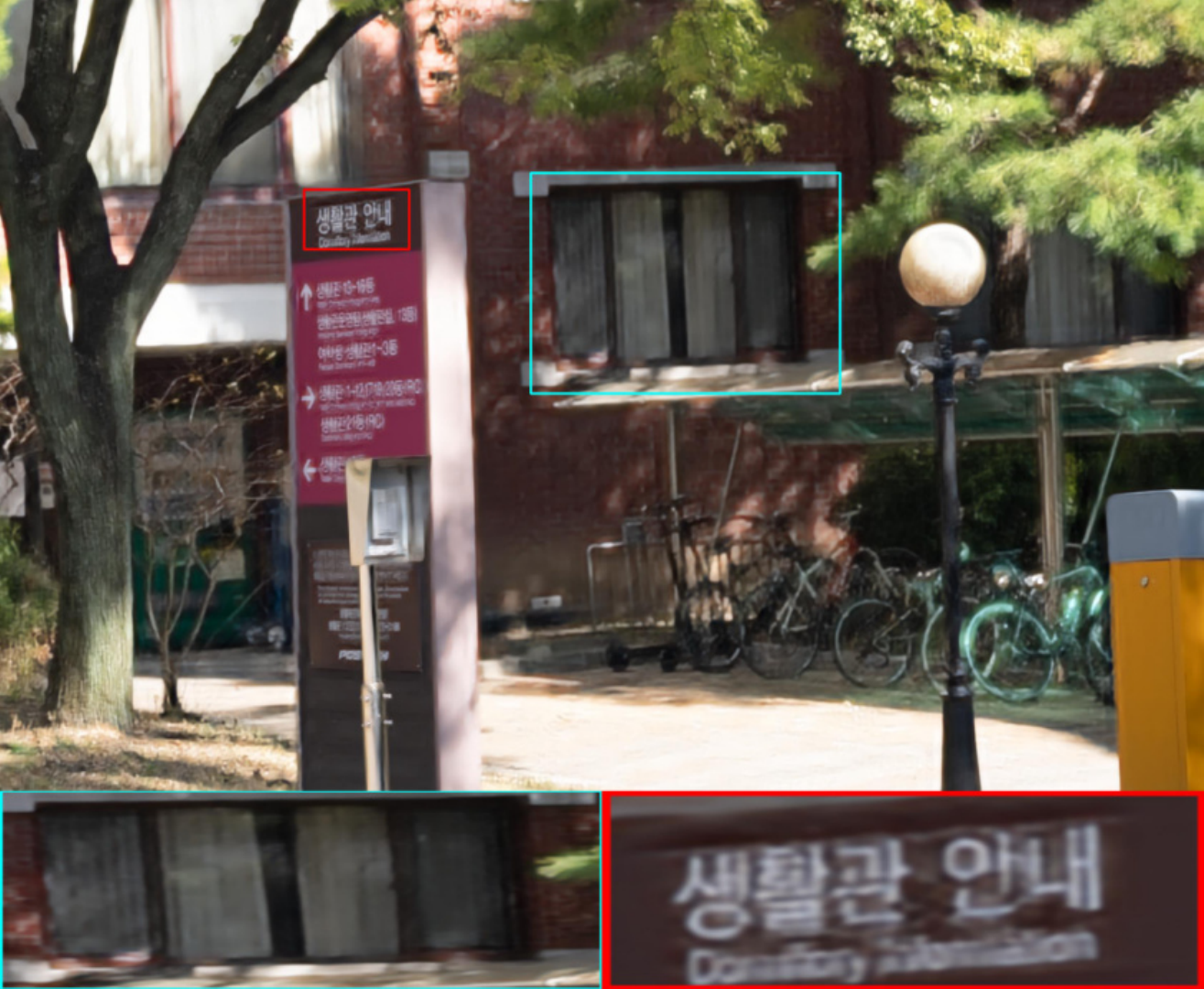}\vspace{4pt}
	\end{subfigure}
	\hfill
	\begin{subfigure}[t]{0.22\linewidth}	
		\includegraphics[width=4cm,height=4cm]{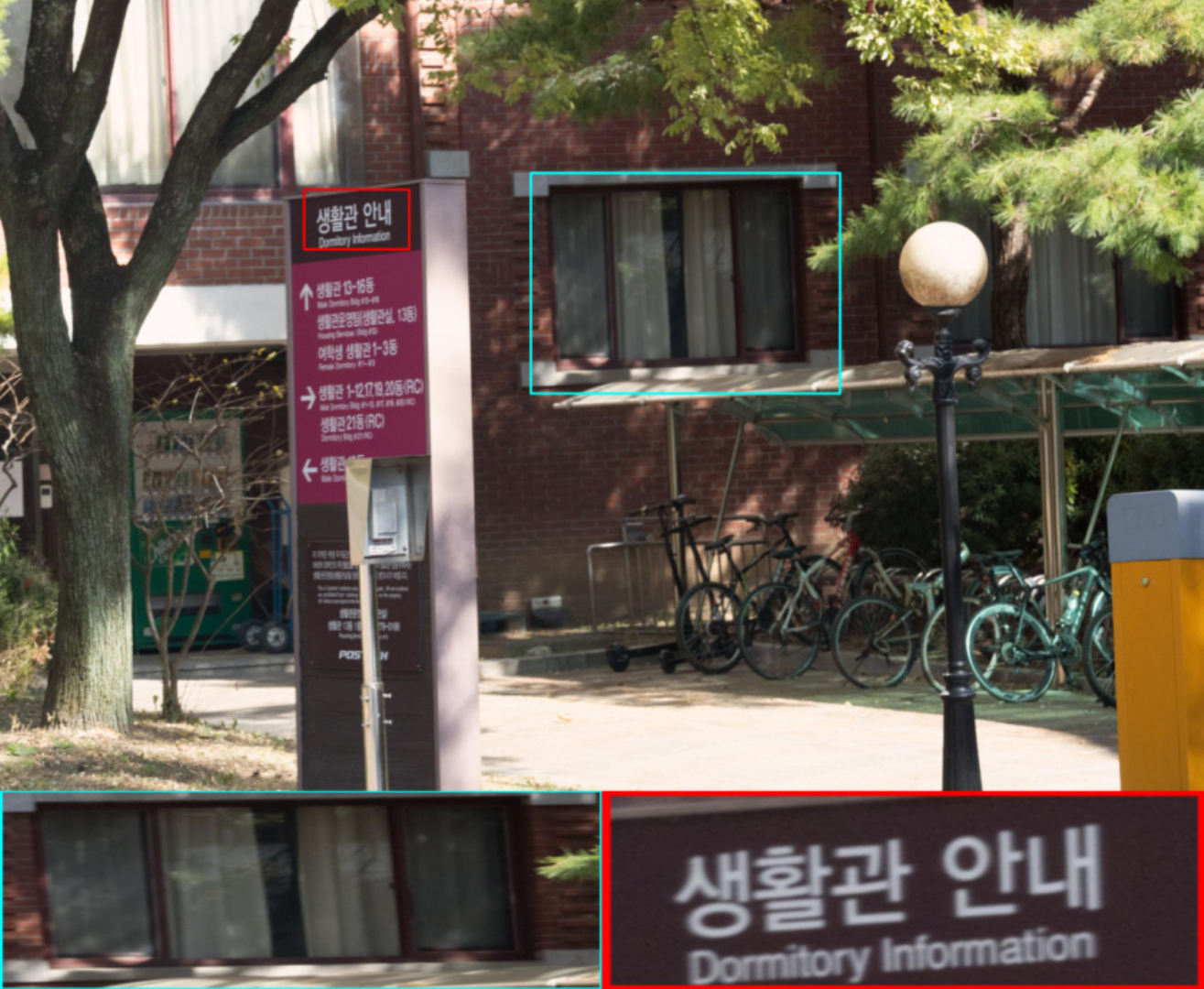}\vspace{4pt}
	\end{subfigure}
	\vfill

	
	\begin{subfigure}[t]{0.22\linewidth}	
		\includegraphics[width=4cm,height=4cm]{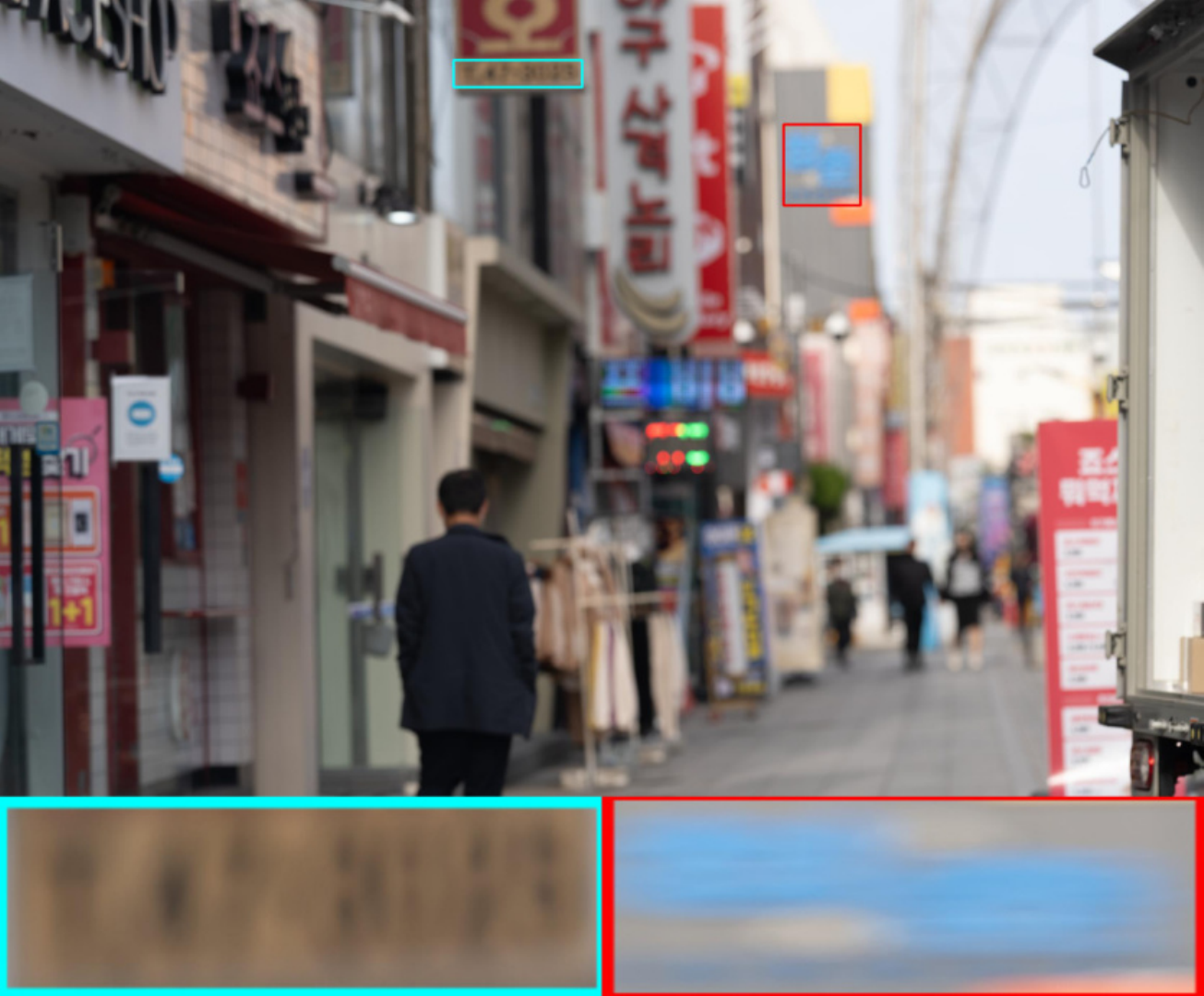}\vspace{4pt}
		\caption{Blurry Input}
	\end{subfigure}
	\hfill
	\begin{subfigure}[t]{0.22\linewidth}	
		\includegraphics[width=4cm,height=4cm]{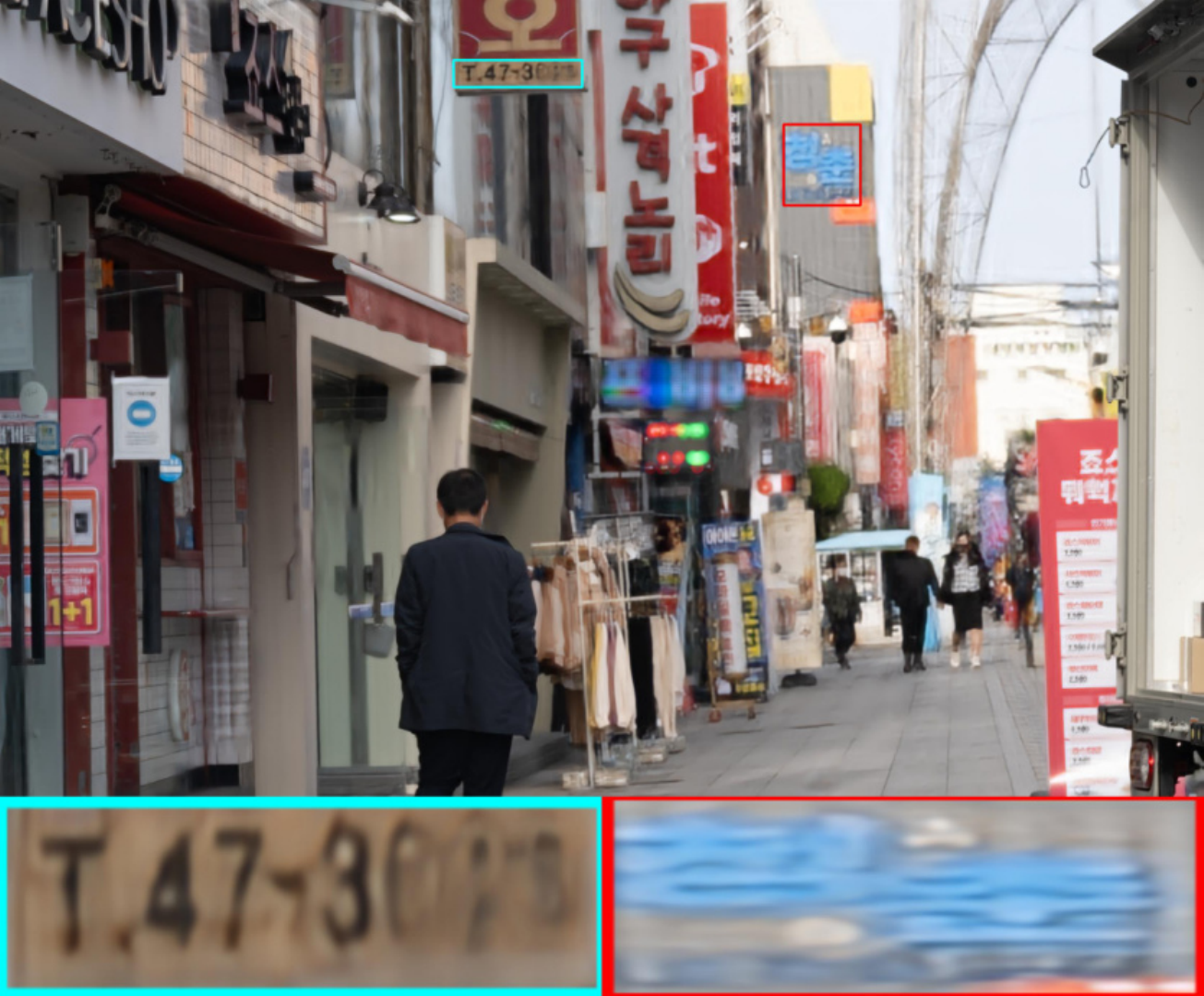}\vspace{4pt}
		\caption{KPAC}
	\end{subfigure}
	\hfill
	\begin{subfigure}[t]{0.22\linewidth}	
		\includegraphics[width=4cm,height=4cm]{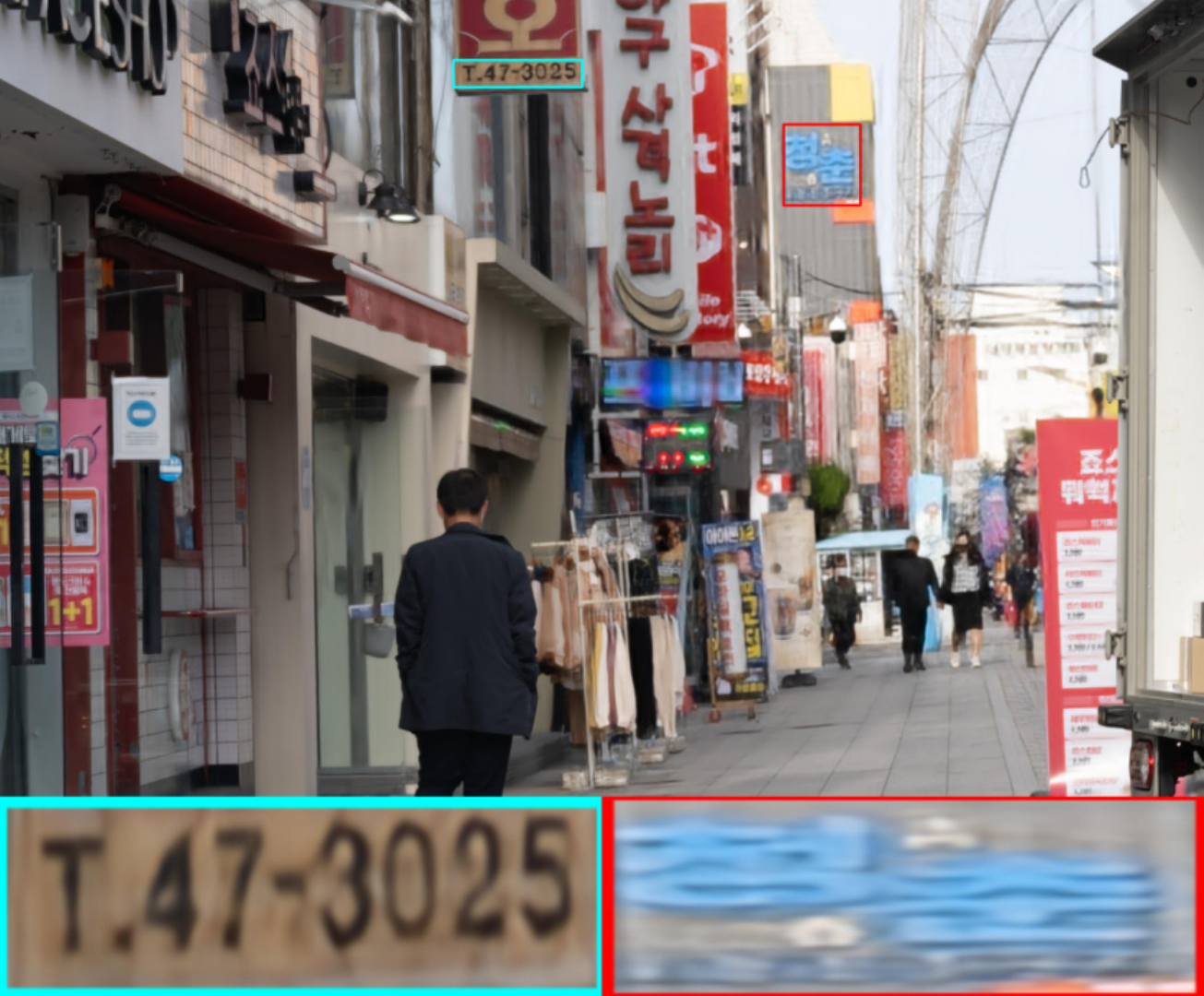}\vspace{4pt}
		\caption{R$^2$KAC}
	\end{subfigure}
	\hfill
	\begin{subfigure}[t]{0.22\linewidth}	
		\includegraphics[width=4cm,height=4cm]{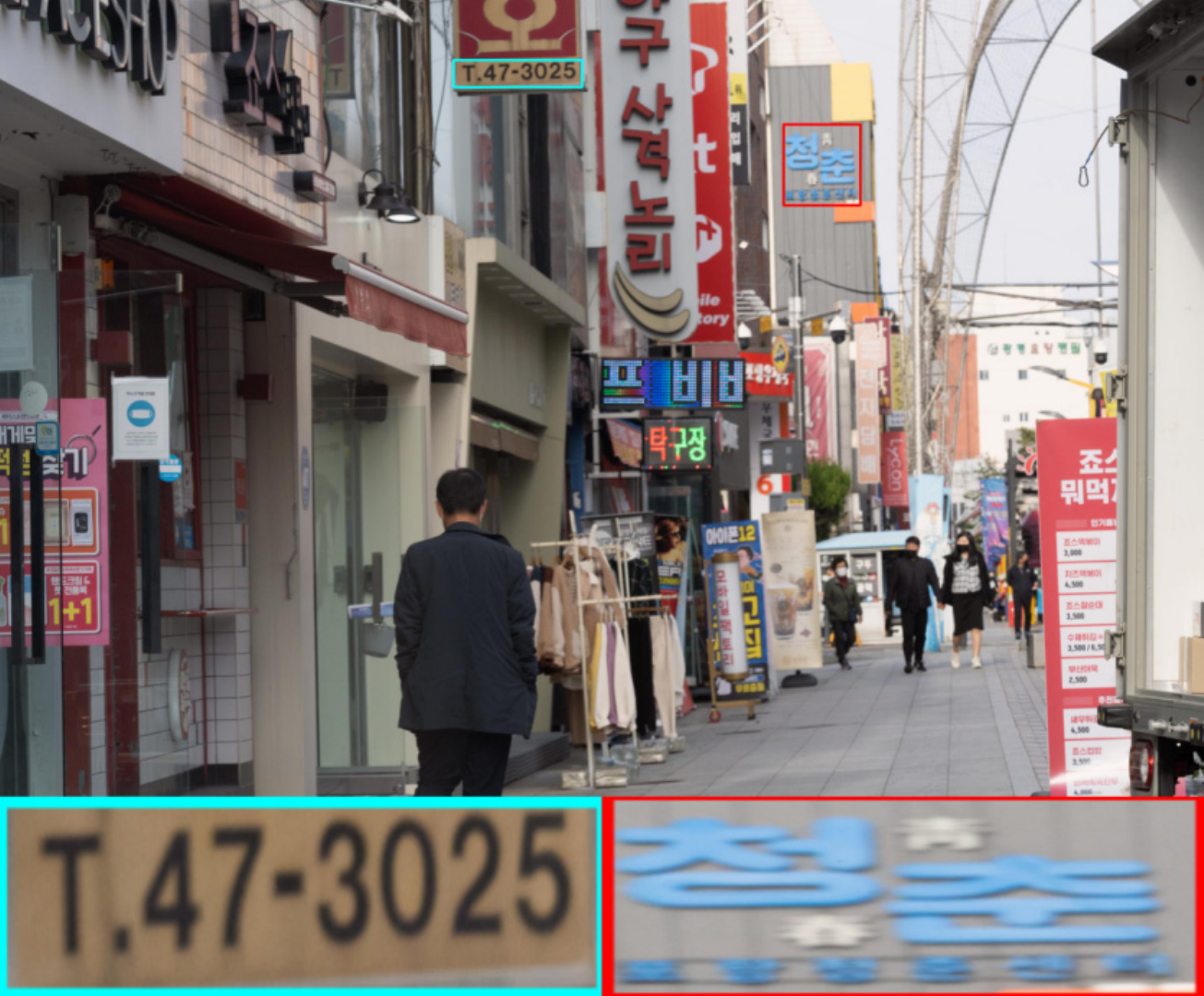}\vspace{4pt}
		
		\caption{Sharp}
	\end{subfigure}
	\vfill
	
	\caption{Visualization of deblurred large blur cases by KPAC and R$^2$KAC,}
	\label{fig11}
\end{figure*}

Except for the quantitative comparisons in the main text, we further show the visualization of deblurred images with large blurs by KPAC and R$^2$KAC in Fig.~\ref{fig11},  
given that R$^2$KAC is proposed to overcome the shortcoming of KPAC, i.e., KPAC cannot handle large blurs\footnote{An image with large blurs means that this image is blurred by large blur kernels. Visually, the image with large blurs loses most details compared to the corresponding sharp image.} well. 
For simplicity, we call the model of replacing R$^2$KAC block with KPAC block in the R$^2$KAC network as KPAC.
Specifically, Fig.~\ref{fig11} shows that R$^2$KAC can restore most details of the blurred images, giving rise to much clearer results than that deblurred by KPAC, demonstrating the superiority of R$^2$KAC in handling large blur cases to KPAC. 

\subsection{More results about experimental validation}
\begin{figure*}[h]
	\centering
	\begin{subfigure}[t]{0.18\linewidth}	
		\includegraphics[width=3cm,height=3.3cm]{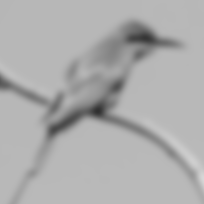}\vspace{4pt}
	\end{subfigure}
	\hfill
	\begin{subfigure}[t]{0.18\linewidth}	
		\includegraphics[width=3cm,height=3.3cm]{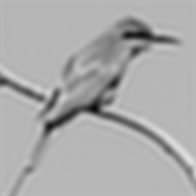}\vspace{4pt}
	\end{subfigure}
	\hfill
	\begin{subfigure}[t]{0.18\linewidth}	
		\includegraphics[width=3cm,height=3.3cm]{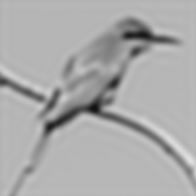}\vspace{4pt}
	\end{subfigure}
	\hfill
	\begin{subfigure}[t]{0.18\linewidth}	
		\includegraphics[width=3cm,height=3.3cm]{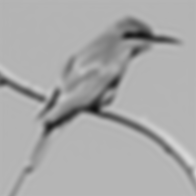}\vspace{4pt}
	\end{subfigure}
	\hfill
	\begin{subfigure}[t]{0.18\linewidth}	
		\includegraphics[width=3cm,height=3.3cm]{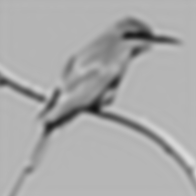}\vspace{4pt}
	\end{subfigure}
	
	\vfill
	\begin{subfigure}[t]{0.18\linewidth}	
		\includegraphics[width=3cm,height=3.3cm]{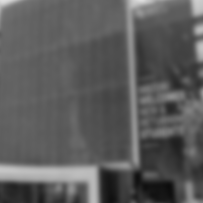}\vspace{4pt}
	\end{subfigure}
	\hfill
	\begin{subfigure}[t]{0.18\linewidth}	
		\includegraphics[width=3cm,height=3.3cm]{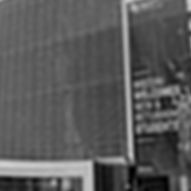}\vspace{4pt}
	\end{subfigure}
	\hfill
	\begin{subfigure}[t]{0.18\linewidth}	
		\includegraphics[width=3cm,height=3.3cm]{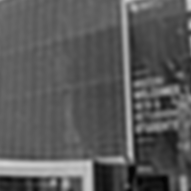}\vspace{4pt}
	\end{subfigure}
	\hfill
	\begin{subfigure}[t]{0.18\linewidth}	
		\includegraphics[width=3cm,height=3.3cm]{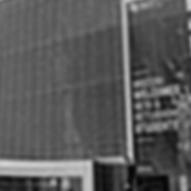}\vspace{4pt}
	\end{subfigure}
	\hfill
	\begin{subfigure}[t]{0.18\linewidth}	
		\includegraphics[width=3cm,height=3.3cm]{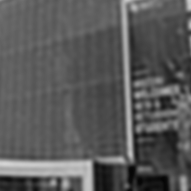}\vspace{4pt}
	\end{subfigure}
	
	\vfill
	\begin{subfigure}[t]{0.18\linewidth}	
		\includegraphics[width=3cm,height=3.3cm]{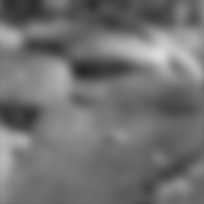}\vspace{4pt}
	\end{subfigure}
	\hfill
	\begin{subfigure}[t]{0.18\linewidth}	
		\includegraphics[width=3cm,height=3.3cm]{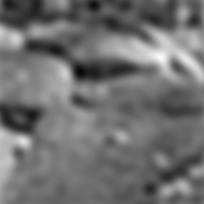}\vspace{4pt}
	\end{subfigure}
	\hfill
	\begin{subfigure}[t]{0.18\linewidth}	
		\includegraphics[width=3cm,height=3.3cm]{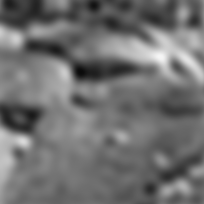}\vspace{4pt}
	\end{subfigure}
	\hfill
	\begin{subfigure}[t]{0.18\linewidth}	
		\includegraphics[width=3cm,height=3.3cm]{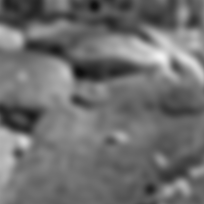}\vspace{4pt}
	\end{subfigure}
	\hfill
	\begin{subfigure}[t]{0.18\linewidth}	
		\includegraphics[width=3cm,height=3.3cm]{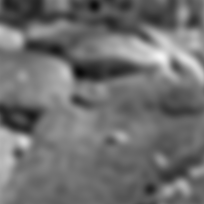}\vspace{4pt}
	\end{subfigure}
	
	\vfill
	\begin{subfigure}[t]{0.18\linewidth}	
		\includegraphics[width=3cm,height=3.3cm]{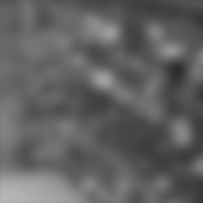}\vspace{4pt}
	\end{subfigure}
	\hfill
	\begin{subfigure}[t]{0.18\linewidth}	
		\includegraphics[width=3cm,height=3.3cm]{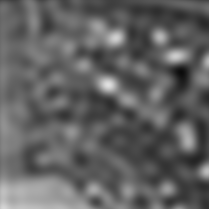}\vspace{4pt}
	\end{subfigure}
	\hfill
	\begin{subfigure}[t]{0.18\linewidth}	
		\includegraphics[width=3cm,height=3.3cm]{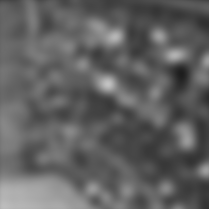}\vspace{4pt}
	\end{subfigure}
	\hfill
	\begin{subfigure}[t]{0.18\linewidth}	
		\includegraphics[width=3cm,height=3.3cm]{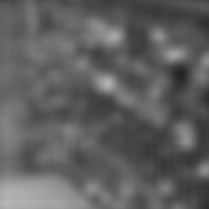}
	\end{subfigure}
	\hfill
	\begin{subfigure}[t]{0.18\linewidth}	
		\includegraphics[width=3cm,height=3.3cm]{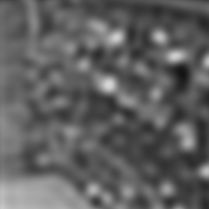}\vspace{4pt}
	\end{subfigure}
	
	\vfill
	\begin{subfigure}[t]{0.18\linewidth}	
		\includegraphics[width=3cm,height=3.3cm]{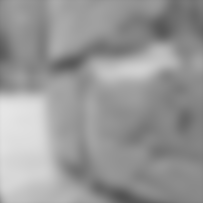}\vspace{4pt}
		\caption{}
	\end{subfigure}
	\hfill
	\begin{subfigure}[t]{0.18\linewidth}	
		\includegraphics[width=3cm,height=3.3cm]{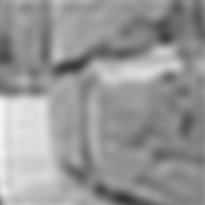}\vspace{4pt}
		\caption{}
	\end{subfigure}
	\hfill
	\begin{subfigure}[t]{0.18\linewidth}	
		\includegraphics[width=3cm,height=3.3cm]{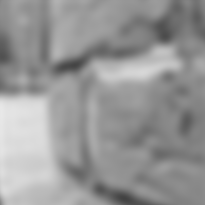}\vspace{4pt}
		\caption{}
	\end{subfigure}
	\hfill
	\begin{subfigure}[t]{0.18\linewidth}	
		\includegraphics[width=3cm,height=3.3cm]{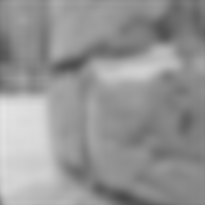}
		\caption{}
	\end{subfigure}
	\hfill
	\begin{subfigure}[t]{0.18\linewidth}	
		\includegraphics[width=3cm,height=3.3cm]{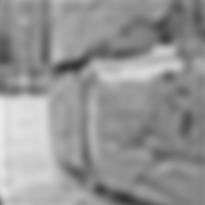}\vspace{4pt}
		\caption{}
	\end{subfigure}
	
	\caption{ (a) blurred by  $\frac{1}{s_t^{2}} k_{\uparrow s_t}$, (b) deblurred by $\frac{1}{s_t^{2}} k^{\dagger}_{\ \uparrow s_t}$, (c) deblurred by KPAC (see Eq. (5) in the main text), (d)  deblurred by RKAC (see Eq. (11) in the main text)   , (e) deblurred by R$^2$KAC (see Eq. (12) in the main text). In these cases, $s_t$ is the target scale factor and the max scaling factor  of (c),(d) and (e) is $5=\max s_i$. $s_t$ are $2.4$, $1.7$, $3.5$, $4.2$ and $3.7$ from top to bottom rows, which represent spatially-varying defocus blur. 
		Approximate accuracies of (c), (d), and (e) from top to bottom are ($98.5\%$, $98.7\%$, $98.7\%$), ( $97.5\%$, $97.3\%$, $97.4\%$,), ($99.2\%$, $97.6\%$, $98.1\%$), ($98.0\%$, $96.1\%$, $97.0\%$) and ($98.8\%$, $97.6\%$, $98.4\%$), respectively. }
	\vspace{-0.2in}
	\label{fig1_1}
\end{figure*}

\begin{figure*}[h]
	\centering
	
	\vfill
	\begin{subfigure}[t]{0.18\linewidth}	
		\includegraphics[width=3cm,height=3.3cm]{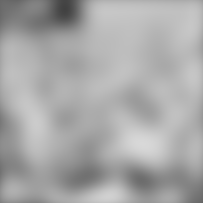}\vspace{4pt}
	\end{subfigure}
	\hfill
	\begin{subfigure}[t]{0.18\linewidth}	
		\includegraphics[width=3cm,height=3.3cm]{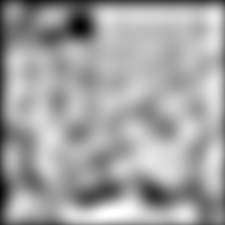}\vspace{4pt}
	\end{subfigure}
	\hfill
	\begin{subfigure}[t]{0.18\linewidth}	
		\includegraphics[width=3cm,height=3.3cm]{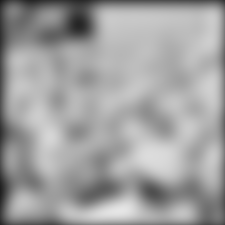}\vspace{4pt}
	\end{subfigure}
	\hfill
	\begin{subfigure}[t]{0.18\linewidth}	
		\includegraphics[width=3cm,height=3.3cm]{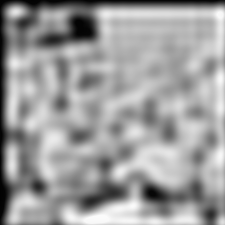}\vspace{4pt}
	\end{subfigure}
	\hfill
	\begin{subfigure}[t]{0.18\linewidth}	
		\includegraphics[width=3cm,height=3.3cm]{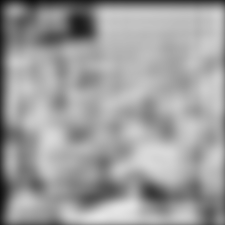}\vspace{4pt}
	\end{subfigure}
	
	\vfill
	\begin{subfigure}[t]{0.18\linewidth}	
		\includegraphics[width=3cm,height=3.3cm]{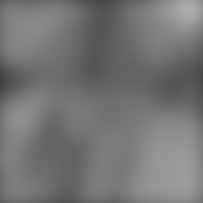}\vspace{4pt}
	\end{subfigure}
	\hfill
	\begin{subfigure}[t]{0.18\linewidth}	
		\includegraphics[width=3cm,height=3.3cm]{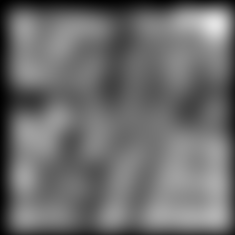}\vspace{4pt}
	\end{subfigure}
	\hfill
	\begin{subfigure}[t]{0.18\linewidth}	
		\includegraphics[width=3cm,height=3.3cm]{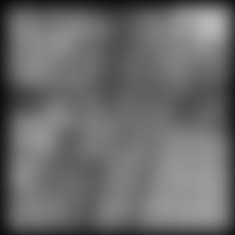}\vspace{4pt}
	\end{subfigure}
	\hfill
	\begin{subfigure}[t]{0.18\linewidth}	
		\includegraphics[width=3cm,height=3.3cm]{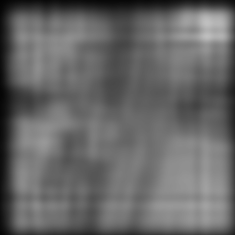}\vspace{4pt}
	\end{subfigure}
	\hfill
	\begin{subfigure}[t]{0.18\linewidth}	
		\includegraphics[width=3cm,height=3.3cm]{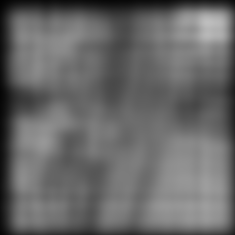}\vspace{4pt}
	\end{subfigure}
	
	\vfill
	\begin{subfigure}[t]{0.18\linewidth}	
		\includegraphics[width=3cm,height=3.3cm]{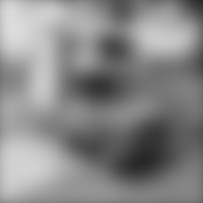}\vspace{4pt}
	\end{subfigure}
	\hfill
	\begin{subfigure}[t]{0.18\linewidth}	
		\includegraphics[width=3cm,height=3.3cm]{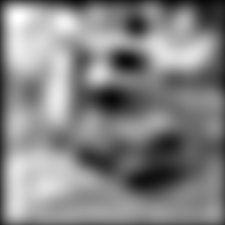}\vspace{4pt}
	\end{subfigure}
	\hfill
	\begin{subfigure}[t]{0.18\linewidth}	
		\includegraphics[width=3cm,height=3.3cm]{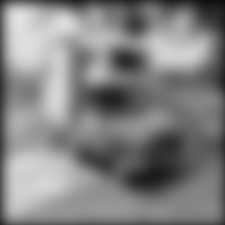}\vspace{4pt}
	\end{subfigure}
	\hfill
	\begin{subfigure}[t]{0.18\linewidth}	
		\includegraphics[width=3cm,height=3.3cm]{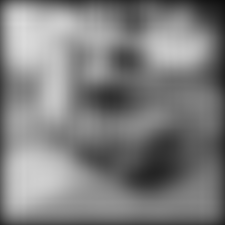}\vspace{4pt}
	\end{subfigure}
	\hfill
	\begin{subfigure}[t]{0.18\linewidth}	
		\includegraphics[width=3cm,height=3.3cm]{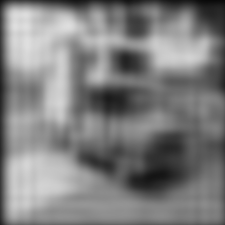}\vspace{4pt}
	\end{subfigure}
	
	\vfill
	\begin{subfigure}[t]{0.18\linewidth}	
		\includegraphics[width=3cm,height=3.3cm]{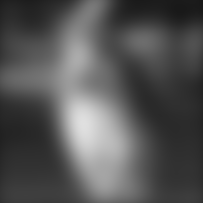}\vspace{4pt}
	\end{subfigure}
	\hfill
	\begin{subfigure}[t]{0.18\linewidth}	
		\includegraphics[width=3cm,height=3.3cm]{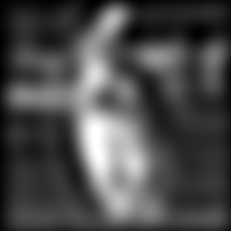}\vspace{4pt}
	\end{subfigure}
	\hfill
	\begin{subfigure}[t]{0.18\linewidth}	
		\includegraphics[width=3cm,height=3.3cm]{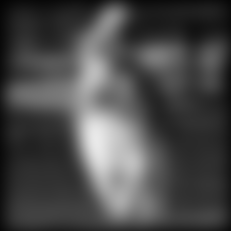}\vspace{4pt}
	\end{subfigure}
	\hfill
	\begin{subfigure}[t]{0.18\linewidth}	
		\includegraphics[width=3cm,height=3.3cm]{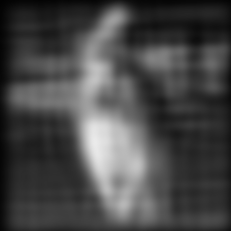}
	\end{subfigure}
	\hfill
	\begin{subfigure}[t]{0.18\linewidth}	
		\includegraphics[width=3cm,height=3.3cm]{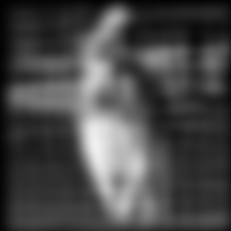}\vspace{4pt}
	\end{subfigure}
	
	\vfill
	\begin{subfigure}[t]{0.18\linewidth}	
		\includegraphics[width=3cm,height=3.3cm]{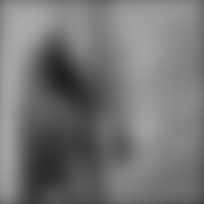}\vspace{4pt}
		\caption{}
	\end{subfigure}
	\hfill
	\begin{subfigure}[t]{0.18\linewidth}	
		\includegraphics[width=3cm,height=3.3cm]{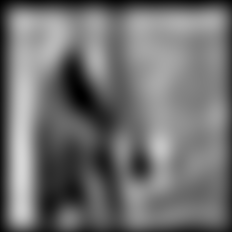}\vspace{4pt}
		\caption{}
	\end{subfigure}
	\hfill
	\begin{subfigure}[t]{0.18\linewidth}	
		\includegraphics[width=3cm,height=3.3cm]{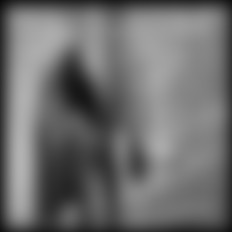}\vspace{4pt}
		\caption{}
	\end{subfigure}
	\hfill
	\begin{subfigure}[t]{0.18\linewidth}	
		\includegraphics[width=3cm,height=3.3cm]{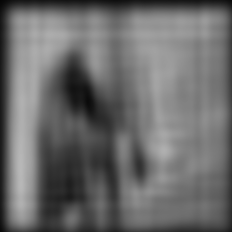}
		\caption{}
	\end{subfigure}
	\hfill
	\begin{subfigure}[t]{0.18\linewidth}	
		\includegraphics[width=3cm,height=3.3cm]{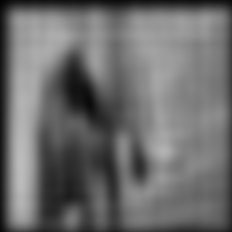}\vspace{4pt}
		\caption{}
	\end{subfigure}
	
	\caption{ (a) blurred by  $\frac{1}{s_t^{2}} k_{\uparrow s_t}$, (b) deblurred by $\frac{1}{s_t^{2}} k^{\dagger}_{\ \uparrow s_t}$, (c) deblurred by KPAC (see Eq. (5) in the main text), (d)  deblurred by RKAC (see Eq. (11) in the main text)   , (e) deblurred by R$^2$KAC (see Eq. (12) in the main text). In these cases, $s_t$ is the target scale factor and the max scaling factor  of (c),(d) and (e) is $5=\max s_i$. $s_t$ are $6.8$, $8.0$, $6.5$, $7.4$ and $7.5$ from top to bottom, which represent the unexpected large blurs $s_t>\max s_i$. 
		Approximate accuracies of (c), (d), and (e) from top to bottom are ($94.6\%$, $94.8\%$, $95.7\%$), ( $91.6\%$, $94.4\%$, $95.1\%$,), ($93.6\%$, $94.3\%$, $95.2\%$), ($89.0\%$, $91.3\%$, $92.6\%$) and ($90.1\%$, $93.0\%$, $94.1\%$), respectively. }
	\vspace{-0.2in}
	\label{fig1_2}
\end{figure*}

In this subection, we present more results regarding KPAC, RKAC, and R$^2$KAC in Figs.~\ref{fig1_1} and~\ref{fig1_2}. 
In the cases of spatially-varying blurs (see Fig.~\ref{fig1_1}), although no kernels fit the actual blur size, KPAC (\ref{fig1_1}c) and R$^2$KAC (\ref{fig1_1}e) still generate comparable results, which are almost equivalent to the deconvolution results obtained by the inverse kernel with the target size $s_t$. 
In the cases of unexpected large blurs (see Fig.~\ref{fig1_2}),  R$^2$KAC and RKAC outperform KPAC in terms of the deblurring quality of output images and approximated accuracy. 
In addition, R$^2$KAC achieves better performance than RKAC, as it removes the ringing artifacts by adding residual deconvolutions.

\subsection{More qualitative results about the comparison}
\begin{figure*}[t]
	\centering
	
	\begin{subfigure}[t]{0.16\linewidth}	
		\includegraphics[width=2.6cm,height=4cm]{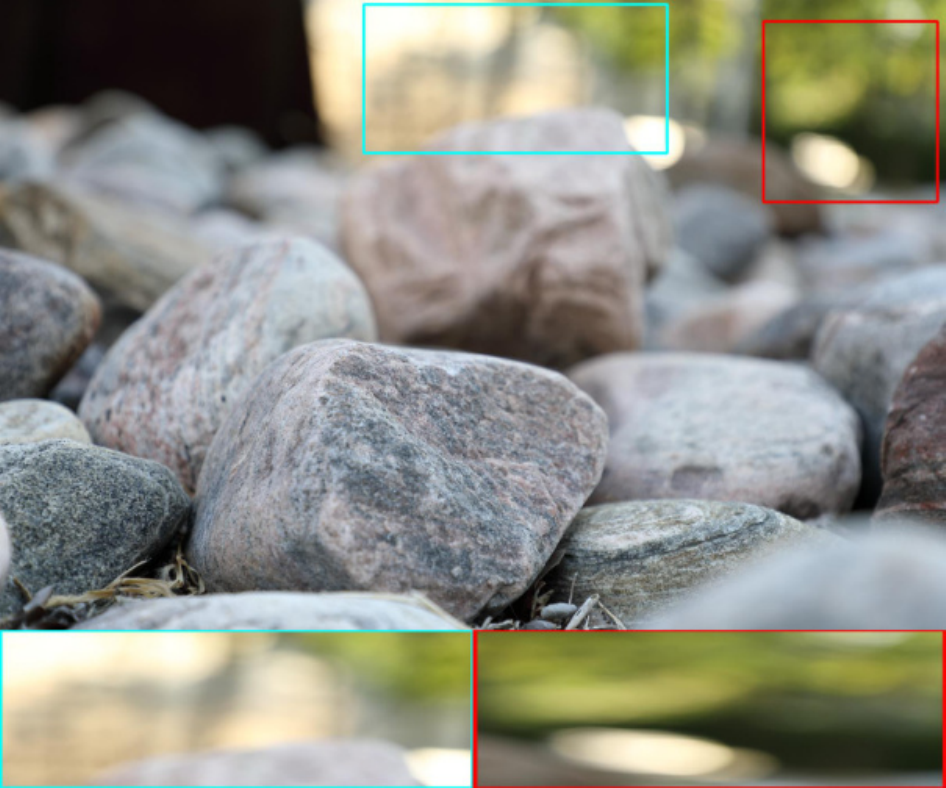}\vspace{4pt}
	\end{subfigure}
	\hfill
	\begin{subfigure}[t]{0.16\linewidth}	
		\includegraphics[width=2.6cm,height=4cm]{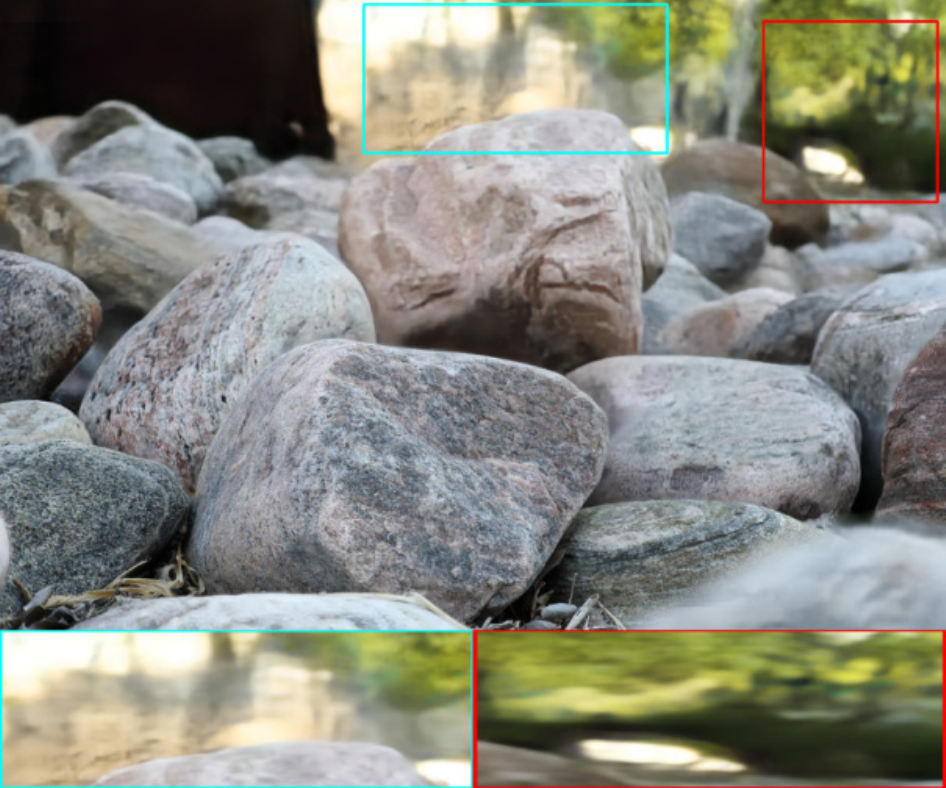}\vspace{4pt}
	\end{subfigure}
	\hfill
	\begin{subfigure}[t]{0.16\linewidth}	
		\includegraphics[width=2.6cm,height=4cm]{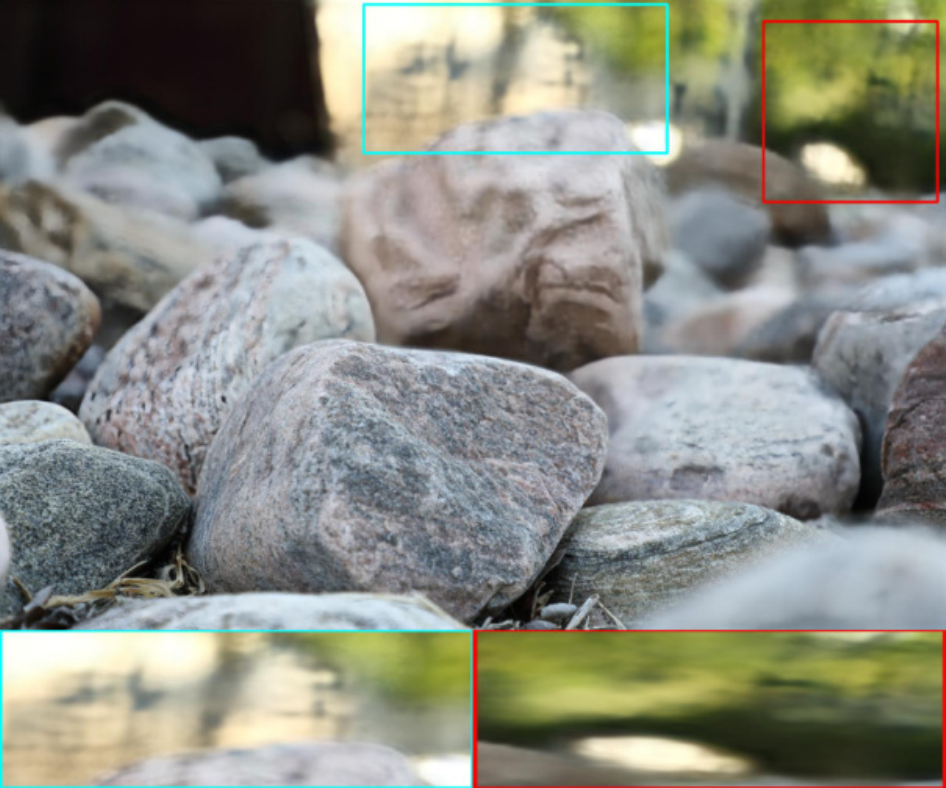}\vspace{4pt}
	\end{subfigure}
	\hfill
	\begin{subfigure}[t]{0.16\linewidth}	
		\includegraphics[width=2.6cm,height=4cm]{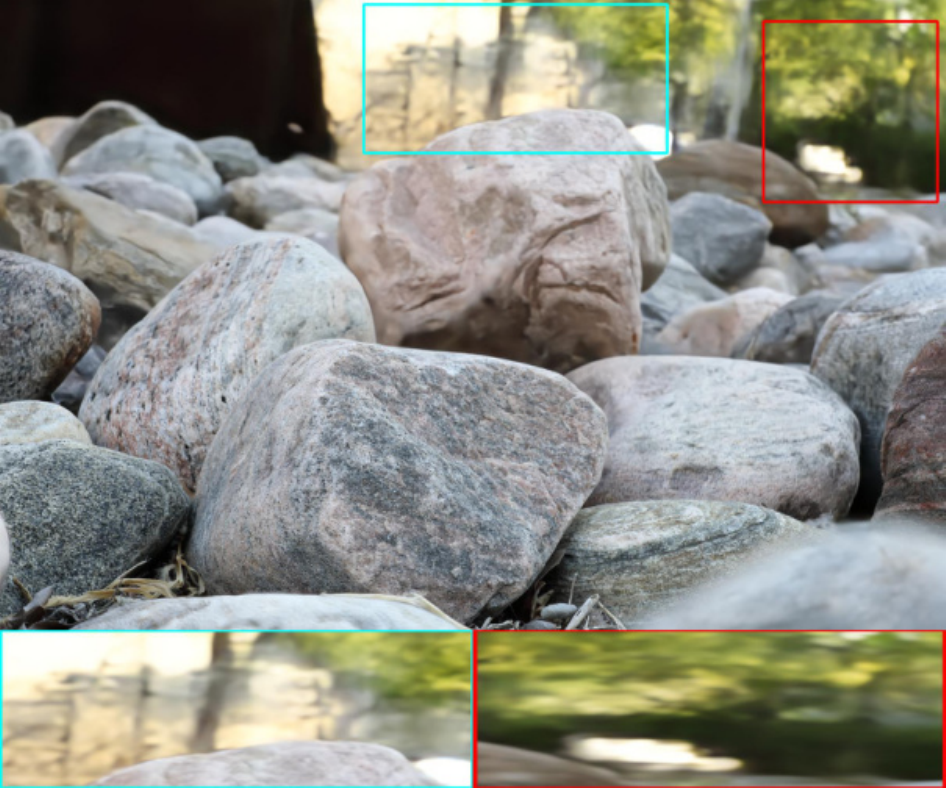}\vspace{4pt}
	\end{subfigure}
	\hfill
	\begin{subfigure}[t]{0.16\linewidth}	
		\includegraphics[width=2.6cm,height=4cm]{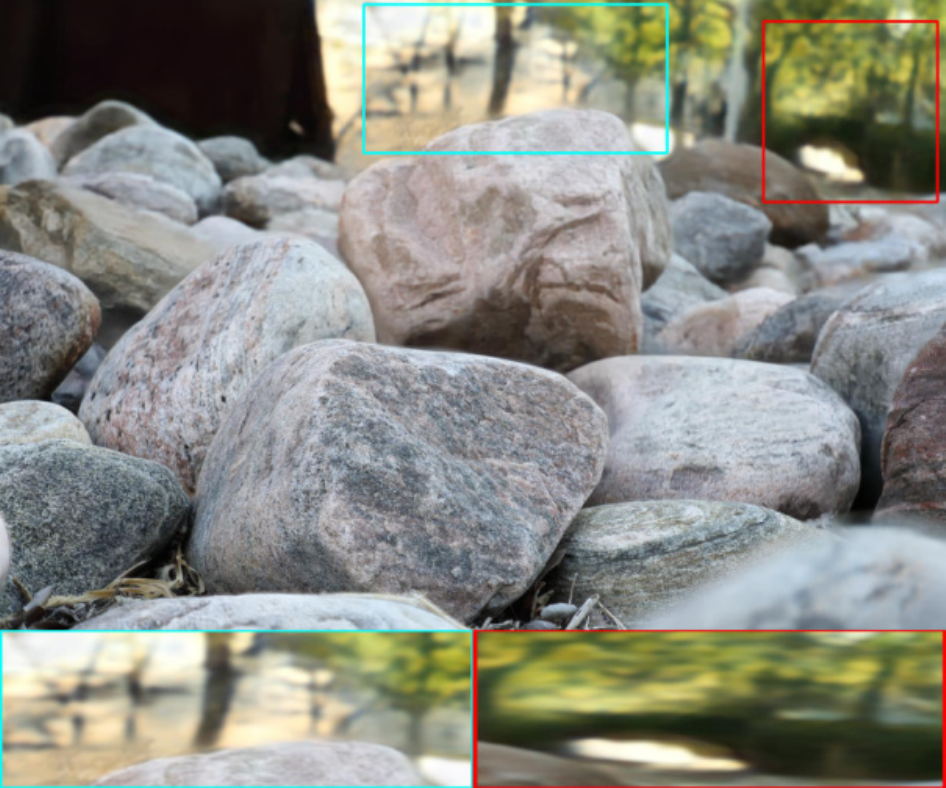}\vspace{4pt}
	\end{subfigure}
	\hfill
	\begin{subfigure}[t]{0.16\linewidth}	
		\includegraphics[width=2.6cm,height=4cm]{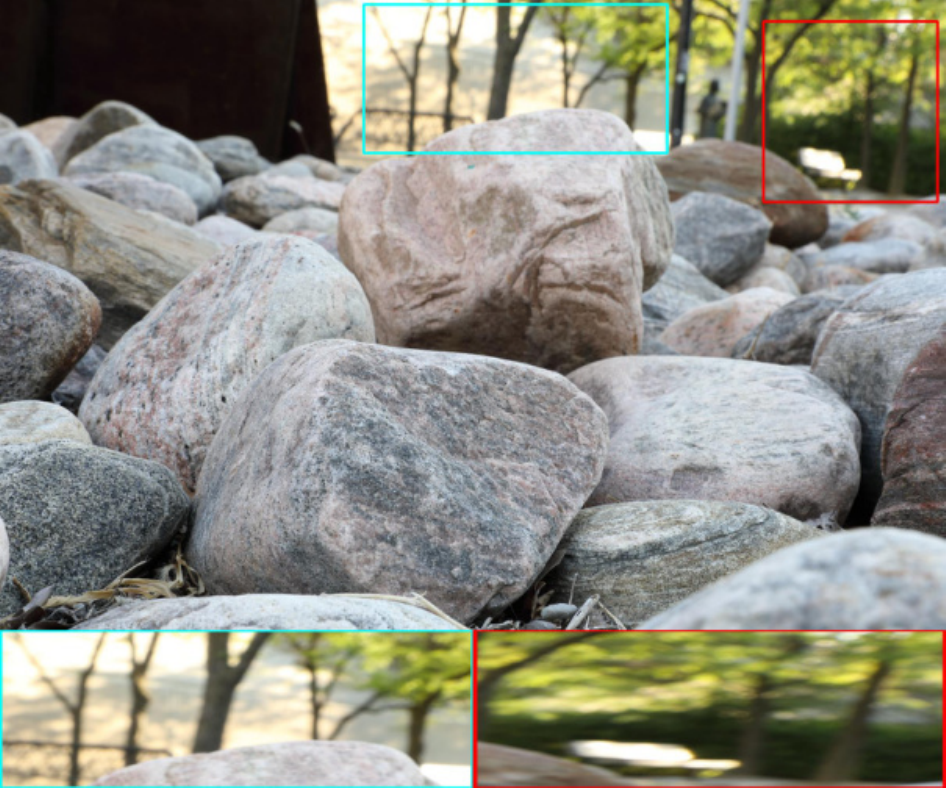}\vspace{4pt}
	\end{subfigure}
	\vfill
	
	\begin{subfigure}[t]{0.16\linewidth}	
		\includegraphics[width=2.6cm,height=4cm]{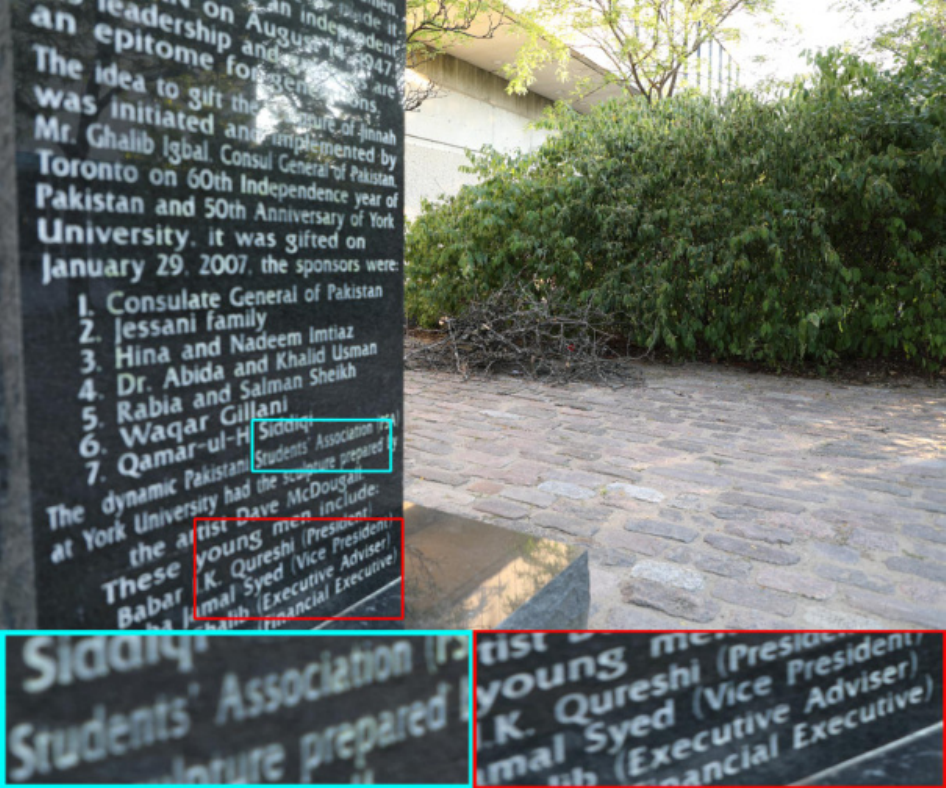}\vspace{4pt}
	\end{subfigure}
	\hfill
	\begin{subfigure}[t]{0.16\linewidth}	
		\includegraphics[width=2.6cm,height=4cm]{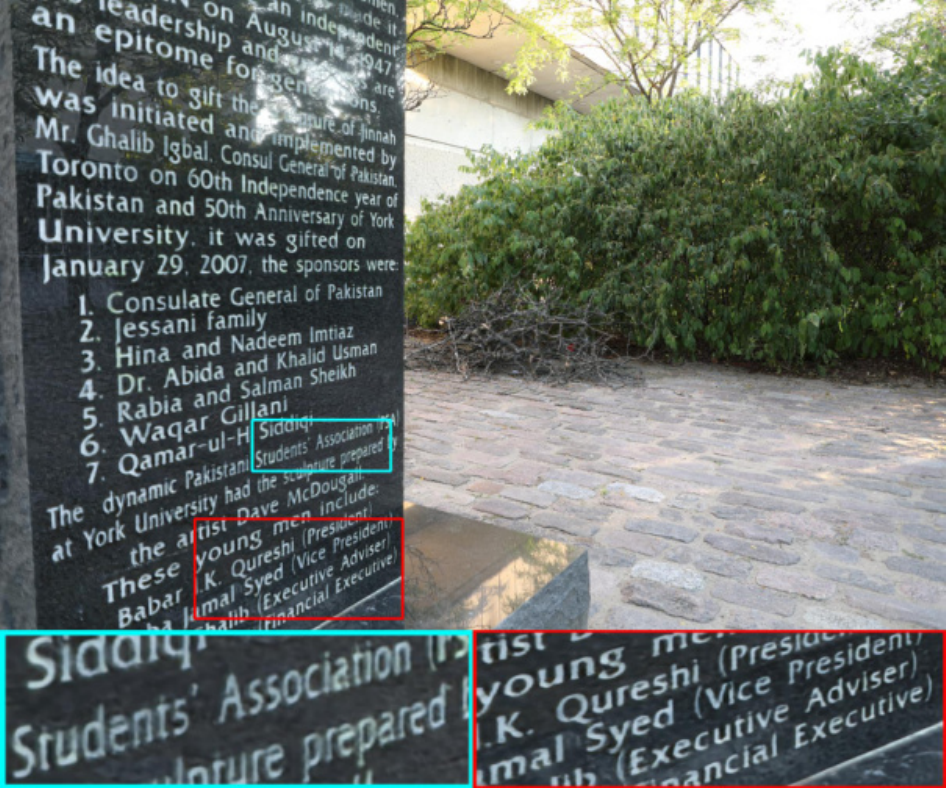}\vspace{4pt}
	\end{subfigure}
	\hfill
	\begin{subfigure}[t]{0.16\linewidth}	
		\includegraphics[width=2.6cm,height=4cm]{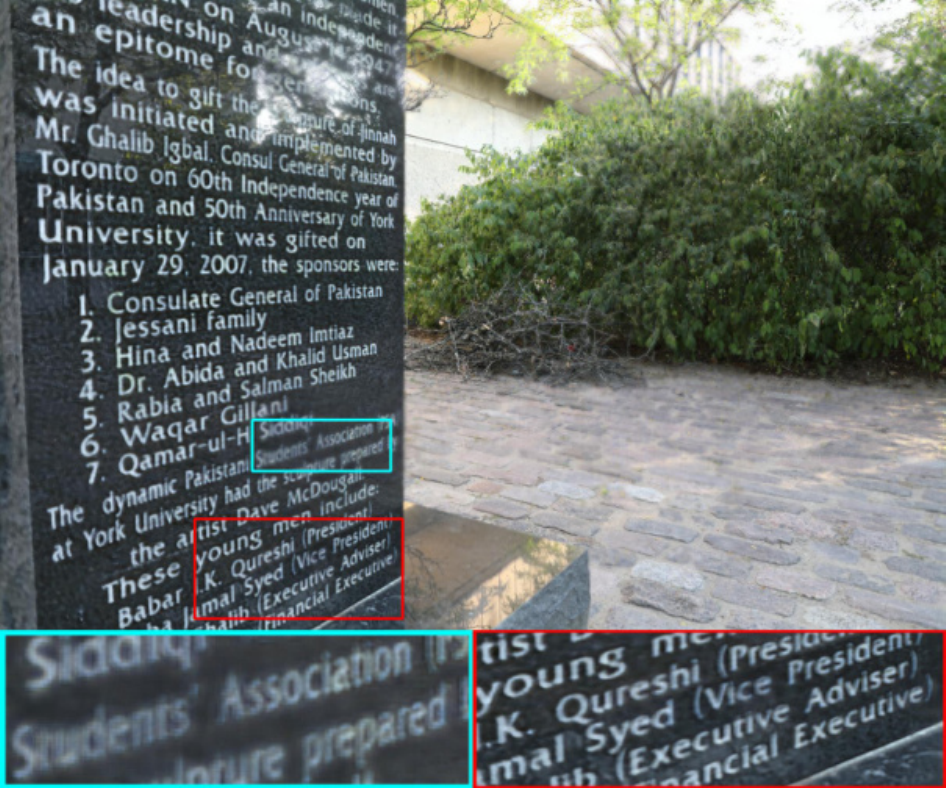}\vspace{4pt}
	\end{subfigure}
	\hfill
	\begin{subfigure}[t]{0.16\linewidth}	
		\includegraphics[width=2.6cm,height=4cm]{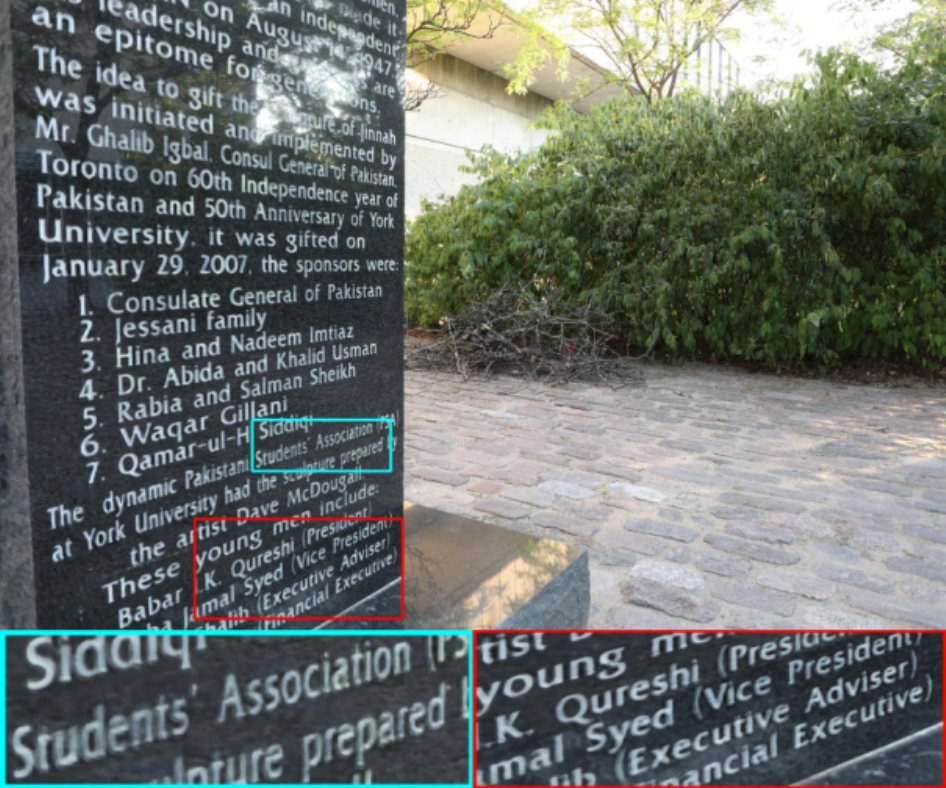}\vspace{4pt}
	\end{subfigure}
	\hfill
	\begin{subfigure}[t]{0.16\linewidth}	
		\includegraphics[width=2.6cm,height=4cm]{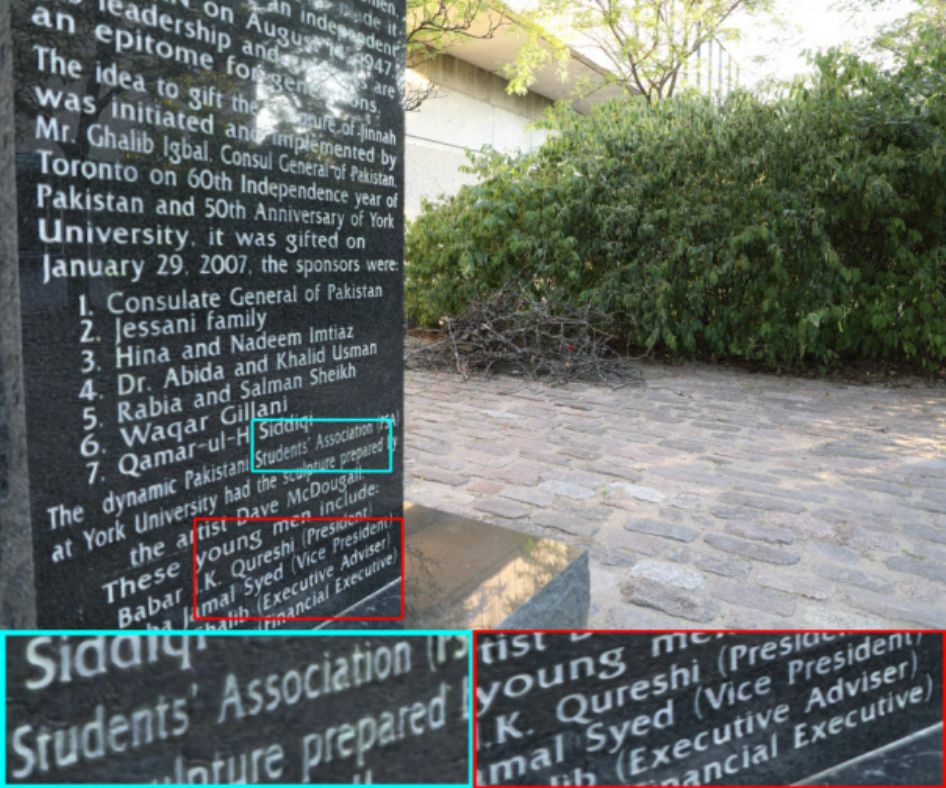}\vspace{4pt}
	\end{subfigure}
	\hfill
	\begin{subfigure}[t]{0.16\linewidth}	
		\includegraphics[width=2.6cm,height=4cm]{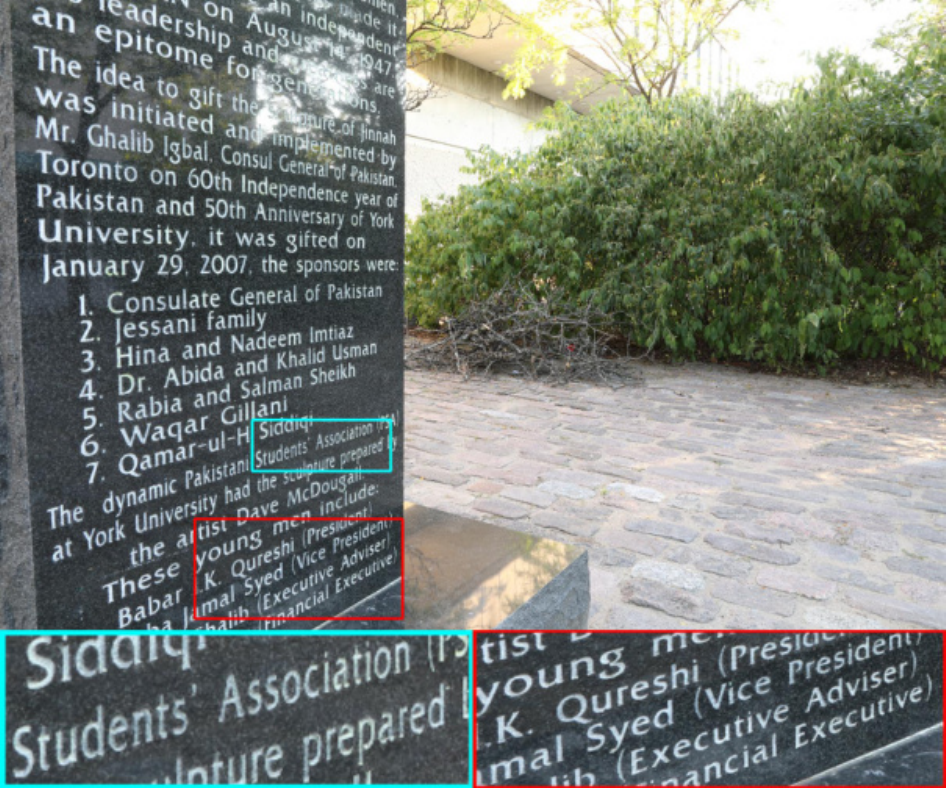}\vspace{4pt}
	\end{subfigure}
	\vfill
	
	\begin{subfigure}[t]{0.16\linewidth}	
		\includegraphics[width=2.6cm,height=4cm]{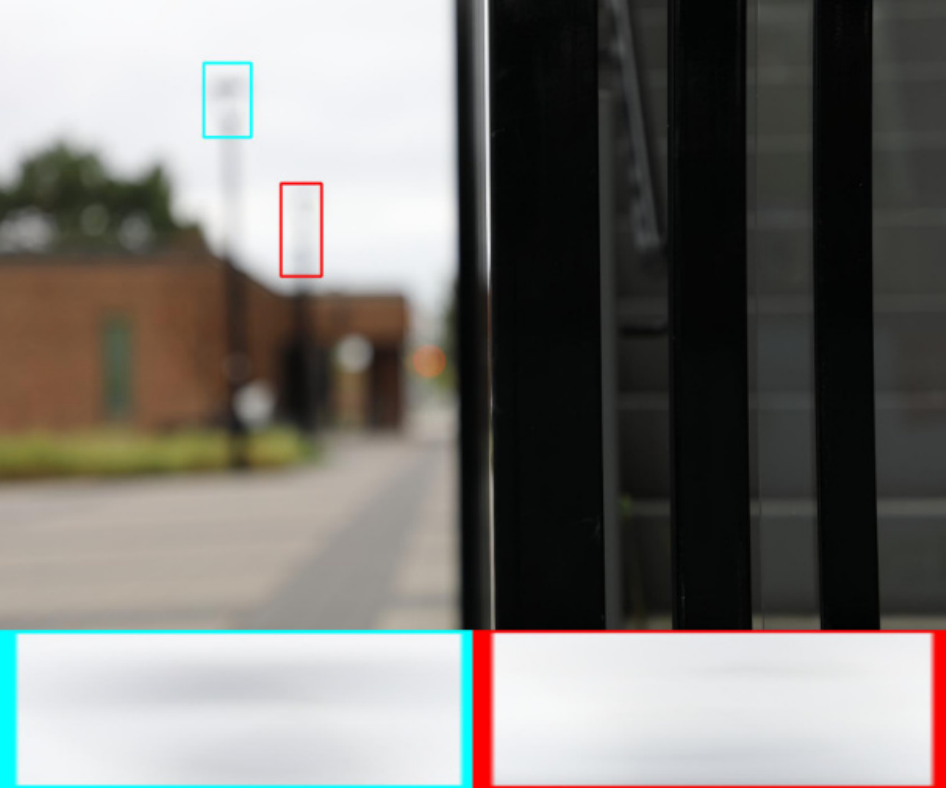}\vspace{4pt}
	\end{subfigure}
	\hfill
	\begin{subfigure}[t]{0.16\linewidth}	
		\includegraphics[width=2.6cm,height=4cm]{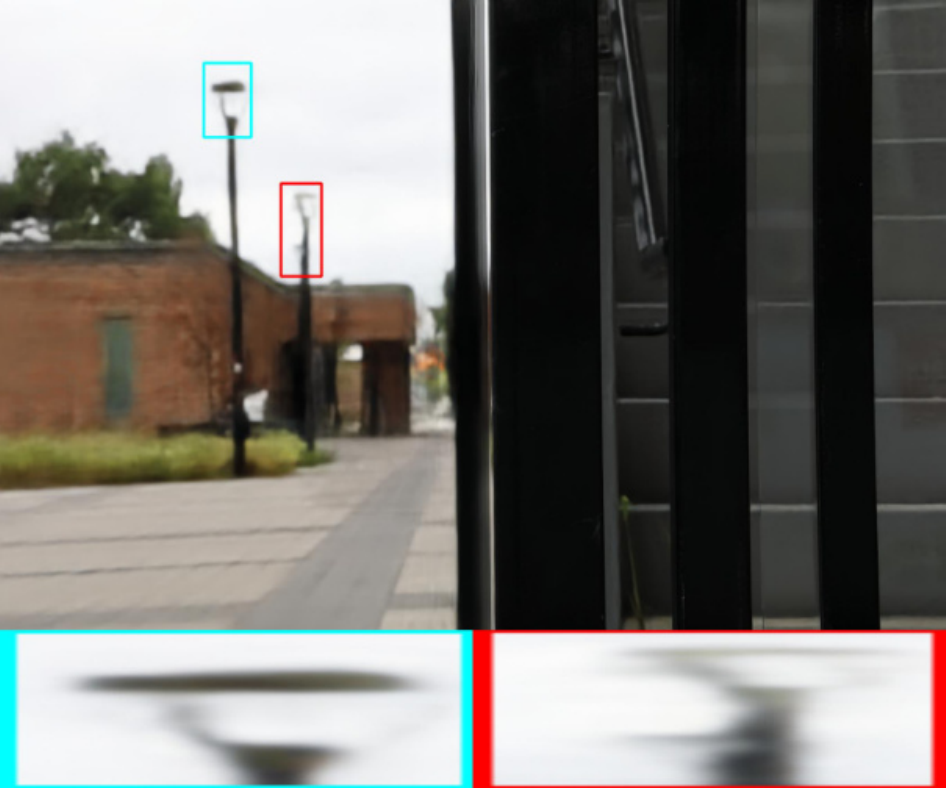}\vspace{4pt}
	\end{subfigure}
	\hfill
	\begin{subfigure}[t]{0.16\linewidth}	
		\includegraphics[width=2.6cm,height=4cm]{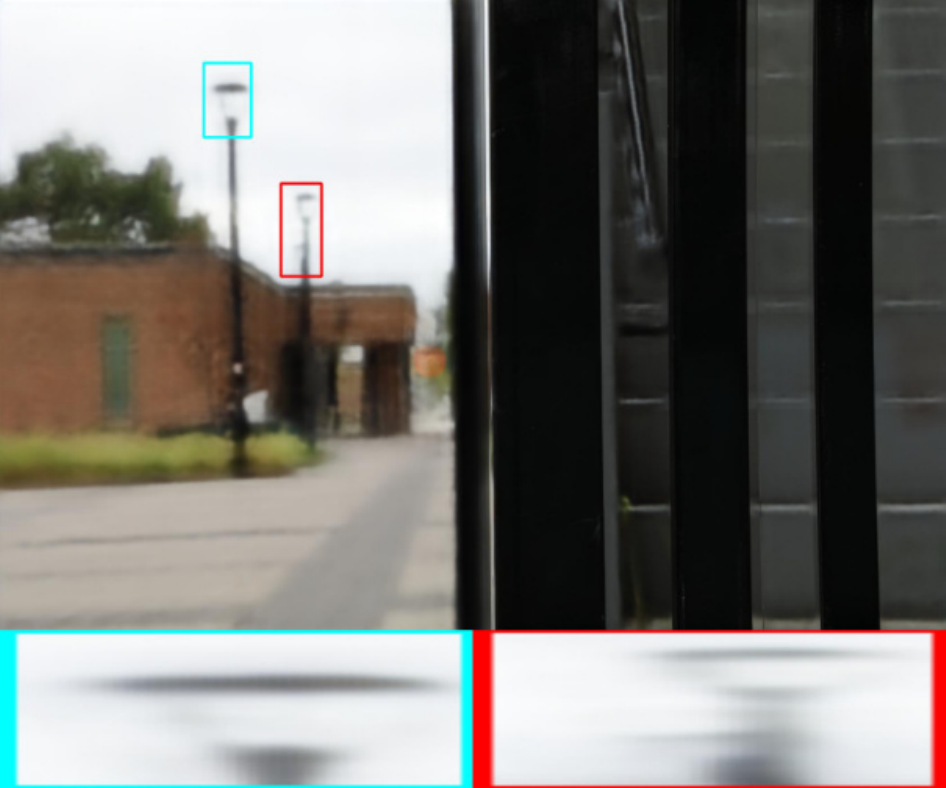}\vspace{4pt}
	\end{subfigure}
	\hfill
	\begin{subfigure}[t]{0.16\linewidth}	
		\includegraphics[width=2.6cm,height=4cm]{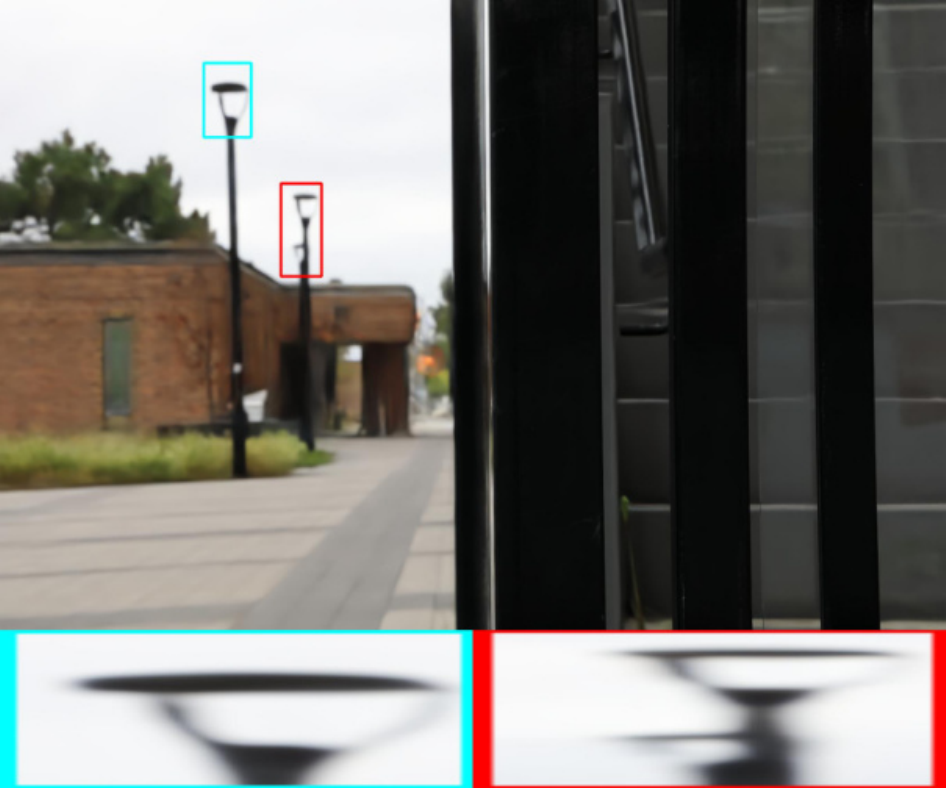}\vspace{4pt}
	\end{subfigure}
	\hfill
	\begin{subfigure}[t]{0.16\linewidth}	
		\includegraphics[width=2.6cm,height=4cm]{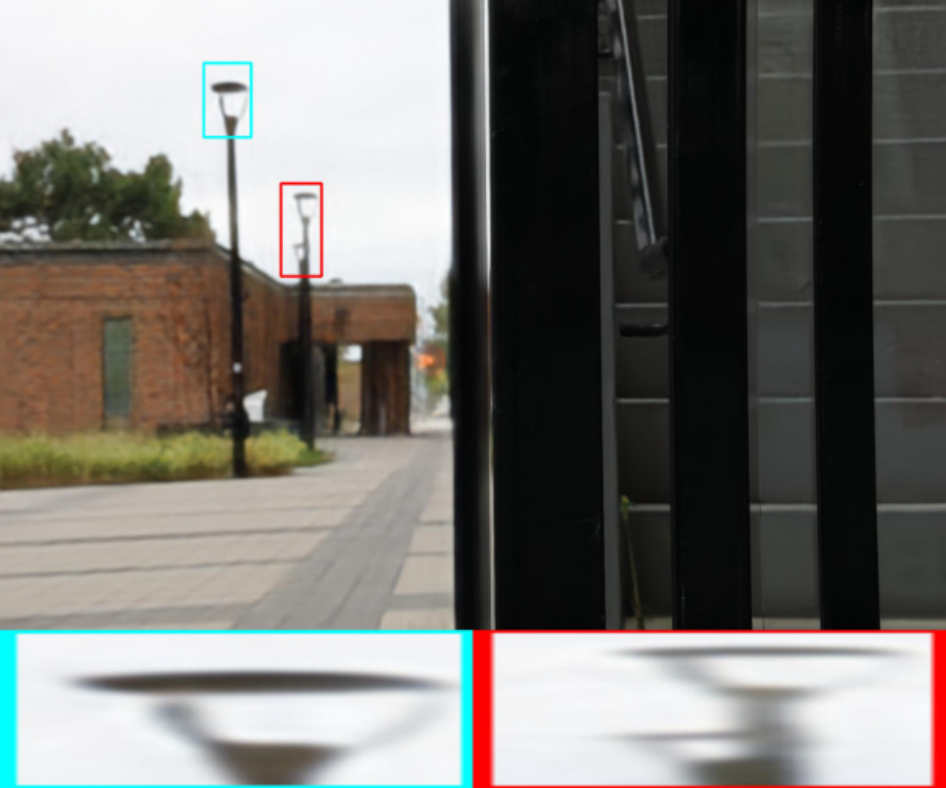}\vspace{4pt}
	\end{subfigure}
	\hfill
	\begin{subfigure}[t]{0.16\linewidth}	
		\includegraphics[width=2.6cm,height=4cm]{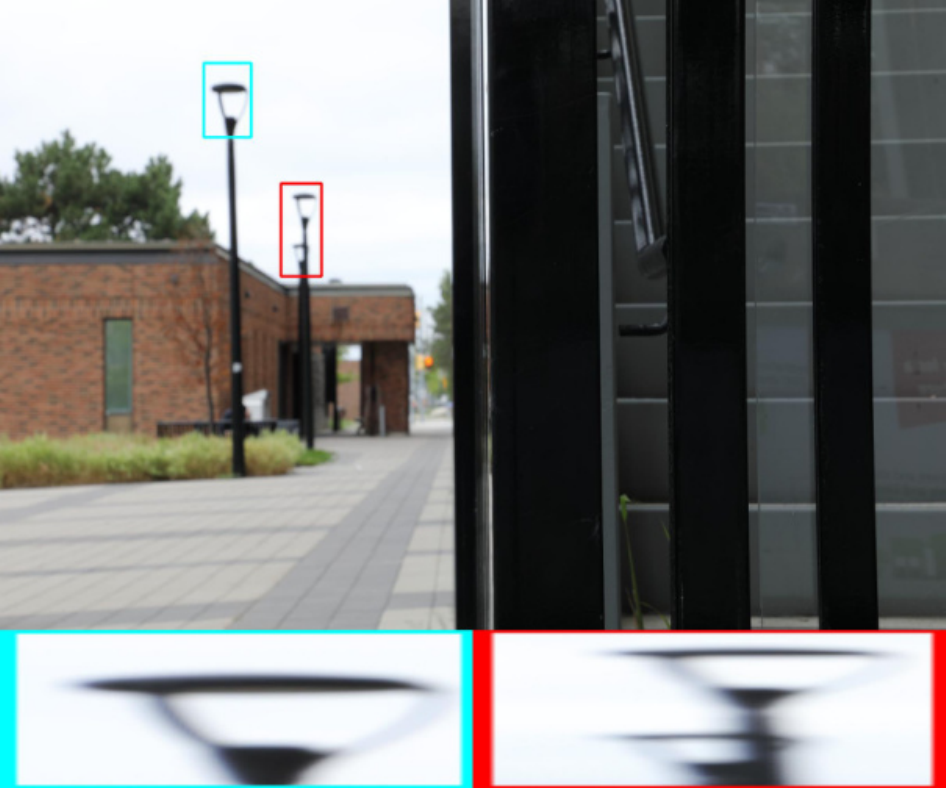}\vspace{4pt}
	\end{subfigure}
	\vfill
	
	\begin{subfigure}[t]{0.16\linewidth}	
		\includegraphics[width=2.6cm,height=4cm]{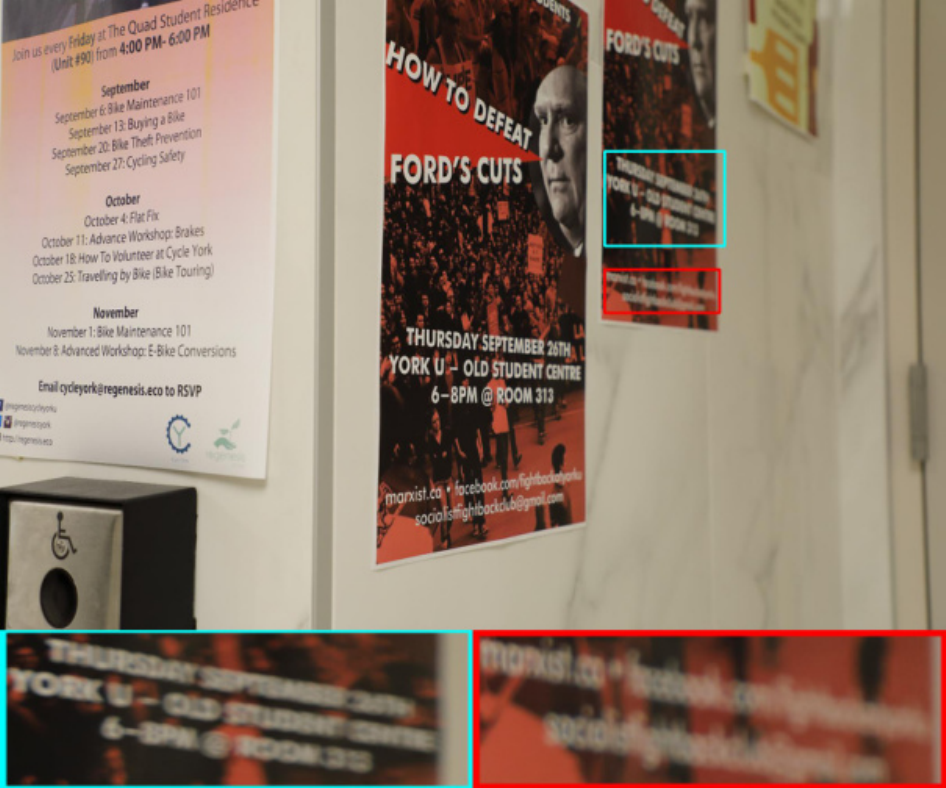}\vspace{4pt}
	\end{subfigure}
	\hfill
	\begin{subfigure}[t]{0.16\linewidth}	
		\includegraphics[width=2.6cm,height=4cm]{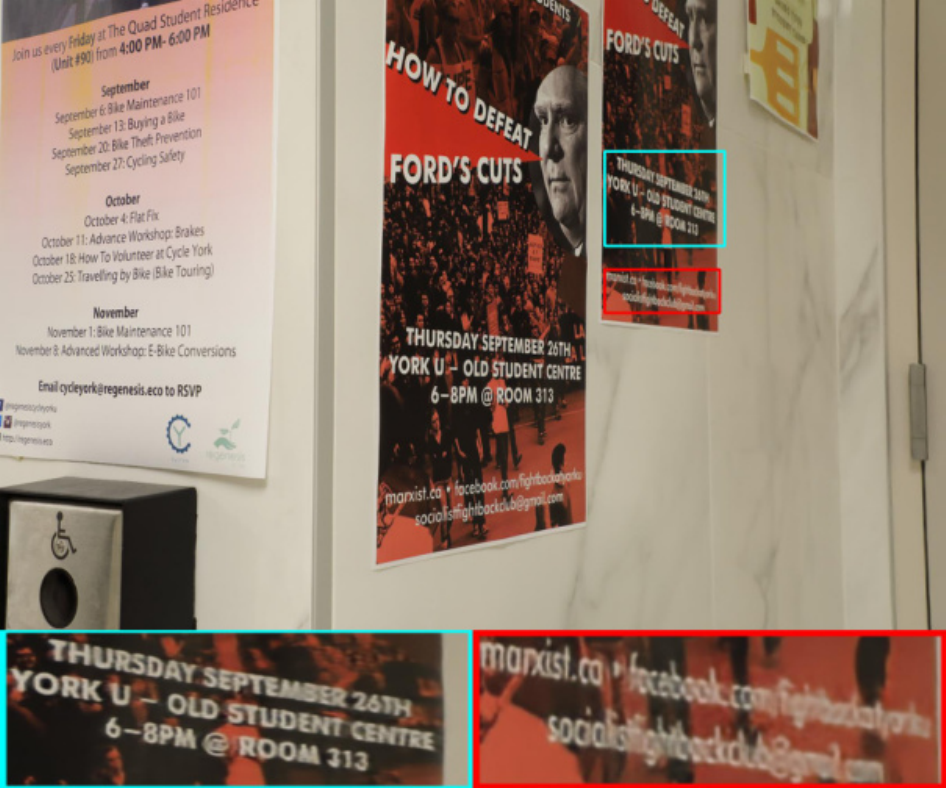}\vspace{4pt}
	\end{subfigure}
	\hfill
	\begin{subfigure}[t]{0.16\linewidth}	
		\includegraphics[width=2.6cm,height=4cm]{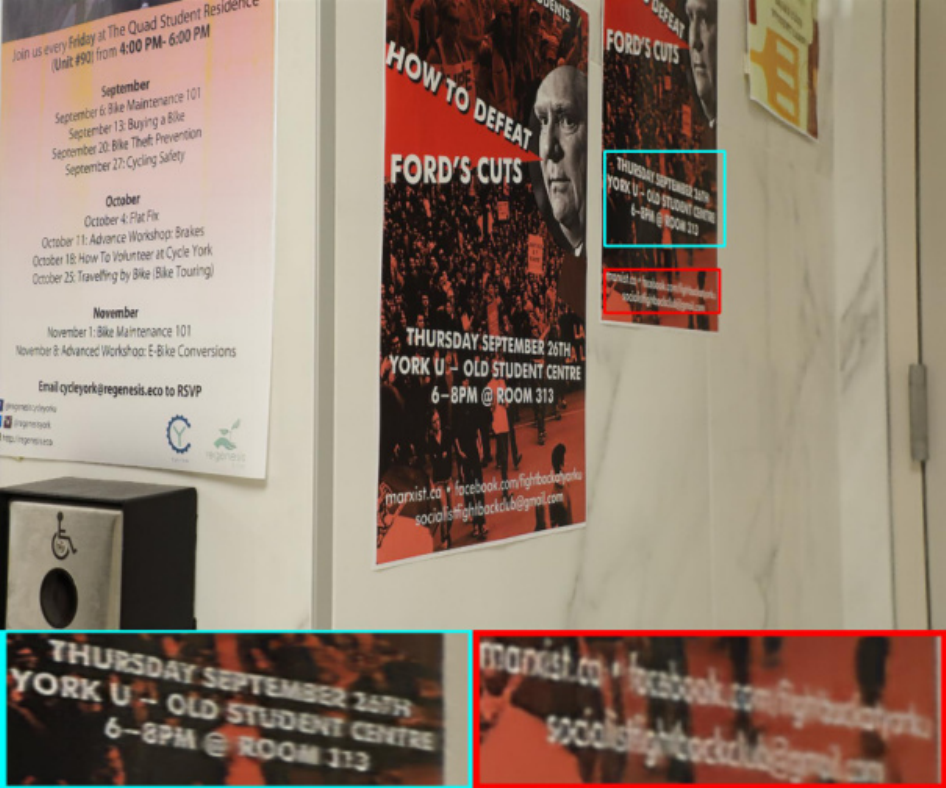}\vspace{4pt}
	\end{subfigure}
	\hfill
	\begin{subfigure}[t]{0.16\linewidth}	
		\includegraphics[width=2.6cm,height=4cm]{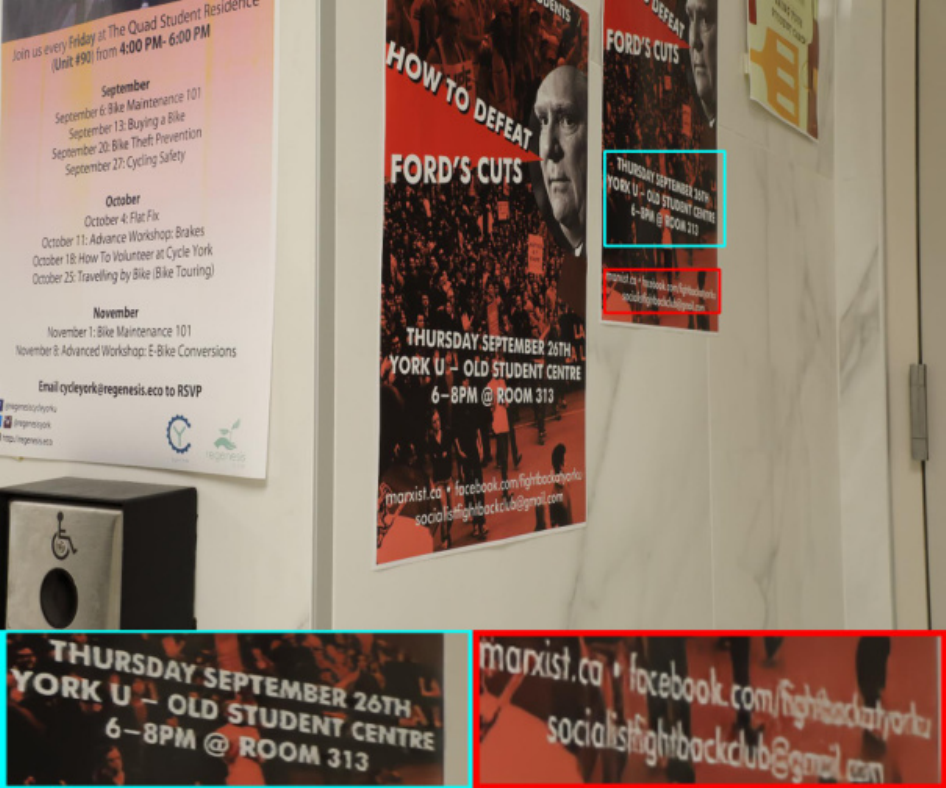}\vspace{4pt}
	\end{subfigure}
	\hfill
	\begin{subfigure}[t]{0.16\linewidth}	
		\includegraphics[width=2.6cm,height=4cm]{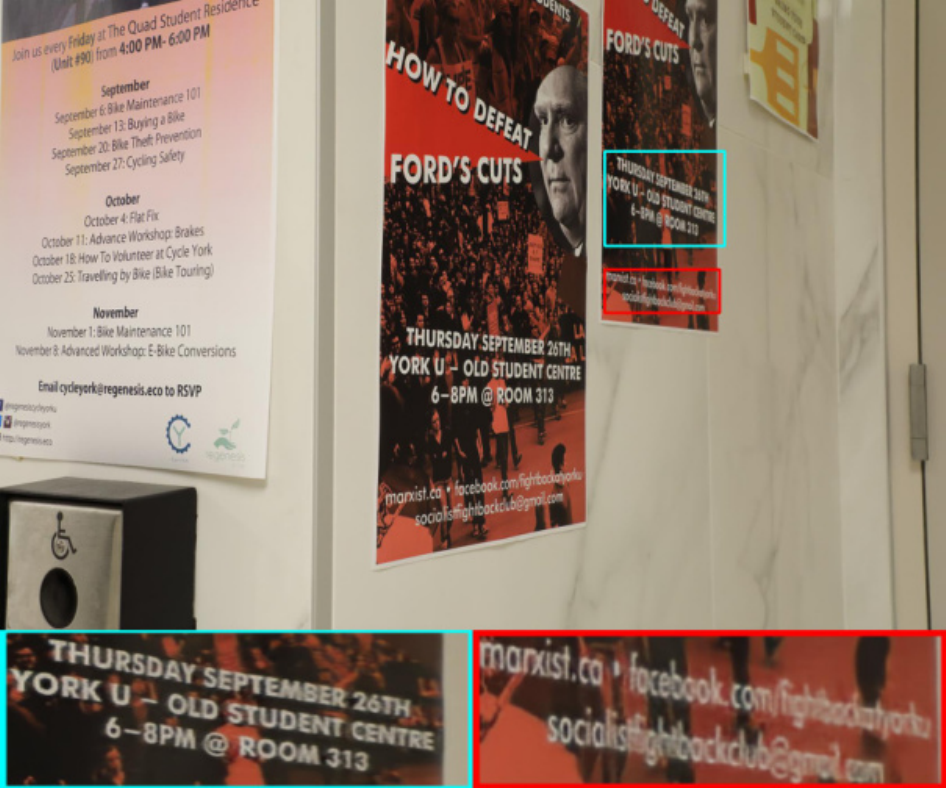}\vspace{4pt}
	\end{subfigure}
	\hfill
	\begin{subfigure}[t]{0.16\linewidth}	
		\includegraphics[width=2.6cm,height=4cm]{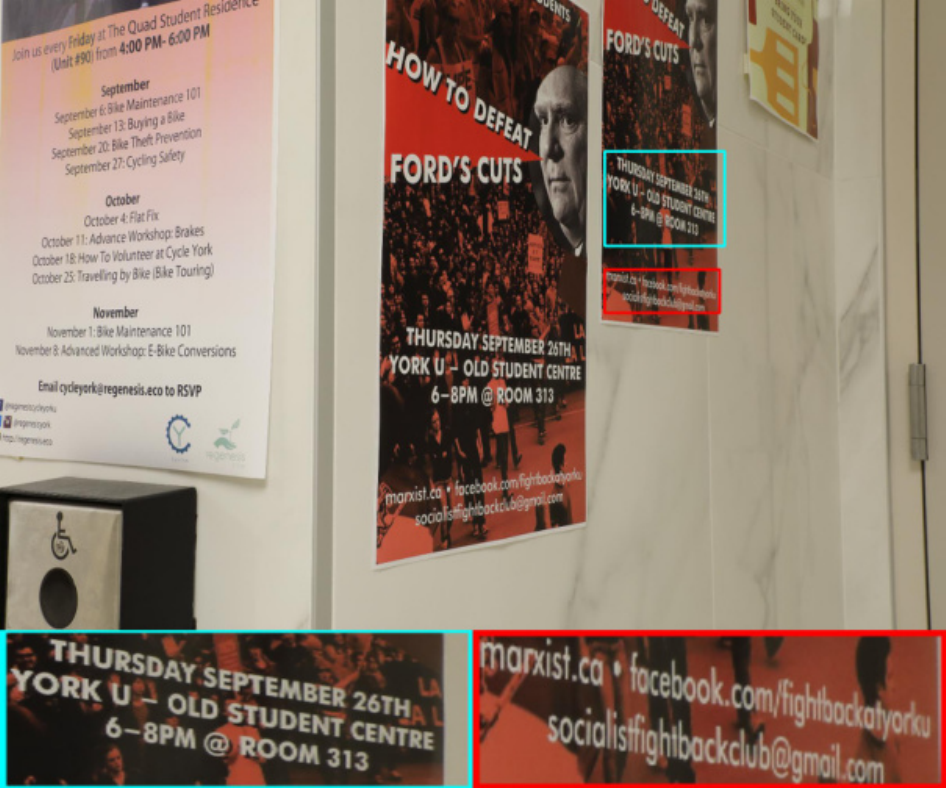}\vspace{4pt}
	\end{subfigure}
	\vfill
	
	\begin{subfigure}[t]{0.16\linewidth}	
		\includegraphics[width=2.6cm,height=4cm]{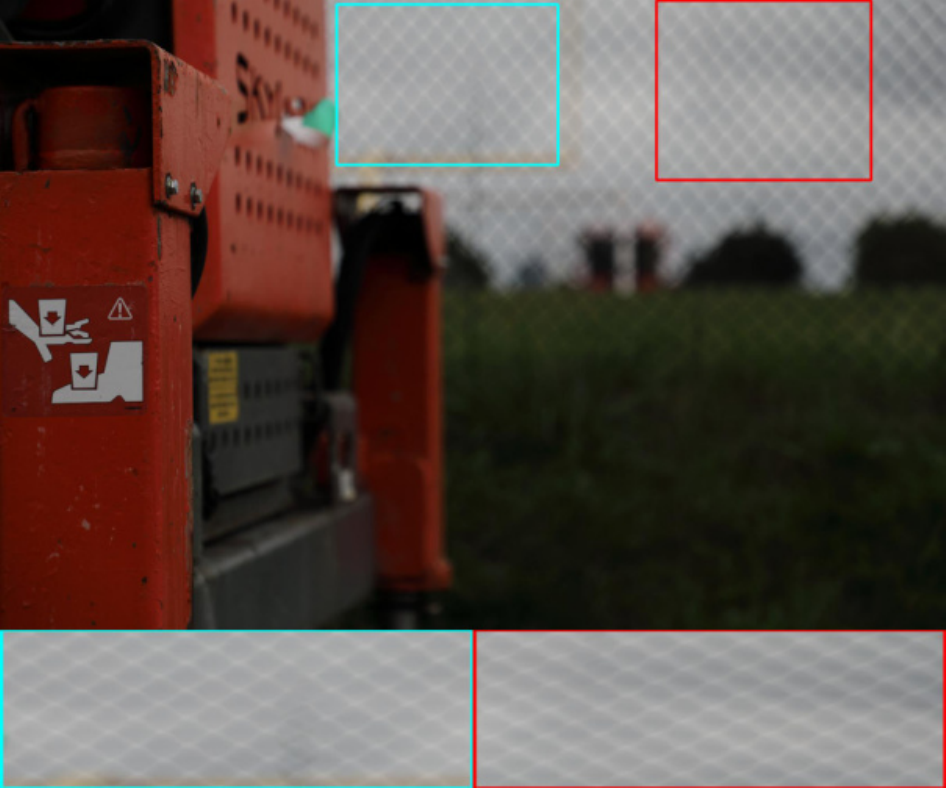}\vspace{4pt}
		\caption{Blurry Input}
	\end{subfigure}
	\hfill
	\begin{subfigure}[t]{0.16\linewidth}	
		\includegraphics[width=2.6cm,height=4cm]{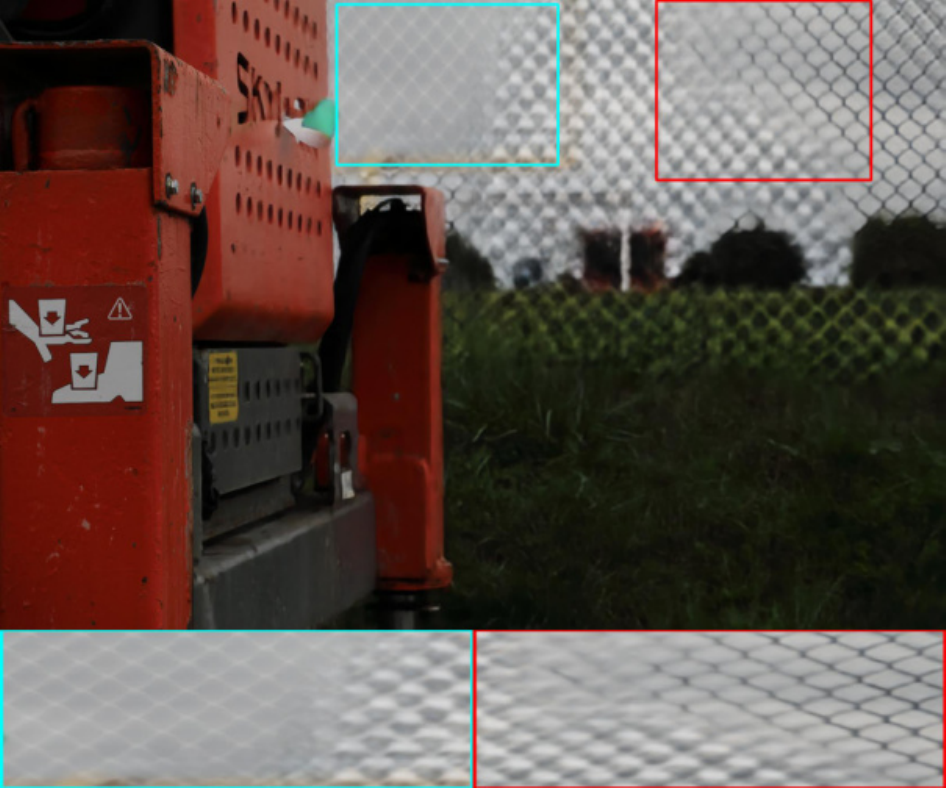}\vspace{4pt}
		\caption{IFAN}
	\end{subfigure}
	\hfill
	\begin{subfigure}[t]{0.16\linewidth}	
		\includegraphics[width=2.6cm,height=4cm]{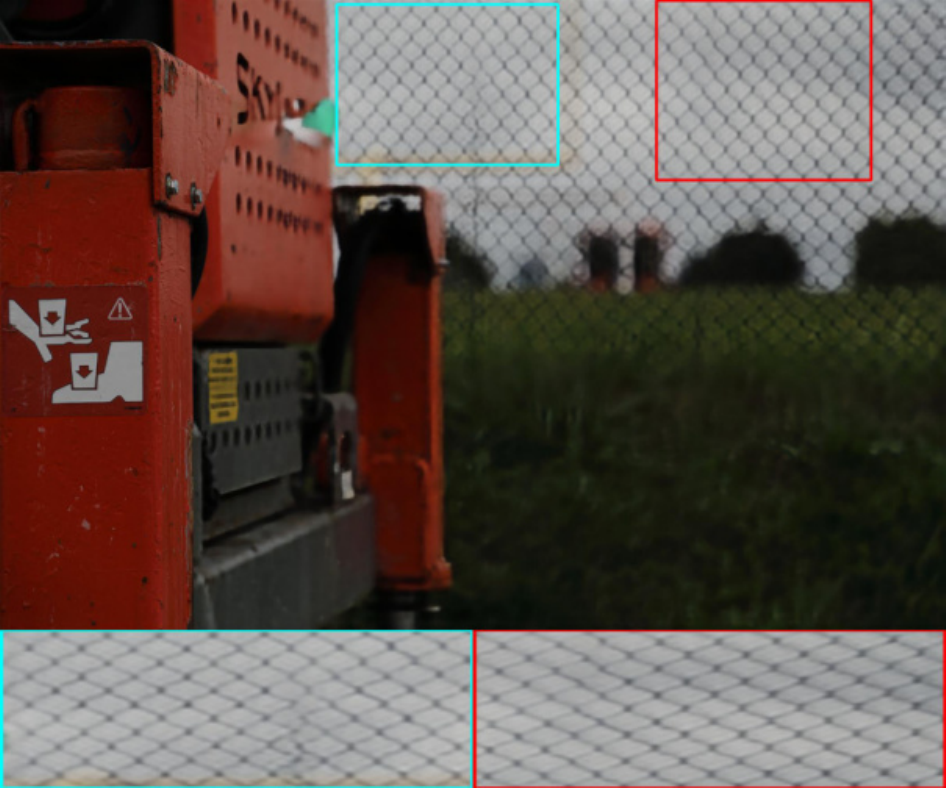}\vspace{4pt}
		\caption{GKMNet}
	\end{subfigure}
	\hfill
	\begin{subfigure}[t]{0.16\linewidth}	
		\includegraphics[width=2.6cm,height=4cm]{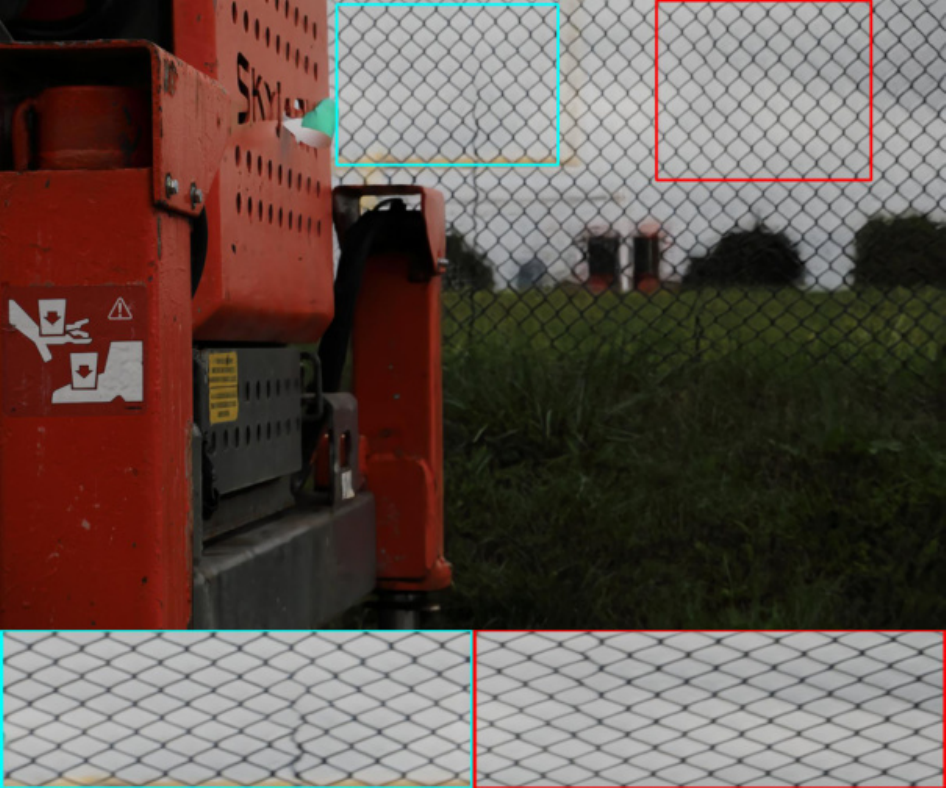}\vspace{4pt}
		\caption{Restormer}
	\end{subfigure}
	\hfill
	\begin{subfigure}[t]{0.16\linewidth}	
		\includegraphics[width=2.6cm,height=4cm]{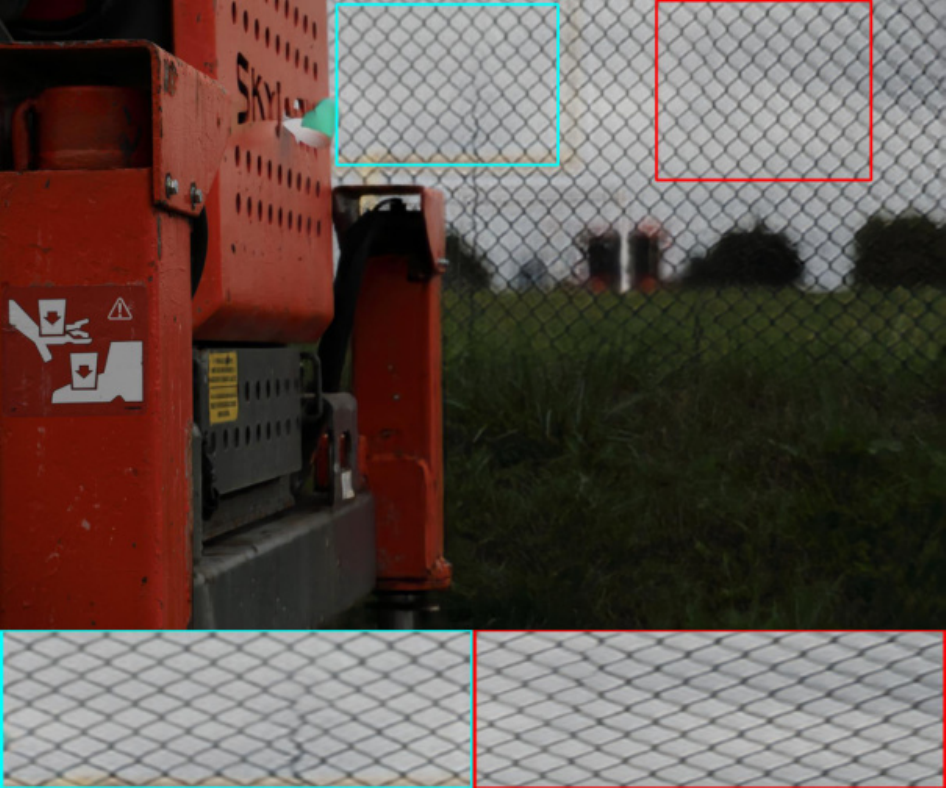}\vspace{4pt}
		\caption{SR-R$^2$KAC-B}
	\end{subfigure}
	\hfill
	\begin{subfigure}[t]{0.16\linewidth}	
		\includegraphics[width=2.6cm,height=4cm]{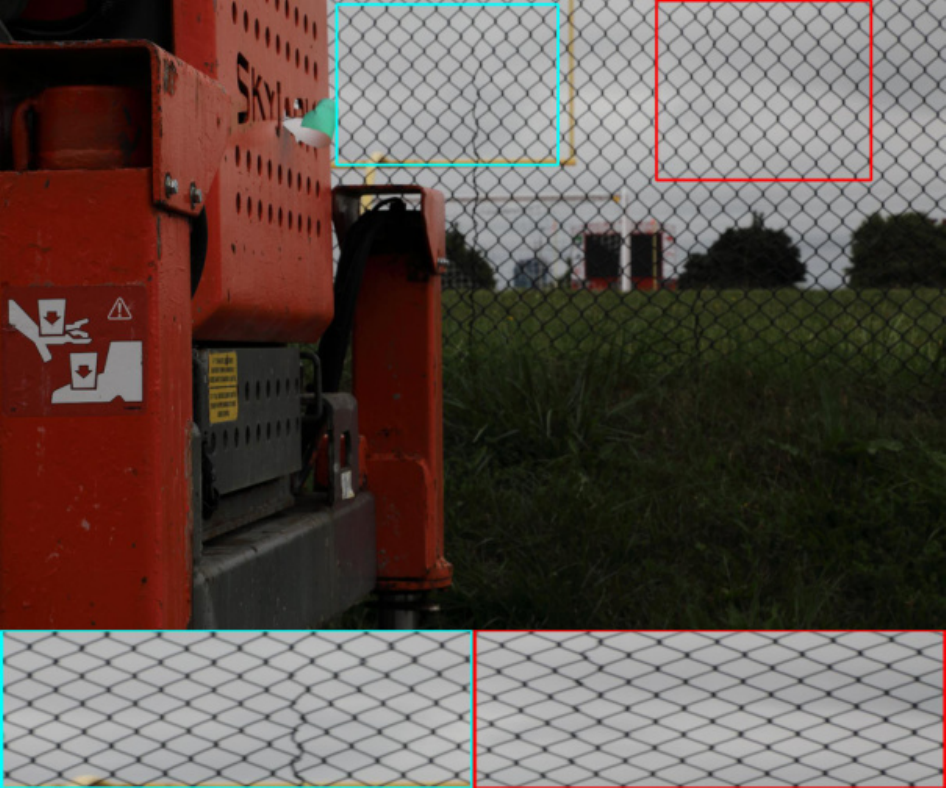}\vspace{4pt}
		\caption{Sharp}
	\end{subfigure}
	\vfill

	\caption{Visualization results of different defocus deblurring methods on the DPDD dataset \cite{Abuolaim2020}.}
	\vspace{-0.2in}
	\label{fig_s_2}
\end{figure*}

\begin{figure*}[t]
	\centering
	
	\begin{subfigure}[t]{0.16\linewidth}	
		\includegraphics[width=2.6cm,height=4cm]{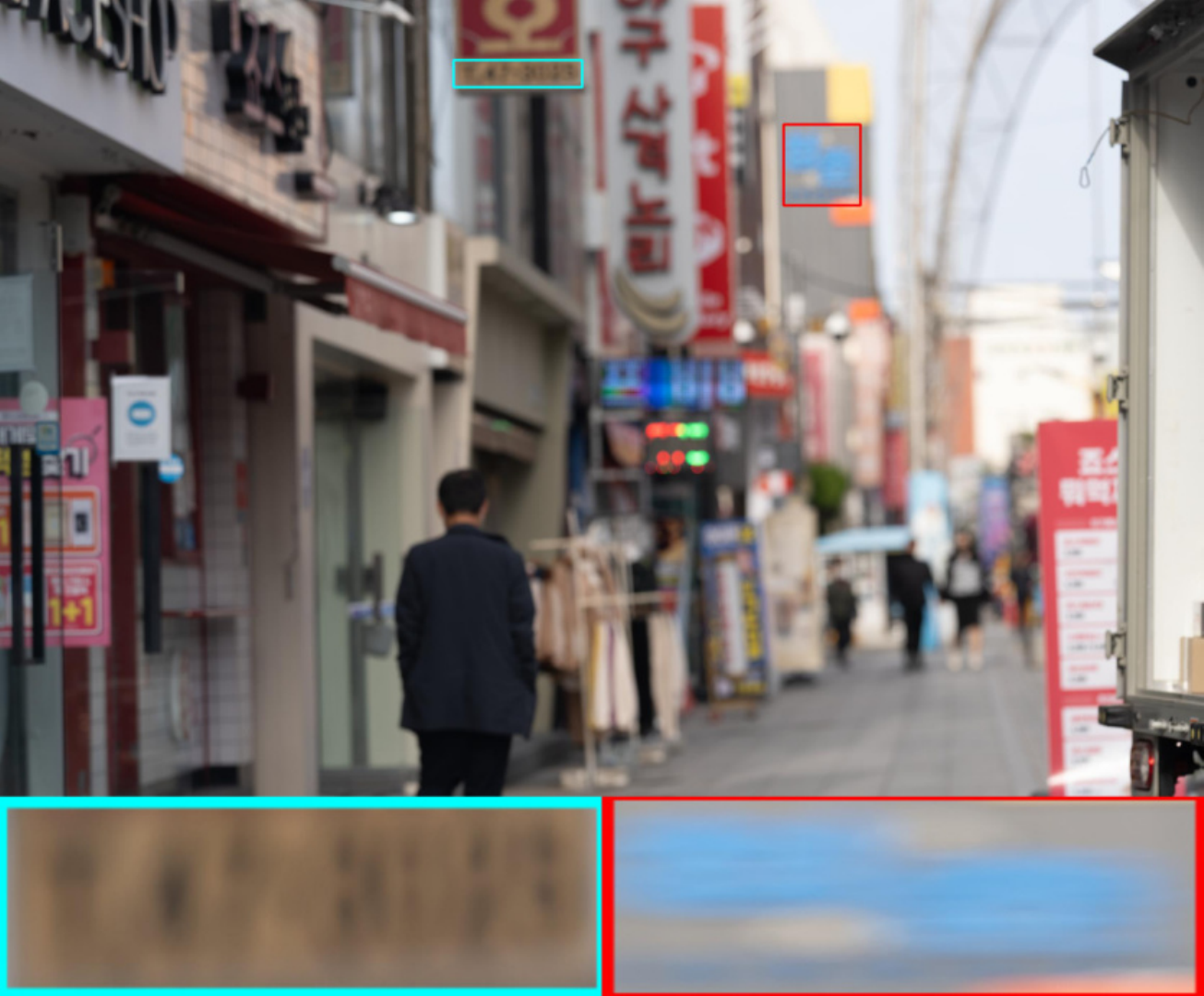}\vspace{4pt}
	\end{subfigure}
	\hfill
	\begin{subfigure}[t]{0.16\linewidth}	
		\includegraphics[width=2.6cm,height=4cm]{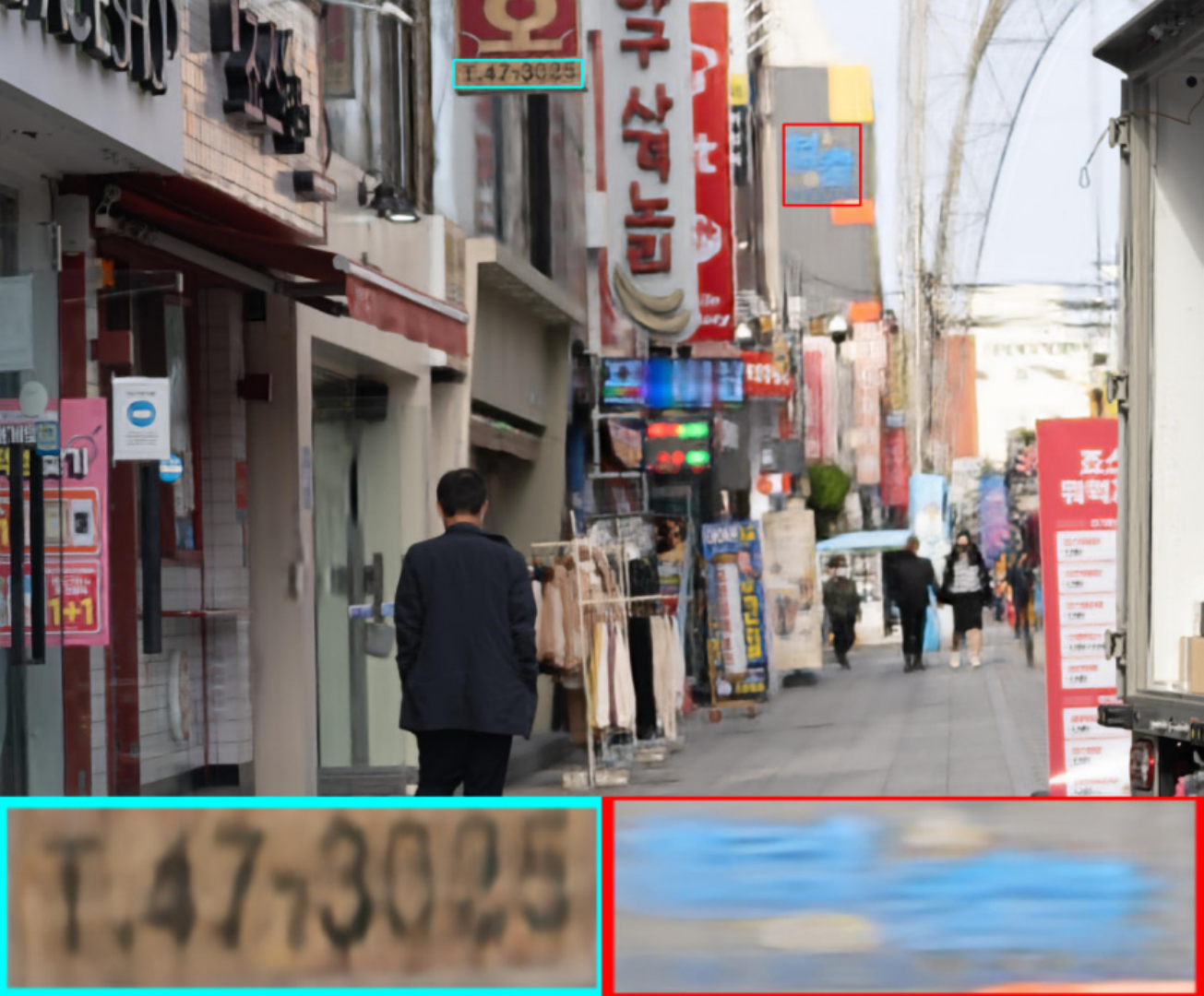}\vspace{4pt}
	\end{subfigure}
	\hfill
	\begin{subfigure}[t]{0.16\linewidth}	
		\includegraphics[width=2.6cm,height=4cm]{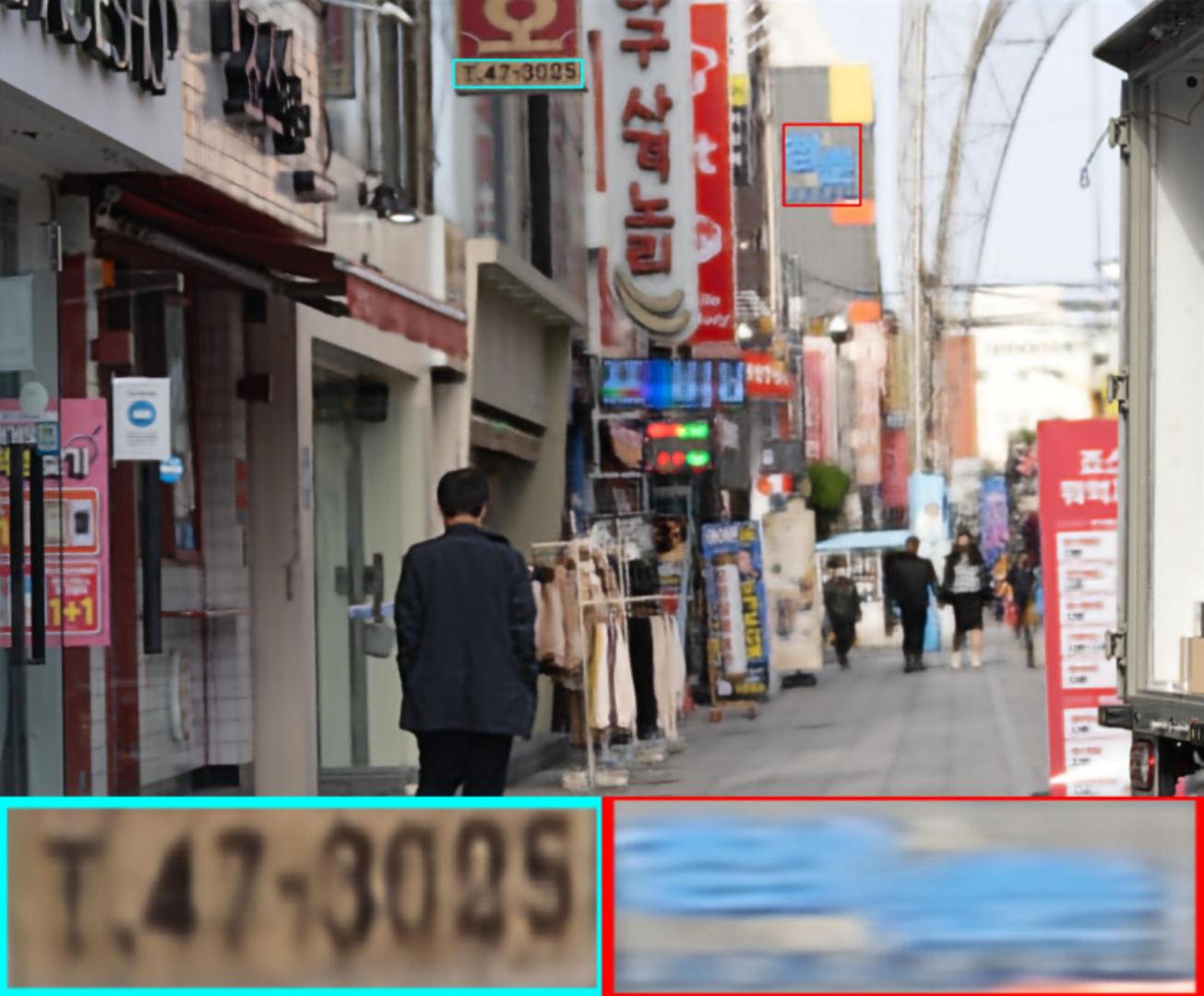}\vspace{4pt}
	\end{subfigure}
	\hfill
	\begin{subfigure}[t]{0.16\linewidth}	
		\includegraphics[width=2.6cm,height=4cm]{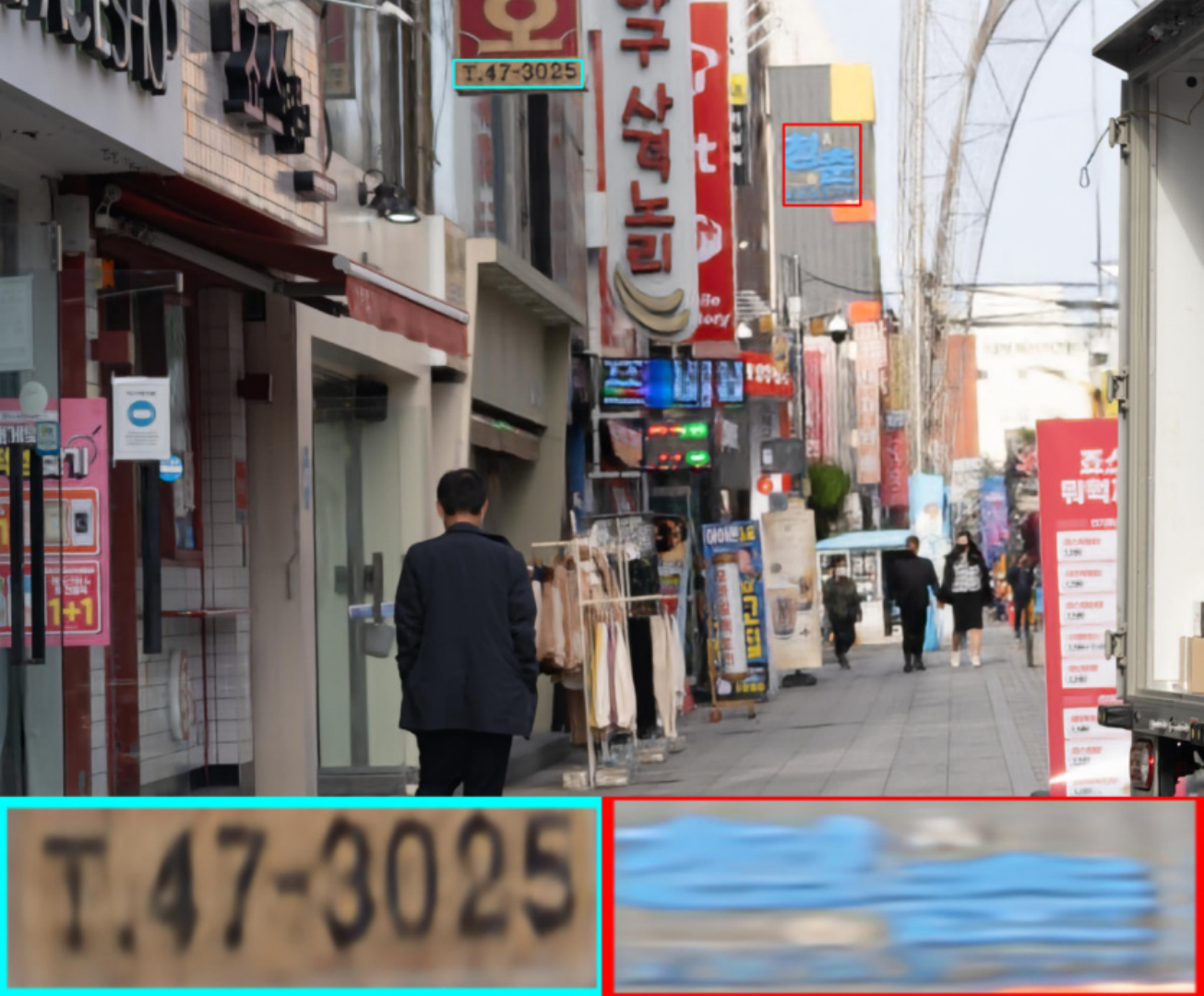}\vspace{4pt}
	\end{subfigure}
	\hfill
	\begin{subfigure}[t]{0.16\linewidth}	
		\includegraphics[width=2.6cm,height=4cm]{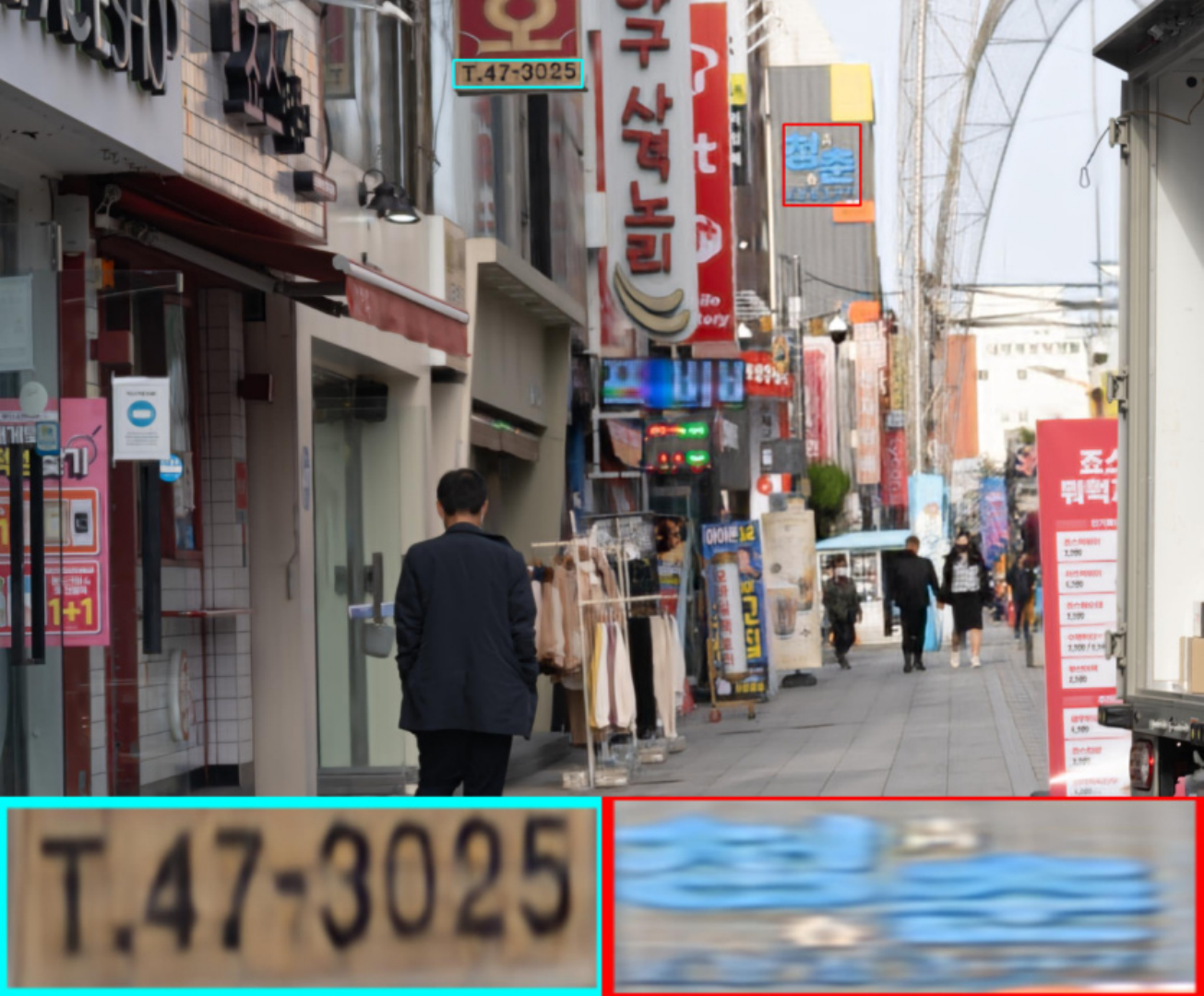}\vspace{4pt}
	\end{subfigure}
	\hfill
	\begin{subfigure}[t]{0.16\linewidth}	
		\includegraphics[width=2.6cm,height=4cm]{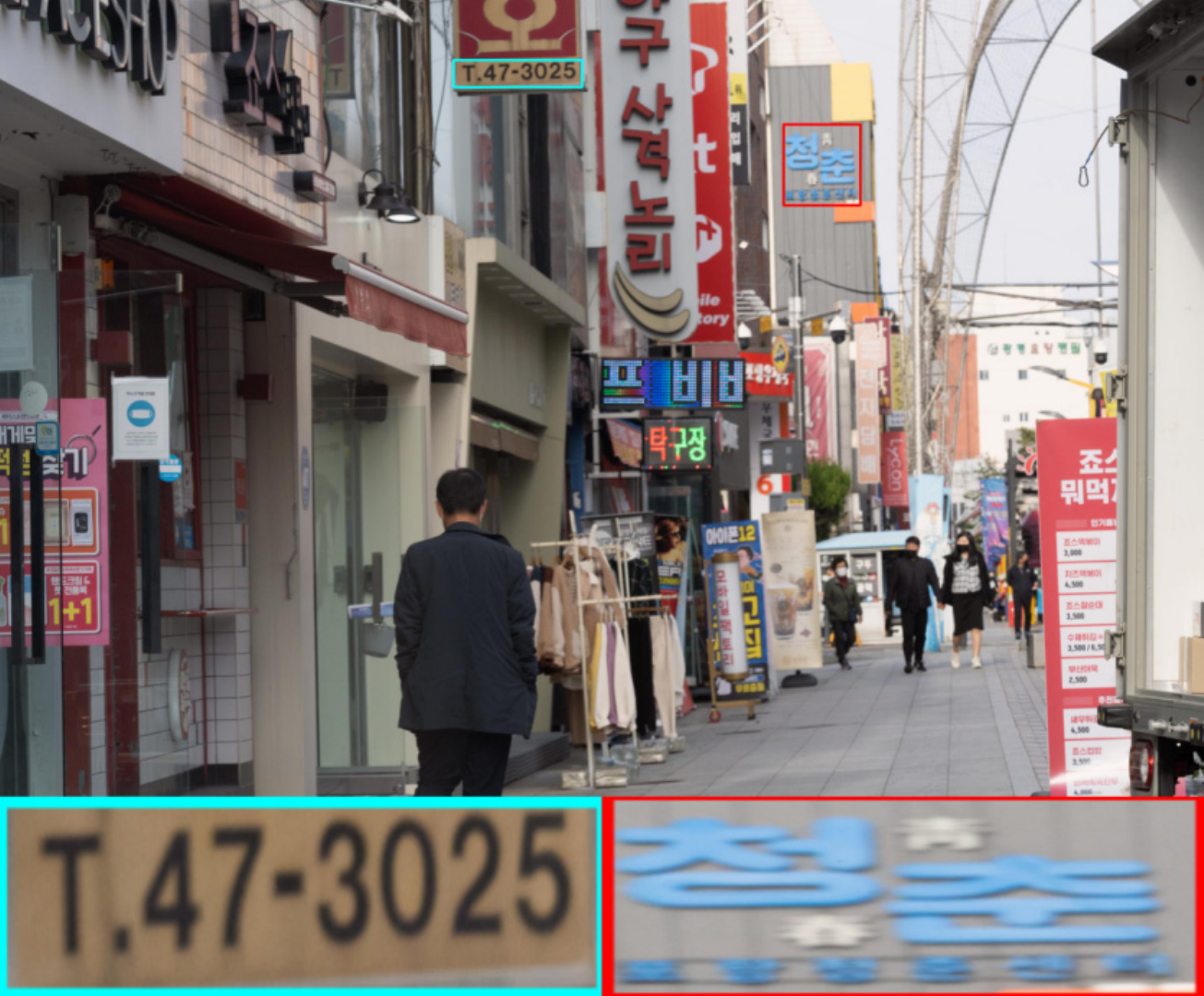}\vspace{4pt}
	\end{subfigure}
	\vfill
	
	\begin{subfigure}[t]{0.16\linewidth}	
		\includegraphics[width=2.6cm,height=4cm]{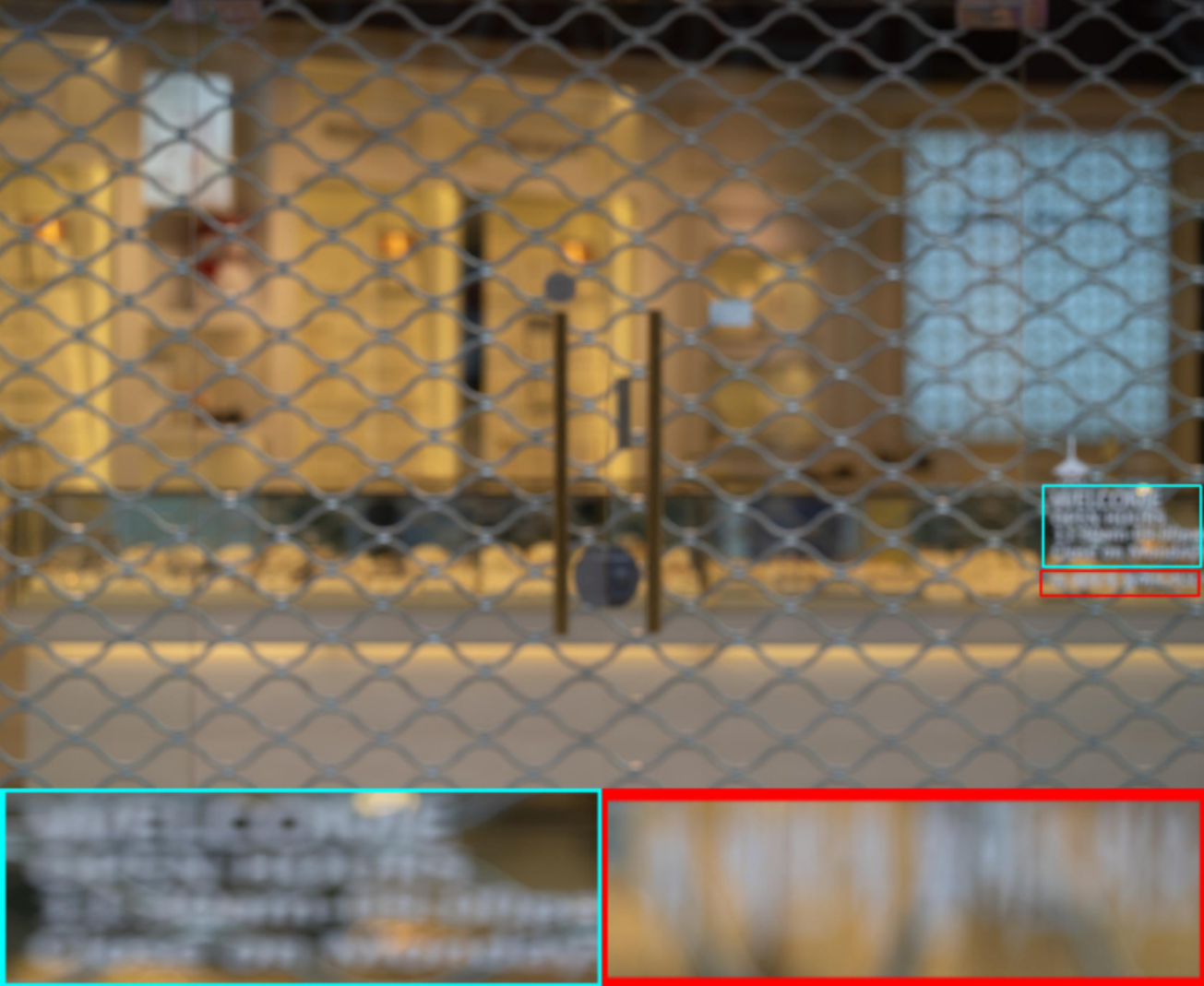}\vspace{4pt}
	\end{subfigure}
	\hfill
	\begin{subfigure}[t]{0.16\linewidth}	
		\includegraphics[width=2.6cm,height=4cm]{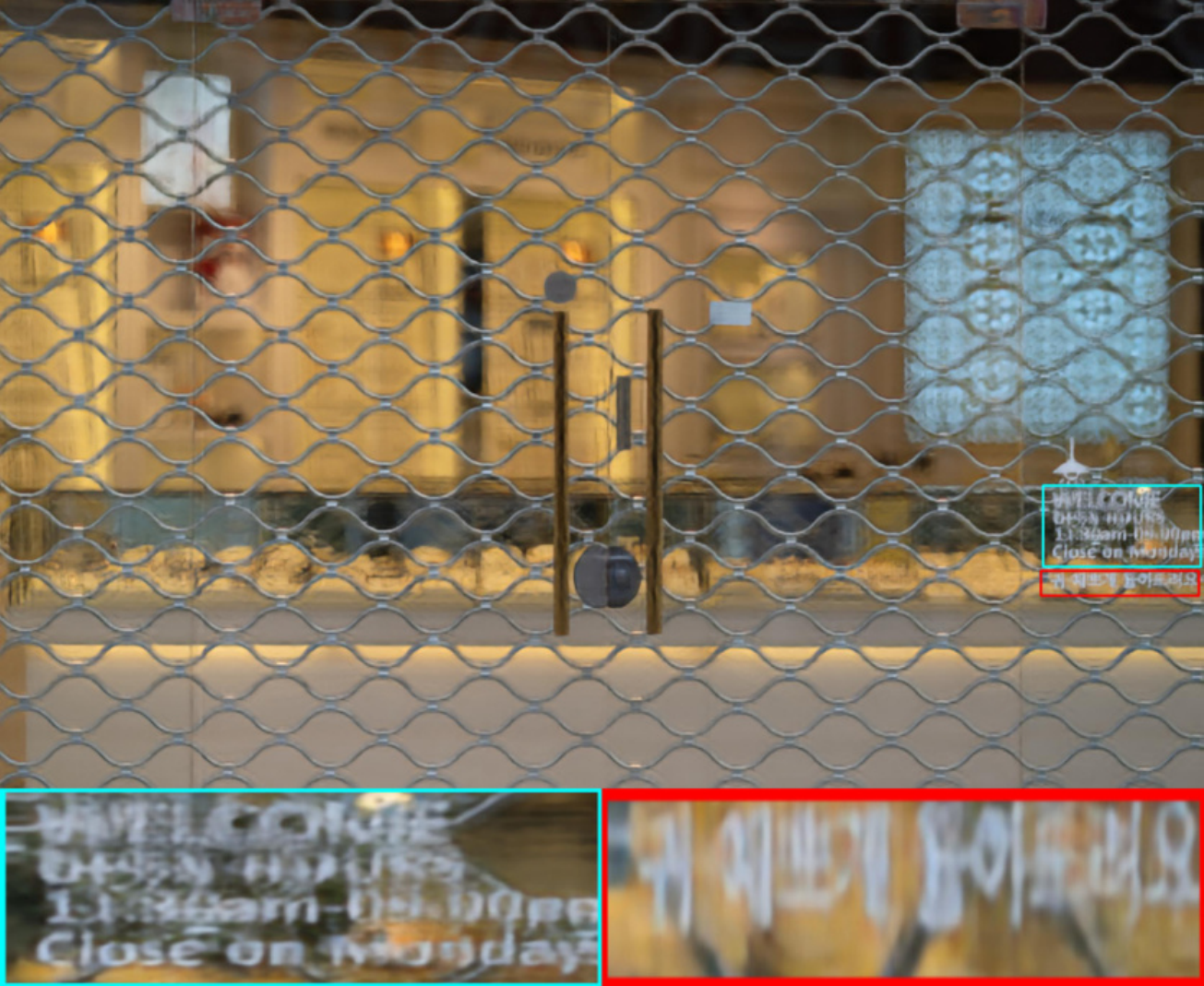}\vspace{4pt}
	\end{subfigure}
	\hfill
	\begin{subfigure}[t]{0.16\linewidth}	
		\includegraphics[width=2.6cm,height=4cm]{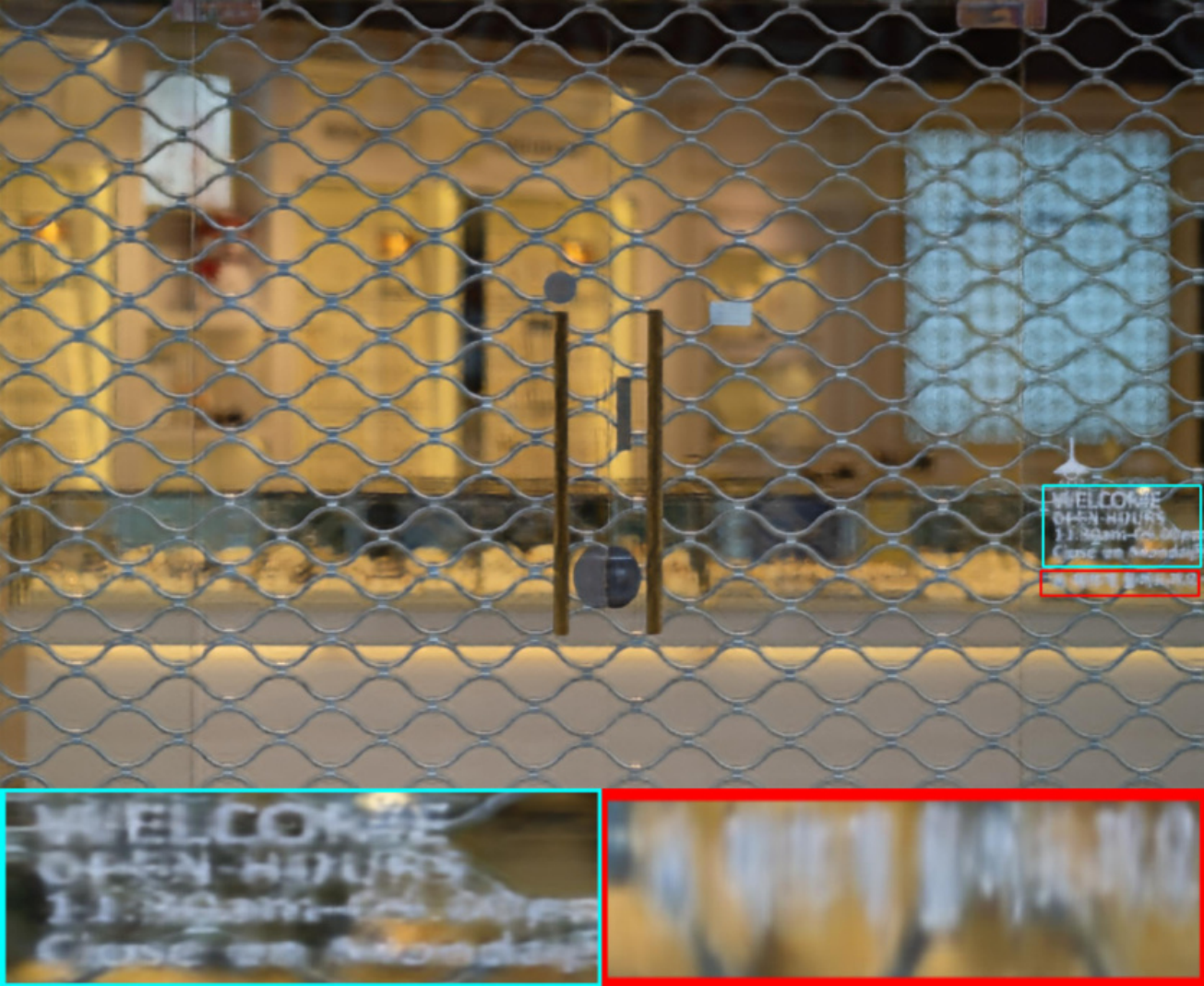}\vspace{4pt}
	\end{subfigure}
	\hfill
	\begin{subfigure}[t]{0.16\linewidth}	
		\includegraphics[width=2.6cm,height=4cm]{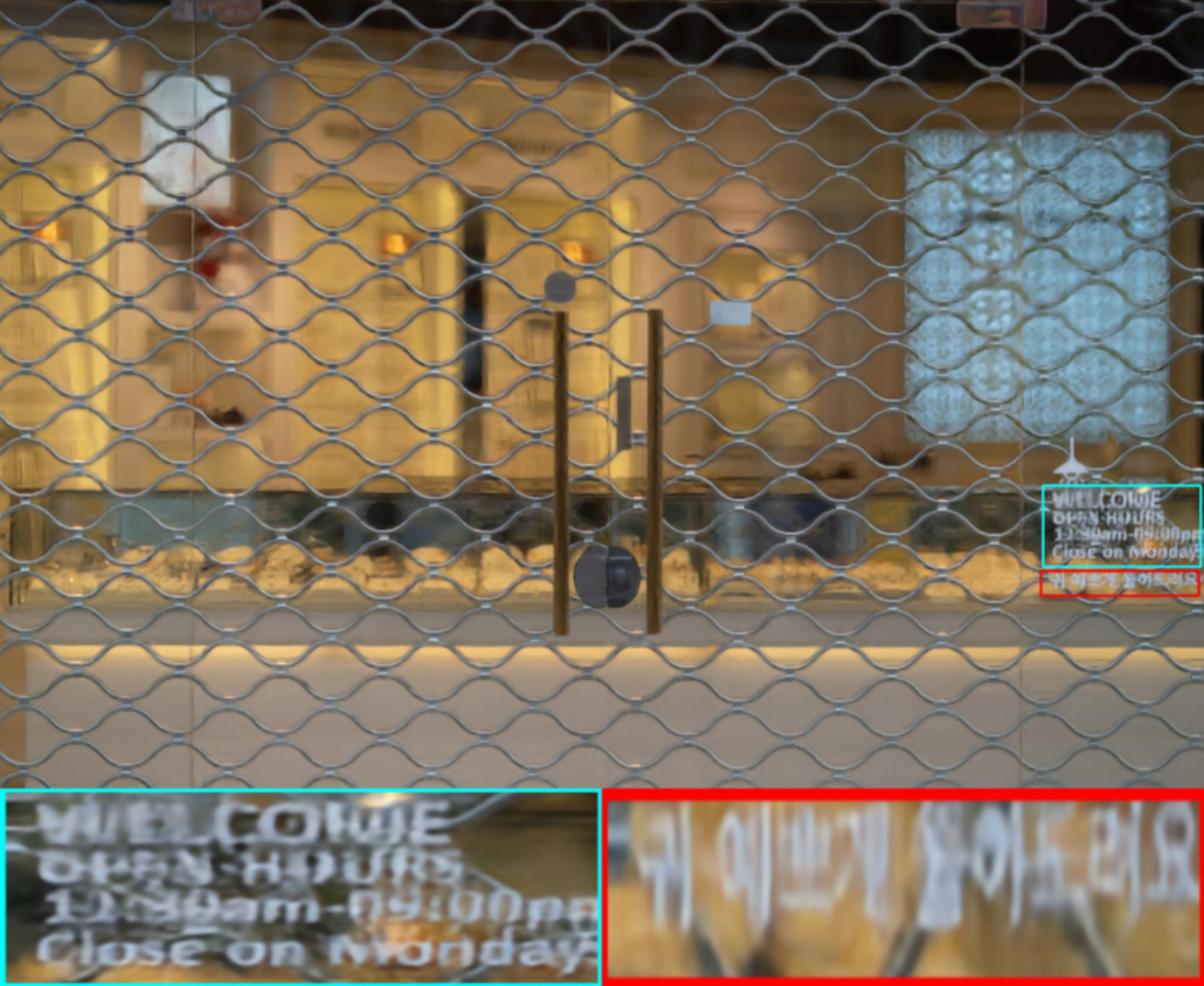}\vspace{4pt}
	\end{subfigure}
	\hfill
	\begin{subfigure}[t]{0.16\linewidth}	
		\includegraphics[width=2.6cm,height=4cm]{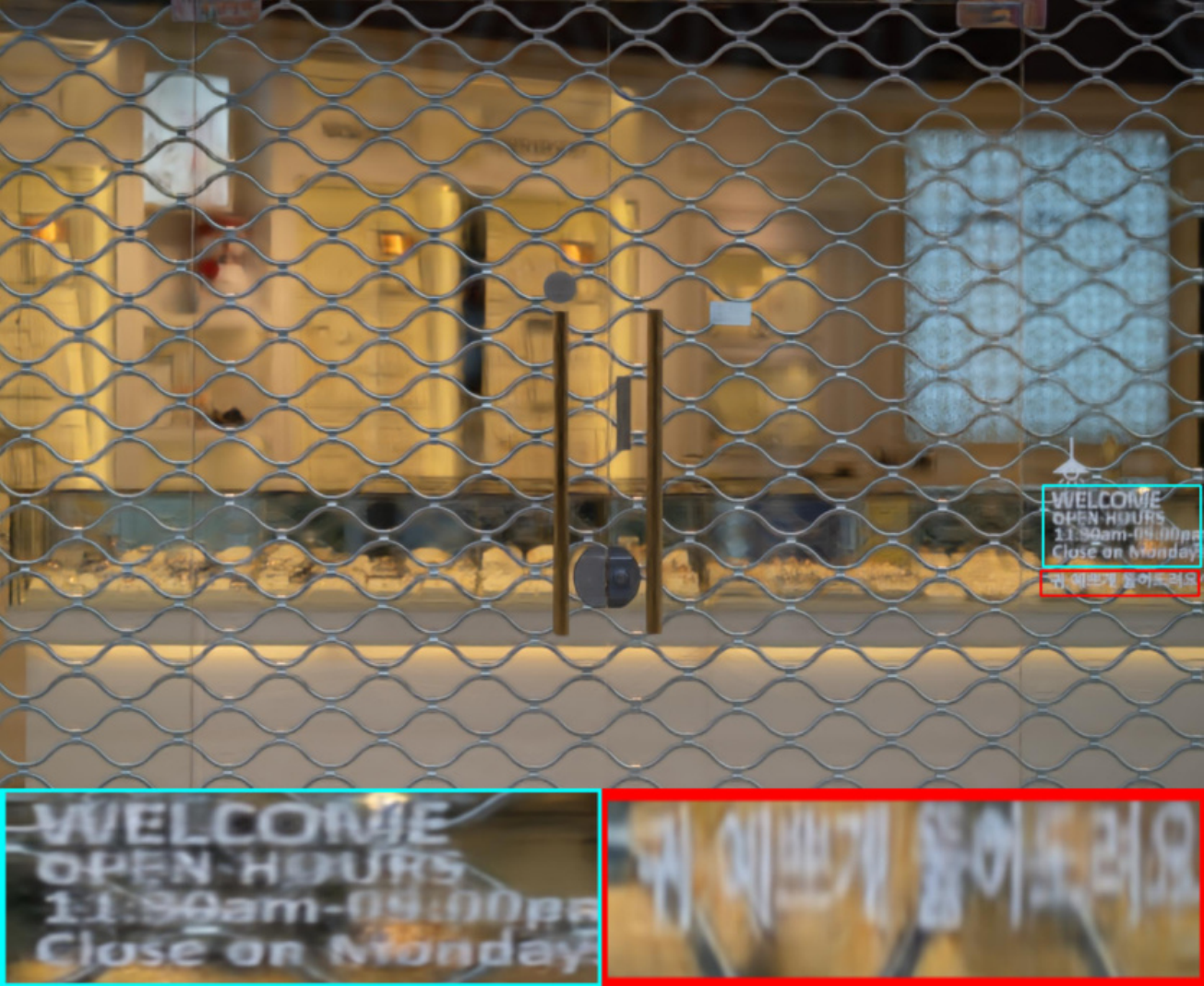}\vspace{4pt}
	\end{subfigure}
	\hfill
	\begin{subfigure}[t]{0.16\linewidth}	
		\includegraphics[width=2.6cm,height=4cm]{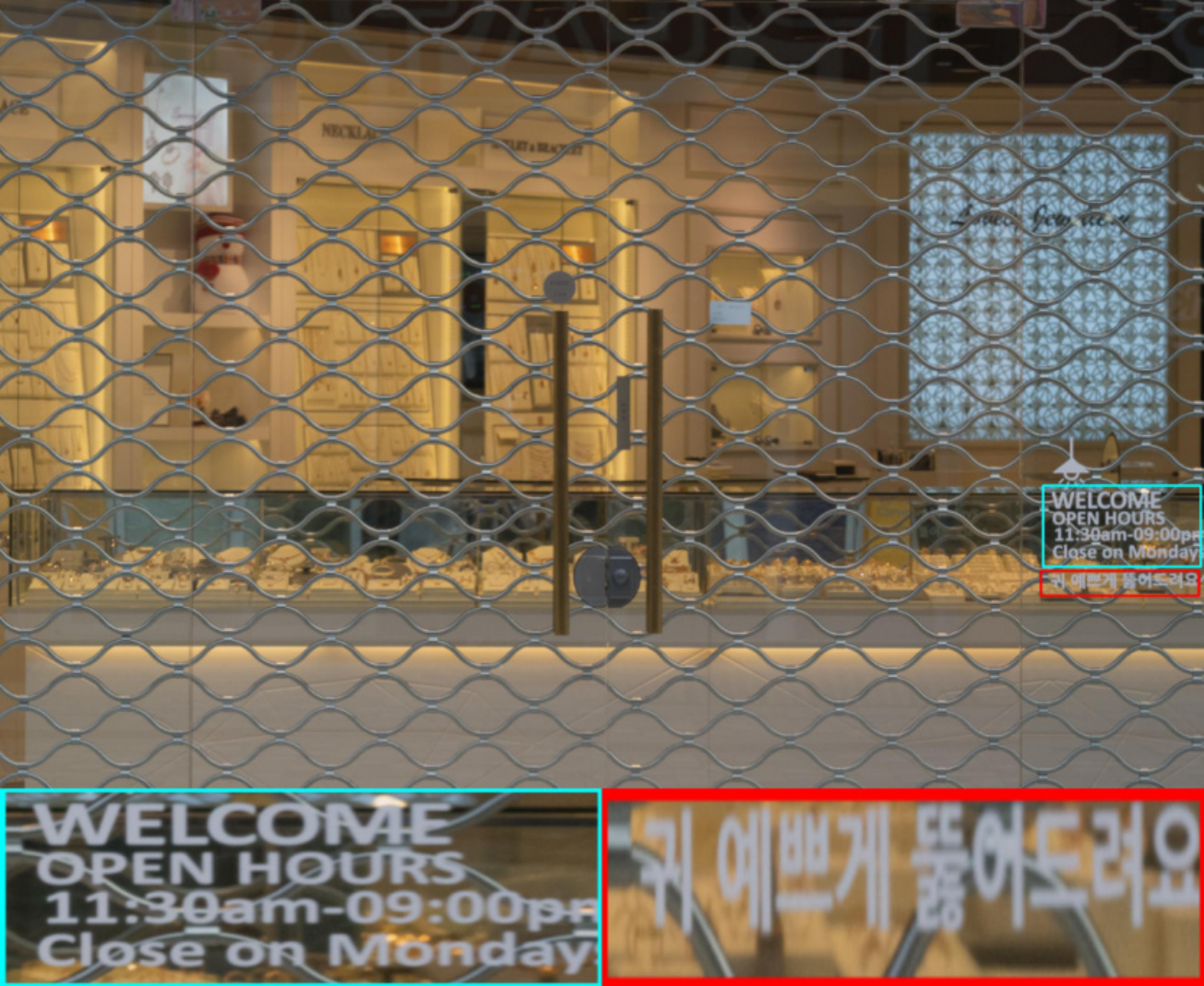}\vspace{4pt}
	\end{subfigure}
	\vfill
	
	\begin{subfigure}[t]{0.16\linewidth}	
		\includegraphics[width=2.6cm,height=4cm]{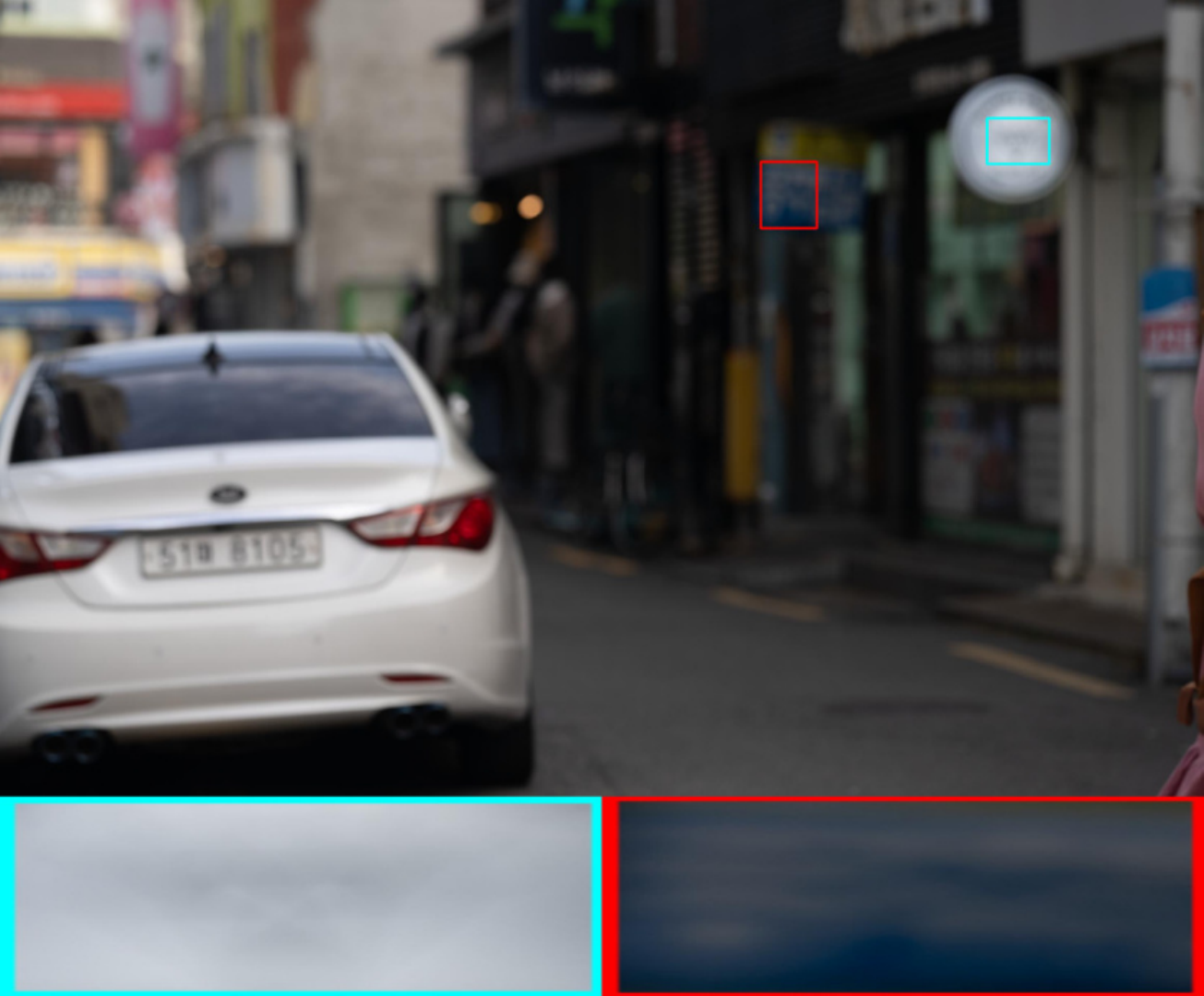}\vspace{4pt}
	\end{subfigure}
	\hfill
	\begin{subfigure}[t]{0.16\linewidth}	
		\includegraphics[width=2.6cm,height=4cm]{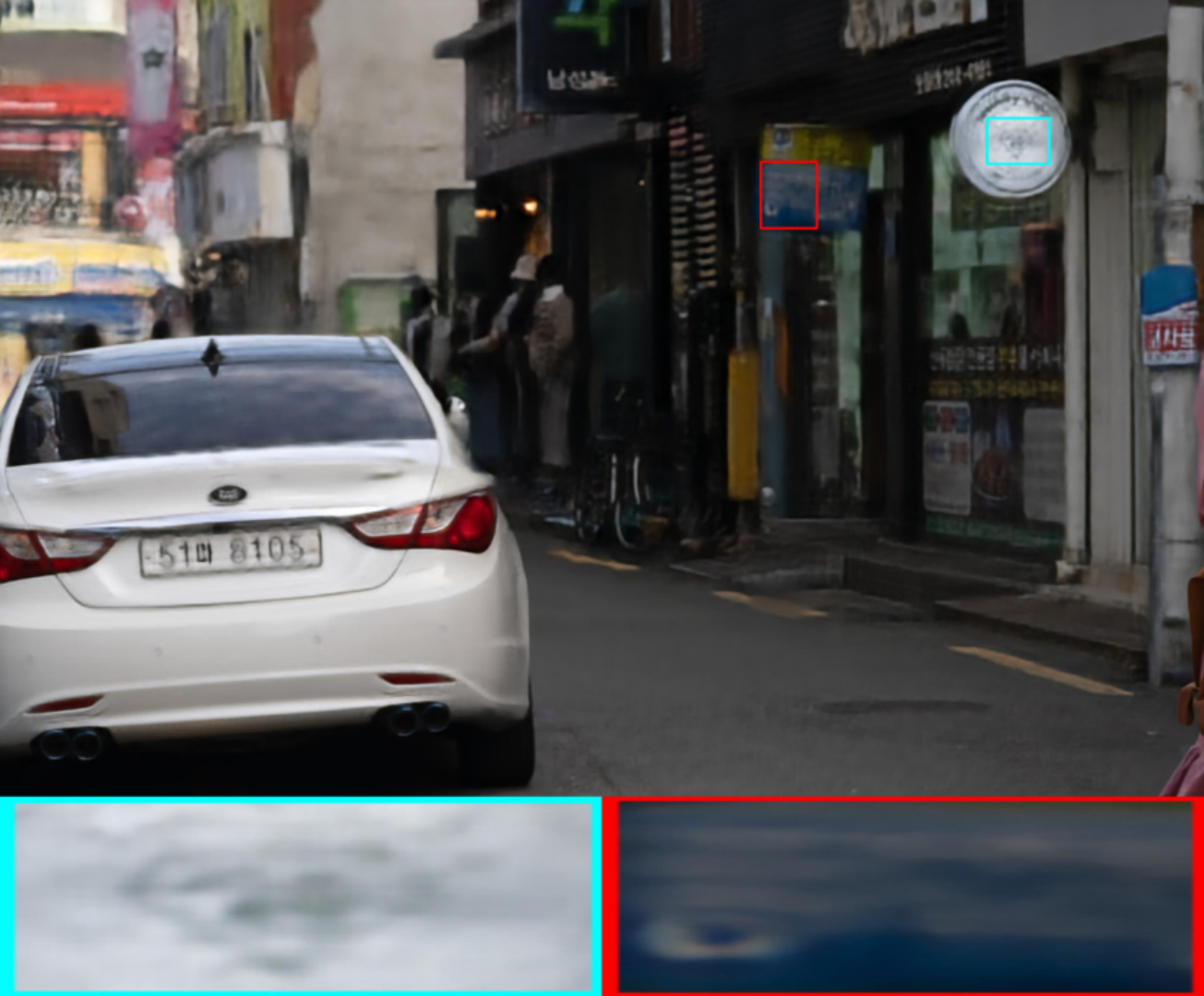}\vspace{4pt}
	\end{subfigure}
	\hfill
	\begin{subfigure}[t]{0.16\linewidth}	
		\includegraphics[width=2.6cm,height=4cm]{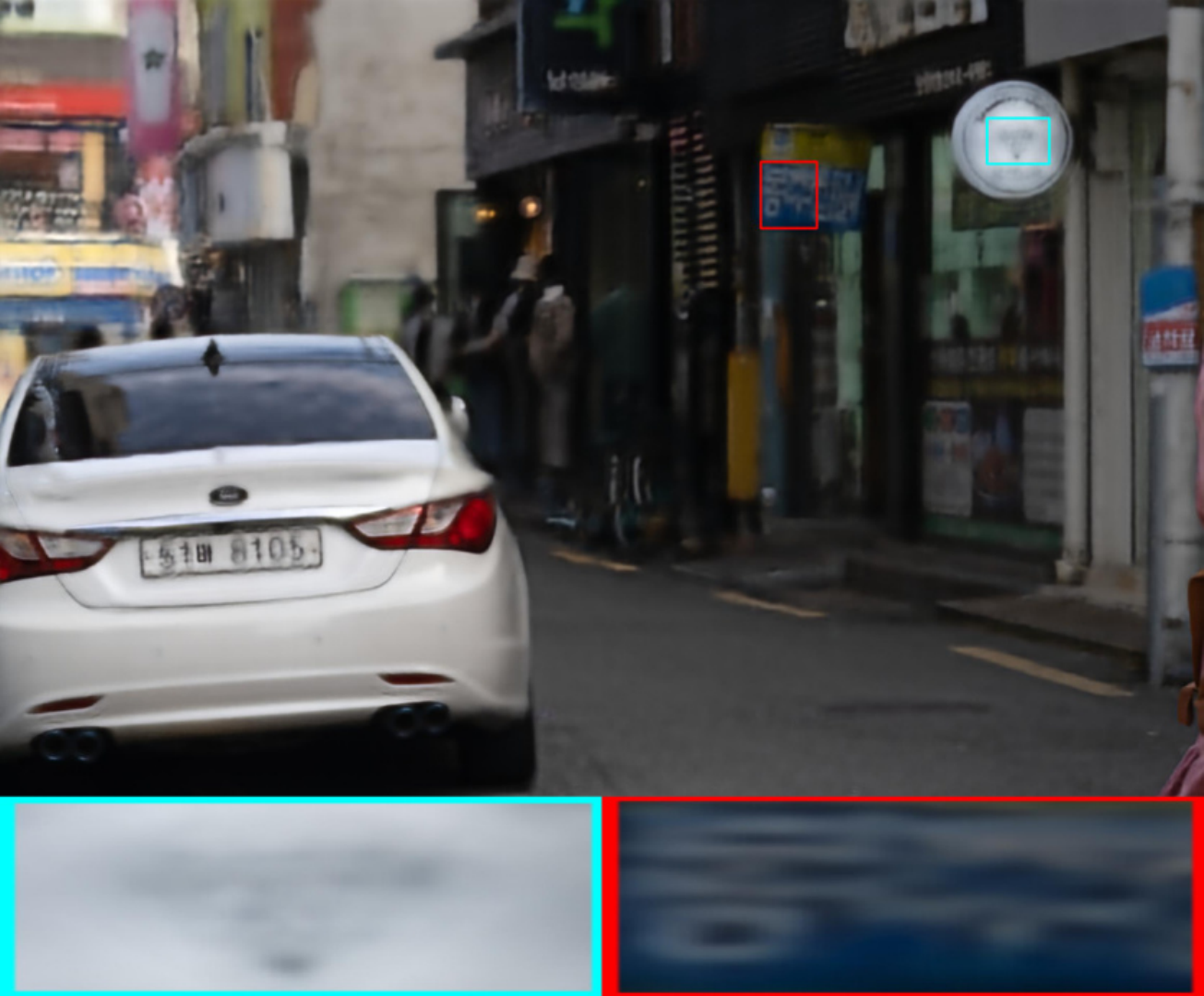}\vspace{4pt}
	\end{subfigure}
	\hfill
	\begin{subfigure}[t]{0.16\linewidth}	
		\includegraphics[width=2.6cm,height=4cm]{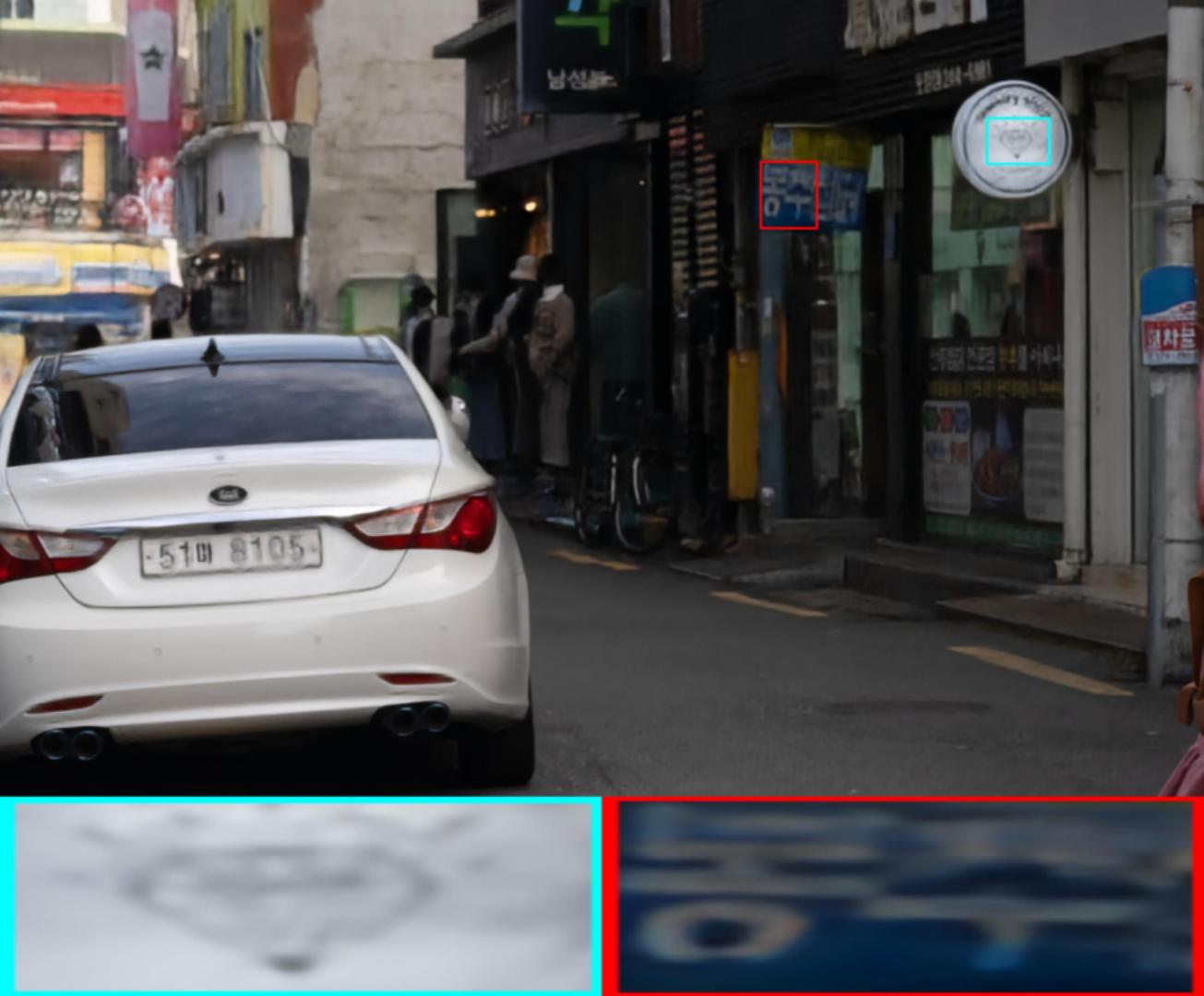}\vspace{4pt}
	\end{subfigure}
	\hfill
	\begin{subfigure}[t]{0.16\linewidth}	
		\includegraphics[width=2.6cm,height=4cm]{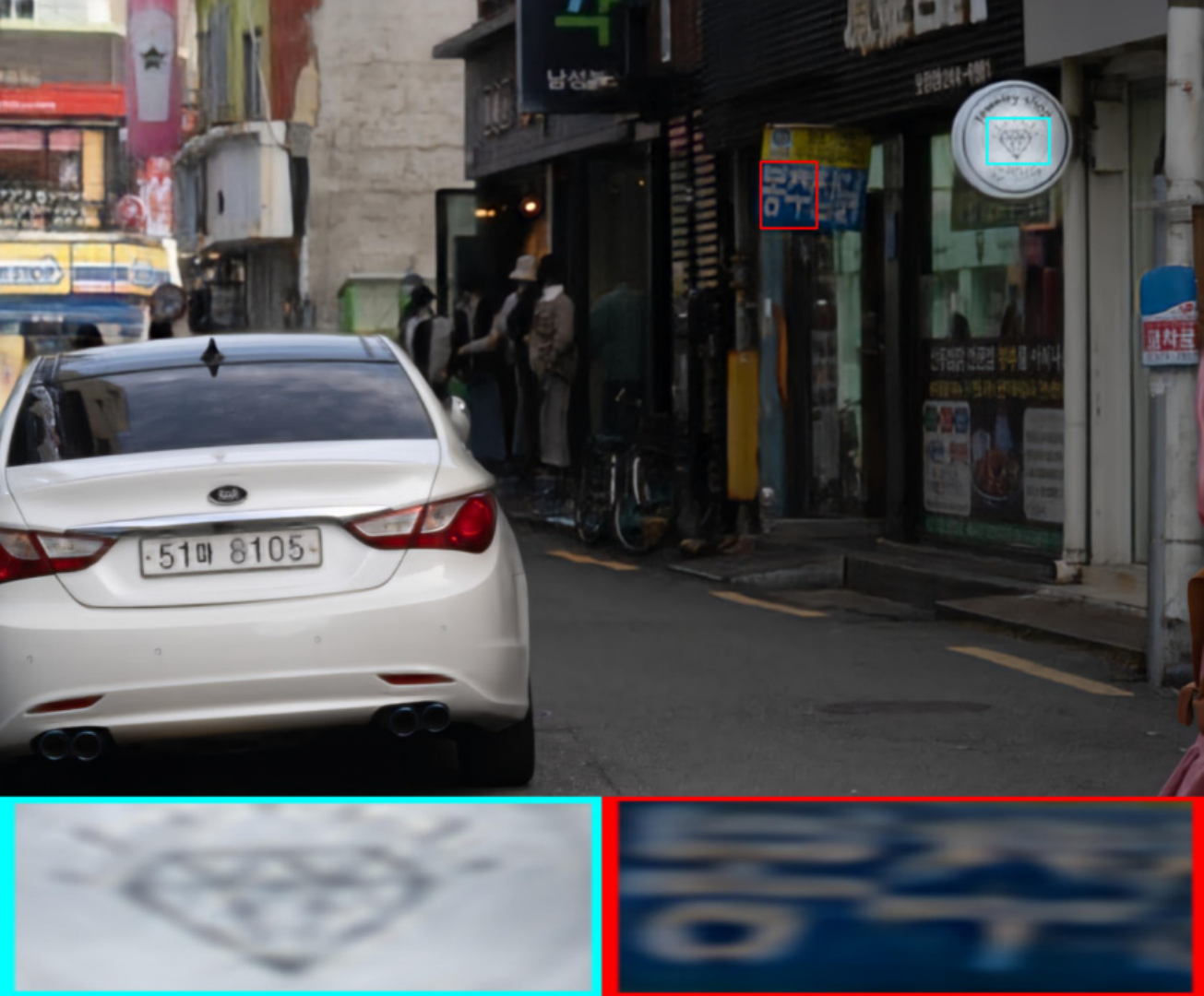}\vspace{4pt}
	\end{subfigure}
	\hfill
	\begin{subfigure}[t]{0.16\linewidth}	
		\includegraphics[width=2.6cm,height=4cm]{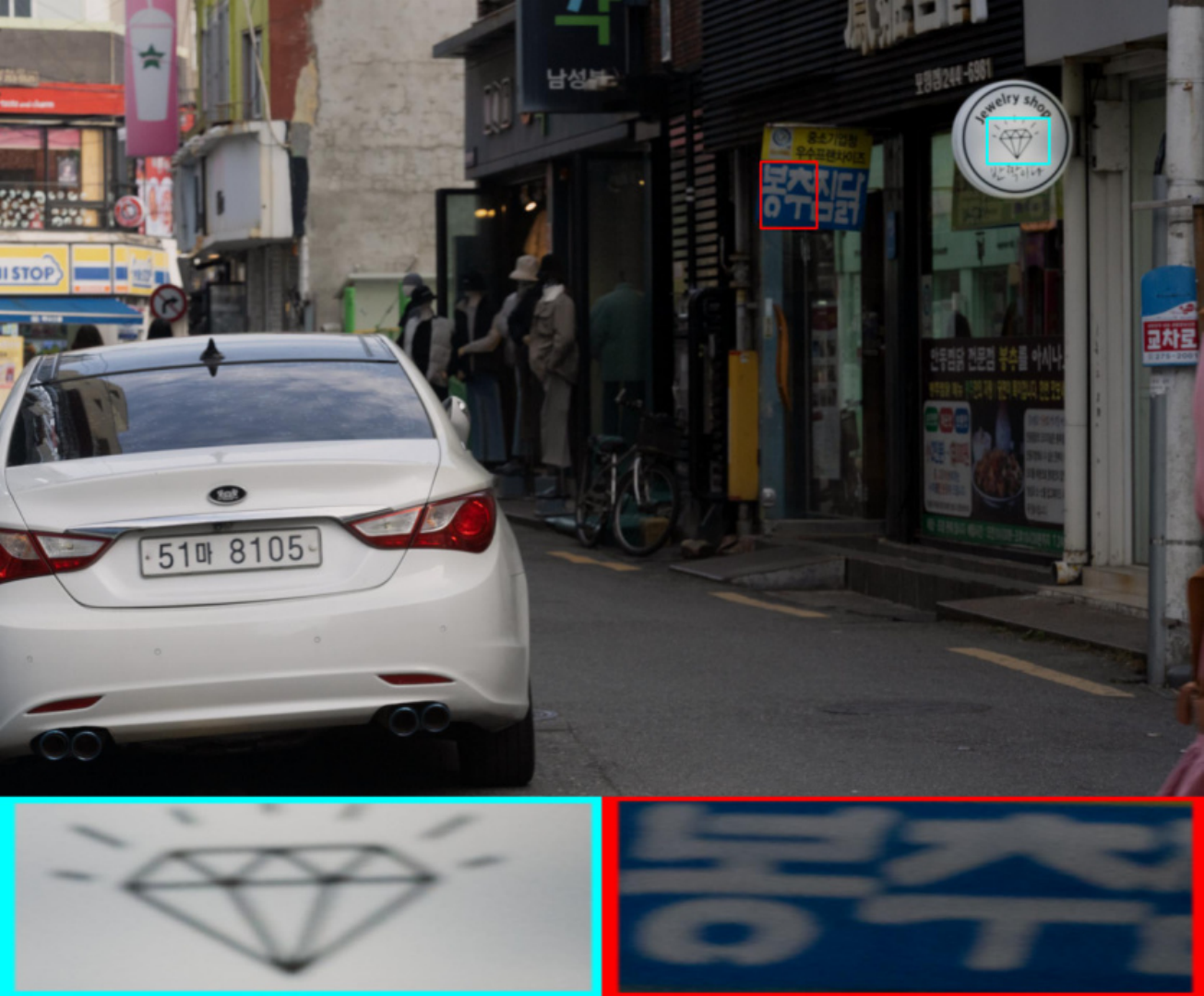}\vspace{4pt}
	\end{subfigure}
	\vfill
	
	\begin{subfigure}[t]{0.16\linewidth}	
		\includegraphics[width=2.6cm,height=4cm]{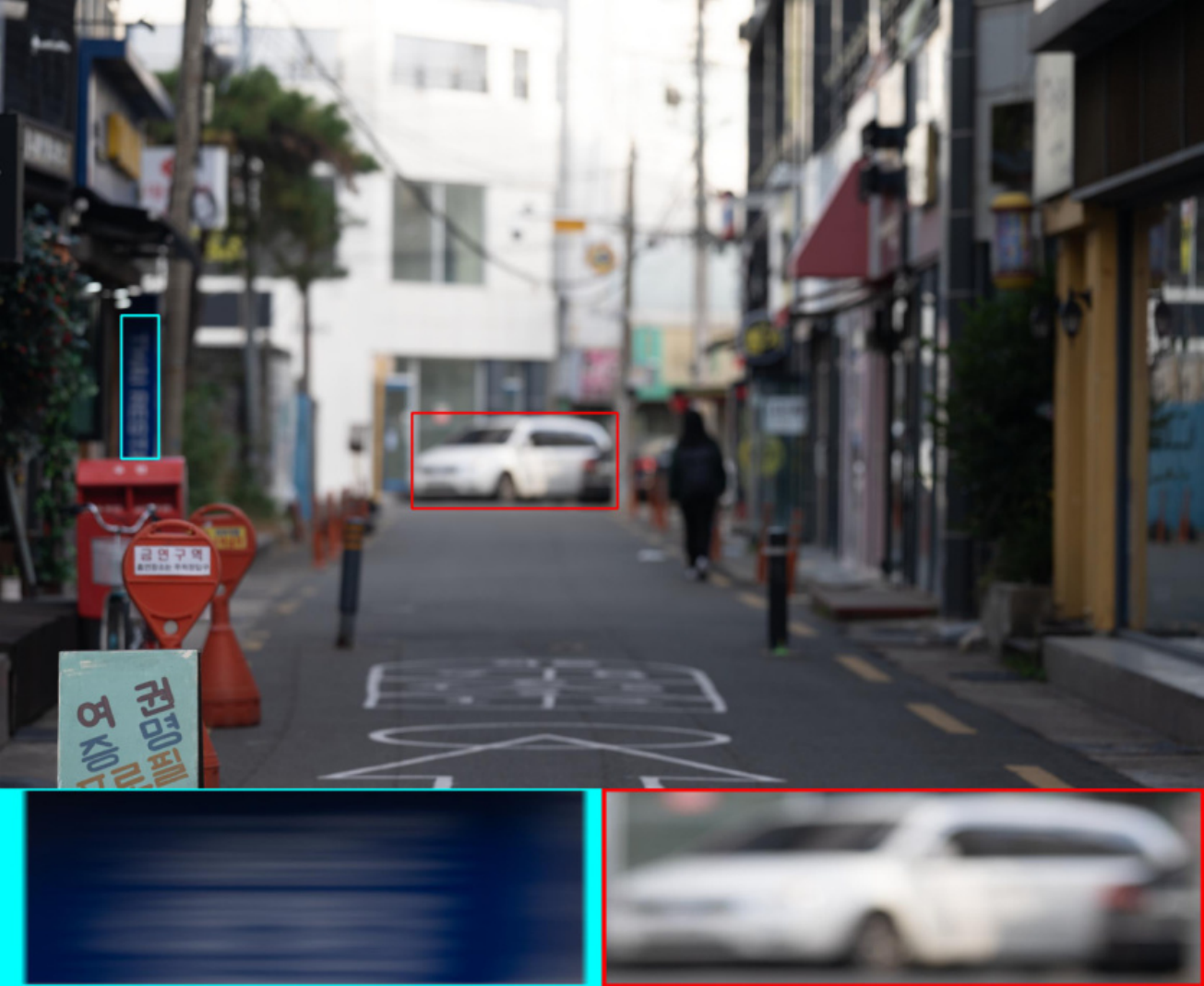}\vspace{4pt}
	\end{subfigure}
	\hfill
	\begin{subfigure}[t]{0.16\linewidth}	
		\includegraphics[width=2.6cm,height=4cm]{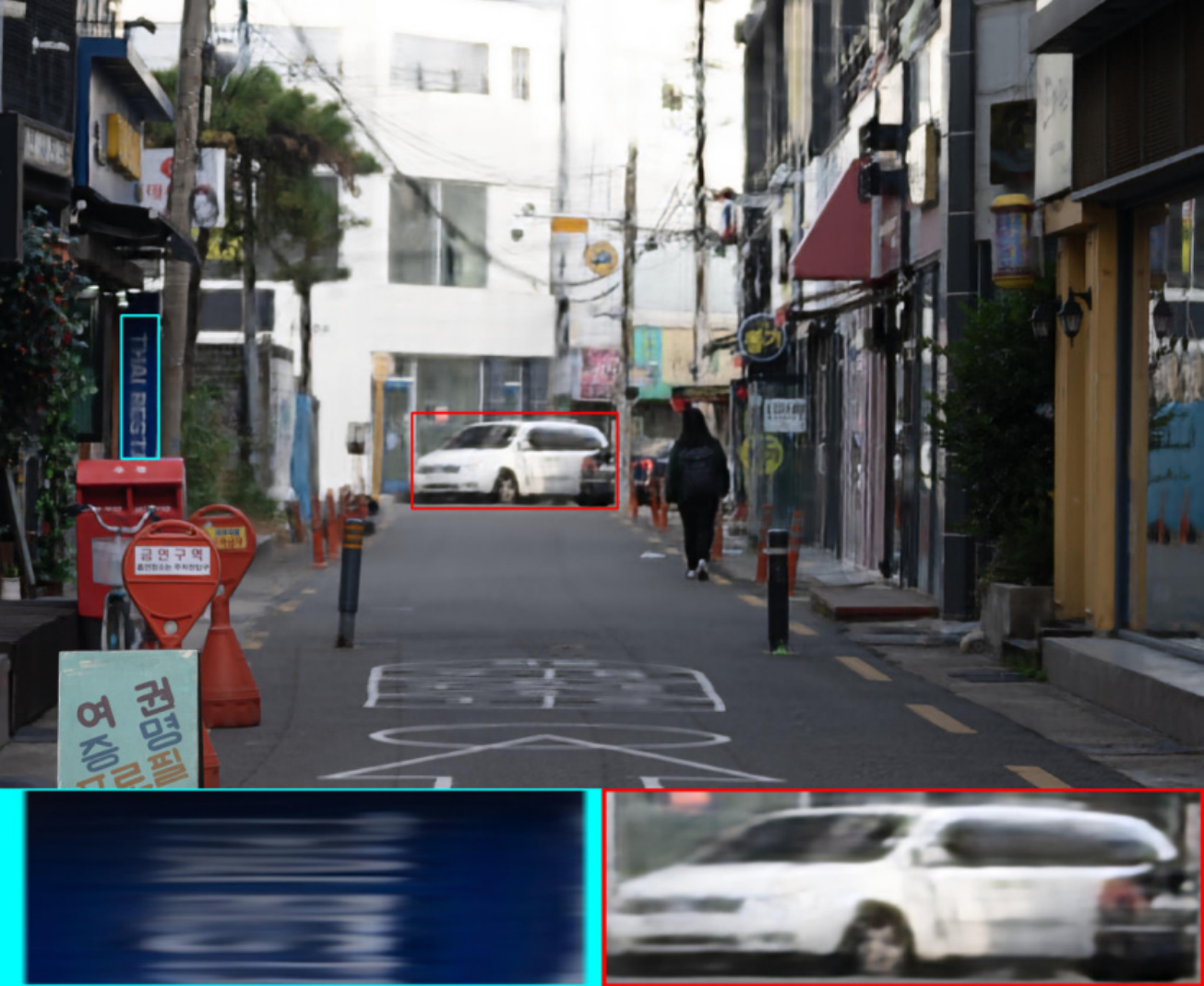}\vspace{4pt}
	\end{subfigure}
	\hfill
	\begin{subfigure}[t]{0.16\linewidth}	
		\includegraphics[width=2.6cm,height=4cm]{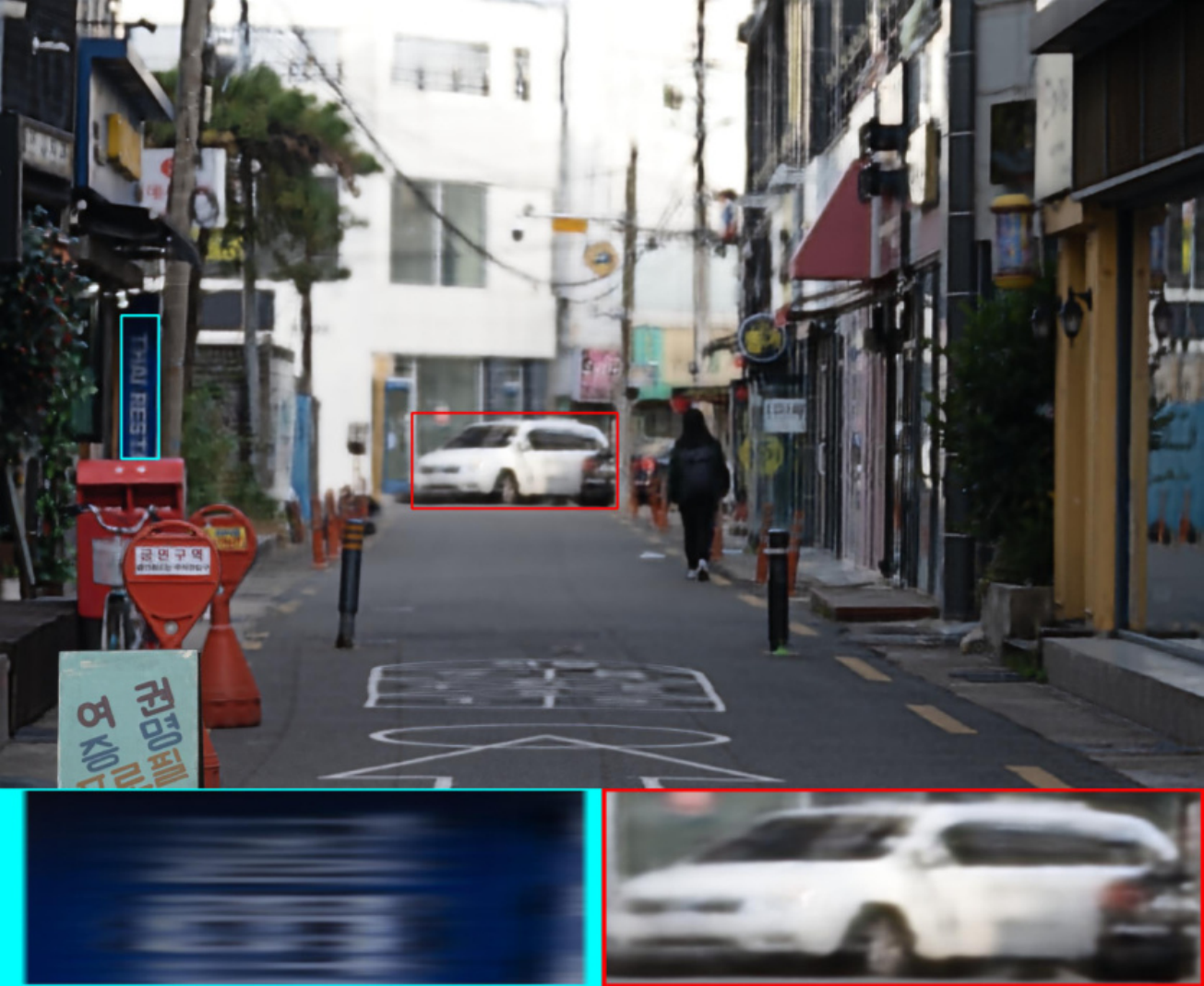}\vspace{4pt}
	\end{subfigure}
	\hfill
	\begin{subfigure}[t]{0.16\linewidth}	
		\includegraphics[width=2.6cm,height=4cm]{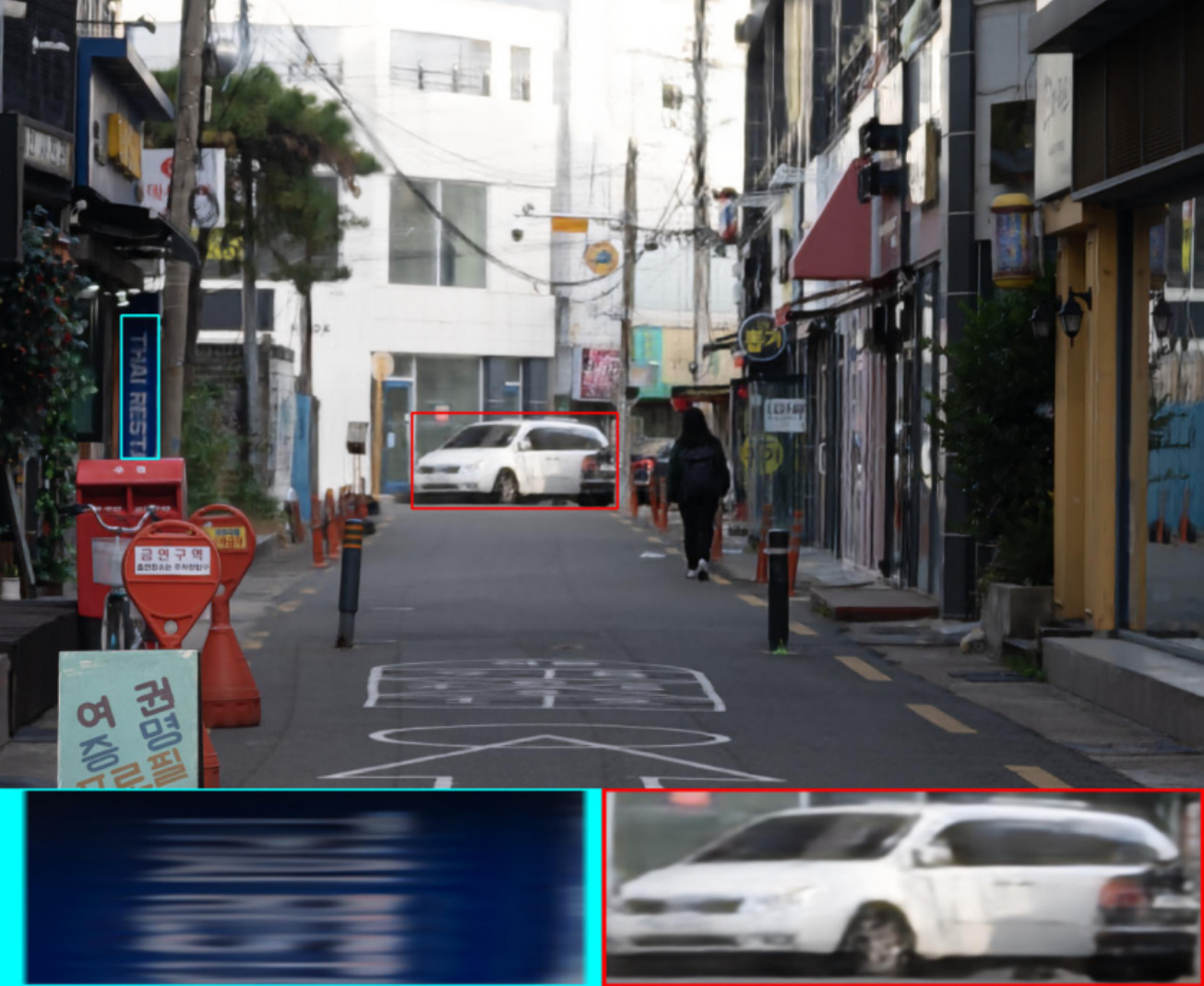}\vspace{4pt}
	\end{subfigure}
	\hfill
	\begin{subfigure}[t]{0.16\linewidth}	
		\includegraphics[width=2.6cm,height=4cm]{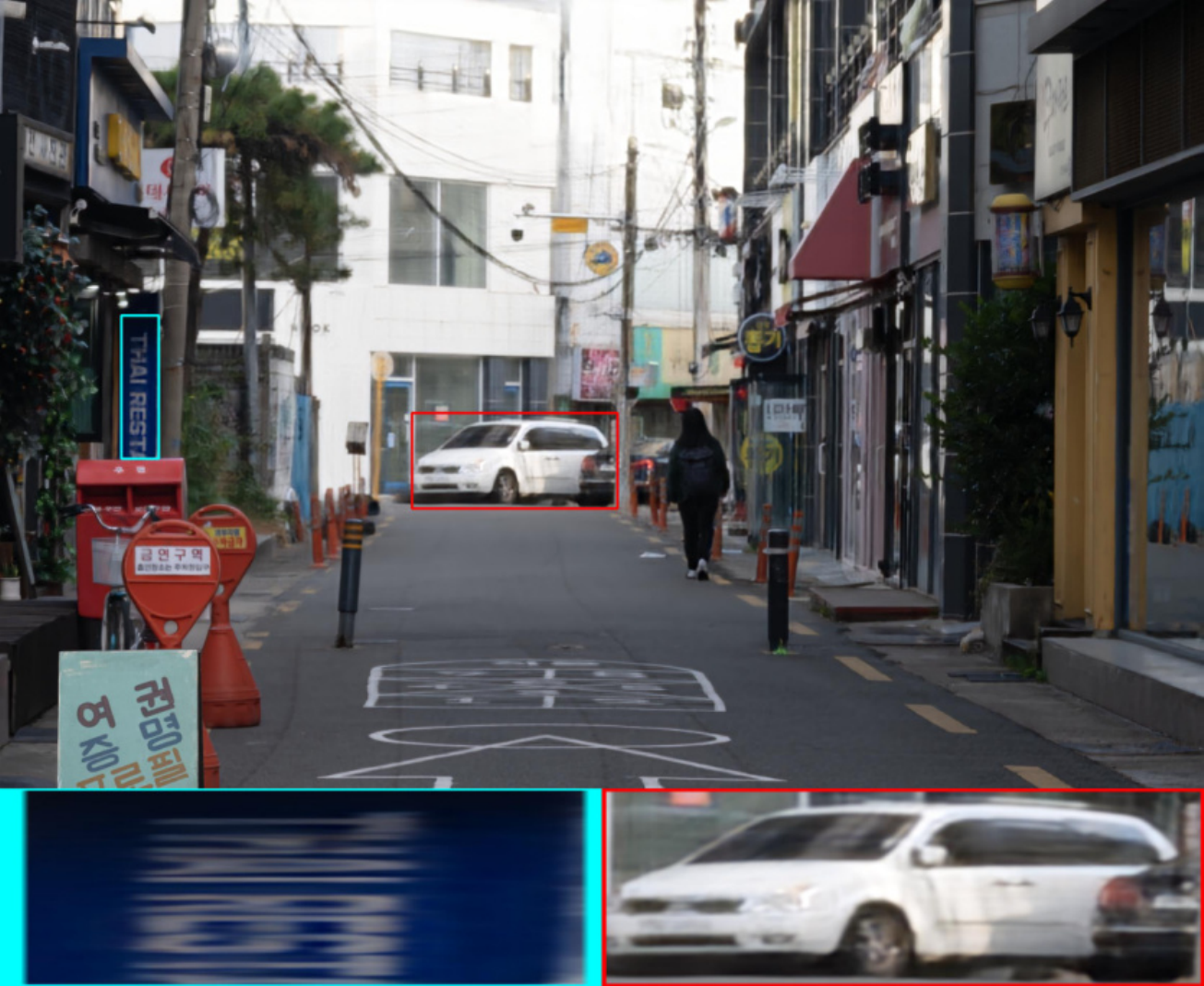}\vspace{4pt}
	\end{subfigure}
	\hfill
	\begin{subfigure}[t]{0.16\linewidth}	
		\includegraphics[width=2.6cm,height=4cm]{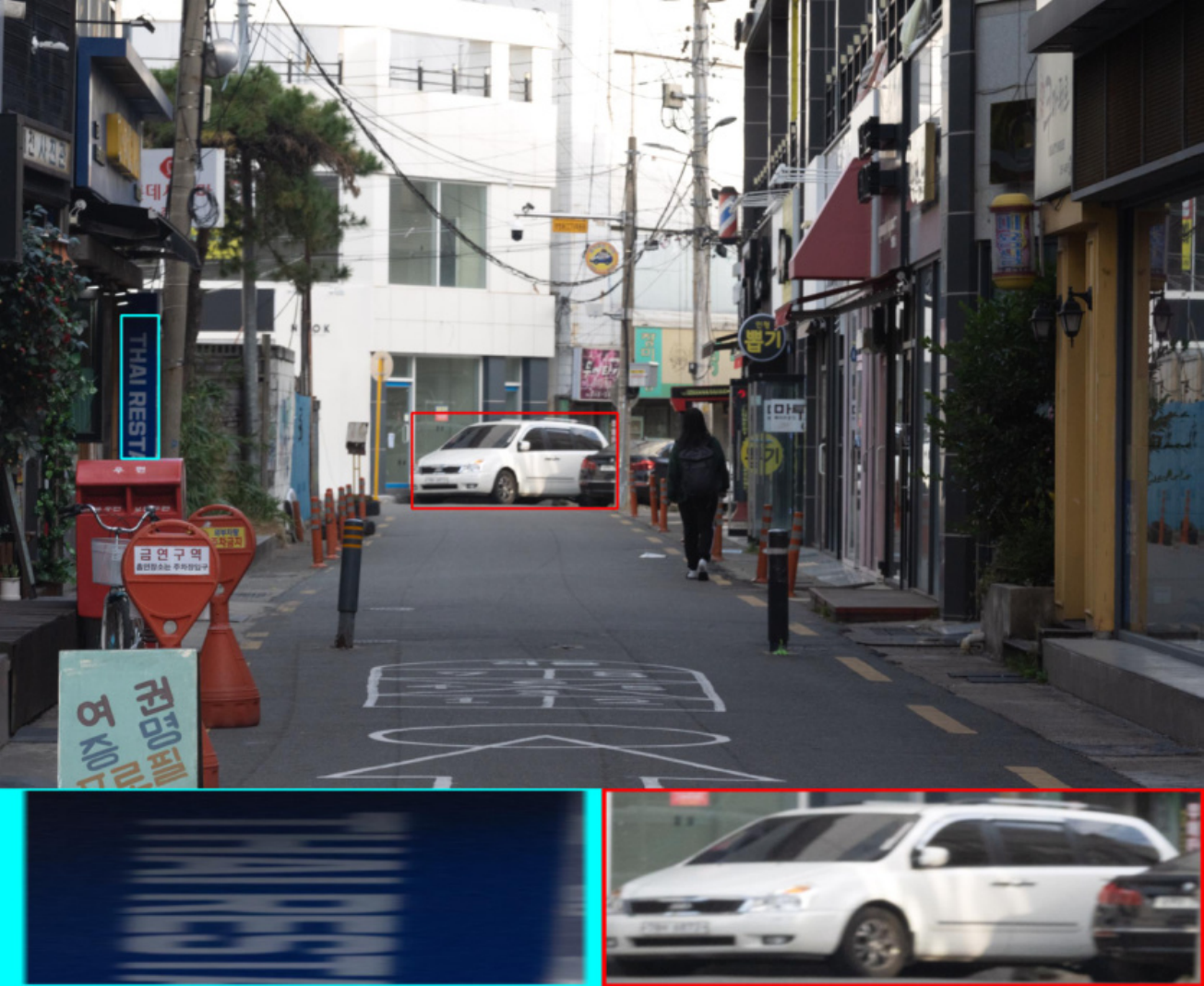}\vspace{4pt}
	\end{subfigure}
	\vfill
	
	\begin{subfigure}[t]{0.16\linewidth}	
		\includegraphics[width=2.6cm,height=4cm]{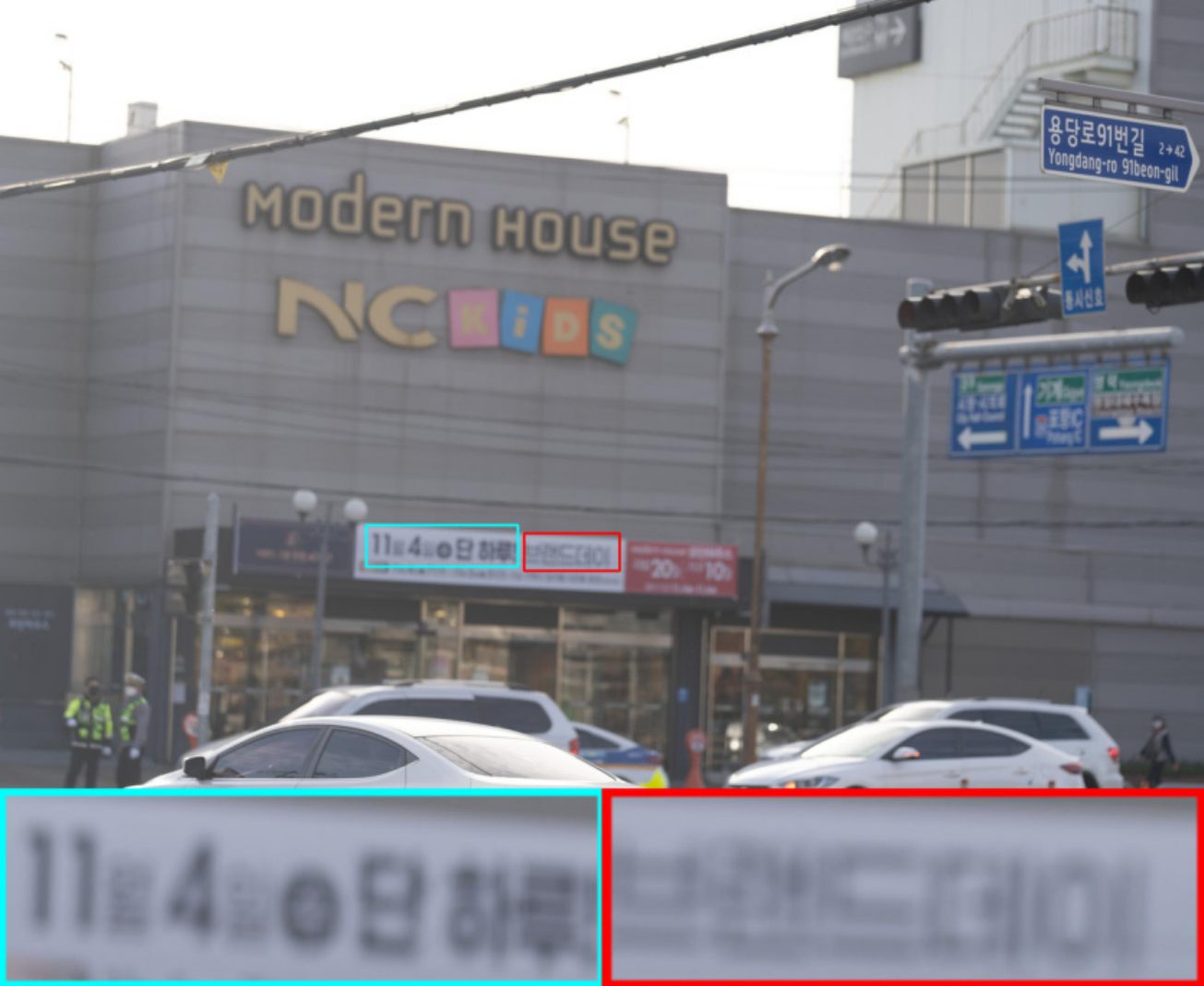}\vspace{4pt}
		\caption{Blurry Input}
	\end{subfigure}
	\hfill
	\begin{subfigure}[t]{0.16\linewidth}	
		\includegraphics[width=2.6cm,height=4cm]{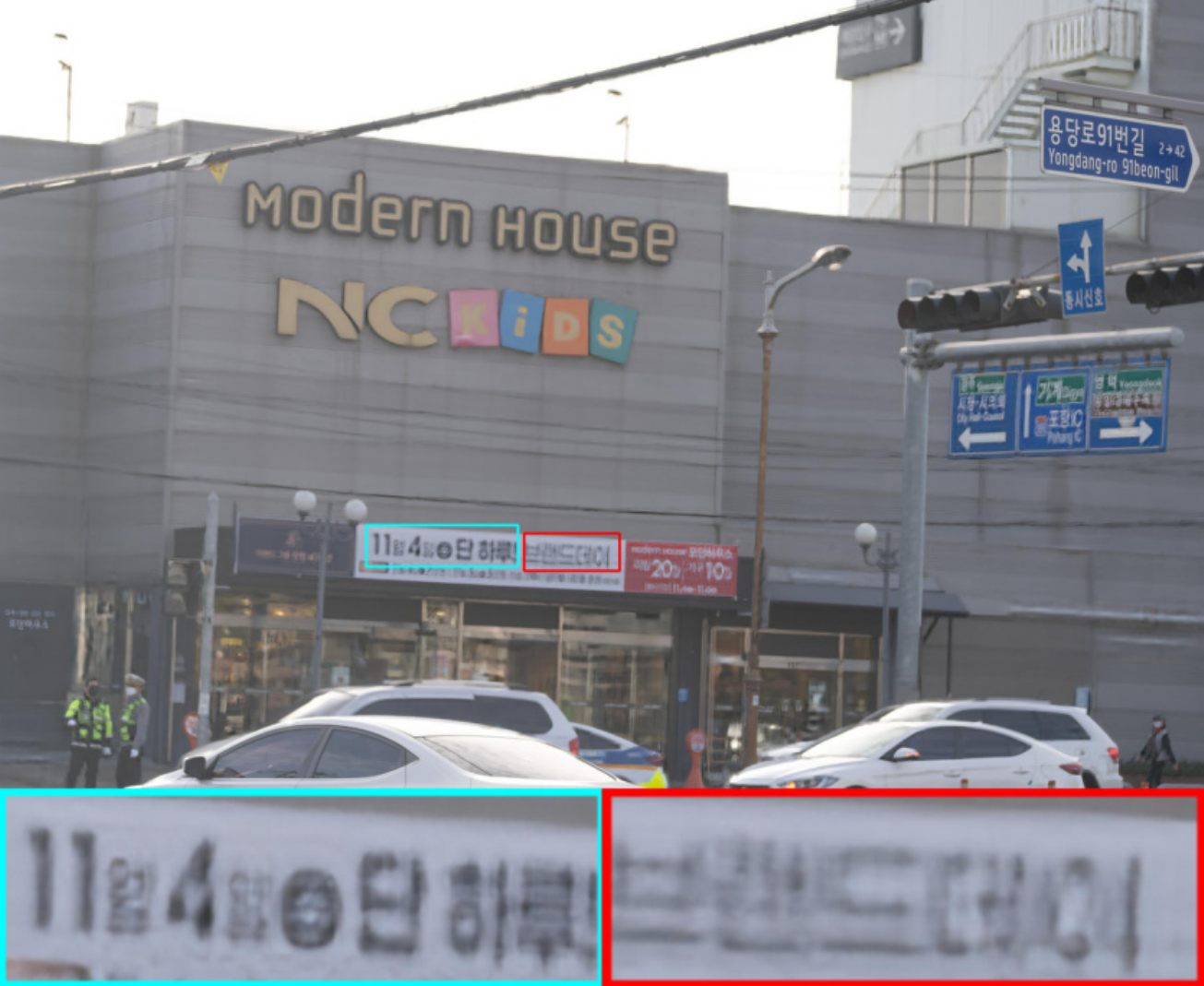}\vspace{4pt}
		\caption{KPAC}
		
	\end{subfigure}
	\hfill
	\begin{subfigure}[t]{0.16\linewidth}	
		\includegraphics[width=2.6cm,height=4cm]{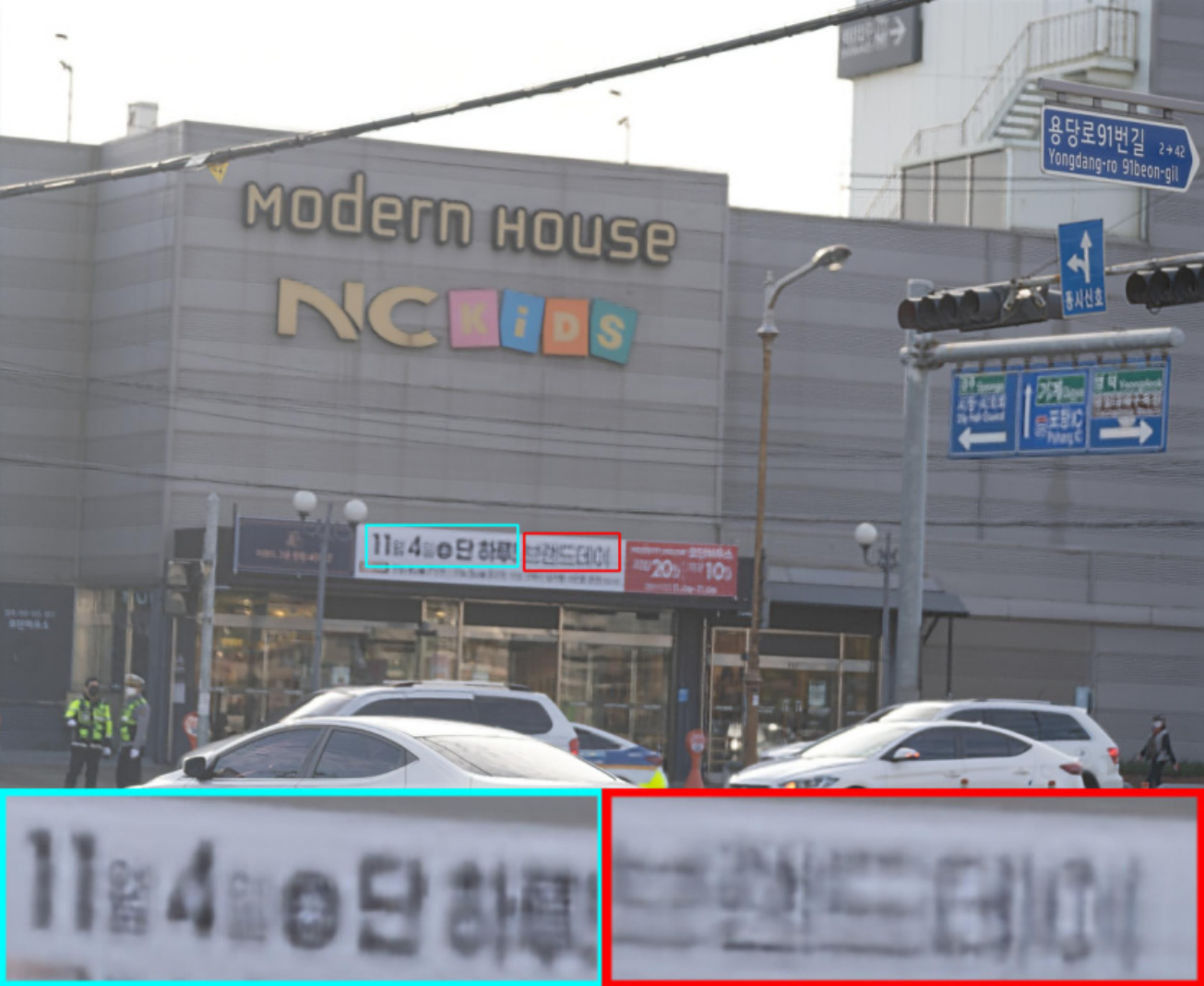}\vspace{4pt}
		\caption{GKMNet}
		
	\end{subfigure}
	\hfill
	\begin{subfigure}[t]{0.16\linewidth}	
		\includegraphics[width=2.6cm,height=4cm]{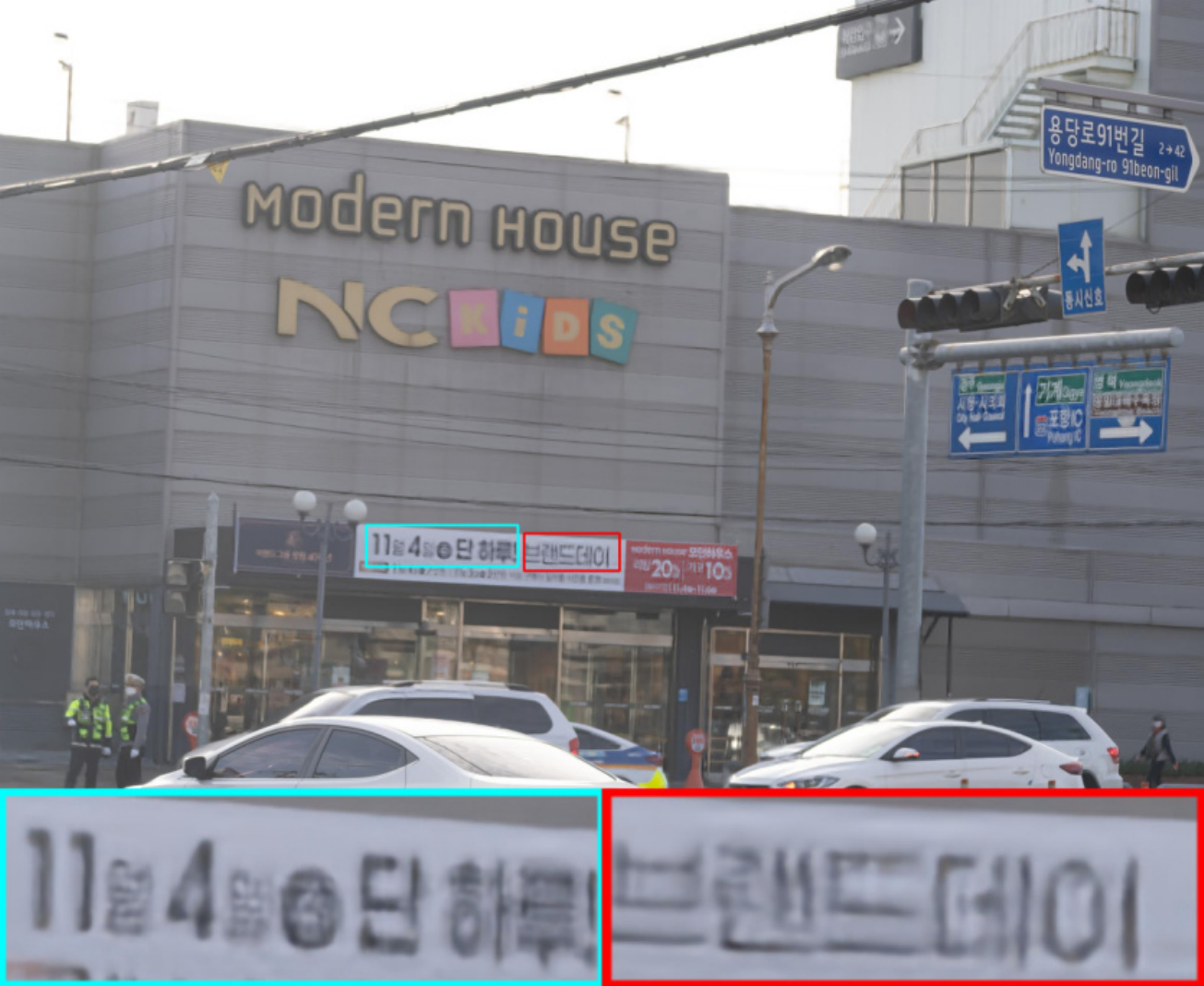}\vspace{4pt}
		\caption{IFAN}
		
	\end{subfigure}
	\hfill
	\begin{subfigure}[t]{0.16\linewidth}	
		\includegraphics[width=2.6cm,height=4cm]{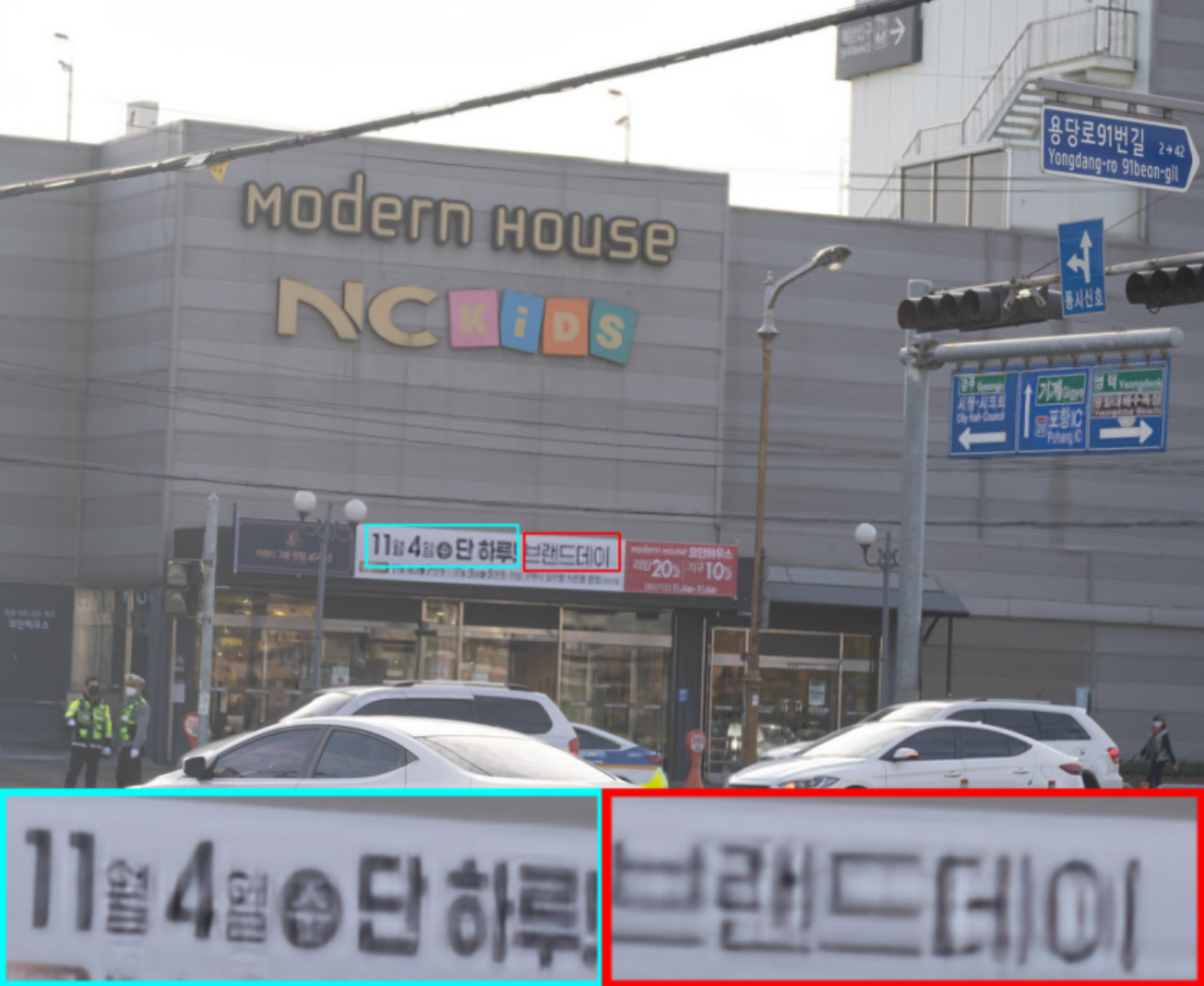}\vspace{4pt}
		\caption{SR-R$^2$KAC-B}
	\end{subfigure}
	\hfill
	\begin{subfigure}[t]{0.16\linewidth}	
		\includegraphics[width=2.6cm,height=4cm]{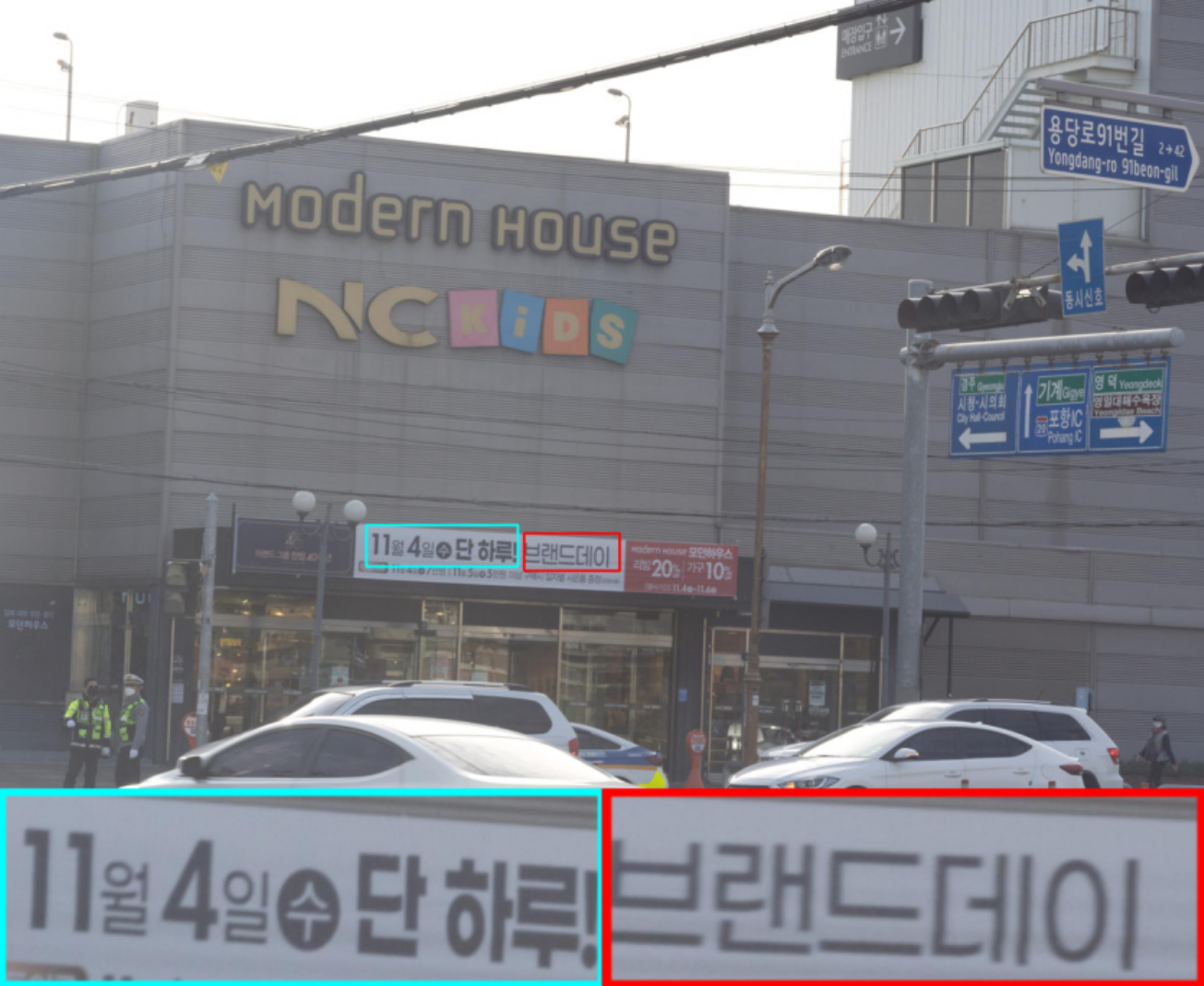}\vspace{4pt}
		\caption{Sharp}
	\end{subfigure}
	\vfill
	
	\caption{Visualization results of different defocus deblurring methods on the RealDoF dataset \cite{Lee2021}.}
	\vspace{-0.2in}
	\label{fig_s_3}
\end{figure*}

\begin{figure*}[h]
	\centering
	
	\begin{subfigure}[t]{0.16\linewidth}	
		\includegraphics[width=2.6cm,height=4cm]{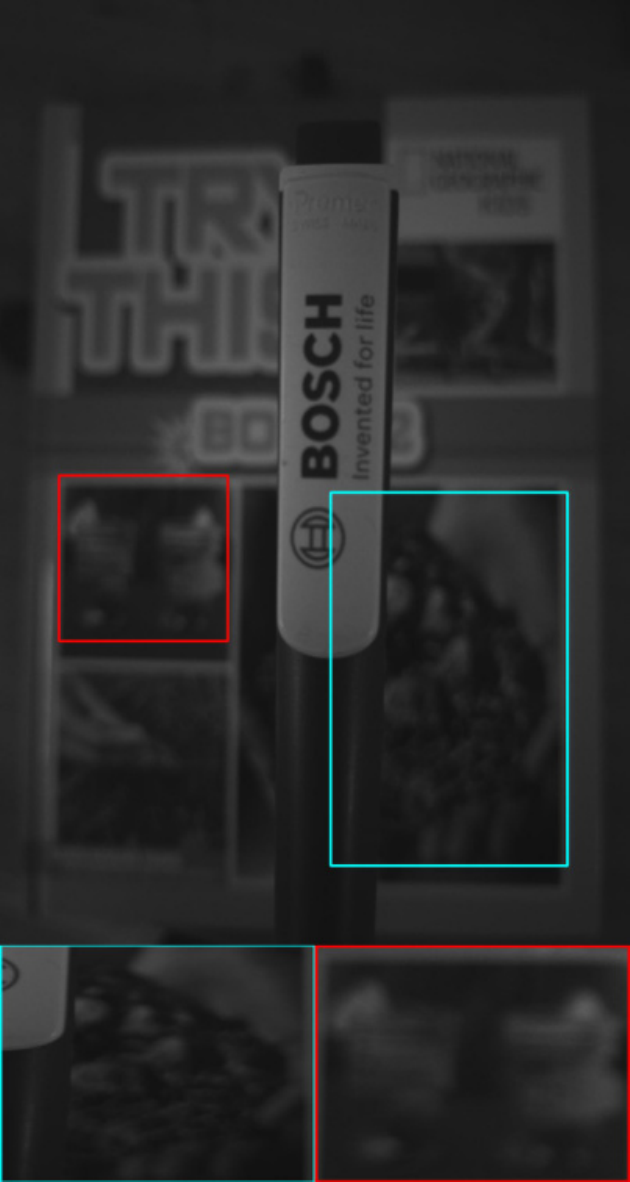}\vspace{4pt}
	\end{subfigure}
	\hfill
	\begin{subfigure}[t]{0.16\linewidth}	
		\includegraphics[width=2.6cm,height=4cm]{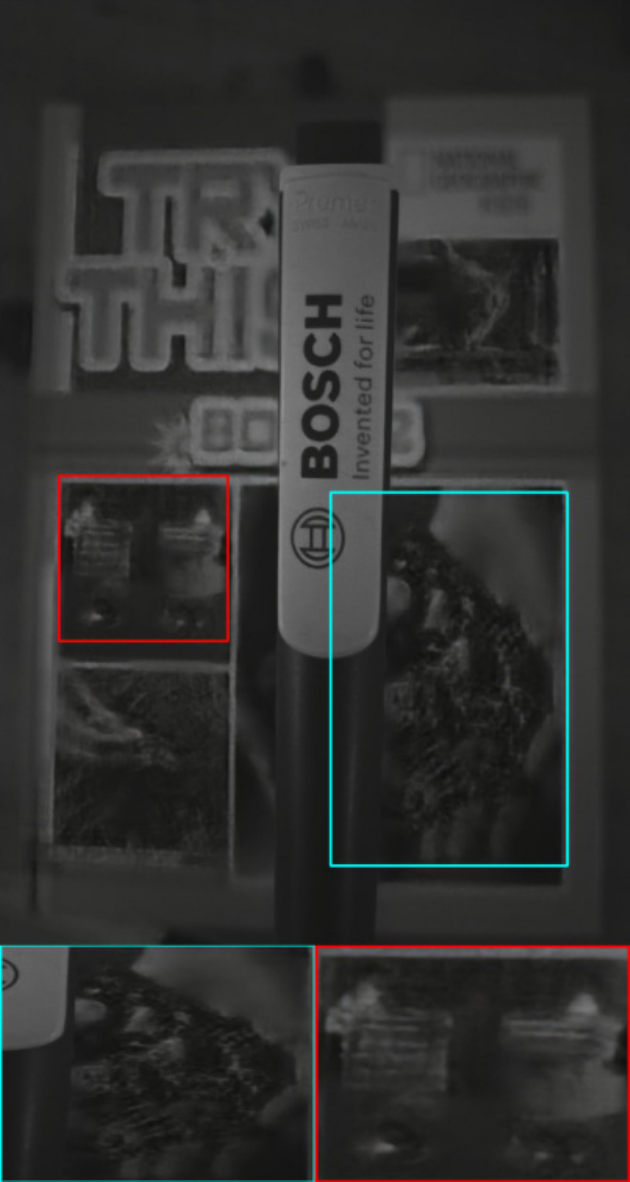}\vspace{4pt}
	\end{subfigure}
	\hfill
	\begin{subfigure}[t]{0.16\linewidth}	
		\includegraphics[width=2.6cm,height=4cm]{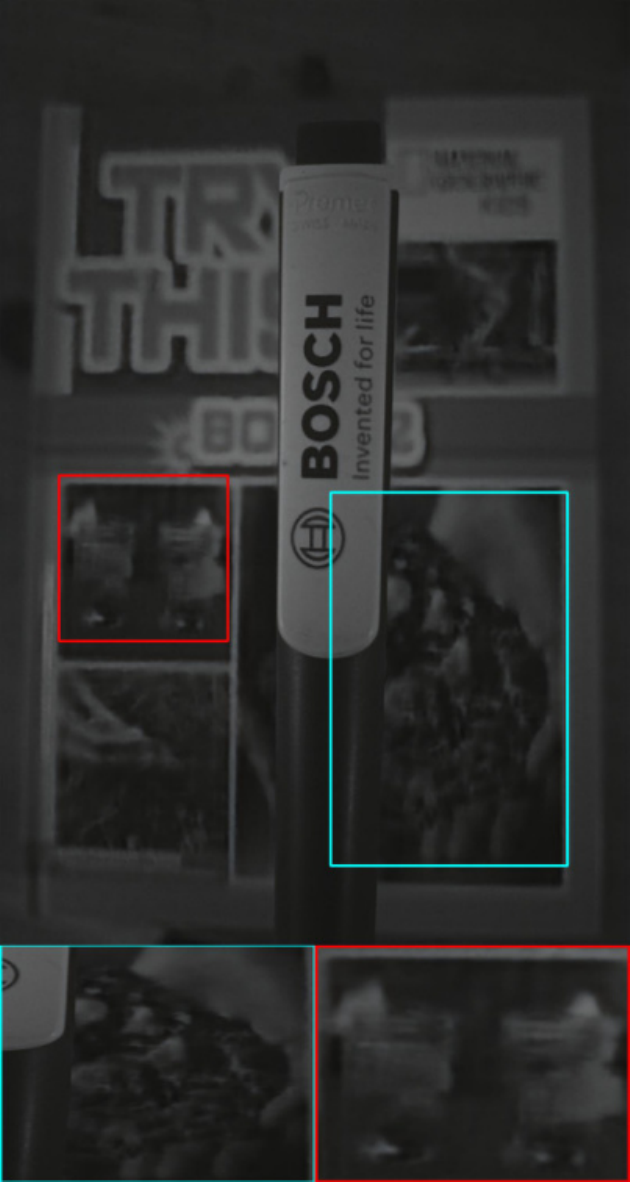}\vspace{4pt}
	\end{subfigure}
	\hfill
	\begin{subfigure}[t]{0.16\linewidth}	
		\includegraphics[width=2.6cm,height=4cm]{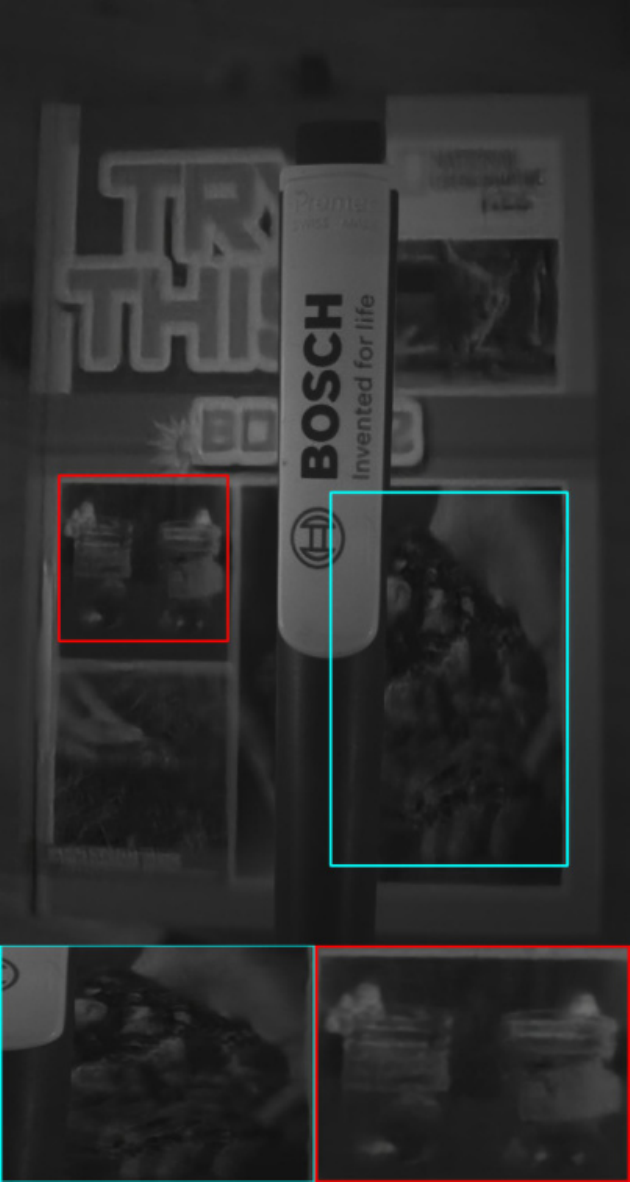}\vspace{4pt}
	\end{subfigure}
	\hfill
	\begin{subfigure}[t]{0.16\linewidth}	
		\includegraphics[width=2.6cm,height=4cm]{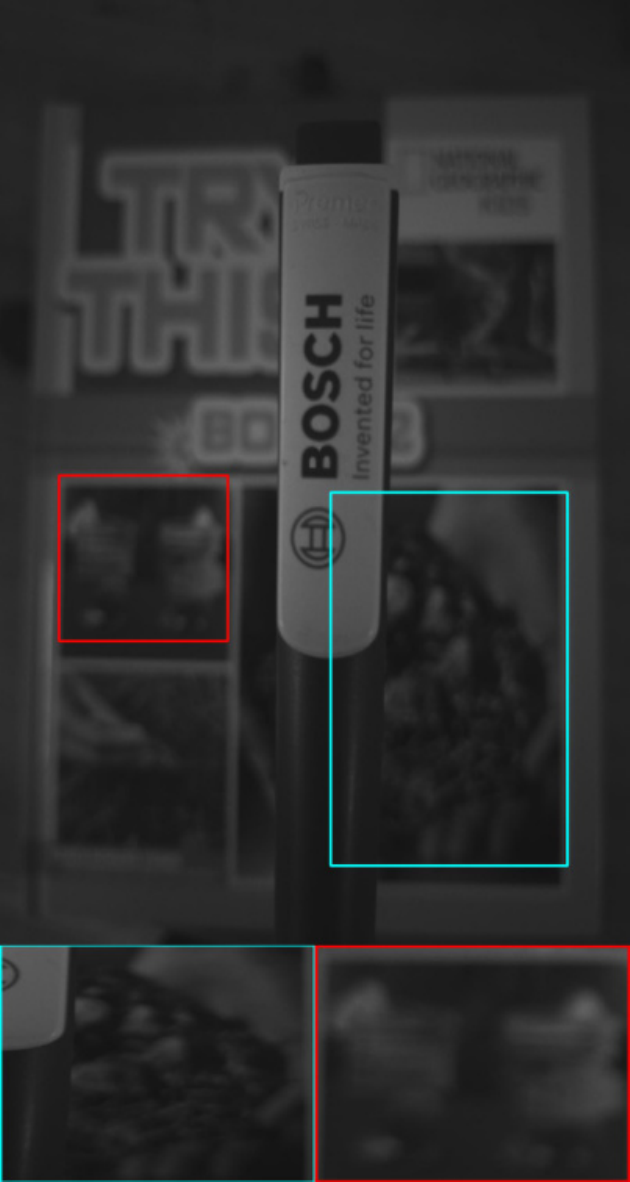}\vspace{4pt}
	\end{subfigure}
	\hfill
	\begin{subfigure}[t]{0.16\linewidth}	
		\includegraphics[width=2.6cm,height=4cm]{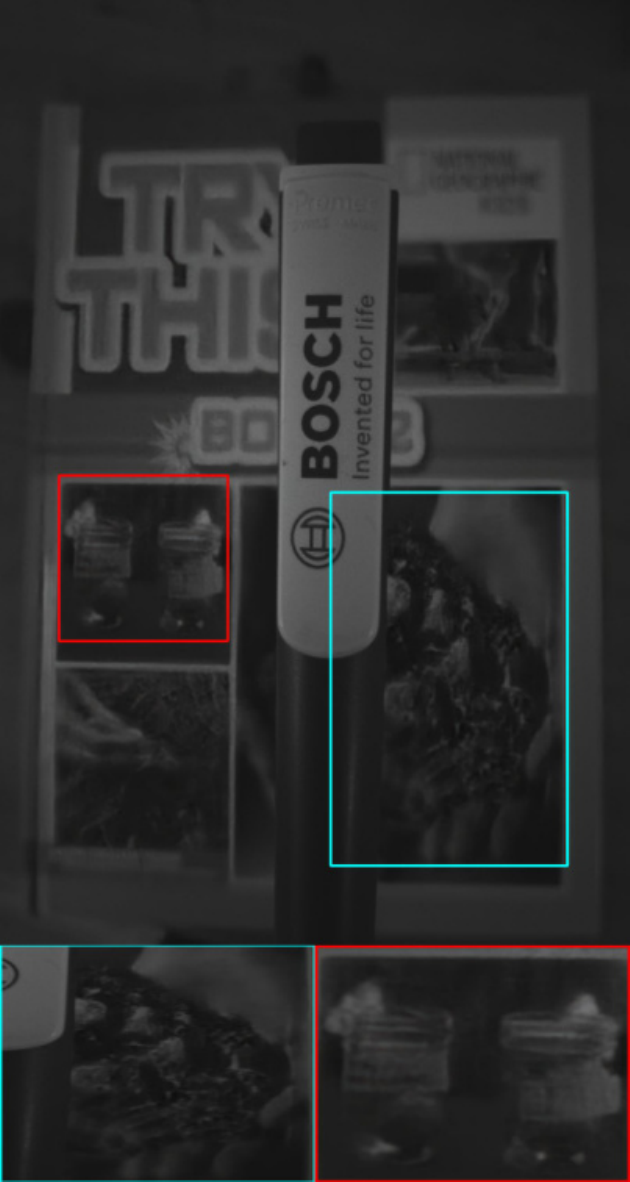}\vspace{4pt}
	\end{subfigure}
	\vfill
	
	\begin{subfigure}[t]{0.16\linewidth}	
		\includegraphics[width=2.6cm,height=4cm]{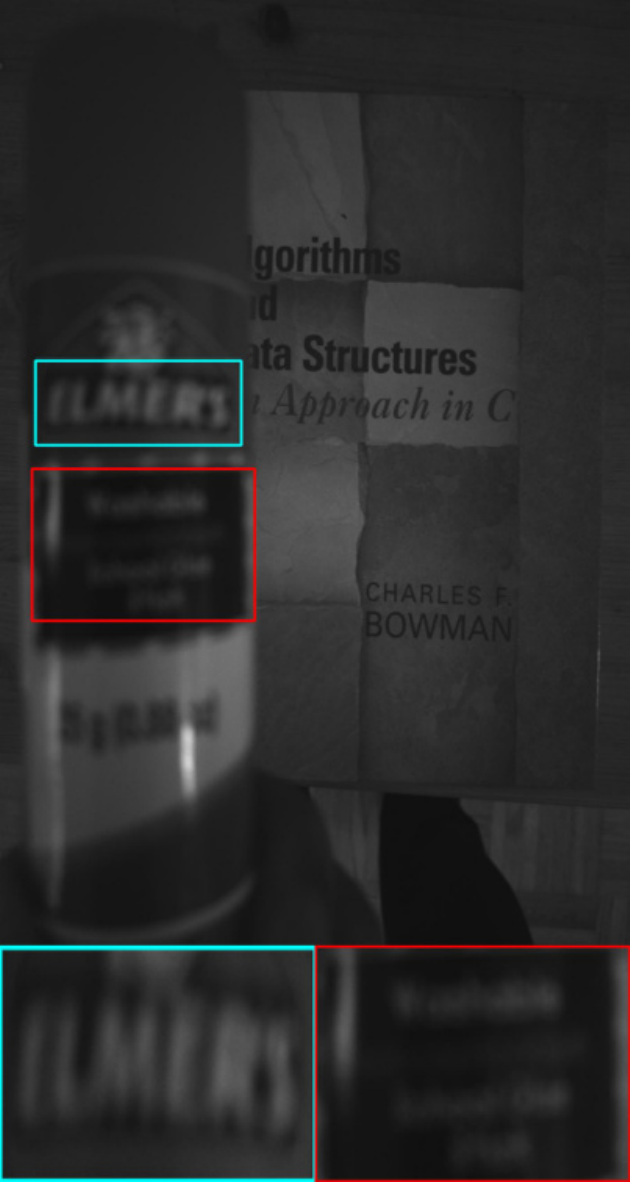}\vspace{4pt}
	\end{subfigure}
	\hfill
	\begin{subfigure}[t]{0.16\linewidth}	
		\includegraphics[width=2.6cm,height=4cm]{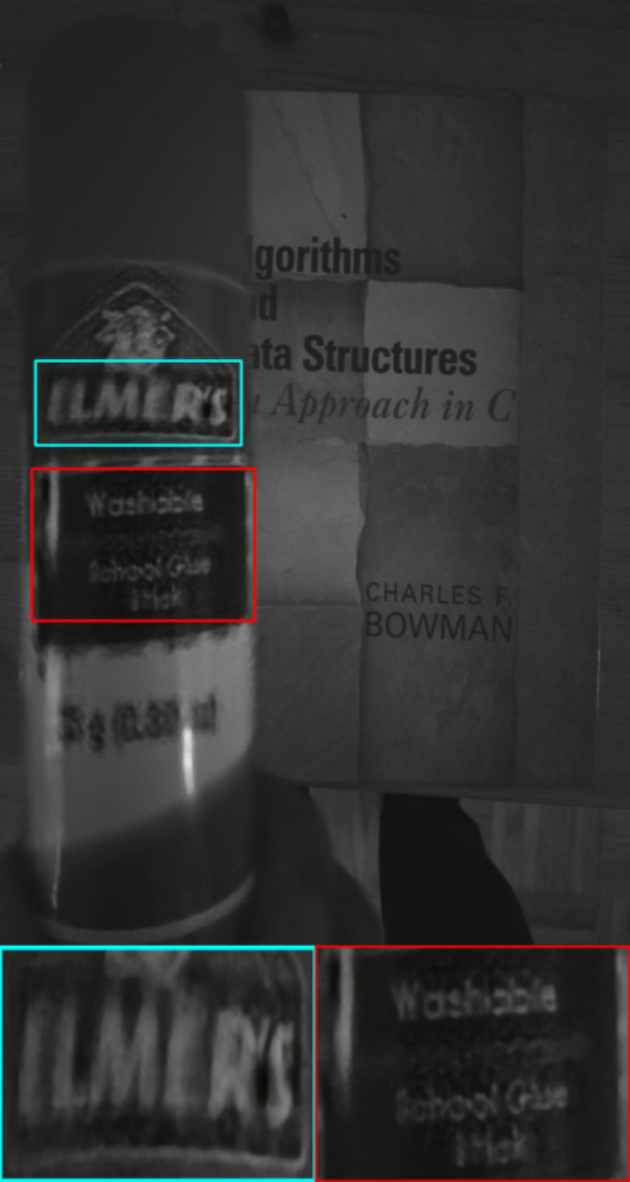}\vspace{4pt}
	\end{subfigure}
	\hfill
	\begin{subfigure}[t]{0.16\linewidth}	
		\includegraphics[width=2.6cm,height=4cm]{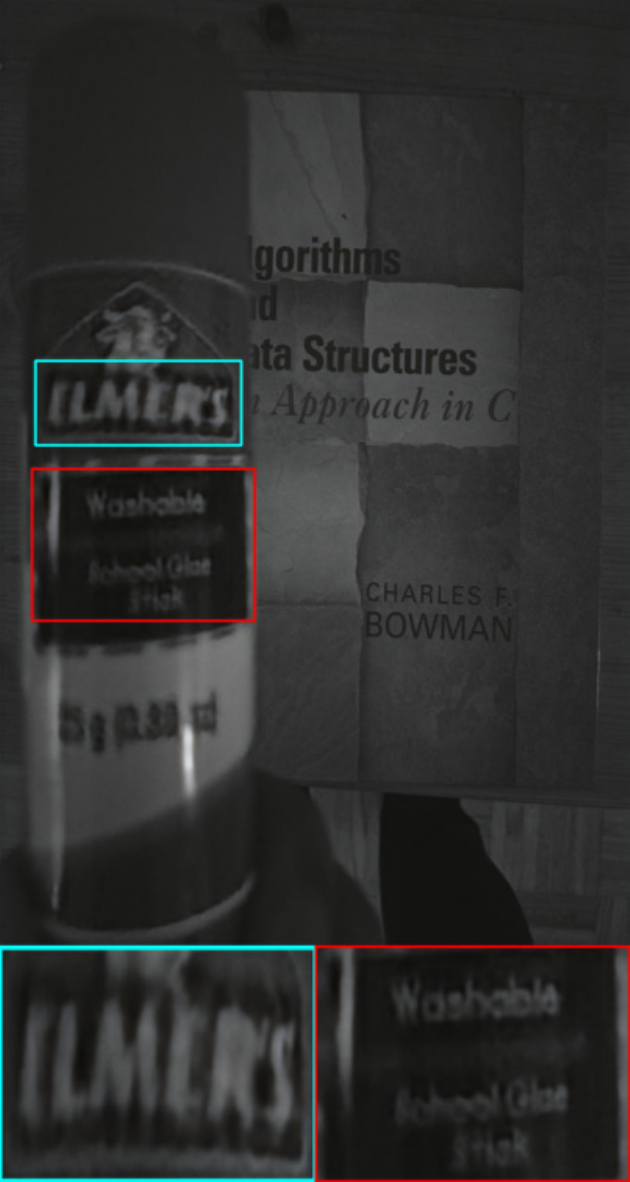}\vspace{4pt}
	\end{subfigure}
	\hfill
	\begin{subfigure}[t]{0.16\linewidth}	
		\includegraphics[width=2.6cm,height=4cm]{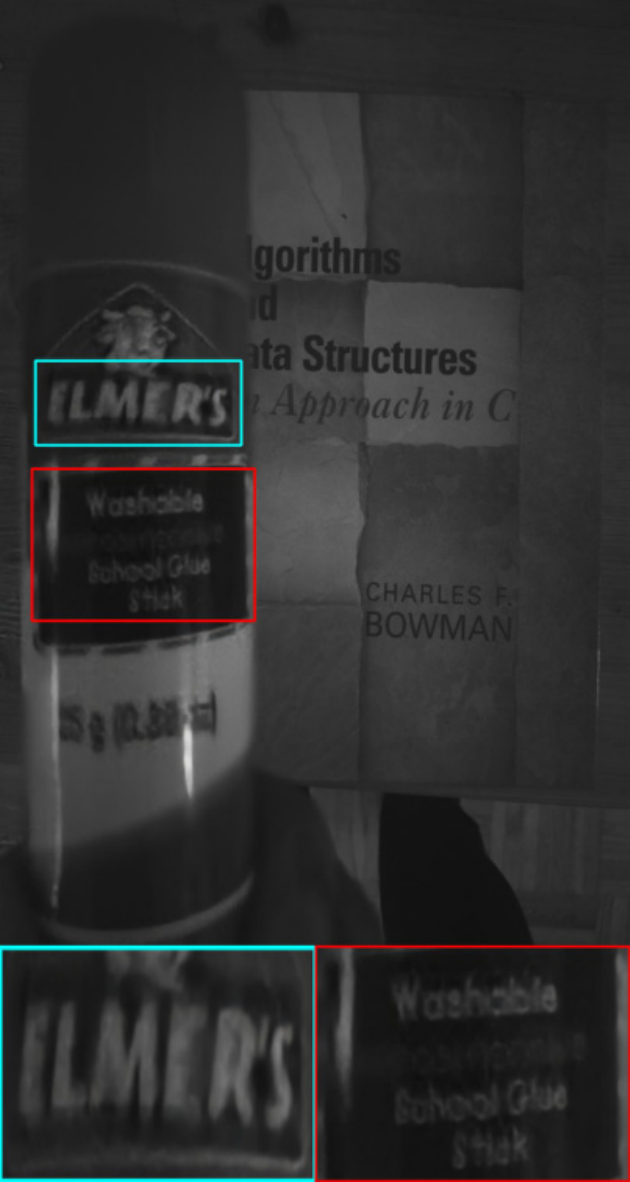}\vspace{4pt}
	\end{subfigure}
	\hfill
	\begin{subfigure}[t]{0.16\linewidth}	
		\includegraphics[width=2.6cm,height=4cm]{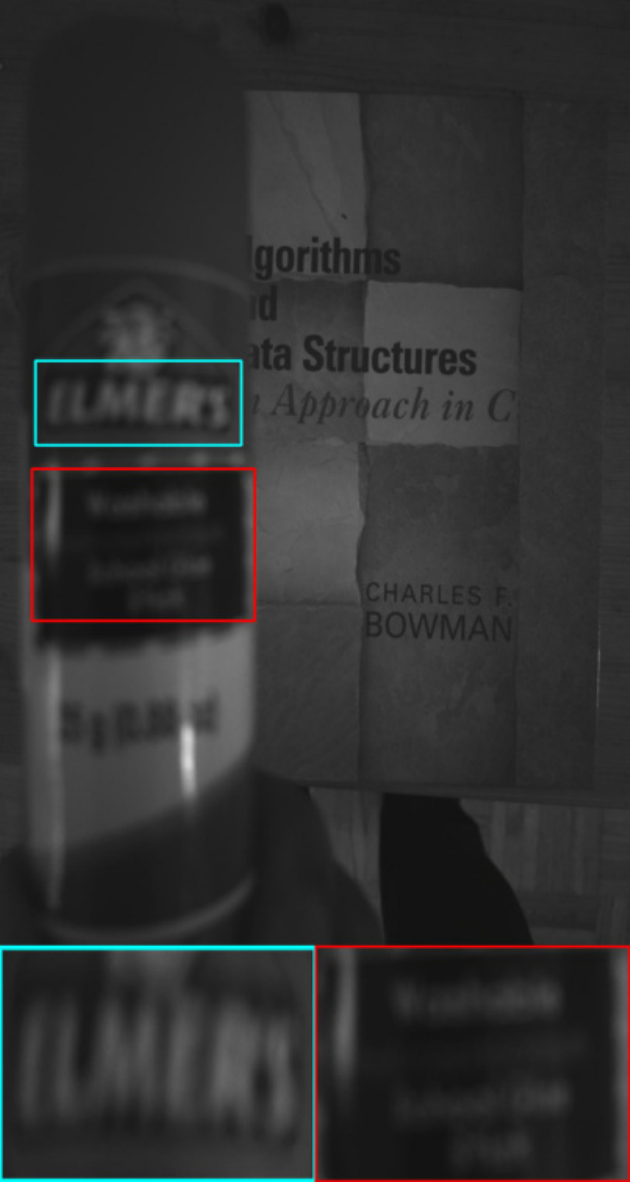}\vspace{4pt}
	\end{subfigure}
	\hfill
	\begin{subfigure}[t]{0.16\linewidth}	
		\includegraphics[width=2.6cm,height=4cm]{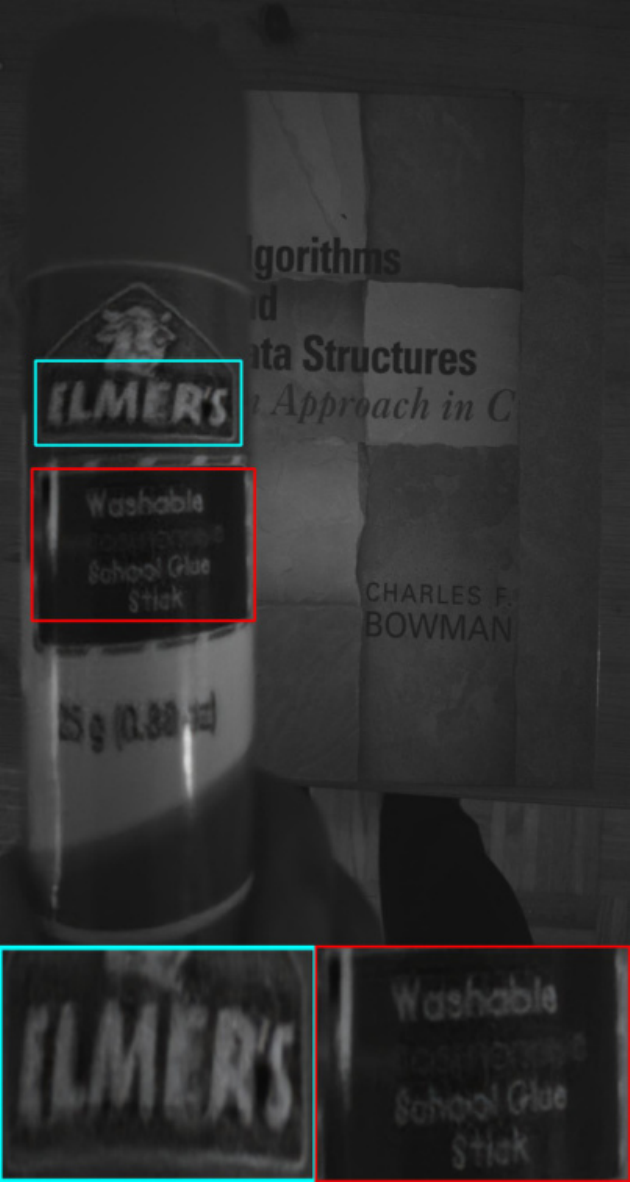}\vspace{4pt}
	\end{subfigure}
	\vfill
	
	\begin{subfigure}[t]{0.16\linewidth}	
		\includegraphics[width=2.6cm,height=4cm]{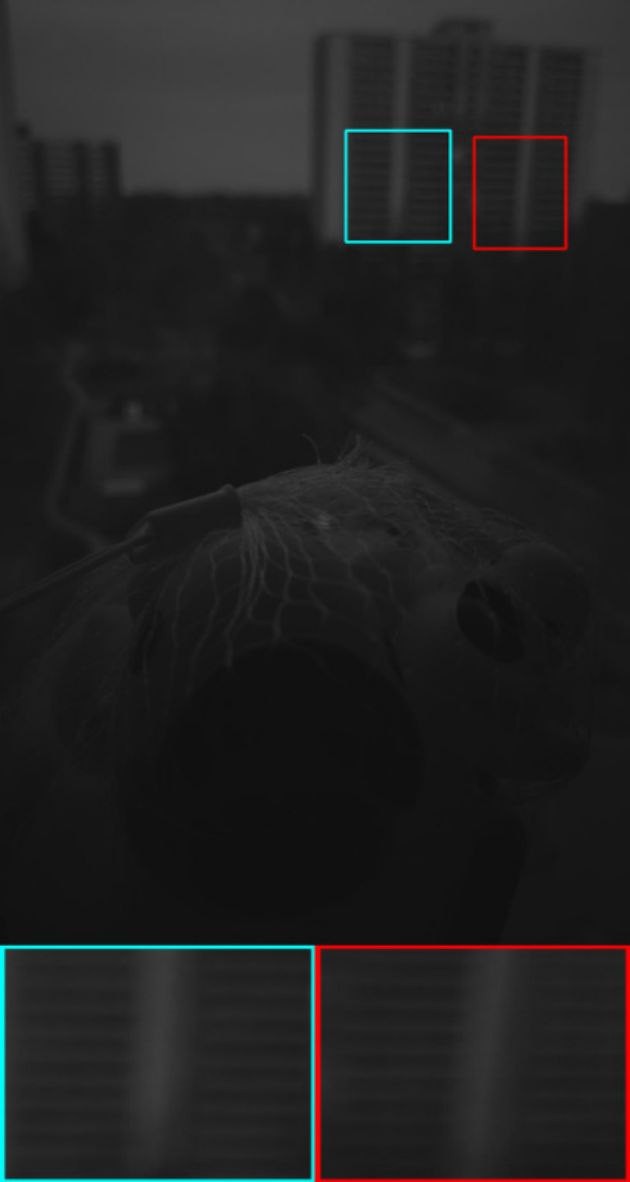}\vspace{4pt}
	\end{subfigure}
	\hfill
	\begin{subfigure}[t]{0.16\linewidth}	
		\includegraphics[width=2.6cm,height=4cm]{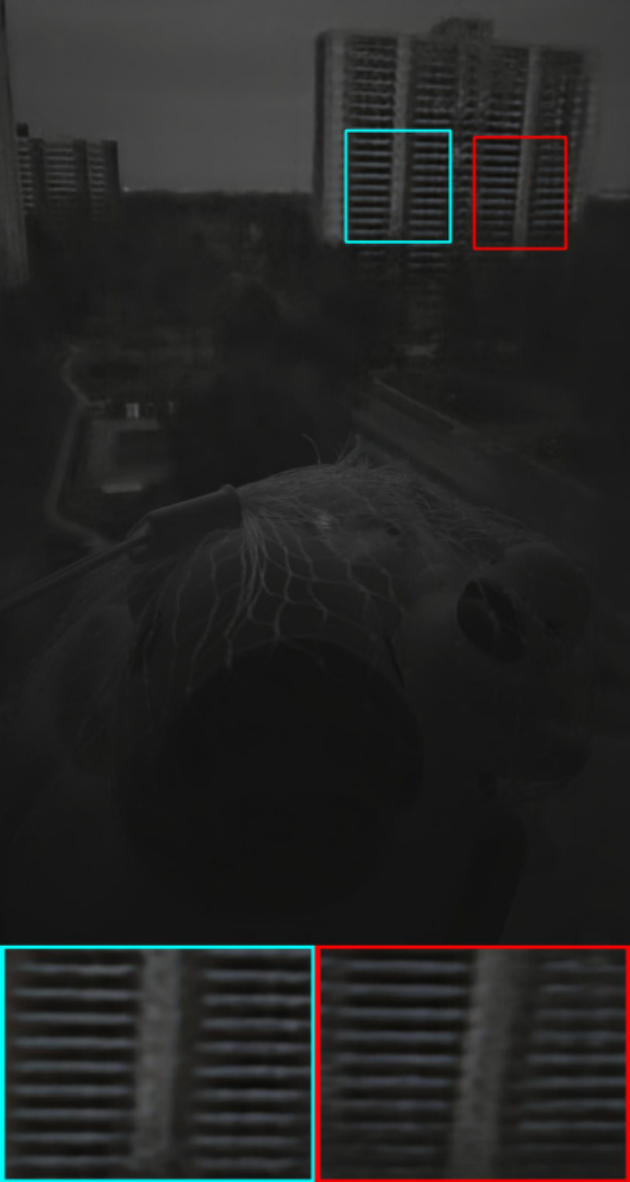}\vspace{4pt}
	\end{subfigure}
	\hfill
	\begin{subfigure}[t]{0.16\linewidth}	
		\includegraphics[width=2.6cm,height=4cm]{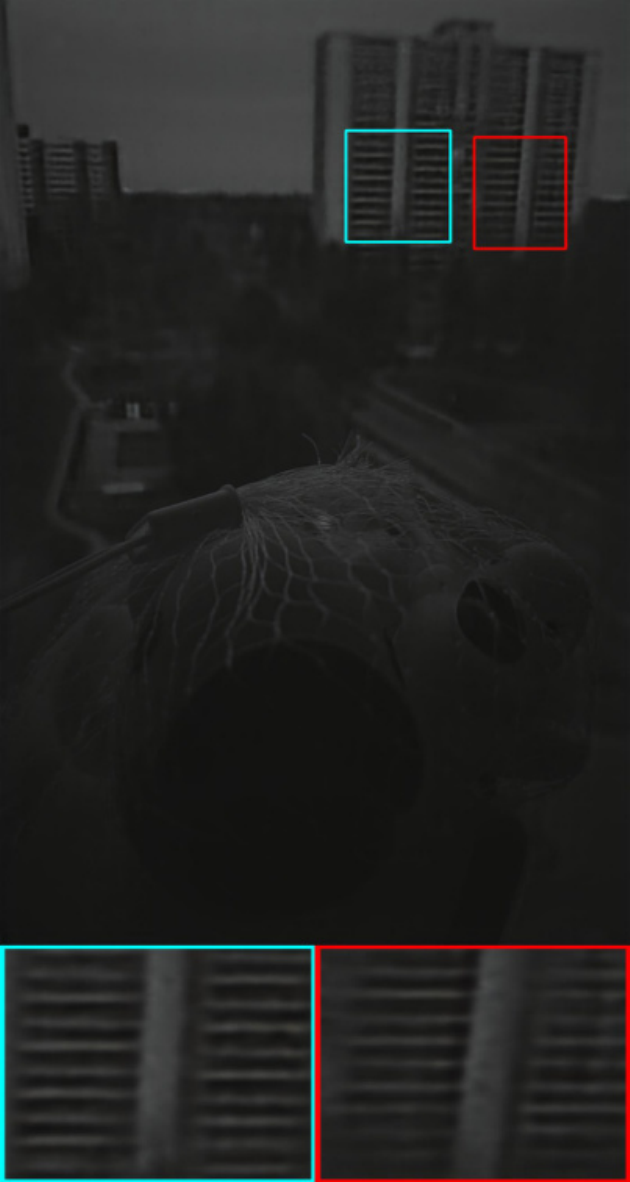}\vspace{4pt}
	\end{subfigure}
	\hfill
	\begin{subfigure}[t]{0.16\linewidth}	
		\includegraphics[width=2.6cm,height=4cm]{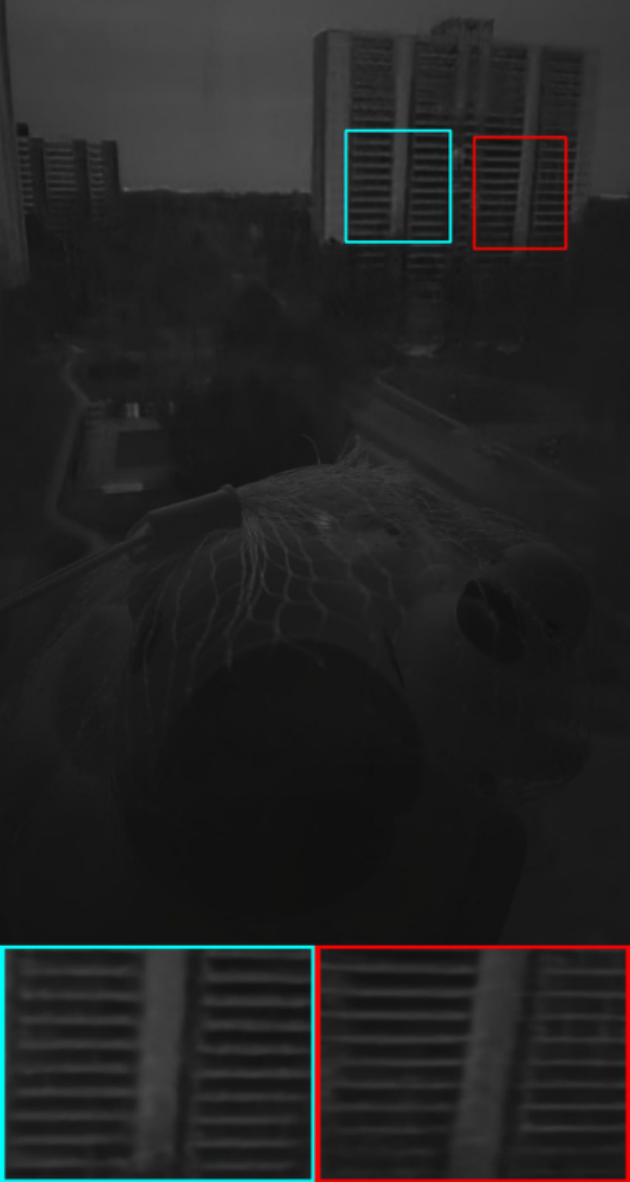}\vspace{4pt}
	\end{subfigure}
	\hfill
	\begin{subfigure}[t]{0.16\linewidth}	
		\includegraphics[width=2.6cm,height=4cm]{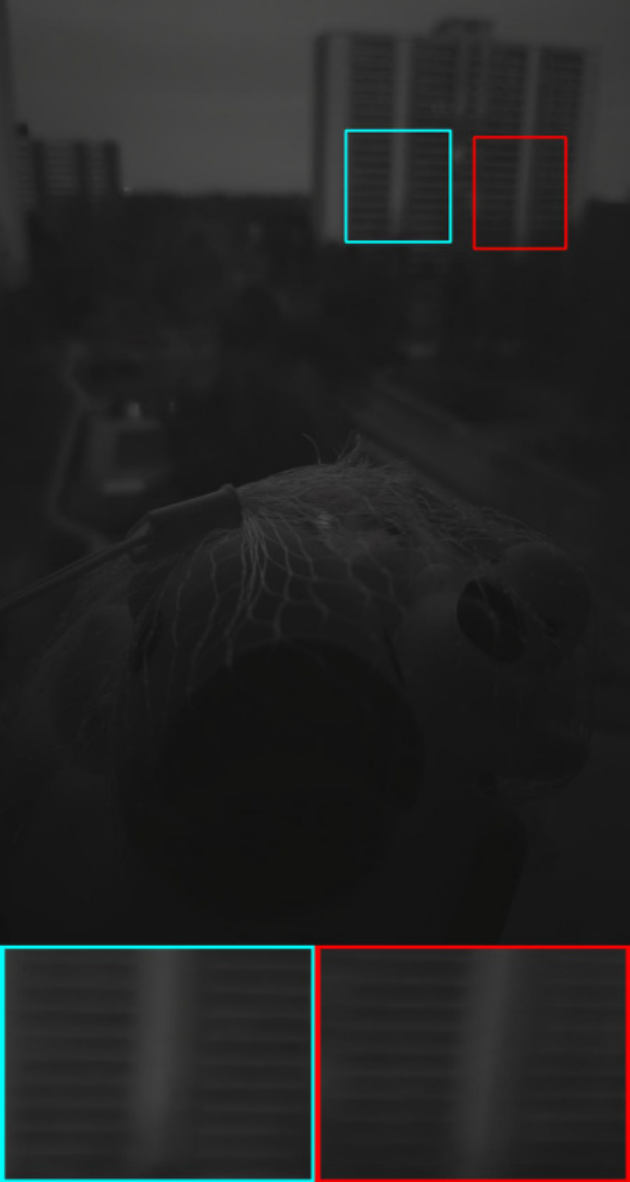}\vspace{4pt}
	\end{subfigure}
	\hfill
	\begin{subfigure}[t]{0.16\linewidth}	
		\includegraphics[width=2.6cm,height=4cm]{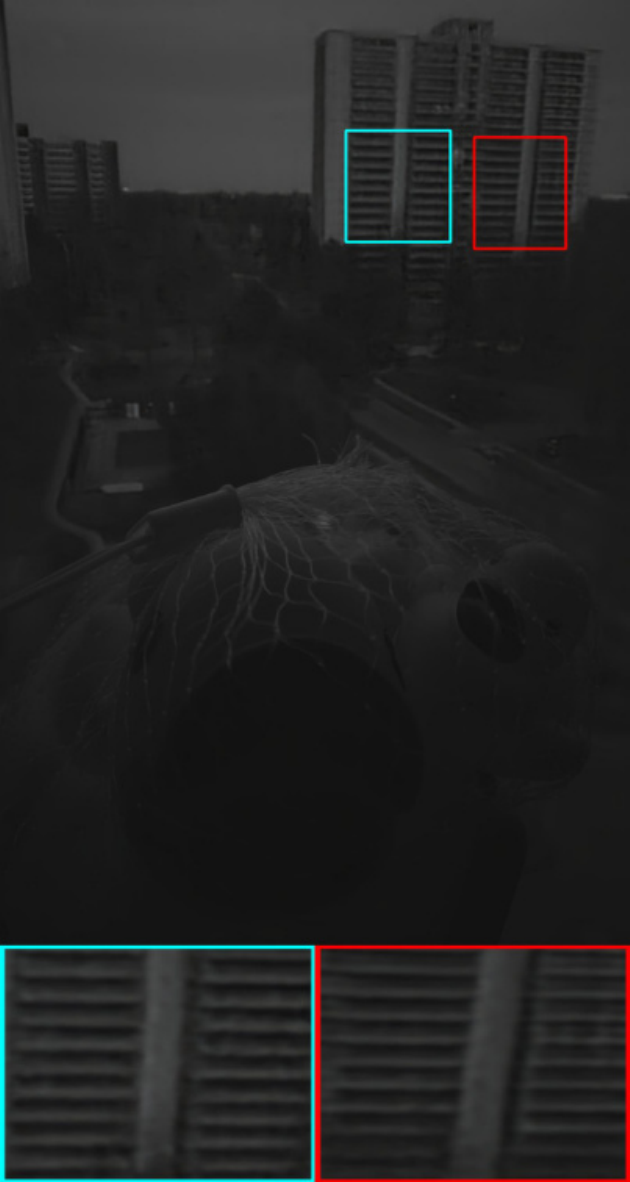}\vspace{4pt}
	\end{subfigure}
	\vfill
	
	\begin{subfigure}[t]{0.16\linewidth}	
		\includegraphics[width=2.6cm,height=4cm]{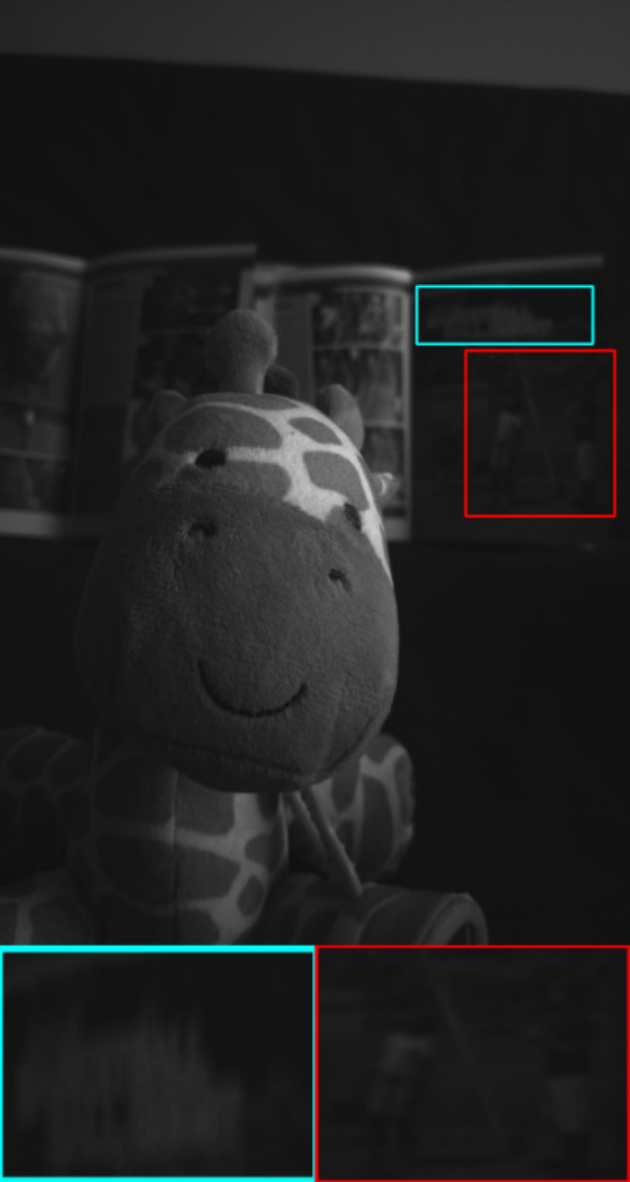}\vspace{4pt}
	\end{subfigure}
	\hfill
	\begin{subfigure}[t]{0.16\linewidth}	
		\includegraphics[width=2.6cm,height=4cm]{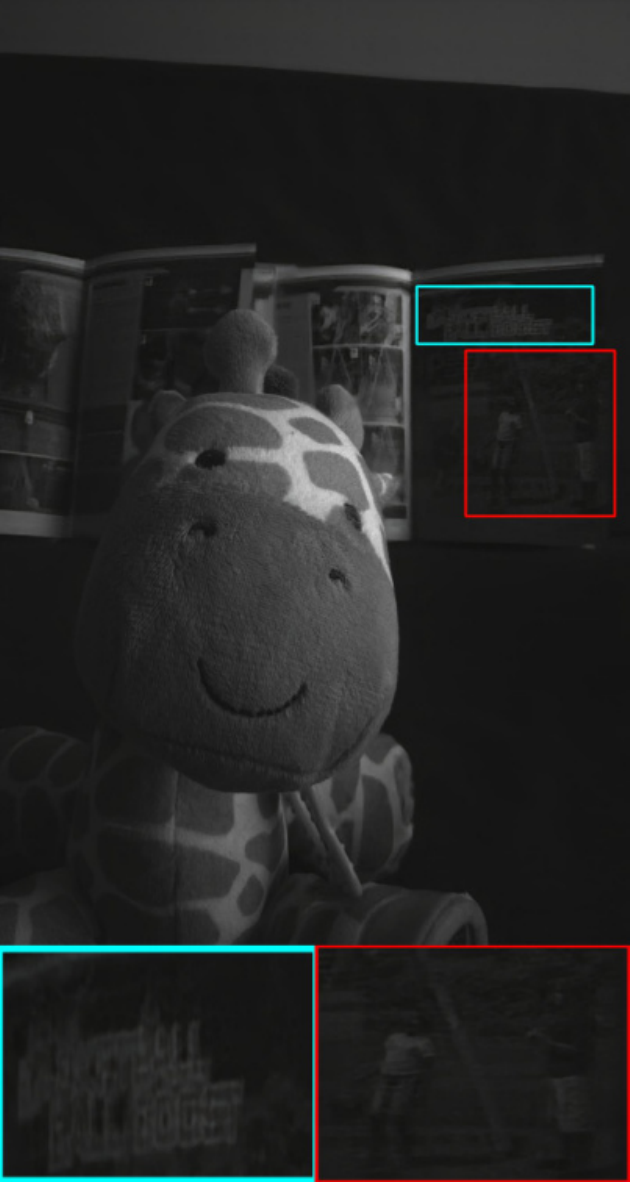}\vspace{4pt}
	\end{subfigure}
	\hfill
	\begin{subfigure}[t]{0.16\linewidth}	
		\includegraphics[width=2.6cm,height=4cm]{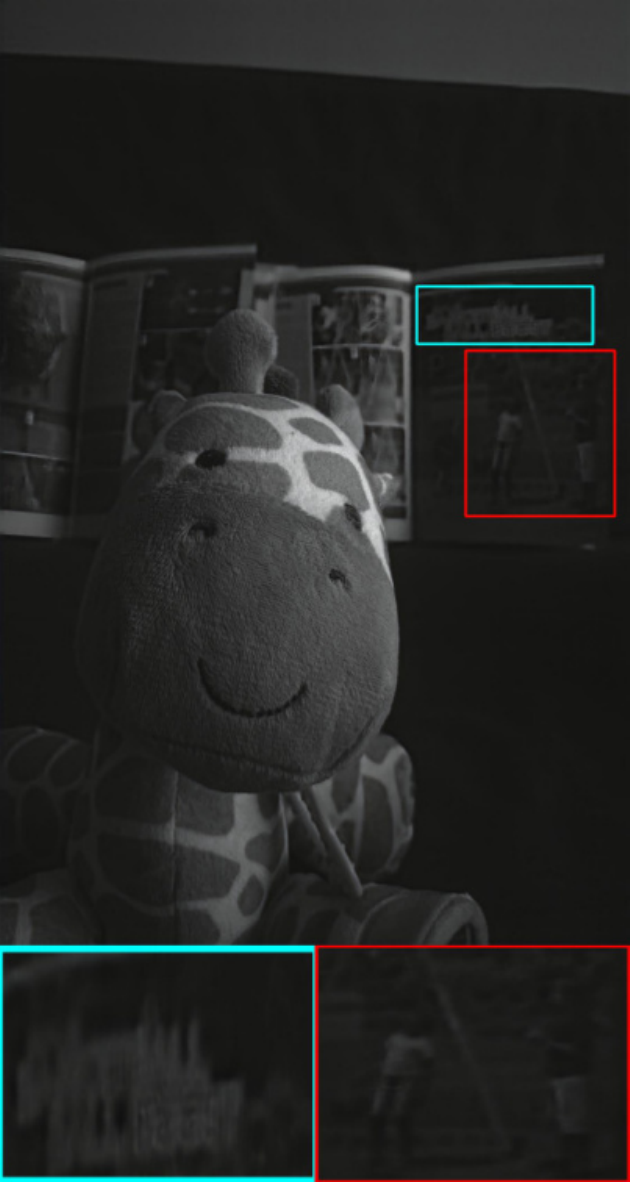}\vspace{4pt}
	\end{subfigure}
	\hfill
	\begin{subfigure}[t]{0.16\linewidth}	
		\includegraphics[width=2.6cm,height=4cm]{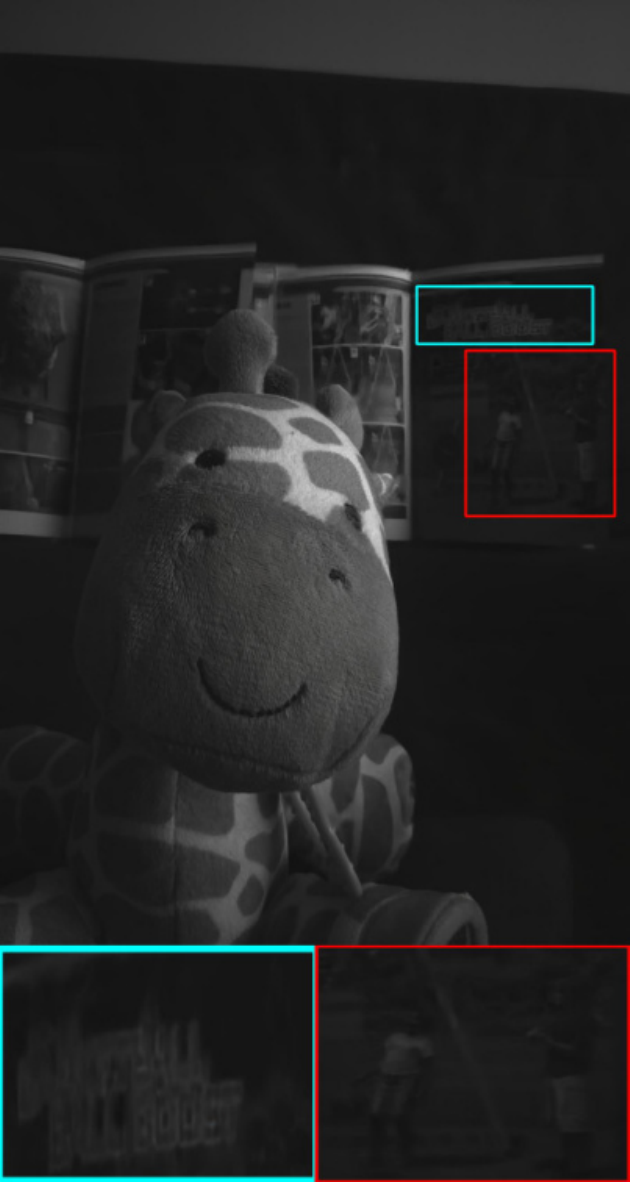}\vspace{4pt}
	\end{subfigure}
	\hfill
	\begin{subfigure}[t]{0.16\linewidth}	
		\includegraphics[width=2.6cm,height=4cm]{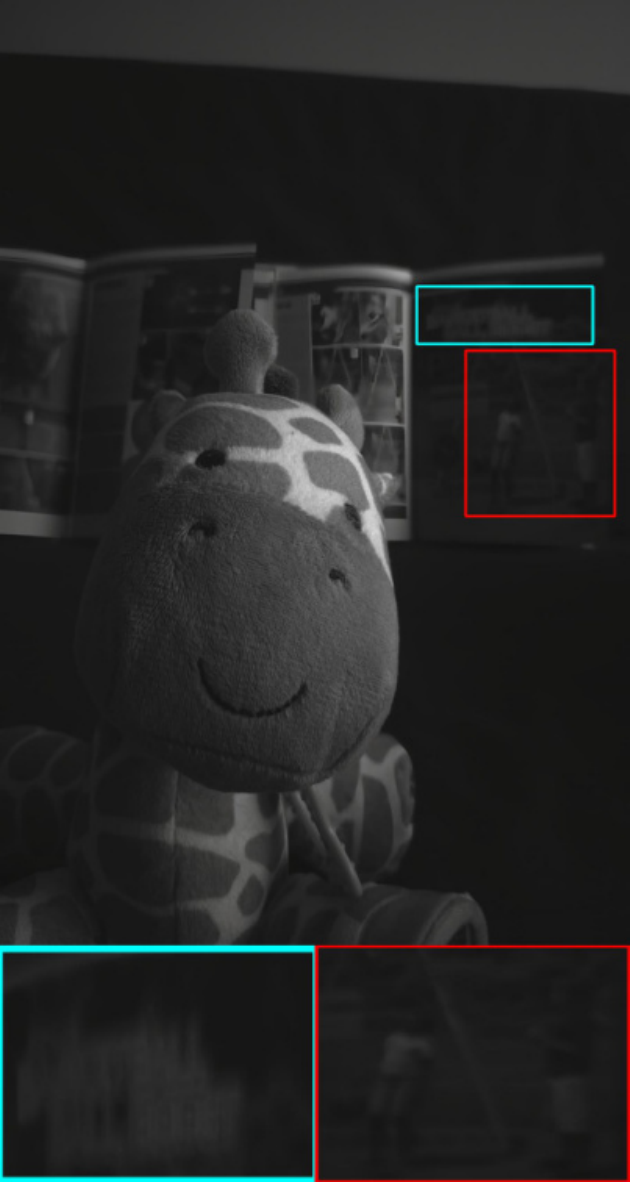}\vspace{4pt}
	\end{subfigure}
	\hfill
	\begin{subfigure}[t]{0.16\linewidth}	
		\includegraphics[width=2.6cm,height=4cm]{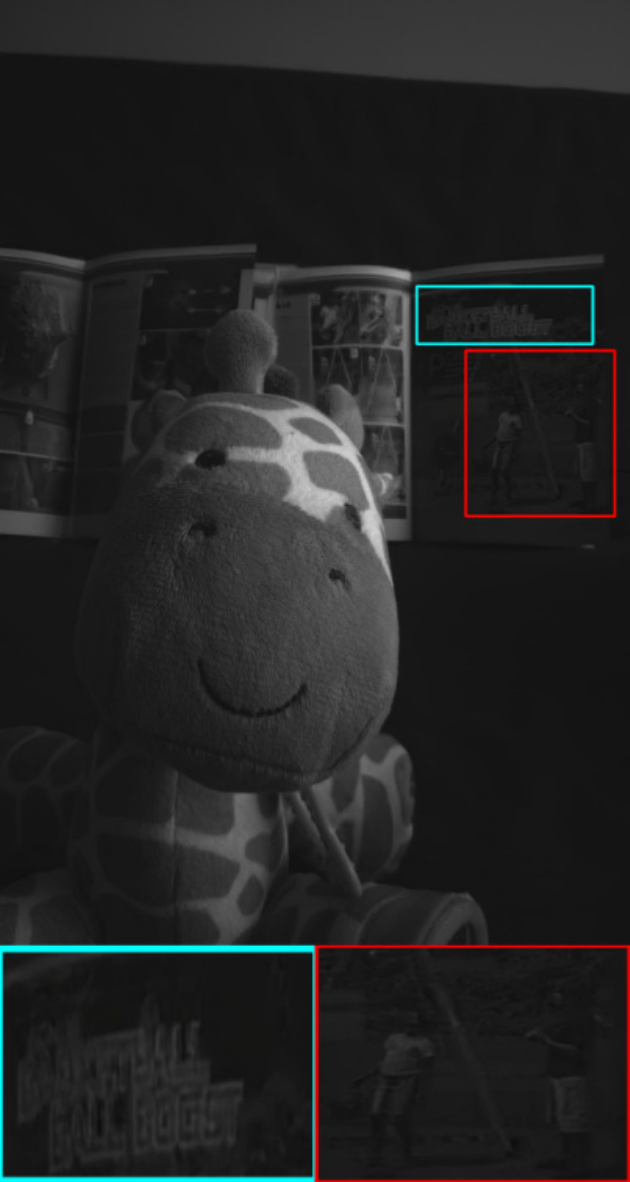}\vspace{4pt}
	\end{subfigure}
	\vfill
	
	\begin{subfigure}[t]{0.16\linewidth}	
		\includegraphics[width=2.6cm,height=4cm]{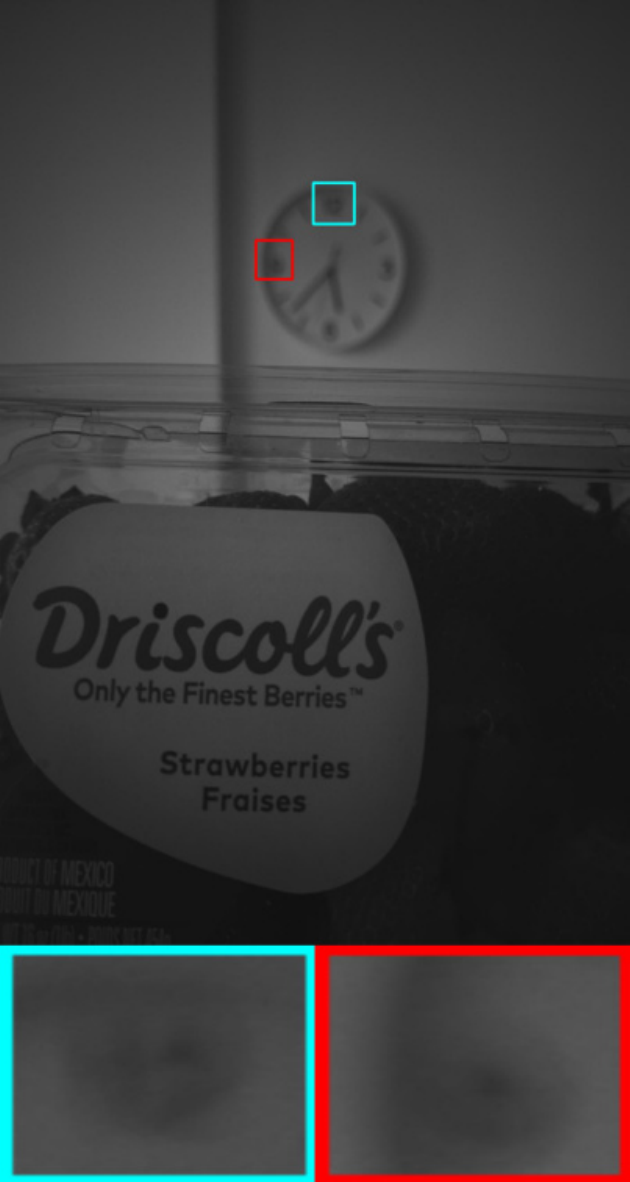}\vspace{4pt}
		\caption{Blurry Input}
	\end{subfigure}
	\hfill
	\begin{subfigure}[t]{0.16\linewidth}	
		\includegraphics[width=2.6cm,height=4cm]{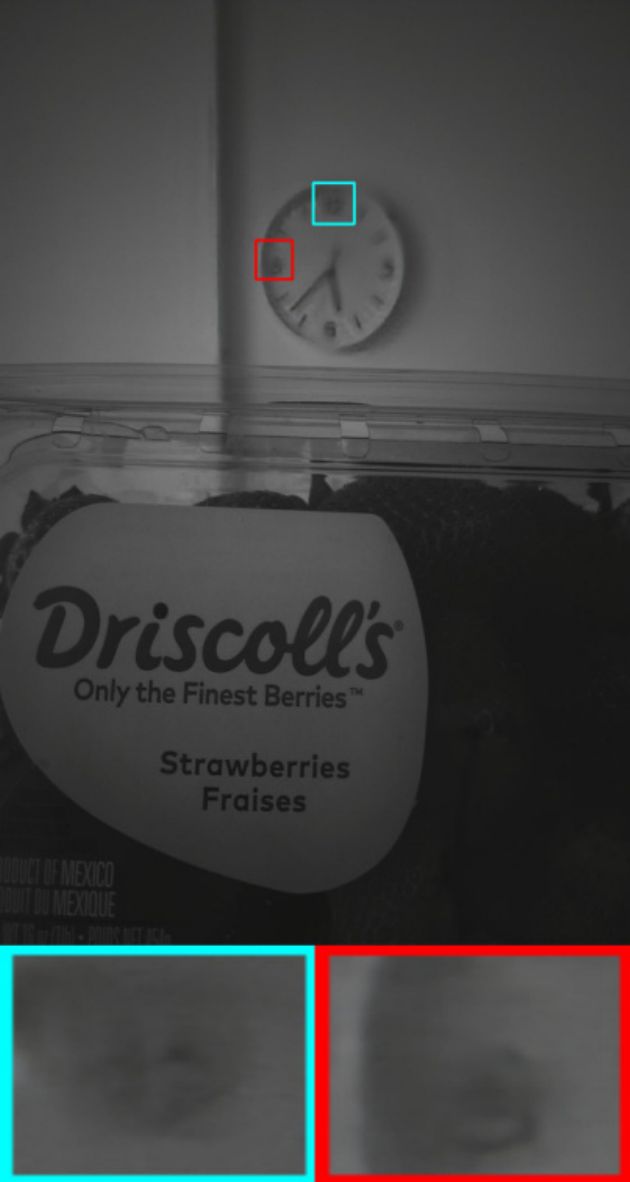}\vspace{4pt}
		\caption{KPAC}
	\end{subfigure}
	\hfill
	\begin{subfigure}[t]{0.16\linewidth}	
		\includegraphics[width=2.6cm,height=4cm]{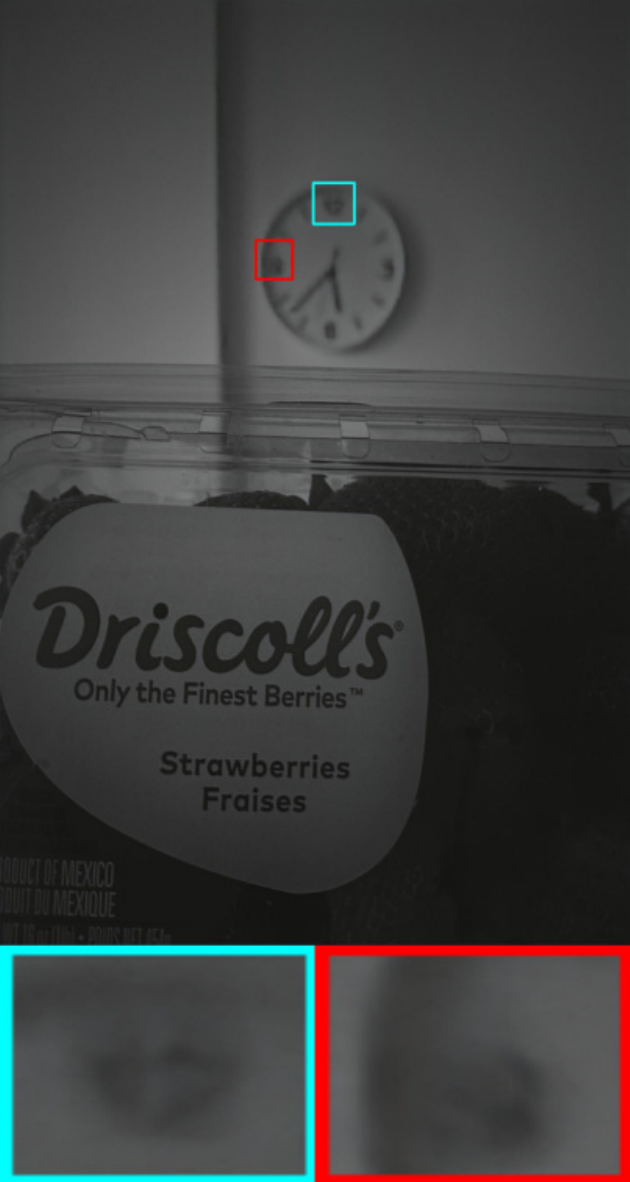}\vspace{4pt}
		\caption{GKMNet}
	\end{subfigure}
	\hfill
	\begin{subfigure}[t]{0.16\linewidth}	
		\includegraphics[width=2.6cm,height=4cm]{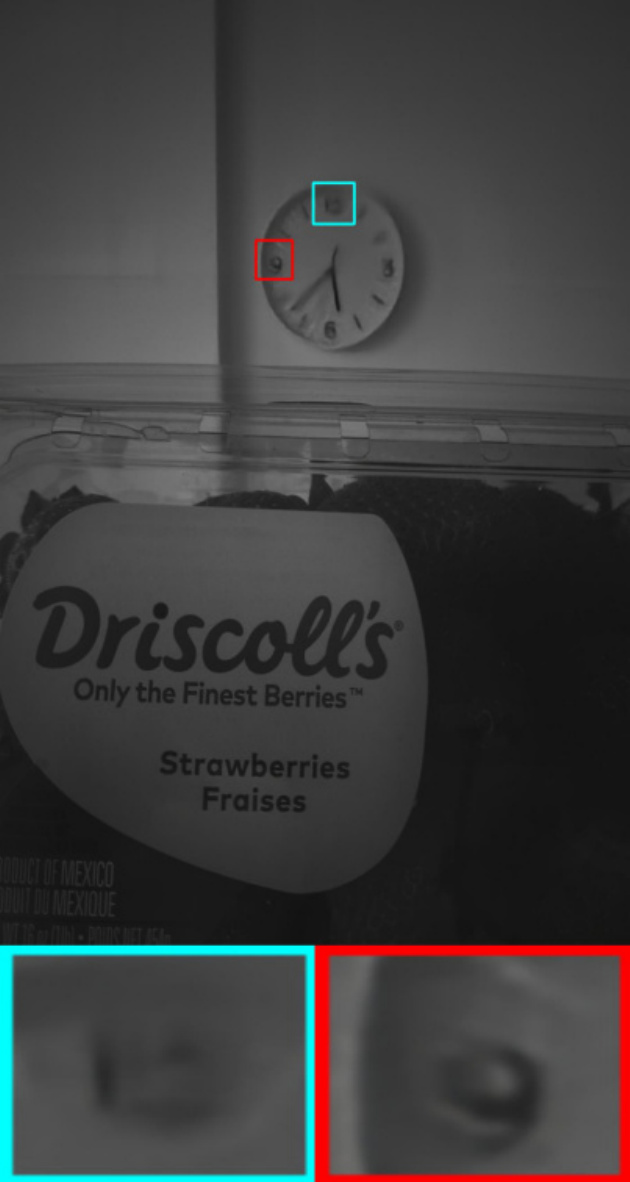}\vspace{4pt}
		\caption{IFAN}
	\end{subfigure}
	\hfill
	\begin{subfigure}[t]{0.16\linewidth}	
		\includegraphics[width=2.6cm,height=4cm]{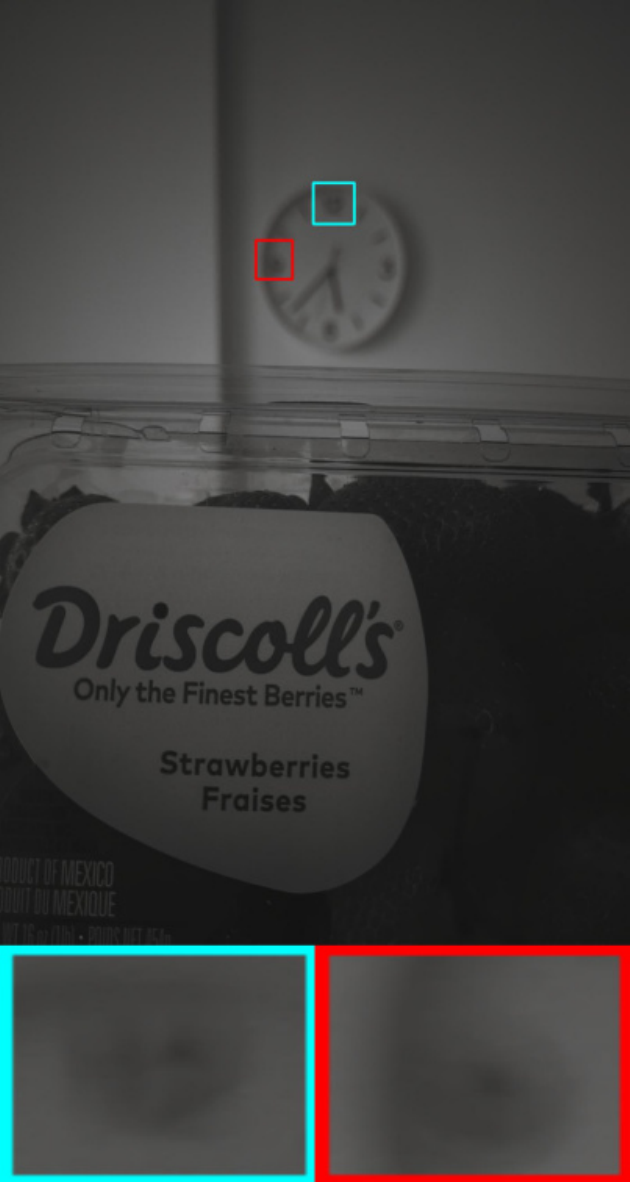}\vspace{4pt}
		\caption{Restormer}
	\end{subfigure}
	\hfill
	\begin{subfigure}[t]{0.16\linewidth}	
		\includegraphics[width=2.6cm,height=4cm]{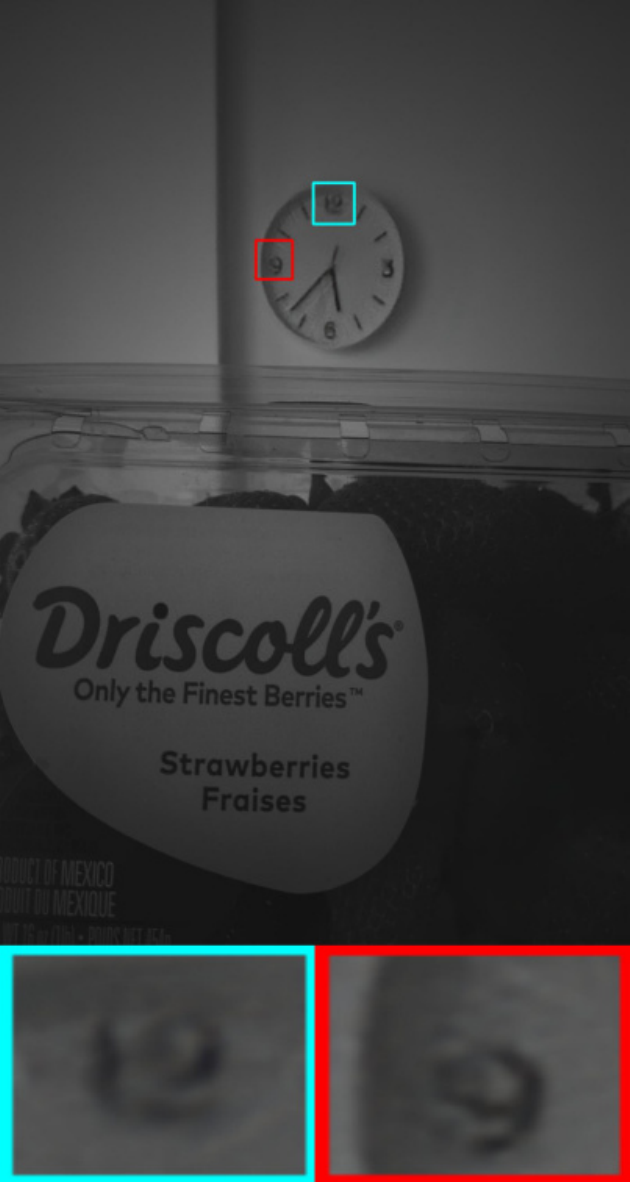}\vspace{4pt}
		\caption{SR-R$^2$KAC-B}
	\end{subfigure}
	\vfill
	
	\caption{Visualization results of different defocus deblurring methods on the PixelDP dataset \cite{Abuolaim2020}.}
	\label{fig_s_4}
\end{figure*}

\begin{figure*}[h]
	\centering
	
	\begin{subfigure}[t]{0.16\linewidth}	
		\includegraphics[width=2.6cm,height=4cm]{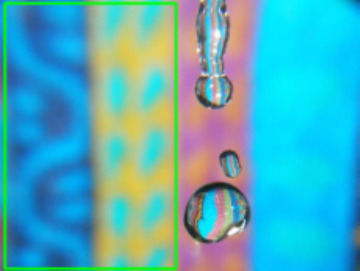}\vspace{4pt}
	\end{subfigure}
	\hfill
	\begin{subfigure}[t]{0.16\linewidth}	
		\includegraphics[width=2.6cm,height=4cm]{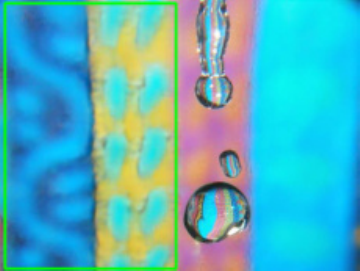}\vspace{4pt}
	\end{subfigure}
	\hfill
	\begin{subfigure}[t]{0.16\linewidth}	
		\includegraphics[width=2.6cm,height=4cm]{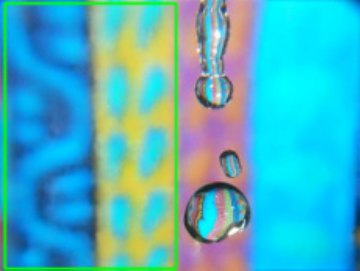}\vspace{4pt}
	\end{subfigure}
	\hfill
	\begin{subfigure}[t]{0.16\linewidth}	
		\includegraphics[width=2.6cm,height=4cm]{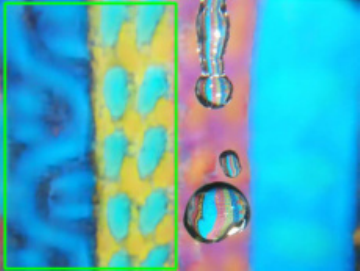}\vspace{4pt}
	\end{subfigure}
	\hfill
	\begin{subfigure}[t]{0.16\linewidth}	
		\includegraphics[width=2.6cm,height=4cm]{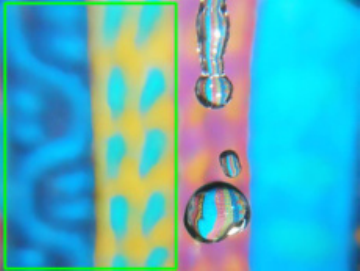}\vspace{4pt}
	\end{subfigure}
	\hfill
	\begin{subfigure}[t]{0.16\linewidth}	
		\includegraphics[width=2.6cm,height=4cm]{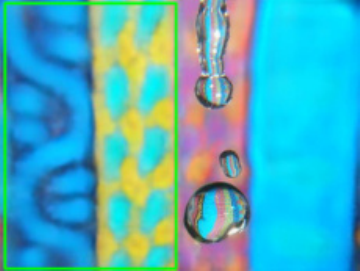}\vspace{4pt}
	\end{subfigure}
	\vfill
	
	\begin{subfigure}[t]{0.16\linewidth}	
		\includegraphics[width=2.6cm,height=4cm]{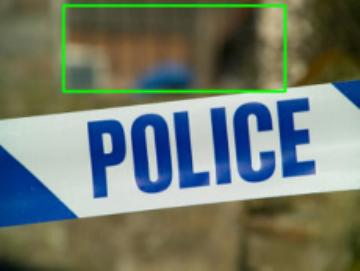}\vspace{4pt}
	\end{subfigure}
	\hfill
	\begin{subfigure}[t]{0.16\linewidth}	
		\includegraphics[width=2.6cm,height=4cm]{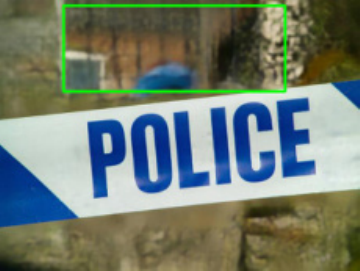}\vspace{4pt}
	\end{subfigure}
	\hfill
	\begin{subfigure}[t]{0.16\linewidth}	
		\includegraphics[width=2.6cm,height=4cm]{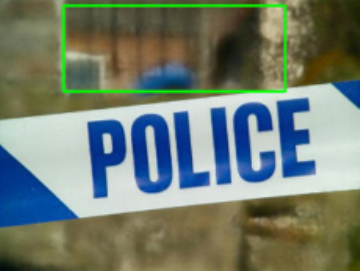}\vspace{4pt}
	\end{subfigure}
	\hfill
	\begin{subfigure}[t]{0.16\linewidth}	
		\includegraphics[width=2.6cm,height=4cm]{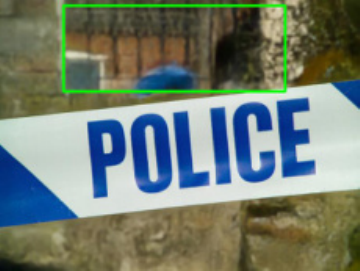}\vspace{4pt}
	\end{subfigure}
	\hfill
	\begin{subfigure}[t]{0.16\linewidth}	
		\includegraphics[width=2.6cm,height=4cm]{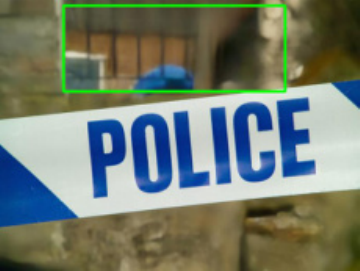}\vspace{4pt}
	\end{subfigure}
	\hfill
	\begin{subfigure}[t]{0.16\linewidth}	
		\includegraphics[width=2.6cm,height=4cm]{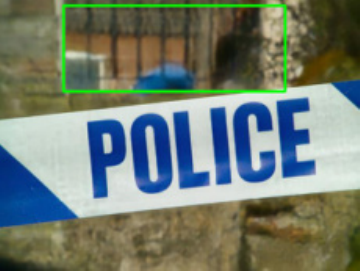}\vspace{4pt}
	\end{subfigure}
	\vfill

	\begin{subfigure}[t]{0.16\linewidth}	
		\includegraphics[width=2.6cm,height=4cm]{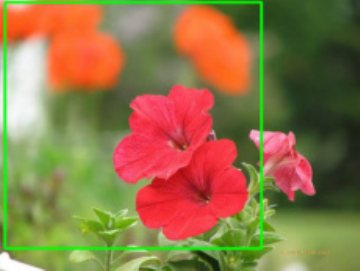}\vspace{4pt}
	\end{subfigure}
	\hfill
	\begin{subfigure}[t]{0.16\linewidth}	
		\includegraphics[width=2.6cm,height=4cm]{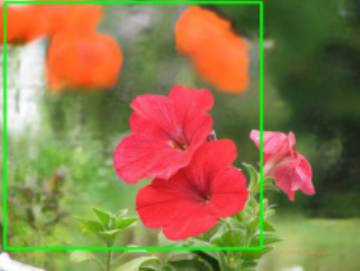}\vspace{4pt}
	\end{subfigure}
	\hfill
	\begin{subfigure}[t]{0.16\linewidth}	
		\includegraphics[width=2.6cm,height=4cm]{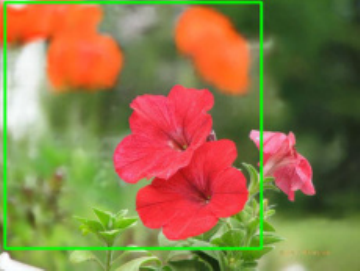}\vspace{4pt}
	\end{subfigure}
	\hfill
	\begin{subfigure}[t]{0.16\linewidth}	
		\includegraphics[width=2.6cm,height=4cm]{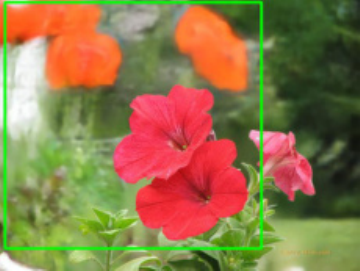}\vspace{4pt}
	\end{subfigure}
	\hfill
	\begin{subfigure}[t]{0.16\linewidth}	
		\includegraphics[width=2.6cm,height=4cm]{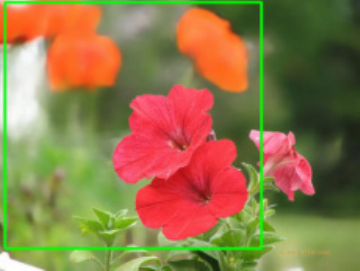}\vspace{4pt}
	\end{subfigure}
	\hfill
	\begin{subfigure}[t]{0.16\linewidth}	
		\includegraphics[width=2.6cm,height=4cm]{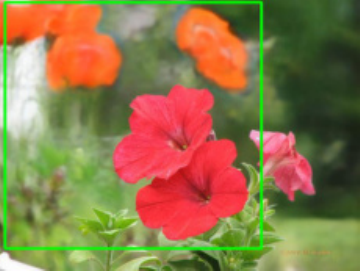}\vspace{4pt}
	\end{subfigure}
	\vfill

	\begin{subfigure}[t]{0.16\linewidth}	
		\includegraphics[width=2.6cm,height=4cm]{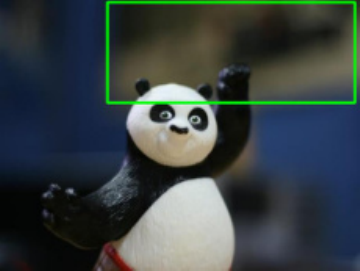}\vspace{4pt}
	\end{subfigure}
	\hfill
	\begin{subfigure}[t]{0.16\linewidth}	
		\includegraphics[width=2.6cm,height=4cm]{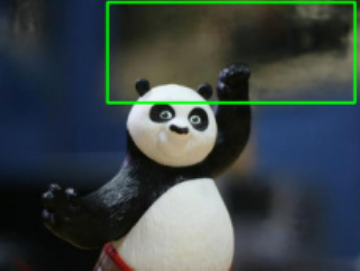}\vspace{4pt}
	\end{subfigure}
	\hfill
	\begin{subfigure}[t]{0.16\linewidth}	
		\includegraphics[width=2.6cm,height=4cm]{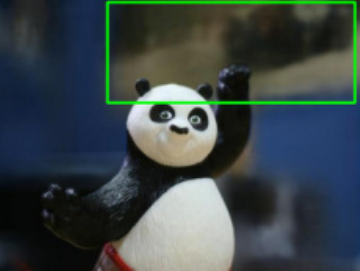}\vspace{4pt}
	\end{subfigure}
	\hfill
	\begin{subfigure}[t]{0.16\linewidth}	
		\includegraphics[width=2.6cm,height=4cm]{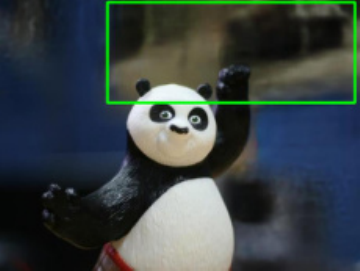}\vspace{4pt}
	\end{subfigure}
	\hfill
	\begin{subfigure}[t]{0.16\linewidth}	
		\includegraphics[width=2.6cm,height=4cm]{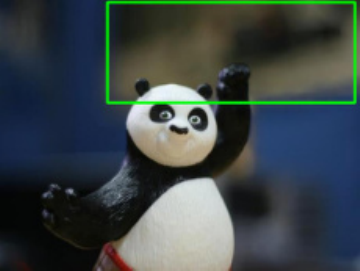}\vspace{4pt}
	\end{subfigure}
	\hfill
	\begin{subfigure}[t]{0.16\linewidth}	
		\includegraphics[width=2.6cm,height=4cm]{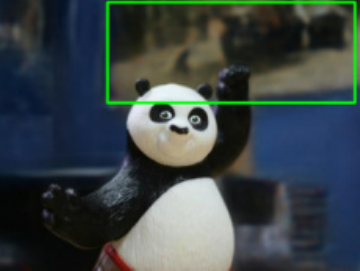}\vspace{4pt}
	\end{subfigure}
	\vfill

	\begin{subfigure}[t]{0.16\linewidth}	
		\includegraphics[width=2.6cm,height=4cm]{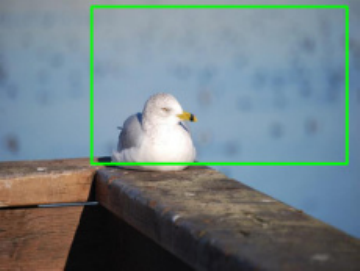}\vspace{4pt}
		\caption{Blurry Input}
	\end{subfigure}
	\hfill
	\begin{subfigure}[t]{0.16\linewidth}	
		\includegraphics[width=2.6cm,height=4cm]{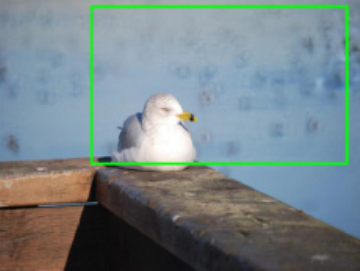}\vspace{4pt}
		\caption{KPAC}
	\end{subfigure}
	\hfill
	\begin{subfigure}[t]{0.16\linewidth}	
		\includegraphics[width=2.6cm,height=4cm]{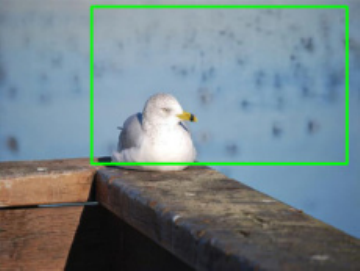}\vspace{4pt}
		\caption{GKMNet}
	\end{subfigure}
	\hfill
	\begin{subfigure}[t]{0.16\linewidth}	
		\includegraphics[width=2.6cm,height=4cm]{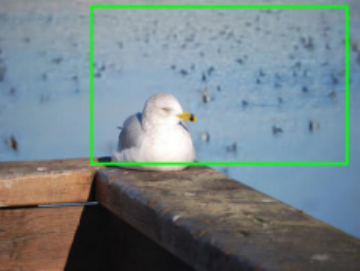}\vspace{4pt}
		\caption{IFAN}
	\end{subfigure}
	\hfill
	\begin{subfigure}[t]{0.16\linewidth}	
		\includegraphics[width=2.6cm,height=4cm]{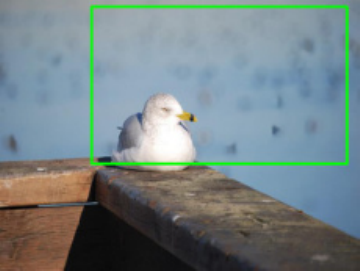}\vspace{4pt}
		\caption{Restormer}
	\end{subfigure}
	\hfill
	\begin{subfigure}[t]{0.16\linewidth}	
		\includegraphics[width=2.6cm,height=4cm]{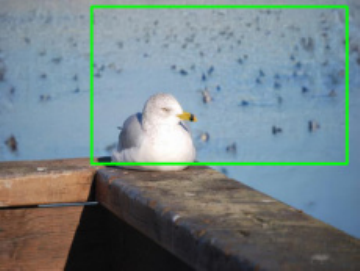}\vspace{4pt}
		\caption{SR-R$^2$KAC-B}
	\end{subfigure}
	\vfill

	\caption{Visualization results of different defocus deblurring methods on the CUHK dataset \cite{Shi2014}. }
	\vspace{-0.2in}
	\label{fig_s_7}
\end{figure*}

For more comprehensive comparison, we present more qualitative results of  KPAC \cite{Son2021}, IFAN \cite{Lee2021}, GKMNet \cite{Quan2021}, Restormer \cite{Zamir2022} and our methods on different datasets, i.e., DPDD \cite{Abuolaim2020}, RealDoF \cite{Lee2021}, PixelDP \cite{Abuolaim2020} and CUHK \cite{Shi2014} datasets. The results about DPDD, RealDoF, PixelDP and CUHK are shown in Fig.~\ref{fig_s_2}-\ref{fig_s_7},
	
\end{document}